\documentclass[referee,pdflatex,sn-mathphys-num]{sn-jnl}

\usepackage{graphicx}%
\usepackage{multirow}%
\usepackage{amsmath,amssymb,amsfonts}%
\usepackage{amsthm}%
\usepackage{mathrsfs}%
\usepackage[title]{appendix}%
\usepackage{xcolor}%
\usepackage{textcomp}%
\usepackage{manyfoot}%
\usepackage{booktabs}%
\usepackage{algorithm}%
\usepackage{algorithmicx}%
\usepackage{algpseudocode}%
\usepackage{listings}%

\usepackage{bm}
\usepackage{cases}
\usepackage{multirow}

\usepackage{multicol}
\usepackage{xcolor}
\usepackage{float,subfig}
\usepackage{caption}
\usepackage{url}
\usepackage{here}
\usepackage{stfloats} 
\usepackage{braket}
\usepackage{multicol}
\usepackage[export]{adjustbox}
\usepackage{hhline}
\usepackage{colortbl}
\usepackage{booktabs}

\usepackage{hyperref}
\hypersetup{
    colorlinks=true,
    linkcolor=blue, 
    urlcolor=magenta,
    }
\makeatletter
\let\saved@hyper@linkurl\hyper@linkurl
\let\saved@hyper@link@\hyper@link@
\AtBeginDocument{%

  \NoHyper 
  
  \let\hyper@linkurl\saved@hyper@linkurl 
  \let\hyper@link@\saved@hyper@link@ 
}
\makeatother

\usepackage{array}
\newcolumntype{C}[1]{>{\hfil}m{#1}<{\hfil}}

\usepackage{lscape} 

\usepackage{tikz}
\usetikzlibrary{arrows.meta,shapes,positioning,shapes.geometric}

\theoremstyle{thmstyleone}%
\theoremstyle{thmstyletwo}%

\theoremstyle{thmstylethree}%

\begin{document}

\title[Article Title]{A Parameter-free Adaptive Resonance Theory-based Topological Clustering Algorithm Capable of Continual Learning}


\author*[1]{\fnm{Naoki} \sur{Masuyama}}\email{masuyama@omu.ac.jp}

\author[1]{\fnm{Takanori} \sur{Takebayashi}}\email{sc23284w@st.omu.ac.jp}

\author[1]{\fnm{Yusuke} \sur{Nojima}}\email{nojima@omu.ac.jp}

\author[2]{\fnm{Chu} \spfx{Kiong} \sur{Loo}}\email{ckloo.um@um.edu.my}

\author[3]{\fnm{Hisao} \sur{Ishibuchi}}\email{hisao@sustech.edu.cn}

\author[4]{\fnm{Stefan} \sur{Wermter}}\email{stefan.wermter@uni-hamburg.de}

\affil*[1]{\orgdiv{Department of Core Informatics}, \orgname{Osaka Metropolitan University}, \orgaddress{\street{1-1 Gakuen-cho Naka-ku}, \city{Sakai-Shi}, \postcode{599-8531}, \state{Osaka}, \country{Japan}}}

\affil[2]{\orgdiv{Department of Artificial Intelligence}, \orgname{Universiti Malaya}, \orgaddress{\city{Kuala Lumpur}, \postcode{50603}, \state{Wilayah Persekutuan Kuala Lumpur}, \country{Malaysia}}}

\affil[3]{\orgdiv{Department of Computer Science and Engineering}, \orgname{Southern University of Science and Technology}, \orgaddress{\city{Shenzhen}, \postcode{518055}, \state{Guangdong Province}, \country{China}}}

\affil[4]{\orgdiv{Department of Informatics}, \orgname{University of Hamburg}, \orgaddress{\street{Vogt-Koelln-Str. 30}, \postcode{22527}, \state{Hamburg}, \country{Germany}}}


\abstract{In general, a similarity threshold (i.e., a vigilance parameter) for a node learning process in Adaptive Resonance Theory (ART)-based algorithms has a significant impact on clustering performance. In addition, an edge deletion threshold in a topological clustering algorithm plays an important role in adaptively generating well-separated clusters during a self-organizing process. In this paper, we propose an ART-based topological clustering algorithm that integrates parameter estimation methods for both the similarity threshold and the edge deletion threshold. The similarity threshold is estimated using a determinantal point process-based criterion, while the edge deletion threshold is defined based on the age of edges. Experimental results with synthetic and real-world datasets show that the proposed algorithm has superior clustering performance to state-of-the-art clustering algorithms without requiring parameter specifications specific to the datasets. Source code is available at \url{https://github.com/Masuyama-lab/CAE}}

\keywords{Clustering, Adaptive Resonance Theory, Continual Learning, Correntropy}



\maketitle

\section{Introduction}
\label{sec:introduction}
Recent advances in the Internet of Things (IoT) have enabled the creation and acquisition of a wide variety of data. Such data are considered important economic resources that can be used in marketing, finance, and the development of IoT solutions. Supervised and unsupervised learning are typical approaches for analyzing acquired data and extracting valuable information. In general, supervised learning algorithms require sufficient labeled training data to achieve high information extraction performance. In contrast, unsupervised learning approaches, such as clustering, can extract valuable information without the need for labeled training data. $k$-means \cite{lloyd82}, Gaussian Mixture Model (GMM) \cite{mclachlan19}, and Self-Organizing Map (SOM) \cite{kohonen82} are representative clustering algorithms. Although $k$-means, GMM, and SOM are simple and highly applicable, these algorithms require the number of clusters or the network size to be specified in advance. This limitation makes it difficult to apply them to data whose distribution is unknown or continually changing.

Typical self-organizing clustering algorithms include Growing Neural Gas (GNG) \cite{fritzke95}, Self-Organizing Incremental Neural Network (SOINN) \cite{furao06}, and Adjusted SOINN (ASOINN) \cite{shen08}. These algorithms can adaptively and continually generate topological networks (i.e., nodes and edges) depending on the distribution of the given data. Adaptive Resonance Theory (ART) \cite{carpenter88, carpenter91b, grossberg23} is widely recognized for addressing the plasticity-stability dilemma in continual learning. Various types of ART-based algorithms have been developed over the years \cite{carpenter91b, vigdor07, wang19, masuyama19b, masuyama22a, masuyama22b}. A notable development is the use of the Correntropy-Induced Metric (CIM) \cite{liu07}, which provides a kernel-based similarity measure. CIM mitigates the problem of distance concentration in high-dimensional spaces and preserves local manifold structures. Thanks to these properties, algorithms employing CIM exhibit faster and more stable self-organizing performance \cite{masuyama19b, masuyama22a, masuyama22b}. Among them, CIM-based ART with Edge and Age (CAEA) \cite{masuyama22a} represents the state-of-the-art in ART-based self-organizing clustering. However, conventional ART-based clustering algorithms depend strongly on manually specified parameters (e.g., vigilance parameter), which are highly sensitive to data characteristics. This sensitivity limits their adaptability in dynamic environments.

To overcome the above-mentioned limitations, this paper proposes a new parameter-free ART-based topological clustering algorithm, called CIM-based ART with Edge (CAE), by introducing two parameter estimation methods to CAEA \cite{masuyama22a}. In the proposed algorithm, a similarity threshold is calculated based on pairwise CIM similarities among a certain number of nodes. The required number of nodes for calculating the similarity threshold is estimated by a Determinantal Point Processes (DPP)-based criterion \cite{kulesza12, holder20}. DPP is a probabilistic model that evaluates the diversity of a set of elements by favoring subsets whose elements are dissimilar to each other. In other words, DPP encourages the selection of diverse and representative subsets, making it suitable for determining an appropriate number of nodes in this context. In addition, an edge deletion threshold is estimated based on the age of each edge, inspired by the edge deletion mechanism of SOINN+ \cite{wiwatcharakoses20}. Due to the parameter estimation methods, the proposed algorithm can adaptively, efficiently, and continually generate a topological network from the given data while maintaining superior clustering performance compared to conventional algorithms. Note that continual learning is generally categorized into three scenarios: domain incremental learning, task incremental learning, and class incremental learning \cite{van19, wiewel19}. This paper focuses on class-incremental learning problems.

The main contributions of this paper are as follows:
\begin{itemize}
	\vspace{-1mm}
	\item[(i)] CAE is proposed as a new parameter-free ART-based clustering algorithm capable of continual learning.
	
	\item[(ii)] An estimation method for the number of nodes used in calculating a similarity threshold is introduced to CAE. The sufficient number of nodes for calculating the threshold is estimated by a DPP-based criterion incorporating CIM.
	
	\item[(iii)] An estimation method of a node deletion threshold is introduced to CAE, which is inspired by the edge deletion method of SOINN+ \cite{wiwatcharakoses20}. The node deletion threshold is estimated based on the age of each edge.
	
	\item[(iv)] Empirical studies show that CAE achieves superior clustering performance over state-of-the-art algorithms, while requiring no parameter specification and supporting continual learning.
\end{itemize}

The paper is organized as follows. Section \ref{sec:literature} presents a literature review. Section \ref{sec:preliminary} presents the preliminary knowledge for a similarity measure and a kernel density estimator used in CAE. Section \ref{sec:proposedAlgorithm} presents the learning procedure of CAE in detail. Section \ref{sec:experiment} presents extensive simulation experiments to evaluate clustering performance by using synthetic and real-world datasets. Section \ref{sec:conclusion} concludes this paper.

\section{Literature Review}
\label{sec:literature}
Continual learning has gained increasing attention as a framework for acquiring knowledge in dynamic, non-stationary environments. A key challenge is overcoming catastrophic forgetting, in which models lose prior knowledge when learning new information. While early studies in continual learning often focused on task-incremental learning for its simplicity in benchmarking catastrophic forgetting, recent research in the deep learning community has increasingly shifted toward class-incremental learning. This setting is considered more realistic and challenging, as no task identifiers are available at test time, and it has become the dominant focus of current continual learning research \cite{zhou24a}. With the rapid progress of deep learning, continual learning has also been actively studied in unsupervised settings \cite{sadeghi24, solomon24, aghasanli25, zhu25}. In such scenarios, the absence of explicit labels introduces additional challenges.

Although many deep learning-based approaches have shown remarkable performance on structured data (e.g., image and text data), their superiority is not always observed with tabular data. In fact, classical machine learning algorithms can still perform competitively on tabular data \cite{shwartz22}. This points out that the superiority of deep learning is not universal across all data modalities. Although recent studies have proposed deep learning-based techniques for tabular data \cite{rabbani25, rauf25}, these approaches often require a large amount of training data and do not typically support continual learning.

In contrast, clustering-based approaches do not require large labeled datasets. They are naturally applicable to unsupervised learning scenarios, especially with tabular data. An important advantage is that many such approaches enable single-pass learning, allowing models to process each instance only once without revisiting previous data. This eliminates the need for repeated training on massive datasets and makes clustering-based approaches especially suitable for dynamic, streaming, or resource-constrained environments \cite{zubarouglu21, oyewole23}.

Classical clustering algorithms such as Gaussian Mixture Model (GMM) \cite{mclachlan19} and $ k $-means \cite{lloyd82} have demonstrated adaptability and applicability in many fields. However, a major drawback of these algorithms is the requirement to pre-specify the number of clusters or partitions. To address this issue, growing self-organizing clustering algorithms such as GNG \cite{fritzke95} and SOINN-based algorithms \cite{furao06, shen08} have been proposed. GNG and SOINN-based algorithms adaptively generate topological networks (i.e., nodes and edges) representing data distributions. However, as these algorithms continually insert new nodes and edges to learn new information, there is a risk of forgetting previously learned information (i.e., catastrophic forgetting). More generally, this phenomenon is known as the plasticity-stability dilemma \cite{carpenter88}. For example, the Grow When Required (GWR) algorithm \cite{marsland02}, a GNG-based method, can mitigate the plasticity-stability dilemma by adding nodes whenever the current network does not sufficiently match the new instance. SOINN+ \cite{wiwatcharakoses20}, an ASOINN-based algorithm, can detect clusters of arbitrary shapes in noisy data streams without pre-defined parameters. However, as the number of network nodes increases, the computational cost of calculating thresholds for each node also increases, thereby reducing learning efficiency.

ART-based clustering algorithms have been explored as a way to mitigate catastrophic forgetting through vigilance control \cite{carpenter91b, vigdor07, wang19, da20, masuyama18, masuyama19a, masuyama19b, masuyamaFTCA}. Nevertheless, even state-of-the-art variants continue to rely on manually tuned vigilance values, making their performance highly dataset-dependent. Several extensions attempted to address this by introducing multi-vigilance mechanisms \cite{da19}, indirect specification \cite{da17,masuyama22a}, or adaptive scaling \cite{meng15}. These algorithms reduce sensitivity to some extent but their dependence on pre-specified parameters limits their practicality in real-world continual learning scenarios. This paper aims to address this gap by introducing automatic calculation of a similarity threshold (i.e., a vigilance parameter) and an edge deletion threshold, thereby advancing beyond prior ART-based clustering algorithms.

From an application standpoint, growing self-organizing clustering algorithms have been widely explored in continual learning \cite{parisi18, george19, hafez23}. They have also been applied in domains such as simultaneous localization and mapping \cite{chin18, toda24} and knowledge acquisition for robots \cite{hafez23b, dawood25}, demonstrating versatility across diverse fields. However, most existing algorithms still require careful parameter tuning to achieve high performance. This limitation underscores the significance of our proposed parameter-free algorithm, which enhances practicality and robustness in dynamic environments.

\section{Preliminary Knowledge}
\label{sec:preliminary}
This section presents preliminary knowledge for a similarity measure and a kernel density estimator used in the proposed CAE algorithm.

\subsection{Correntropy and Correntropy-induced Metric}
\label{sec:cimDefinition}
Correntropy \cite{liu07} provides a generalized similarity measure between two arbitrary data points $ \mathbf{x} = (x_{1},x_{2},\ldots,x_{d}) $ and $ \mathbf{y} = (y_{1},y_{2},\ldots,y_{d}) $ as follows:
\begin{equation}
	C(\mathbf{x}, \mathbf{y}) = \textbf{E} \left[ \kappa_{\sigma} \left( \mathbf{x}, \mathbf{y} \right) \right],
\end{equation}
where $ \textbf{E} \left[ \cdot \right] $ is the expectation operation, and $ \kappa_{\sigma} \left( \cdot \right) $ denotes a positive definite kernel with a bandwidth $ \sigma $. The correntropy is estimated as follows:
\begin{equation}
	\hat{C}(\mathbf{x}, \mathbf{y}, \sigma) = \frac{1}{d} \sum_{i=1}^{d} \kappa_{\sigma} \left( x_{i}, y_{i} \right).
	\label{eq:correntropy}
\end{equation}

In this paper, we use the following Gaussian kernel in the correntropy:
\begin{equation}
	\kappa_{\sigma} \left( x_{i}, y_{i} \right) = \exp \left[ - \frac{\left( x_{i} - y_{i} \right)^{2} }{2 \sigma^{2}} \right].
	\label{eq:gaussian}
\end{equation}

A nonlinear metric called CIM is derived from the correntropy \cite{liu07}. CIM quantifies the similarity between two data points $ \mathbf{x} $ and $ \mathbf{y} $ as follows:
\begin{equation}
	\mathrm{CIM}\left(\mathbf{x}, \mathbf{y}, \sigma \right) = \left[ 1 - \hat{C}(\mathbf{x}, \mathbf{y}, \sigma) \right]^{\frac{1}{2}},
	\label{eq:defcim}
\end{equation}
here, since the Gaussian kernel in (\ref{eq:gaussian}) does not have the coefficient $ \frac{1}{\sqrt{2\pi}\sigma} $, the range of CIM is limited to $ \left[0,1 \right] $.

In general, the Euclidean distance suffers from the curse of dimensionality. However, CIM reduces this drawback since the correntropy calculates the similarity between two data points by using a kernel function. Moreover, it has also been shown that CIM with the Gaussian kernel has a high outlier rejection ability \cite{liu07}.

\subsection{Kernel Density Estimator}
\label{sec:kde}
A similarity measurement between an instance and a node has a large impact on the performance of clustering algorithms. CAE uses the CIM as a similarity measurement. As defined in (\ref{eq:defcim}), the state of the CIM is controlled by a kernel bandwidth $ \sigma $ which is a data-dependent parameter. In CAE, the initial state of $ \sigma $ is defined by training instances.

When a new node $ \mathbf{y}_{K+1} $ is generated from $ \mathbf{x}_{n} $, a kernel bandwidth $ \sigma_{K+1} $ is estimated from the past $ \lambda $ instances, i.e., $ \left( \mathbf{x}_{n-1},\mathbf{x}_{n-2},\ldots, \mathbf{x}_{n-\lambda} \right) $, by using a kernel density estimation with the Gaussian kernel, which is defined as follows:
\begin{equation}
	\mathbf{\Sigma} = \left( \frac{4}{2+d} \right)^{\frac{1}{4+d}}  \mathbf{\Gamma}  \lambda^{-\frac{1}{4+d}},
	\label{eq:SIGMA}
\end{equation}
where $ \mathbf{\Gamma} $ denotes a rescale operator which is defined by a standard deviation of each attribute value among the $ \lambda $ instances. Here, $ \mathbf{\Sigma} $ contains the bandwidth of each attribute. In this paper, the median of $ \mathbf{\Sigma} $ is selected as a representative kernel bandwidth for the new node $ \mathbf{y}_{K+1} $, i.e.,
\begin{equation}
	\sigma_{K+1} = \mathrm{median} \left( \mathbf{\Sigma} \right).
	\label{eq:sigma}
\end{equation}

Equation (\ref{eq:SIGMA}) is known as the Silverman's rule \cite{silverman18}.

\section{Proposed Algorithm}
\label{sec:proposedAlgorithm}

\subsection{Overview}
\label{sec:overview}
The proposed CAE algorithm is an extension of CAEA \cite{masuyama22a}, which represents the state-of-the-art in ART-based topological clustering. CAEA requires two data-dependent parameters to be specified in advance: (i) the number of nodes $\lambda$ used to calculate the similarity threshold, and (ii) the threshold $a_{\text{max}}$ for the edge deletion process. In contrast to CAEA, the CAE algorithm estimates these parameters during the learning process. Specifically, the sufficient number of nodes $\lambda$ for calculating the similarity threshold is estimated by a DPP-based criterion incorporating CIM, and the node deletion threshold $a_{\text{max}}$ is determined based on the age of each edge.

\begin{figure}[htbp]
	\centering
	\includegraphics[width=3.5in]{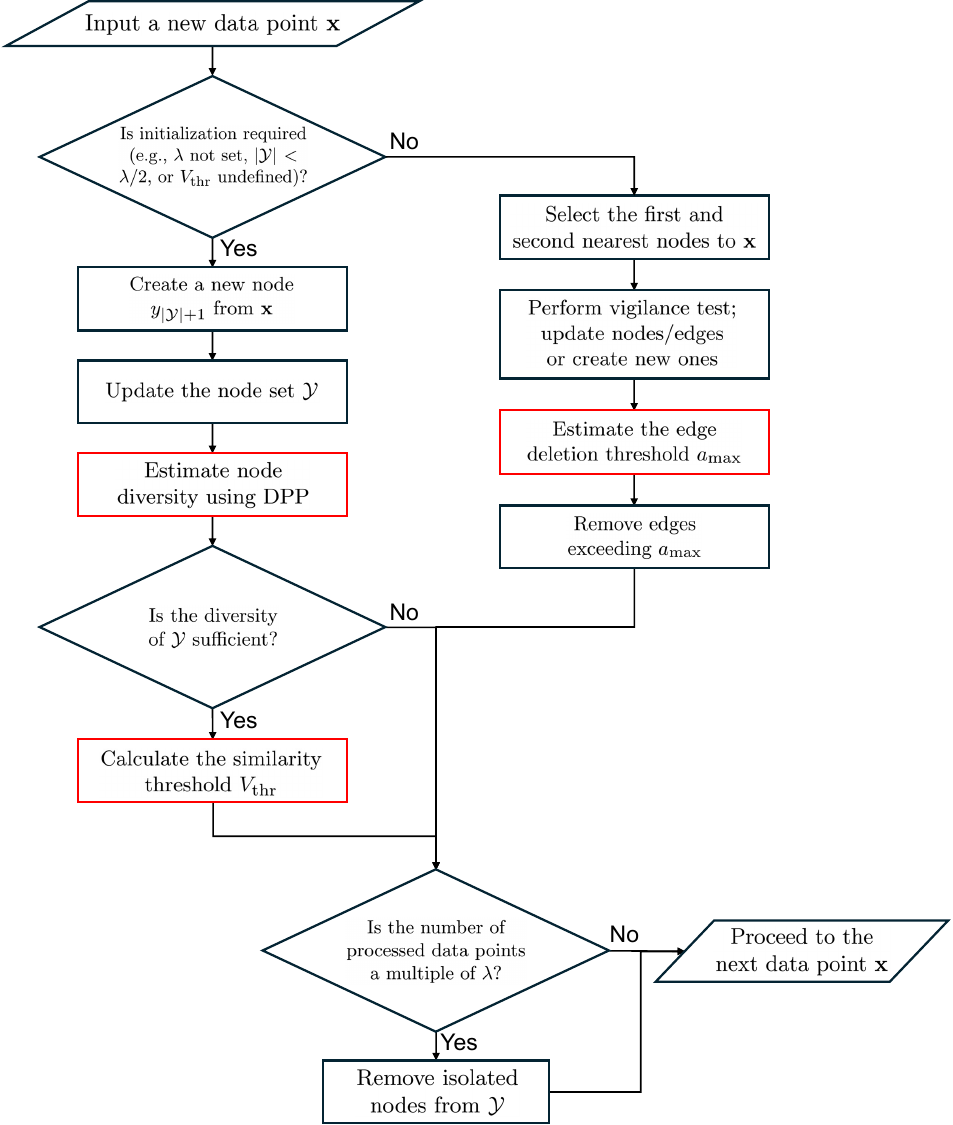}
	\vspace{3mm}
	\caption{Flowchart of the CAE learning procedure. The procedure is identical to the CAEA learning process except for the parameter estimation steps highlighted in red.}
	\label{fig:flowchart_cae}
\end{figure}

Table \ref{tab:notations} summarizes the main notations used in this paper. The following subsections describe the learning procedures of CAE step by step. Algorithm \ref{alg:pseudocodeCAE} presents the overall learning procedure. To complement the algorithmic description, Fig.\ref{fig:flowchart_cae} illustrates the flowchart of the CAE learning process, in which CAE-specific parameter estimation steps are visually distinguished from the shared processes of CAEA.

\begin{table}[htbp]
	\centering
	\caption{Summary of notations}
	\renewcommand{\arraystretch}{1.2}
	\label{tab:notations}
	\begin{tabular}{ll}
		\hline\hline
		Notation               & Description  \\
		\hline
		$\mathbf{x}$           & $d$-dimensional data point  \\
		$\mathcal{X} $         & set of data points ($\mathbf{x} \in \mathcal{X}$) \\
		$\mathbf{y}_{k}$       & $k$-th node \\
		$\mathcal{Y}$          & set of nodes ($\mathbf{y}_{k} \in \mathcal{Y}$) \\
		$|\mathcal{Y}|$        & number of nodes in $\mathcal{Y}$  \\
		$\mathbf{y}_{s_{1}}$   & 1st winner node \\
		$\mathbf{y}_{s_{2}}$   & 2nd winner node  \\
		$\sigma$               & bandwidth of a kernel function  \\
		$\mathcal{S}$          & set of bandwidths of a kernel function \\
		$\mathrm{CIM}\left(\mathbf{x}, \mathbf{y}, \sigma \right)$ & A similarity measure between two data points $ \mathbf{x} $ and $ \mathbf{y} $ \\
		$ \mathcal{N}_{k} $    & set of neighbor nodes of node $ \mathbf{y}_{k} $   \\
		$\lambda$              & number of active nodes  \\
		$\mathcal{A}$          & set of active nodes \\
		$D$          		   & diversity of active nodes \\
		$V_{s_{1}}$            & CIM value between a data point $\mathbf{x}$ and $\mathbf{y}_{s_{1}}$ \\
		$V_{s_{2}}$            & CIM value between a data point $\mathbf{x}$ and $\mathbf{y}_{s_{2}}$ \\
		$\mathbf{R}$           & matrix of pairwise similarities \\
		$V_{\text{threshold}}$ & similarity threshold (a vigilance parameter)   \\
		$ M_{k} $              & number of winning counts of $ \mathbf{y}_{k} $ \\
		$ \mathcal{M} $        & set of winning counts $ M_{k} $ ($ M_{k} \in \mathcal{M}$)  \\
		$ a(\mathbf{y}_{k},\mathbf{y}_{l}) $   & age of an edge between nodes $\mathbf{y}_{k}$ and $\mathbf{y}_{l} \in \mathcal{Y} \setminus \mathbf{y}_{k}$ \\
		$ \mathcal{E} $   	   & set of ages of edges ($a(\mathbf{y}_{k},\mathbf{y}_{l}) \in \mathcal{E}$) \\
		$\alpha_{\text{del}}$  & set of ages of deleted edges  \\
		$a_{\text{max}}$       & edge deletion threshold  \\
		\hline\hline
	\end{tabular}
\end{table}

\begin{algorithm*}
  \caption{Learning procedure of CAE}\label{alg:pseudocodeCAE}
  \begin{algorithmic}[1]
  \Require
  \Statex A set of data points $\mathcal{X}$.
  \Ensure
  \Statex A set of nodes $\mathcal{Y}$,
  \Statex A set of ages of edges $\mathcal{E}$.
  
  \While{existing data points to be trained}
      \State Input a data point $\mathbf{x}$ ($\mathbf{x} \in \mathcal{X}$)
      \If{the number of active nodes $\lambda$ is not defined \textbf{or} $|\mathcal{Y}| < \lambda/2$ \textbf{or} $V_{\text{threshold}}$ is not calculated} 
          \State Create a node as $\mathbf{y}_{|\mathcal{Y}|+1} = \mathbf{x}$
          \State Update a set of nodes: $\mathcal{Y} \gets \mathcal{Y} \cup \{ \mathbf{y}_{|\mathcal{Y}|+1} \}$
          \State Estimate the diversity of nodes \Comment{Algorithm \ref{alg:ActiveNode}}
          \If{the diversity of nodes is sufficient (i.e., $ D <  1.0\mathrm{e}{-6}$)}
              \State Calculate a similarity threshold $V_{\text{threshold}}$
          \EndIf
      \Else
          \State Select the 1st and 2nd nearest nodes from $\mathbf{x}$ (i.e., $\mathbf{y}_{s_{1}}$ and $\mathbf{y}_{s_{2}}$) based on CIM
          \State Perform vigilance test, and create/update nodes and edges \Comment{Algorithm \ref{alg:UpdateNodesEdges}}
          \State Estimate an edge deletion threshold $ a_{\text{max}} $ \Comment{Algorithm \ref{alg:EstEdgeThreshold}}
          \State Delete edges using $ a_{\text{max}} $ \Comment{Algorithm \ref{alg:DeleteEdges}}
      \EndIf
      \If{the number of presented data points is a multiple of $\lambda$}
          \State Delete isolated nodes
      \EndIf
  \EndWhile
  \end{algorithmic}
\end{algorithm*}

\subsection{Estimation of Diversity of Nodes}
\label{sec:estDiversity}

In CAE, a similarity threshold is defined by pairwise similarities among nodes (i.e., $\mathcal{Y}$) (see Section \ref{sec:calSimilarity} for details). Therefore, the diversity of nodes for calculating the similarity threshold is important to obtain an appropriate threshold value, which leads to good clustering performance.

The diversity $D$ of the active node set $\mathcal{A}$ is estimated by a DPP-based criterion \cite{kulesza12,holder20} incorporating CIM as follows:
\begin{equation}
	D =  \det( \mathbf{R} ),
	\label{eq:div}
\end{equation}
where
\begin{equation}
	\mathbf{R} =  \left[ \exp\left(1 - \mathrm{CIM}(\mathbf{y}_{i}, \mathbf{y}_{j}, \sigma)\right) \right]_{1 \leq i, j \leq |\mathcal{A}|}.
	\label{eq:SimilarityMat}
\end{equation}
here, $\det( \mathbf{R} )$ is the determinant of the matrix $\mathbf{R}$, and $\mathbf{R}$ is a matrix of pairwise similarities between nodes in $\mathcal{A}$. A bandwidth $\sigma$ for CIM is calculated from $ \mathbf{H} $ in (\ref{eq:SIGMA}) by using the node set $\mathcal{A}$. As in (\ref{eq:SIGMA}), $ \mathbf{H} $ contains the bandwidth of a kernel function in CIM. In this paper, the median of $ \mathbf{H} $ is used as the bandwidth of the Gaussian kernel in CIM, i.e.,
\begin{equation}
	\sigma = \mathrm{median} \left( \mathbf{H} \right).
	\label{eq:sigma}
\end{equation}

In general, the diversity $D = 0$ means that the node set $\mathcal{A}$ is not diverse while $D > 0$ means $\mathcal{A}$ is diverse. In other words, the value of $D$ becomes close to zero when a new node is created around the existing nodes.

\begin{algorithm}[htbp]
  \caption{Estimate diversity of nodes}\label{alg:ActiveNode}
  \begin{algorithmic}[1]
  \Require
  \Statex A data point $\mathbf{x}$,
  \Statex A set of active nodes $\mathcal{A}$,
  \Statex The number of active nodes $\lambda$.
  \Ensure
  \Statex Diversity $D$,
  \Statex The number of active nodes $\lambda$.
  
  \State Calculate a pairwise similarity matrix $\mathbf{R}$ \Comment{(\ref{eq:SimilarityMat})}
  \State Calculate the diversity as $D = \det( \mathbf{R} )$ \Comment{(\ref{eq:div})}
  \If{$ D <  1.0\mathrm{e}{-6}$}
      \State $\lambda \gets 2|\mathcal{A}|$
  \Else
      \State $\lambda \gets \infty$
  \EndIf
  \end{algorithmic}
\end{algorithm}

Algorithm \ref{alg:ActiveNode} summarizes the estimation process of the diversity of nodes. In CAE, the value of $\lambda$ is set as twice the number of nodes (i.e., $2|\mathcal{A}|$) when the diversity $D$ satisfies $D < 1.0\mathrm{e}{-6}$. If the number of nodes becomes smaller than $\lambda/2$ after a node deletion process, $\lambda$ is calculated again in line 6 of Algorithm \ref{alg:pseudocodeCAE} using Algorithm \ref{alg:ActiveNode}.

As shown in lines 3-4 of Algorithm \ref{alg:pseudocodeCAE}, the first $\lambda/2$ data points (i.e., $\{ \mathbf{x}_{1}, \mathbf{x}_{2},\ldots \mathbf{x}_{\lambda/2} \}$) directly become nodes, i.e., $ \mathcal{Y} = \{\mathbf{y}_{1}, \mathbf{y}_{2}, \ldots, \mathbf{y}_{\lambda/2}\} $ where $ \mathbf{y}_{k} = \mathbf{x}_{k} $ $ ( k = 1, 2, \ldots, \lambda/2 ) $. In addition, the bandwidth for the Gaussian kernel in CIM is assigned to each node, i.e., $ \mathcal{S} = \{ \sigma_{1},\sigma_{2},\ldots, \sigma_{\lambda/2}\} $ where $ \sigma_{1} = \sigma_{2} = \cdots = \sigma_{\lambda/2} $. 

The value of $\lambda$ is automatically updated by Algorithm \ref{alg:ActiveNode} in the proposed CAE algorithm. In an active node set $\mathcal{A}$, $\lambda$ nodes are stored. When a new node is added to $\mathcal{A}$, an old node is removed to maintain the active node set size as $\lambda$. The addition of a new node and the removal of an old node are explained later.

\subsection{Calculation of Similarity Threshold}
\label{sec:calSimilarity}
The similarity threshold $ V_{\text{threshold}} $ is calculated by the average of the minimum pairwise CIM values in the active node set $ \mathcal{A}$ as follows:
\begin{equation}
	V_{\text{threshold}} = \frac{1}{\lambda} \sum_{\mathbf{y}_{i} \in \mathcal{A}} \min_{\mathbf{y}_{j} \in \mathcal{A} \setminus \mathbf{y}_{i}} \left[\mathrm{CIM}\left(\mathbf{y}_{i}, \mathbf{y}_{j}, \mathrm{mean}(\mathcal{S})\right)\right],
	\label{eq:pairwiseCIM}
\end{equation}
where $ \mathcal{S} $ is a set of bandwidths of the Gaussian kernel in CIM for $ \mathcal{A} $. The bandwidth of each node in $ \mathcal{A} $ is calculated by using (\ref{eq:SIGMA}) and (\ref{eq:sigma}) when a new node is created.

\subsection{Selection of Winner Nodes}
\label{sec:winnerNodes}

During the learning process of CAE, every time a data point $ \mathbf{x} $ is given, two nodes that have a similar state to $ \mathbf{x} $ are selected from $\mathcal{Y}$, namely the 1st winner node $ \mathbf{y}_{s_{1}} $ and the 2nd winner node $ \mathbf{y}_{s_{2}} $. The winner nodes are determined based on the value of CIM in line 11 of Algorithm \ref{alg:pseudocodeCAE} as follows:
\begin{equation}
	\hspace{-4.5mm} s_{1} = \arg\min_{i \in \lambda} \left[ \mathrm{CIM}\left(\mathbf{x}, \mathbf{y}_{i}, \mathrm{mean}(\mathcal{S}) \right) \right],
	\label{eq:winnerCIM1}
\end{equation}
\begin{equation}
	s_{2} = \arg\min_{\substack{i \in \lambda \\ i \neq s_1}} \left[ \mathrm{CIM}\left(\mathbf{x}, \mathbf{y}_{i}, \mathrm{mean}(\mathcal{S}) \right) \right],
	\label{eq:winnerCIM2}
\end{equation}

\noindent where $ s_{1} $ and $ s_{2} $ denote the indexes of the 1st and 2nd winner nodes, respectively. $ \mathcal{S} = \{ \sigma_{1},\sigma_{2},\ldots,\sigma_{|\mathcal{Y}|} \} $ is a set of bandwidths of the Gaussian kernel in CIM corresponding to a set of nodes $ \mathcal{Y} $.

Note that the 1st winner node $\mathbf{y}_{s_{1}}$ becomes a new active node, and the oldest node in the active node set $\mathcal{A}$ (i.e., $\lambda$ nodes in $\mathcal{Y}$) is replaced by the new one.

\subsection{Vigilance Test}
\label{sec:vigilanceTest}
Similarities between the data point $ \mathbf{x} $ and each of the 1st and 2nd winner nodes are defined in line 11 of Algorithm \ref{alg:pseudocodeCAE} as inputs for Algorithm \ref{alg:UpdateNodesEdges} as follows:
\begin{equation}
	V_{s_{1}} =  \mathrm{CIM}\left(\mathbf{x}, \mathbf{y}_{s_{1}}, \mathrm{mean}(\mathcal{S}) \right),
	\label{eq:cim1}
\end{equation}
\vspace{-8pt}
\begin{equation}
	V_{s_{2}} = \mathrm{CIM}\left(\mathbf{x}, \mathbf{y}_{s_{2}}, \mathrm{mean}(\mathcal{S}) \right).
	\label{eq:cim2}
\end{equation}

The vigilance test classifies the relationship between the data point $\mathbf{x}$ and the two winner nodes into three cases by using the similarity threshold $ V_{\text{threshold}} $, i.e.,

\begin{itemize}
	
	\item Case I \\
	\indent The similarity between the data point $ \mathbf{x} $ and the 1st winner node $ \mathbf{y}_{s_{1}} $ is larger (i.e., less similar) than $ V_{\text{threshold}} $, namely:
	\begin{equation}
		V_{\text{threshold}} < V_{s_{1}} \leq V_{s_{2}}.
		\label{eq:case1}
	\end{equation}
	
	\vspace{0.8mm}
	
	\item Case I\hspace{-.1em}I \\
	\indent The similarity between $ \mathbf{x} $ and the 1st winner node $ \mathbf{y}_{s_{1}} $ is smaller (i.e., more similar) than $ V_{\text{threshold}} $, and the similarity between $ \mathbf{x} $ and the 2nd winner node $ \mathbf{y}_{s_{2}} $ is larger (i.e., less similar) than $ V_{\text{threshold}} $, namely:
	\begin{equation}
		V_{s_{1}} \leq V_{\text{threshold}} < V_{s_{2}}.
		\label{eq:case2}
	\end{equation}
	
	\vspace{0.8mm}
	
	\item Case I\hspace{-.1em}I\hspace{-.1em}I \\
	\indent The similarities between $ \mathbf{x} $ and the 1st and 2nd winner nodes  (i.e., $ \mathbf{y}_{s_{1}} $ and $ \mathbf{y}_{s_{2}} $) are both smaller (i.e., more similar) than $ V_{\text{threshold}} $, namely:
	\begin{equation}
		V_{s_{1}} \leq V_{s_{2}} \leq V_{\text{threshold}}.
		\label{eq:case3}
	\end{equation}
	
\end{itemize}

\subsection{Creation/Update of Nodes and Edges}
\label{sec:update}
Depending on the result of the vigilance test, a different operation is performed.

If the data point $ \mathbf{x} $ is classified as Case I by the vigilance test (i.e., (\ref{eq:case1}) is satisfied), a new node is created as $\mathbf{y}_{|\mathcal{Y}|+1} = \mathbf{x}$, and updated a set of nodes as $ \mathcal{Y} \leftarrow \mathcal{Y} \cup \{ \mathbf{y}_{|\mathcal{Y}|+1} \} $. Here, the node $\mathbf{y}_{|\mathcal{Y}|+1}$ becomes a new active node, and the oldest node in the active node set $\mathcal{A}$ (i.e., $\lambda$ nodes in $\mathcal{Y}$) is replaced by the new one. In addition, a bandwidth $ \sigma_{|\mathcal{Y}|+1} $ for $ \mathbf{y}_{|\mathcal{Y}|+1} $ is calculated by (\ref{eq:SIGMA}) and (\ref{eq:sigma}) with the active node set $\mathcal{A}$, and the winning counts of $ \mathbf{y}_{|\mathcal{Y}|+1} $ is initialized as $ M_{|\mathcal{Y}|+1} = 1 $.

If the data point $ \mathbf{x} $ is classified as Case I\hspace{-1pt}I by the vigilance test (i.e., (\ref{eq:case2}) is satisfied), first, the winning counts of $ \mathbf{y}_{s_{1}} $ is updated as follows:
\begin{equation}
	M_{s_{1}} \leftarrow M_{s_{1}} + 1.
	\label{eq:countNode}
\end{equation}

Then, $ \mathbf{y}_{s_{1}} $ is updated as follows:
\begin{equation}
	\mathbf{y}_{s_{1}} \leftarrow \mathbf{y}_{s_{1}} + \frac{1}{M_{s_{1}}} \left( \mathbf{x} - \mathbf{y}_{s_{1}} \right),
	\label{eq:updateNodeWeight1}
\end{equation}
where, the node $\mathbf{y}_{s_{1}}$ becomes a new active node, and the oldest node in the active node set $\mathcal{A}$ (i.e., $\lambda$ nodes in $\mathcal{Y}$) is replaced by the new one.

When updating the node, the difference between $\mathbf{x}$ and $\mathbf{y}$ is divided by $M_{s_{1}}$, so the update step becomes smaller as more samples are accumulated. As a result, the node position becomes stable in dense regions, allowing it to retain the accumulated local information rather than being strongly affected by a single new sample.

The age of each edge connected to the 1st winner node $ \mathbf{y}_{k_{1}} $ is also updated as follows:
\begin{equation}
	a{(\mathbf{y}_{s_{1}},\mathbf{y}_{k})} \leftarrow a{(\mathbf{y}_{s_{1}},\mathbf{y}_{k})} + 1 \quad (\mathbf{y}_{k} \in \mathcal{N}_{s_{1}}),
	\label{eq:edd_age}
\end{equation}
where $ \mathcal{N}_{s_{1}} $ is a set of all neighbor nodes of the node $ \mathbf{y}_{s_{1}} $.

If the data point $ {\mathbf{x}} $ is classified as Case I\hspace{-1pt}I\hspace{-1pt}I by the vigilance test (i.e., (\ref{eq:case3}) is satisfied), the same operations as Case I\hspace{-1pt}I (i.e., (\ref{eq:countNode})-(\ref{eq:edd_age})) are performed. In addition, if there is an edge between $ \mathbf{y}_{s_{1}} $ and $ \mathbf{y}_{s_{2}} $, an age of the edge is reset as follows:
\begin{equation}
	a{(\mathbf{y}_{s_{1}},\mathbf{y}_{s_{2}})} \leftarrow 1.
	\label{eq:ageInit}
\end{equation} 

In the case that there is no edge between $ \mathbf{y}_{s_{1}} $ and $ \mathbf{y}_{s_{2}} $, a new edge is defined with an age of the edge by (\ref{eq:ageInit}).

After updated the edge information, the neighbor nodes of $ \mathbf{y}_{s_{1}} $ are updated as follows:
\begin{equation}
	{\mathbf{y}}_{k} \gets {\mathbf{y}}_{k} + \frac{1}{10 M_{k}} \left({\mathbf{x}}-{\mathbf{y}}_{k}\right) \quad \left(\mathbf{y}_{k} \in \mathcal{N}_{s_{1}}\right).
	\label{eq:updateNodeWeight2}
\end{equation}

Apart from the above operations in Cases I-I\hspace{-1pt}I\hspace{-1pt}I, the nodes with no edges are deleted (and removed from the active node set $\mathcal{A}$) every $ \lambda $ data points for the noise reduction purpose (i.e., the node deletion interval is the presentation of $ \lambda $ data points), which is performed in lines 13-14 of Algorithm \ref{alg:pseudocodeCAE}.

With respect to the active node set $\mathcal{A}$, its update rules are summarized as follows. In Case I, a new node is directly created by the data point $ {\mathbf{x}} $ and added to $\mathcal{A}$. In Case I\hspace{-1pt}I and Case I\hspace{-1pt}I\hspace{-1pt}I, the updated winner node in (\ref{eq:updateNodeWeight1}) is added to $\mathcal{A}$. In all cases, the oldest active node is removed from $\mathcal{A}$. Then, in lines 13-14 of Algorithm \ref{alg:pseudocodeCAE}, all active nodes with no edges are removed. After this removal procedure, the number of active nodes can be smaller than $\lambda$.

\begin{algorithm}[htbp]
  \caption{Update nodes and edges}\label{alg:UpdateNodesEdges}
  \begin{algorithmic}[1] 
  \Require
    \Statex A data point $\mathbf{x}$,
    \Statex A set of nodes $\mathcal{Y}$,
    \Statex A set of active nodes $\mathcal{A}$,
    \Statex A set of bandwidths of a kernel function $\mathcal{S}$,
    \Statex A set of ages of edges $\mathcal{E}$,
    \Statex A set of winning counts $\mathcal{M}$,
    \Statex The 1st and 2nd winner nodes $\mathbf{y}_{s_{1}}, \mathbf{y}_{s_{2}}$,
    \Statex The similarities $V_{s_{1}}, V_{s_{2}}$ between $\mathbf{x}$ and $\mathbf{y}_{s_{1}}, \mathbf{y}_{s_{2}}$,
    \Statex The similarity threshold $V_{\text{threshold}}$.

    \Ensure
    \Statex A set of nodes $\mathcal{Y}$,
    \Statex A set of active nodes $\mathcal{A}$,
    \Statex A set of bandwidths of a kernel function $\mathcal{S}$,
    \Statex A set of ages of edges $\mathcal{E}$,
    \Statex A set of winning counts $\mathcal{M}$.
  
  \If{$V_{\text{threshold}} < V_{s_{1}}$} \Comment{Case I}
      \State $\mathbf{y}_{|\mathcal{Y}|+1} \gets \mathbf{x}$
      \State $\mathcal{Y} \gets \mathcal{Y} \cup \{ \mathbf{y}_{|\mathcal{Y}|+1} \}$
      \State $M_{|\mathcal{Y}|+1} \gets 1$, $\mathcal{M} \gets \mathcal{M} \cup \{ M_{|\mathcal{Y}|+1} \}$
      \State Calculate $\sigma_{|\mathcal{Y}|+1}$ using (\ref{eq:SIGMA}) and (\ref{eq:sigma}) with active node set $\mathcal{A}$
      \State $\mathcal{S} \gets \mathcal{S} \cup \{ \sigma_{|\mathcal{Y}|+1} \}$
      \State Update the active node set $\mathcal{A}$
  \Else \Comment{Case II}
      \State $M_{s_1} \gets M_{s_1} + 1$ \Comment{(\ref{eq:countNode})}
      \State $\mathbf{y}_{s_{1}} \gets \mathbf{y}_{s_{1}} + \frac{1}{M_{s_1}}( \mathbf{x} - \mathbf{y}_{s_{1}})$ \Comment{(\ref{eq:updateNodeWeight1})}
      \State Update the active node set $\mathcal{A}$
      \For{$\mathbf{y}_{k} \in \mathcal{N}_{s_{1}}$}
          \State $a{(\mathbf{y}_{s_{1}},\mathbf{y}_{k})} \gets a{(\mathbf{y}_{s_{1}},\mathbf{y}_{k})} + 1$ \Comment{(\ref{eq:edd_age})}
      \EndFor
      \If{$V_{s_{2}} \leq V_{\text{threshold}}$} \Comment{Case III}
          \State $a(\mathbf{y}_{s_{1}},\mathbf{y}_{s_{2}}) \gets 1$ \Comment{(\ref{eq:ageInit})}
          \For{$\mathbf{y}_{k} \in \mathcal{N}_{s_{1}}$}
              \State $\mathbf{y}_{k} \gets \mathbf{y}_{k} + \frac{1}{10M_{k}}( \mathbf{x}- \mathbf{y}_{k})$ \Comment{(\ref{eq:updateNodeWeight2})}
          \EndFor
      \EndIf
  \EndIf
  \end{algorithmic}
\end{algorithm}

Algorithm \ref{alg:UpdateNodesEdges} summarizes the creation/update processes for nodes and edges.

\subsection{Estimation of Edge Deletion Threshold}
\label{sec:EstEdgeThreshold}

CAE estimates an edge deletion threshold based on the ages of the current edges and the deleted edges, which is inspired by the edge deletion mechanism of SOINN+ \cite{wiwatcharakoses20}.

The edge deletion threshold $a_{\text{max}}$ is defined as follows:
\begin{equation}
	a_{\text{max}} = \bar{\alpha}_{\text{del}} \frac{|\alpha_{\text{del}}|}{|\alpha_{\text{del}}|+|\alpha|} + a_{\text{thr}} \left(1-\frac{|\alpha_{\text{del}}|}{|\alpha_{\text{del}}|+|\alpha|} \right),
	\label{eq:CulA_max}
\end{equation}
where $\alpha_{\text{del}}$ is the set of ages of all the deleted edges during the learning process, $|\alpha_{\text{del}}|$ is the number of elements in $\alpha_{\text{del}}$. $\bar{\alpha}_{\text{del}}$ is the arithmetic mean of $\alpha_{\text{del}}$, $\alpha$ is a set of ages of edges that connect to $\mathbf{y}_{s_{1}}$ ($\alpha \subset \mathcal{E}$), and $|\alpha|$ is the number of elements in $\alpha$. The coefficient $a_{\text{thr}}$ is defined as follows:
\begin{equation}
	a_{\text{thr}} =\alpha_{0.75} + \text{IQR}(\alpha),
	\label{eq:outliers}
\end{equation}
where $\alpha_{0.75}$ is the 75th percentile of elements in $\alpha$, and $\text{IQR}(\alpha)$ is the interquartile range.

The edge deletion threshold $a_{\text{max}}$ is updated each time the age of an edge increases. Algorithm \ref{alg:EstEdgeThreshold} summarizes the estimation process of the edge deletion threshold $a_{\text{max}}$.

\begin{algorithm}[htbp]
  \caption{Estimate edge deletion threshold}\label{alg:EstEdgeThreshold}
  \begin{algorithmic}[1]
  \Require
  \Statex The 1st winner node $\mathbf{y}_{s_{1}}$,
  \Statex A set of ages of edges $\mathcal{E}$,
  \Statex A set of ages of deleted edges $\alpha_{\text{del}}$.
  \Ensure
  \Statex The edge deletion threshold $a_{\text{max}}$.
  
  \State $\alpha \gets$ a set of ages of edges which connect to $\mathbf{y}_{s_{1}}$
  \State $\alpha_{0.75} \gets$ the 75th percentile of elements in $\alpha$
  \State $a_{\text{thr}} \gets \alpha_{0.75} + \text{IQR}(\alpha)$ \Comment{(\ref{eq:outliers})}
  \State $a_{\text{max}} \gets \bar{\alpha}_{\text{del}} \frac{|\alpha_{\text{del}}|}{|\alpha_{\text{del}}|+|\alpha|} + a_{\text{thr}}  \left(1 -  \frac{|\alpha_{\text{del}}|}{|\alpha_{\text{del}}|+|\alpha|} \right)$ \Comment{(\ref{eq:CulA_max})}
  \end{algorithmic}
\end{algorithm}

The differences between the above method and SOINN+ are summarized as follows. 

\begin{itemize}
  \item The coefficient of $\text{IQR}(\alpha)$ in (\ref{eq:outliers}) is set to 2 in SOINN+, while it is set to 1 in CAE. The motivation for this change is that an ART-based topological clustering algorithm tends to create fewer edges than a SOINN-based algorithm, especially in the early stage of learning. Therefore, the coefficient of $\text{IQR}(\alpha)$ in (\ref{eq:outliers}) is set as 1 to weaken the influence of $a_{\text{thr}}$ while emphasizing the information of deleted edges. 
	
	\vspace{1mm}
	\item SOINN+ considers all edges that can reach $\mathbf{y}_{s_{1}}$, while CAE only considers edges which connect to $\mathbf{y}_{s_{1}}$. The motivation for this change is to reduce computation time. SOINN+ searches all the nodes and edges connected to $\mathbf{y}_{s_{1}}$ whenever $a_{\text{max}}$ is estimated. In contrast, CAE only uses edges connected to $\mathbf{y}_{s_{1}}$ (i.e., line 1 in Algorithm \ref{alg:EstEdgeThreshold}). Thus, the computation speed of CAE is faster than SOINN+.
\end{itemize}

\subsection{Deletion of Edges}
\label{sec:deleteEdge}

If there is an edge whose age is greater than the edge deletion threshold $a_{\text{max}}$, the edge is deleted and the set of ages of deleted edges $\alpha_{\text{del}}$ is updated.

Algorithm \ref{alg:DeleteEdges} summarizes the edge deletion process.

\begin{algorithm}[htbp]
  \caption{Delete edges}\label{alg:DeleteEdges}
  \begin{algorithmic}[1]
  \Require
  \Statex The edge deletion threshold $a_{\text{max}}$,
  \Statex The 1st winner node $\mathbf{y}_{s_{1}}$,
  \Statex A set of neighbors of the 1st winner node $\mathcal{N}_{s_{1}}$,
  \Statex A set of ages of deleted edges $\alpha_{\text{del}}$.
  \Ensure
  \Statex A set of ages of edges $\mathcal{E}$,
  \Statex A set of ages of deleted edges $\alpha_{\text{del}}$.
  
  \For{$\mathbf{y}_{k} \in \mathcal{N}_{s_{1}}$}
      \If{$a(\mathbf{y}_{s_{1}},\mathbf{y}_{k}) > a_{\text{max}}$}
          \State $\alpha_{\text{del}} \gets \alpha_{\text{del}} \cup \{ a(\mathbf{y}_{s_{1}},\mathbf{y}_{k}) \}$
          \State Delete the edge between $\mathbf{y}_{s_{1}}$ and $\mathbf{y}_{k}$
      \EndIf
  \EndFor
  \end{algorithmic}
\end{algorithm}

\section{Simulation Experiments}
\label{sec:experiment}
In this section, we evaluate the performance of CAE from multiple perspectives and compare it to state-of-the-art clustering algorithms. First, we qualitatively and quantitatively assess clustering performance using a two-dimensional synthetic dataset in both stationary and non-stationary environments. Second, we quantitatively evaluate clustering performance across a broad range of real-world datasets. Third, we investigate continual learning ability by measuring metrics such as Incremental Average (IA) and Backward Transfer (BWT), which directly quantify knowledge retention and forgetting on real-world datasets. Fourth, we analyze the validity of the number of nodes automatically estimated by the DPP-based criterion incorporating CIM. Finally, we examine the computational complexity of CAE.

All experiments are conducted using MATLAB 2020a on a 2.2GHz Xeon Gold 6238R processor with 768GB of RAM under Windows OS.

\subsection{Compared Algorithms}
\label{sec:algorithm_2d}
We compare six algorithms, namely AutoCloud \cite{bezerra20}, ASOINN \cite{shen08}, SOINN+ \cite{wiwatcharakoses20}, TCA \cite{masuyama19b}, CAEA \cite{masuyama22a}, and CAE. Note that AutoCloud organizes clusters by using fuzzy concepts, i.e., the algorithm allows each data point to belong to multiple clusters simultaneously. ASOINN and SOINN+ are GNG-based algorithms while TCA, CAEA, and CAE are ART-based algorithms. The source code of AutoCloud\footnote{\url{https://github.com/BrunuXCosta/AutoCloud}}, ASOINN\footnote{\url{https://cs.nju.edu.cn/rinc/Soinn.html}}, SOINN+\footnote{\url{https://osf.io/6dqu9/}}, TCA\footnote{\url{https://github.com/Masuyama-lab/TCA}}, and CAEA\footnote{\url{https://github.com/Masuyama-lab/HCAEA}} is provided by the authors of the original papers. The source code of CAE is available on GitHub\footnote{\url{https://github.com/Masuyama-lab/CAE}}.

SOINN+ and CAE require no parameter tuning, whereas AutoCloud, ASOINN, TCA, and CAEA each include parameters that influence their clustering performance. For these algorithms, we determine parameter values using grid search. Table \ref{tab:paramAlgorithms} summarizes the parameter ranges used for grid search in each algorithm, which are based on the experimental settings reported in the original publications. As in the original paper \cite{bezerra20}, we use the Euclidean distance as the distance function for AutoCloud.

\begin{table*}[htbp]
	\caption{Range of grid search for AutoCloud, ASOINN, TCA, and CAEA}
	\label{tab:paramAlgorithms}
	\footnotesize
	\centering
	\renewcommand{\arraystretch}{1.2}
  \resizebox{\textwidth}{!}{
	\begin{tabular}{l|ll|l}
		\hline\hline
		Algorithm & \multicolumn{2}{l|}{Parameter} & Grid Range  \\
		\hline
		AutoCloud    & $m$ &: Value for determining a threshold.  & \{1, 2, 3\} \\                                                                           
		\hline
		ASOINN    & $ \lambda $                   & : Deletion cycle of isolated nodes. & \{50, 100, 150, 200, 250, 300, 350, 400, 450, 500\}   \\
				& $ a_{\mathrm{max}} $          & : Maximum age of a node.             & \{10, 20, 30, 40, 50, 60, 70, 80, 90, 100\}           \\
		\hline
		TCA       & $ \lambda $                   & : Topology construction cycle.       & \{50, 100, 150, 200, 250, 300, 350, 400, 450, 500\}   \\
				& $ \sigma_{\mathrm{init}} $    & : Initial kernel bandwidth.          & \{0.1, 0.2, 0.3, 0.4, 0.5, 0.6, 0.7, 0.8, 0.9, 1.0\}  \\
		\hline
		CAEA      & $ \lambda $                   & : Deletion cycle of isolated nodes. & \{50, 100, 150, 200, 250, 300, 350, 400, 450, 500\}   \\
				& $ a_{\mathrm{max}} $          & : Maximum age of edge.               & \{10, 20, 30, 40, 50, 60, 70, 80, 90, 100\}           \\
		\hline\hline
	\end{tabular}
  }
\end{table*}

\subsection{Evaluation on Synthetic Dataset}
\label{sec:synthetic}
First, we evaluate the clustering performance of CAE using a two-dimensional synthetic dataset in both stationary and non-stationary environments.

The synthetic dataset is primarily used to visualize the network topology constructed by each algorithm and to offer qualitative insights into the self-organizing process. Quantitative comparisons of clustering performance are presented in Section \ref{sec:realworld} using real-world datasets.

\subsubsection{Dataset}
\label{sec:dataset_2d}
\begin{figure}[htbp]
	\centering
	\includegraphics[width=1.4in]{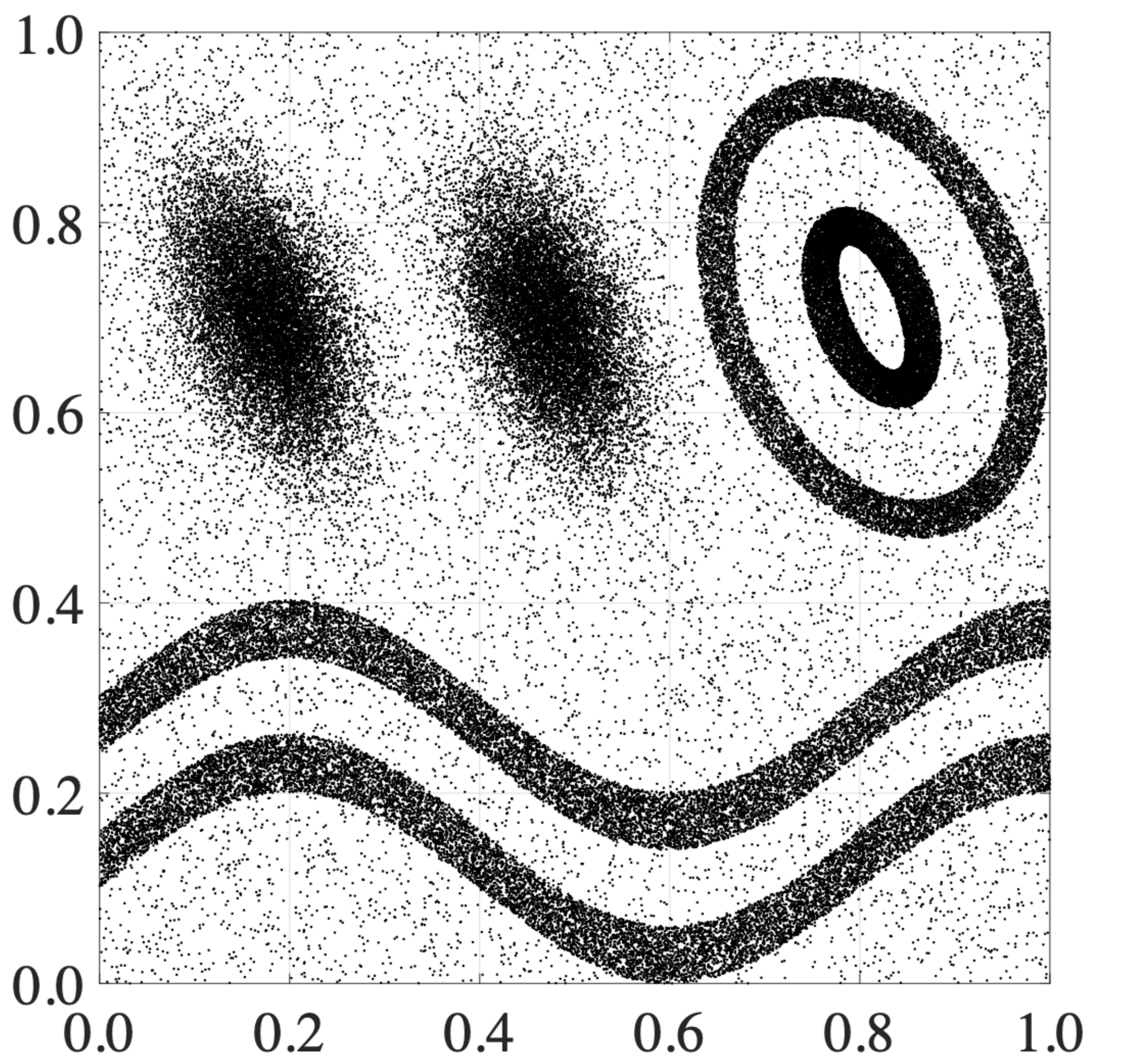}
	\caption{Two-dimensional synthetic dataset.}
	\label{fig:Data_2D}
\end{figure}

\begin{figure}[htbp]
	\vspace{-10mm}
	\centering
	\subfloat[Stream \#1]{
		\includegraphics[width=1.4in]{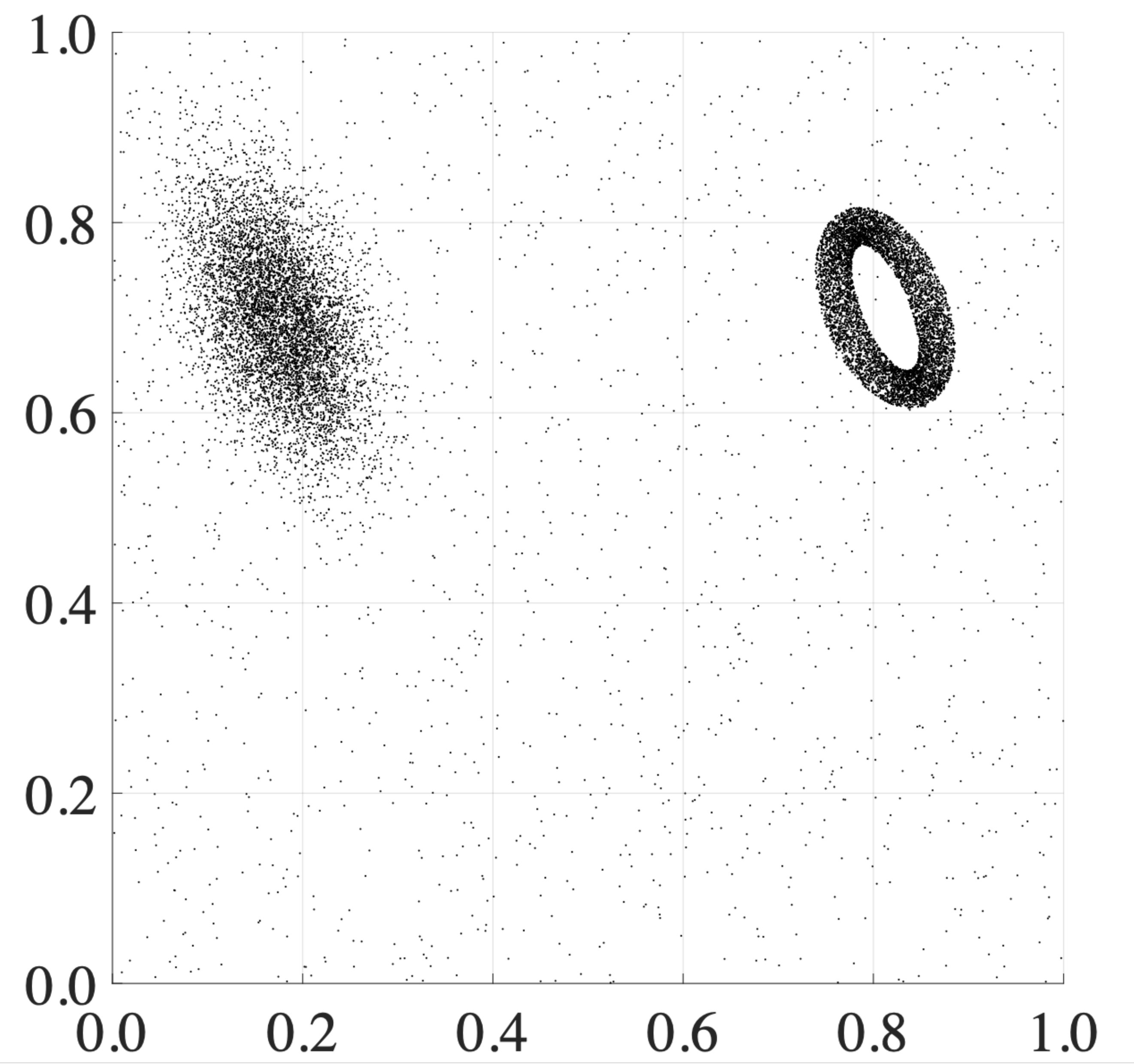}
		\label{fig:Stream_S1}
	}\hfil
	\subfloat[Stream \#2]{
		\includegraphics[width=1.4in]{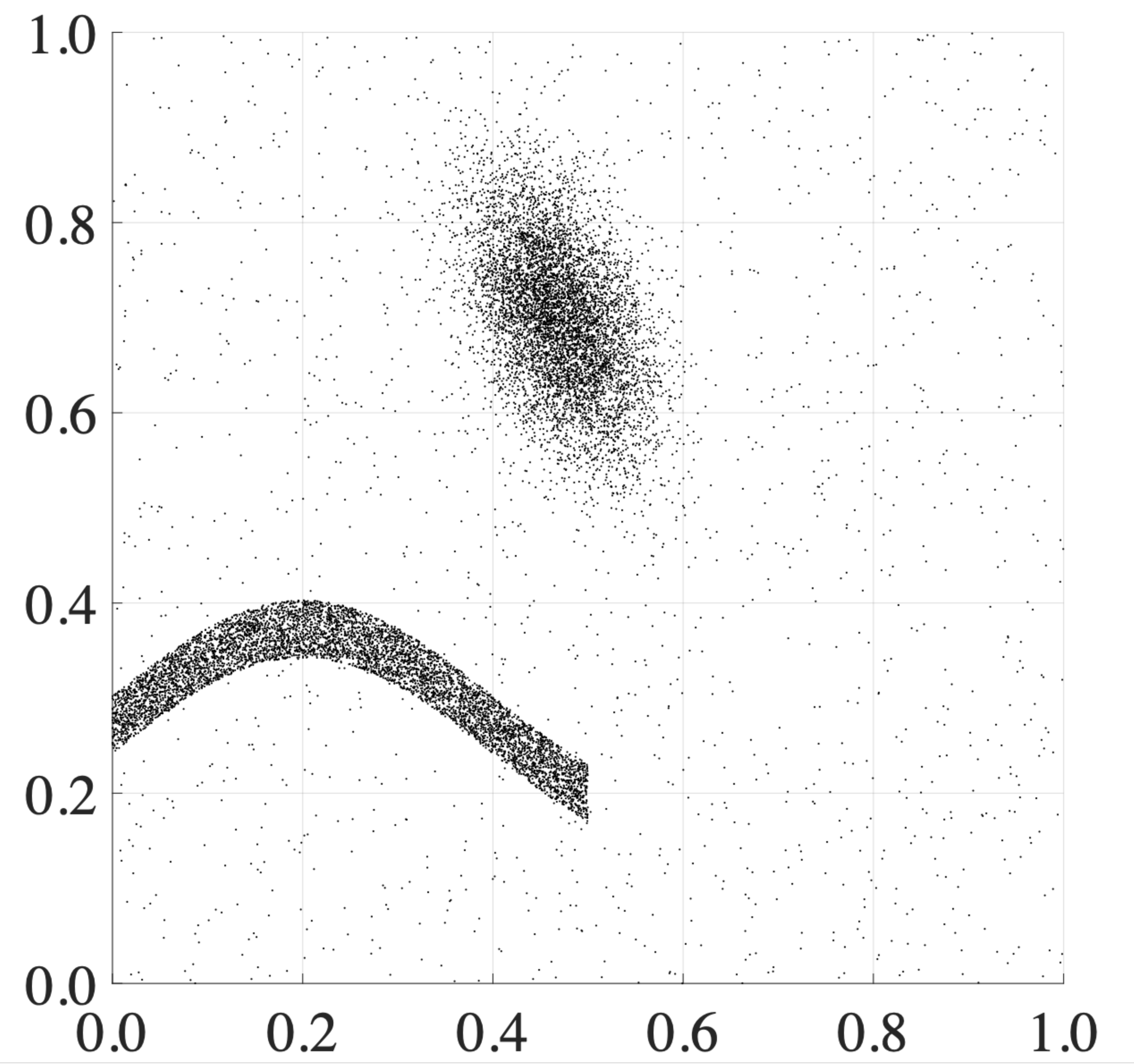}
		\label{fig:Stream_S2}
	}\hfil
	\subfloat[Stream \#3]{
		\includegraphics[width=1.4in]{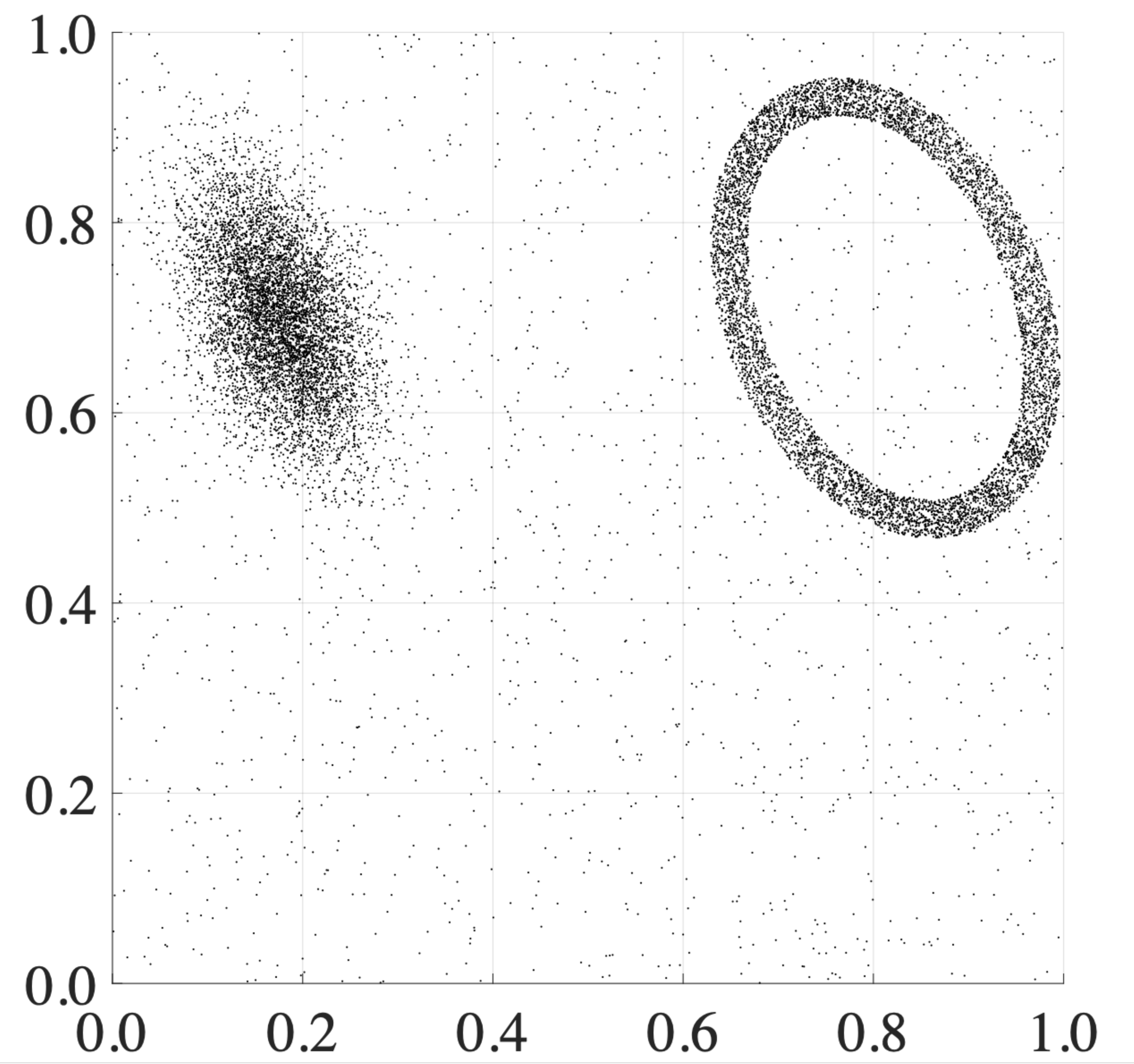}
		\label{fig:Stream_S3}
	} \\
	\subfloat[Stream \#4]{
		\includegraphics[width=1.4in]{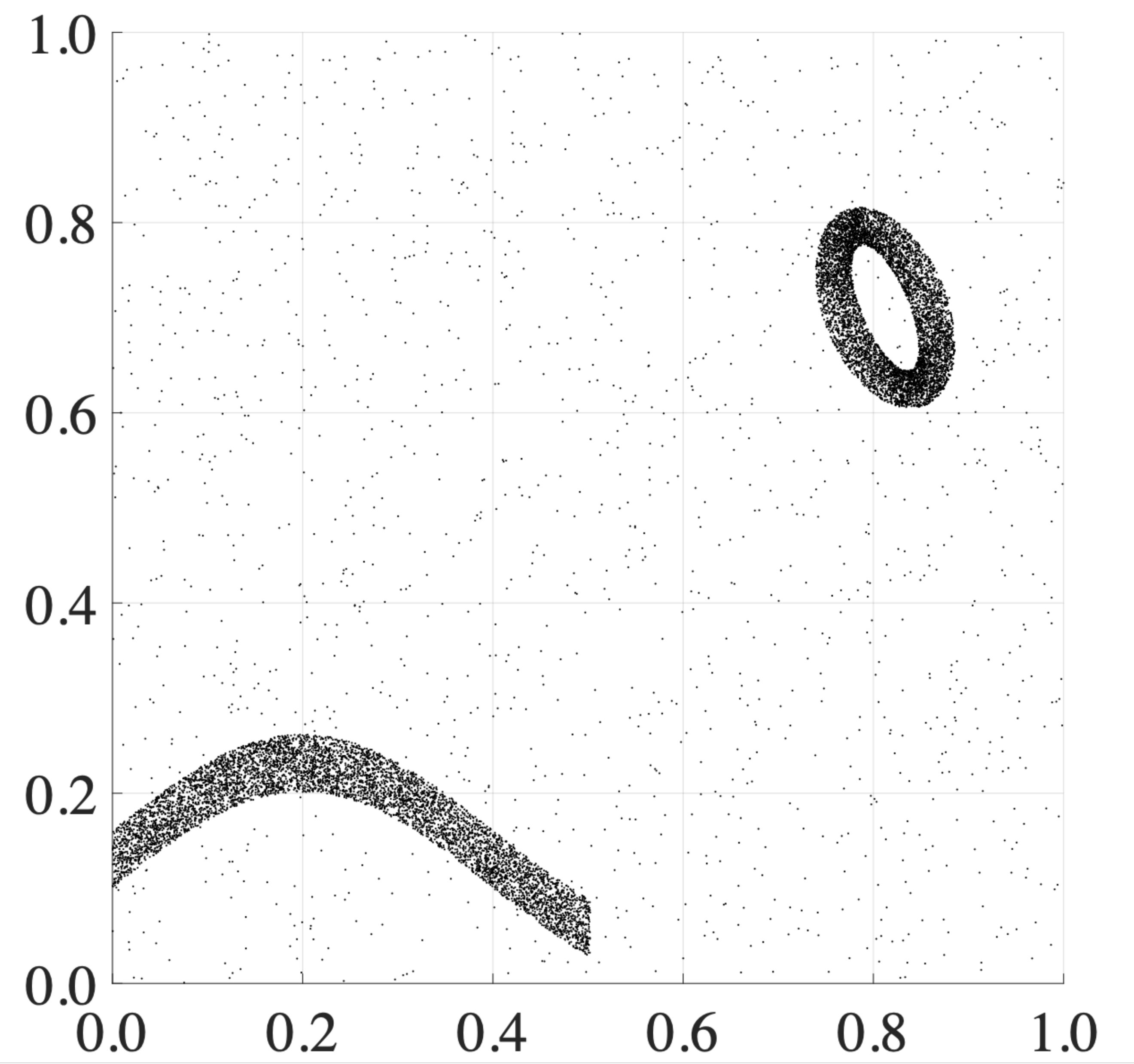}
		\label{fig:Stream_S4}
	}\hfil
	\subfloat[Stream \#5]{
		\includegraphics[width=1.4in]{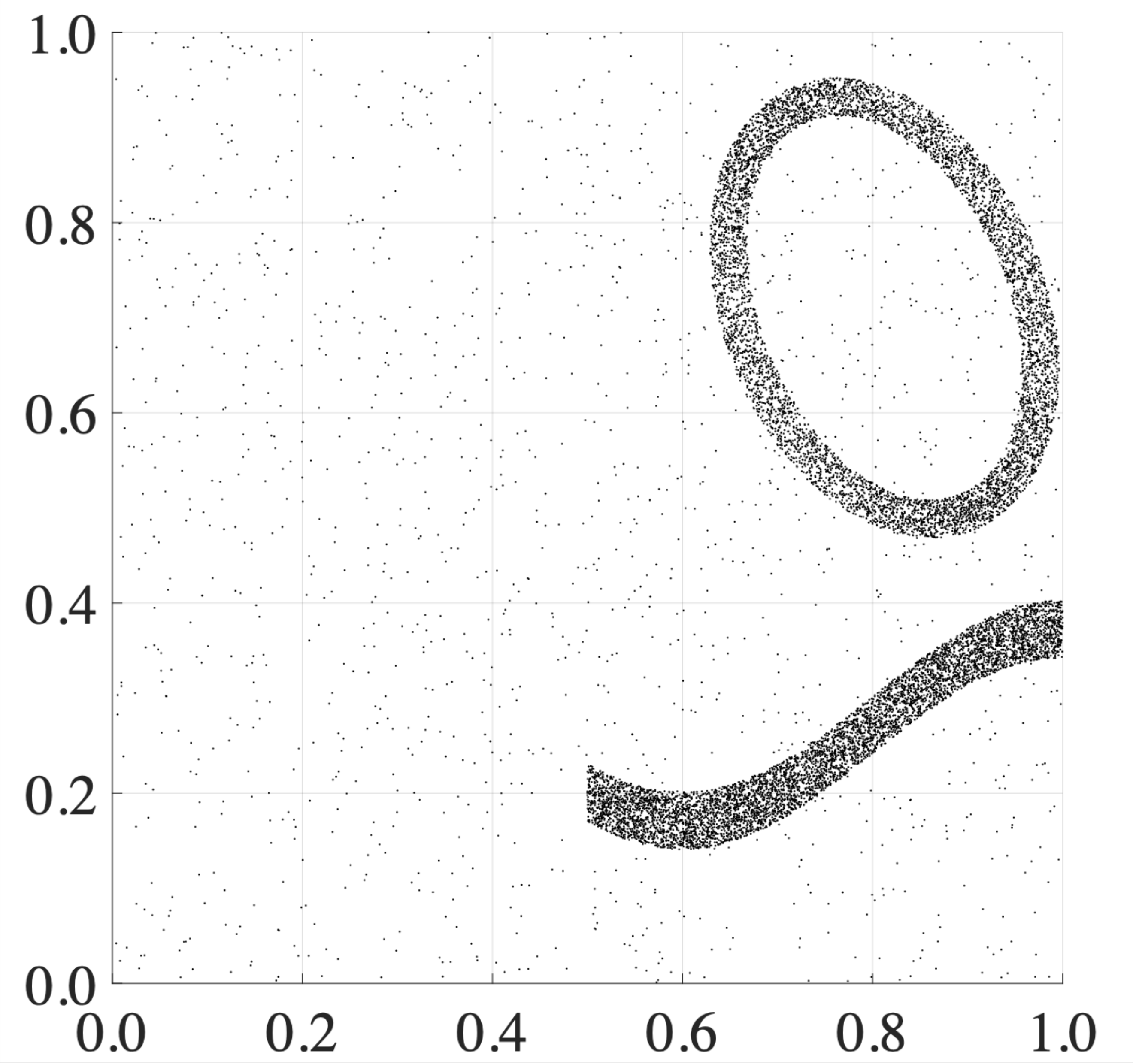}
		\label{fig:Stream_S5}
	}\hfil
	\subfloat[Stream \#6]{
		\includegraphics[width=1.4in]{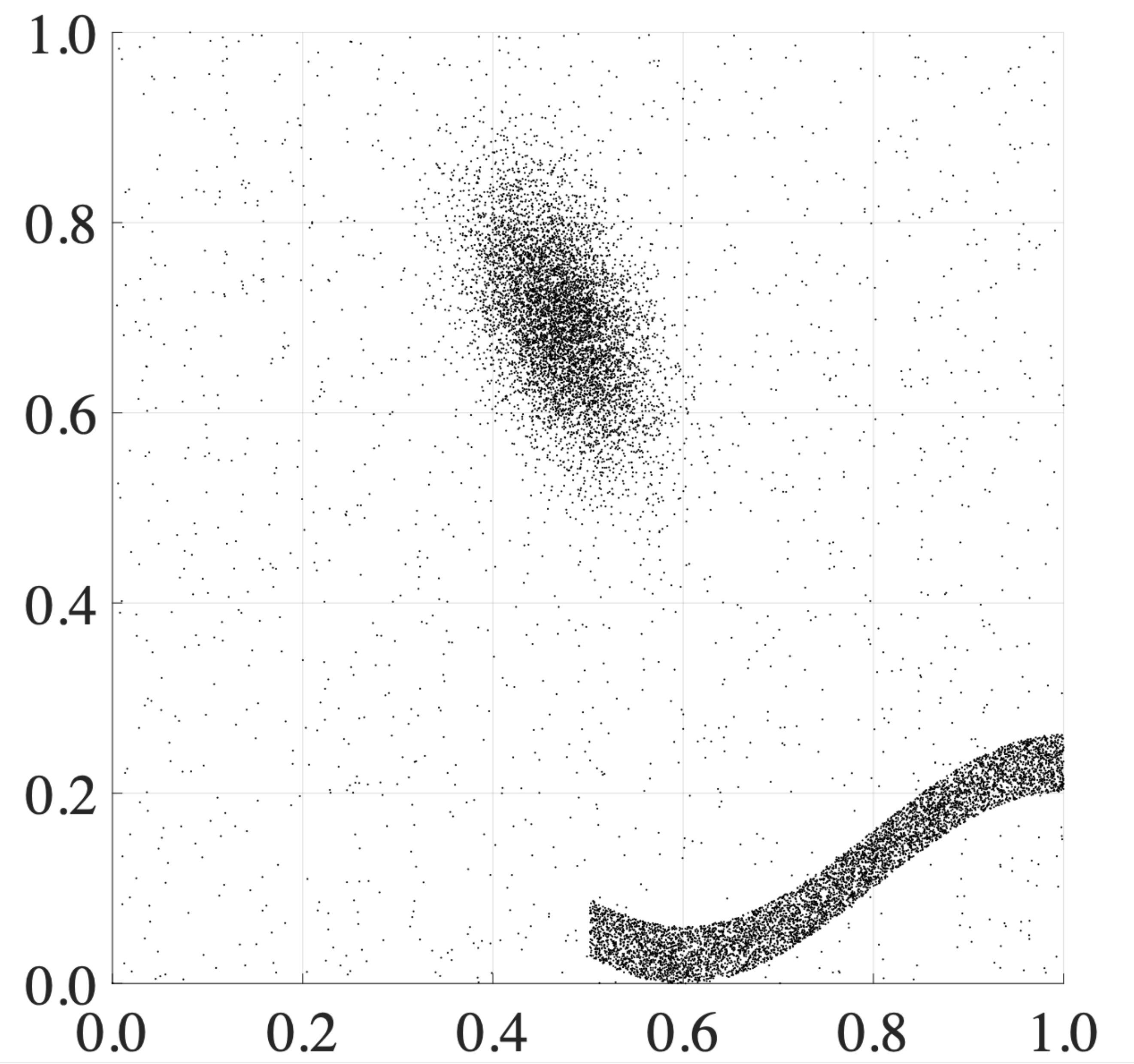}
		\label{fig:Stream_S6}
	}
	\caption{Visualization of the synthetic dataset in sequential order (i.e., (a) to (f)).}
	\label{fig:Stream}
\end{figure}

Fig. \ref{fig:Data_2D} shows the entire dataset in the stationary environment, where the value range for each dimension is [0, 1]. The dataset comprises six distributions, each containing 15,000 data points. Additionally, 10\% uniformly distributed noise is added, resulting in a total of 1,500$\times$6 noise points. Fig. \ref{fig:Stream} shows the dataset in the non-stationary environment, where the data points from Fig. \ref{fig:Data_2D} are divided into six streams, presented sequentially from stream \#1 to \#6. During our experiments, each dataset is presented to each algorithm only once without any pre-processing. In the stationary environment, data points are randomly sampled from the entire dataset. In the non-stationary environment, data points are randomly sampled from a specific distribution, with the class distribution shifting sequentially. As CAE is a clustering algorithm, we use the same data points for both training and testing; that is, each algorithm is trained on all data points in the dataset and evaluated using the same set of data.

\subsubsection{Results}
\label{sec:results_2d}
During grid search, each algorithm is trained using all available data points, and the Normalized Mutual Information (NMI) \cite{strehl02} score is computed on the same set used for training. This evaluation is repeated 15 times with different random seeds for data selection. For comparison, we use the parameter setting corresponding to the median NMI score among the 15 runs. Table \ref{tab:paramGrid_2d} summarizes the grid search results for the synthetic dataset. These results indicate that the optimal parameter values for each algorithm depend on the environment.

\begin{table*}[htbp]
	\vspace{2mm}
	\caption{Parameter specifications by grid search for synthetic dataset}
	\label{tab:paramGrid_2d}
	\footnotesize
	\centering
	\renewcommand{\arraystretch}{1.2}
	\scalebox{1.0}{
		\begin{tabular}{l|c|cc|cc|cc}
			\hline\hline
			\multirow{2}{*}{Environment} & AutoCloud & \multicolumn{2}{c|}{ASOINN} & \multicolumn{2}{c|}{TCA} & \multicolumn{2}{c}{CAEA} \\
			&  $m$ & $\lambda$ & $ a_{\mathrm{max}} $ & $\lambda$ & $ \sigma_{\mathrm{init}} $ & $\lambda$ & $ a_{\mathrm{max}} $ \\
			\hline
			Stationary & 2 & 400 & 10 & 450 & 0.1 & 50 & 30  \\
			Non-stationary & 2 & 250 & 10 & 500 & 0.1 & 50 & 40  \\
			\hline\hline
		\end{tabular}
	}
\end{table*}

\begin{figure*}[htbp]
	\centering
	\subfloat[AutoCloud]{
		\includegraphics[width=1.4in]{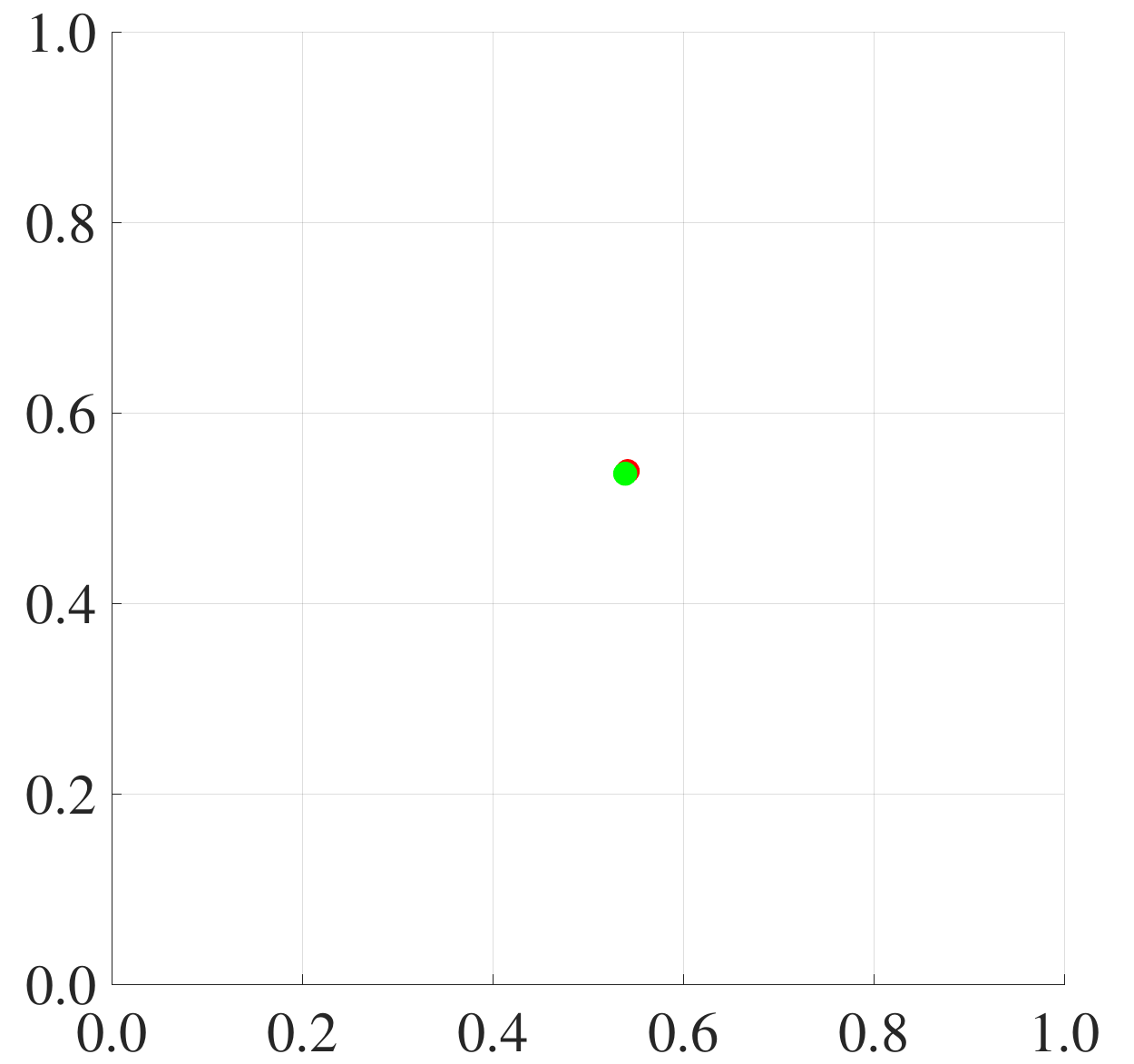}
		\label{fig:Stationary_AutoCloud}
	}\hfil
	\subfloat[ASOINN]{
		\includegraphics[width=1.4in]{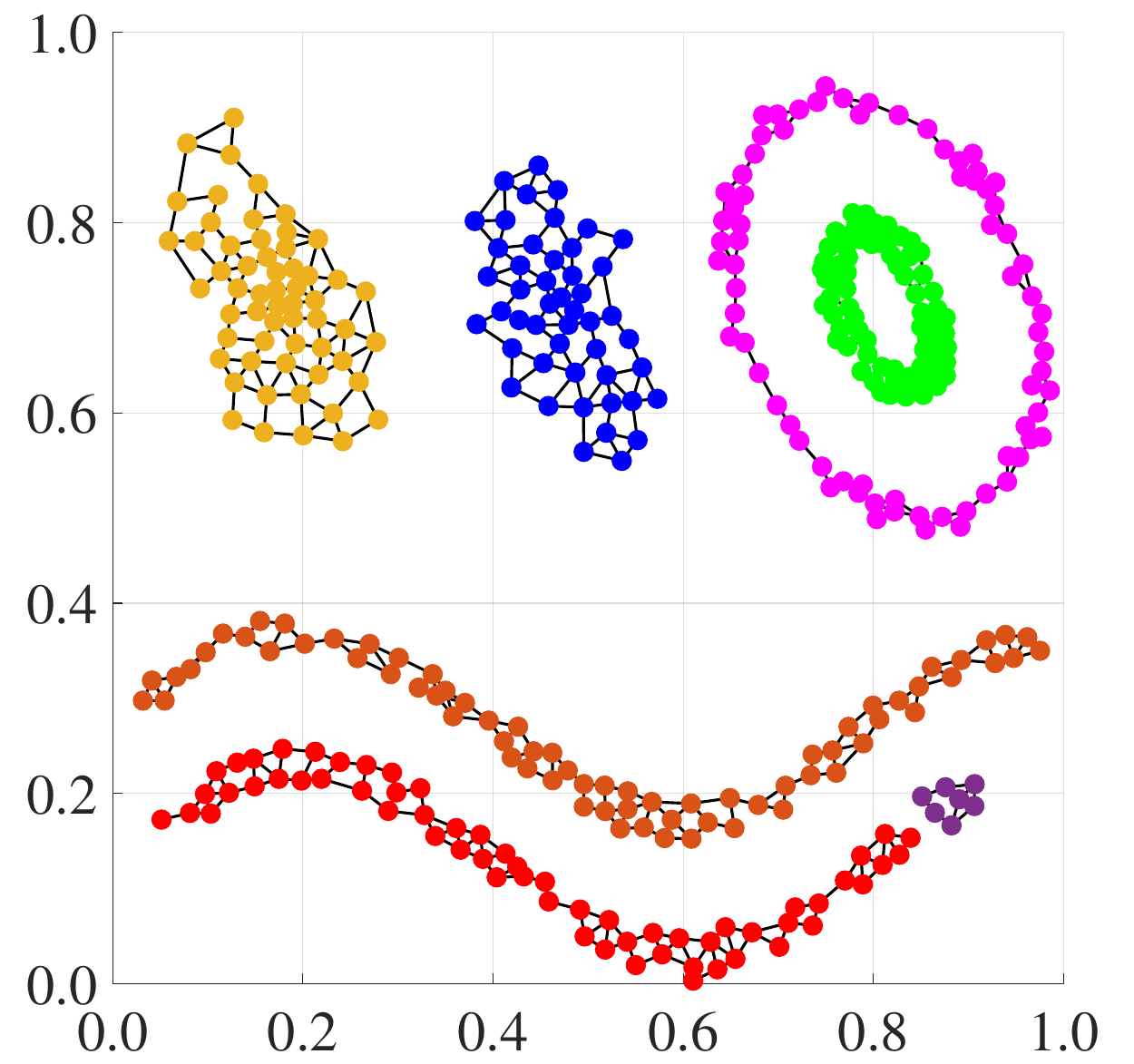}
		\label{fig:Stationary_ASOINN}
	}\hfil
	\subfloat[SOINN+]{
		\includegraphics[width=1.4in]{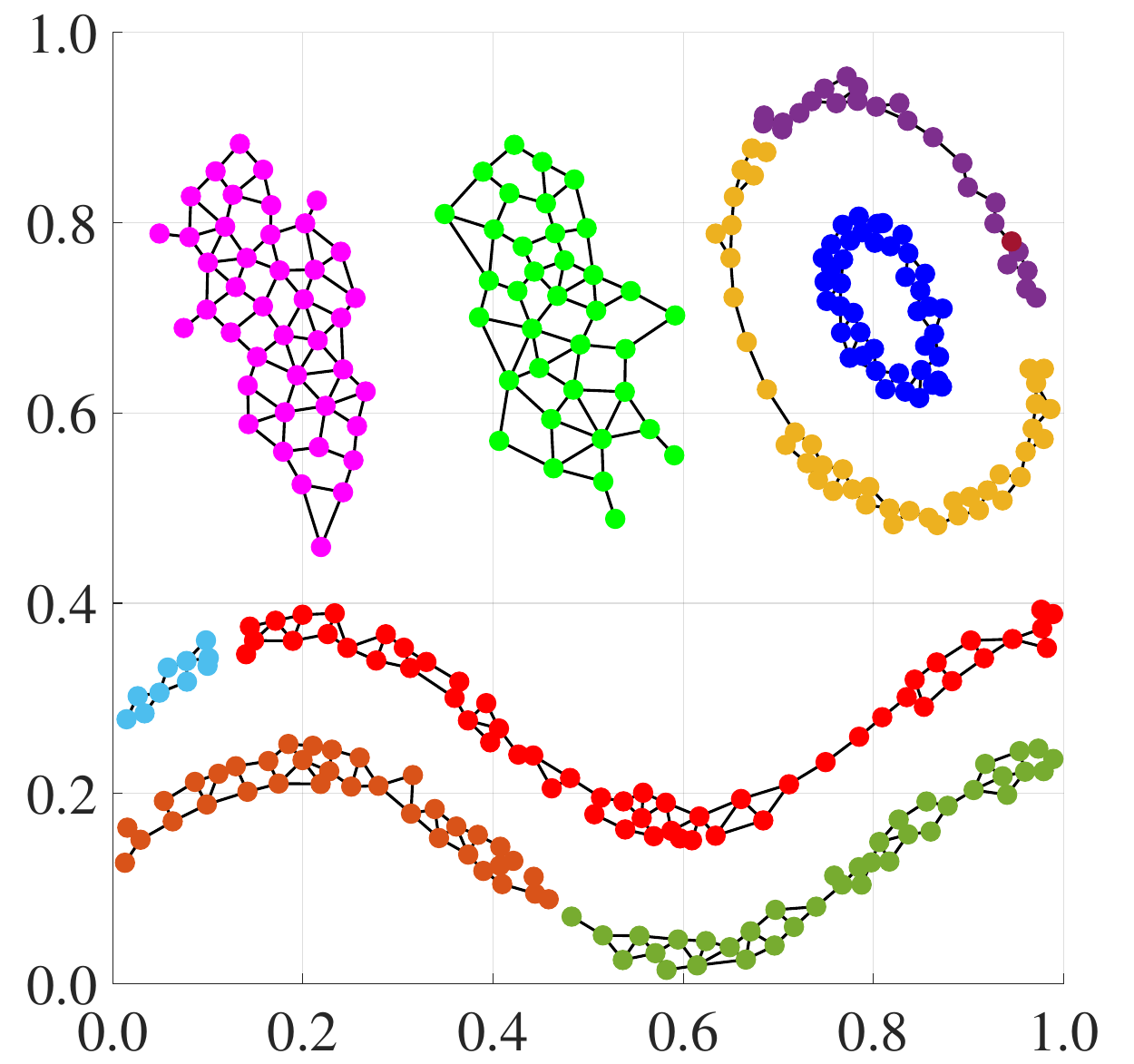}
		\label{fig:Stationary_SOINNplus}
	}
	\vspace{-2mm}
	\hfil
	\\
	\subfloat[TCA]{
		\includegraphics[width=1.4in]{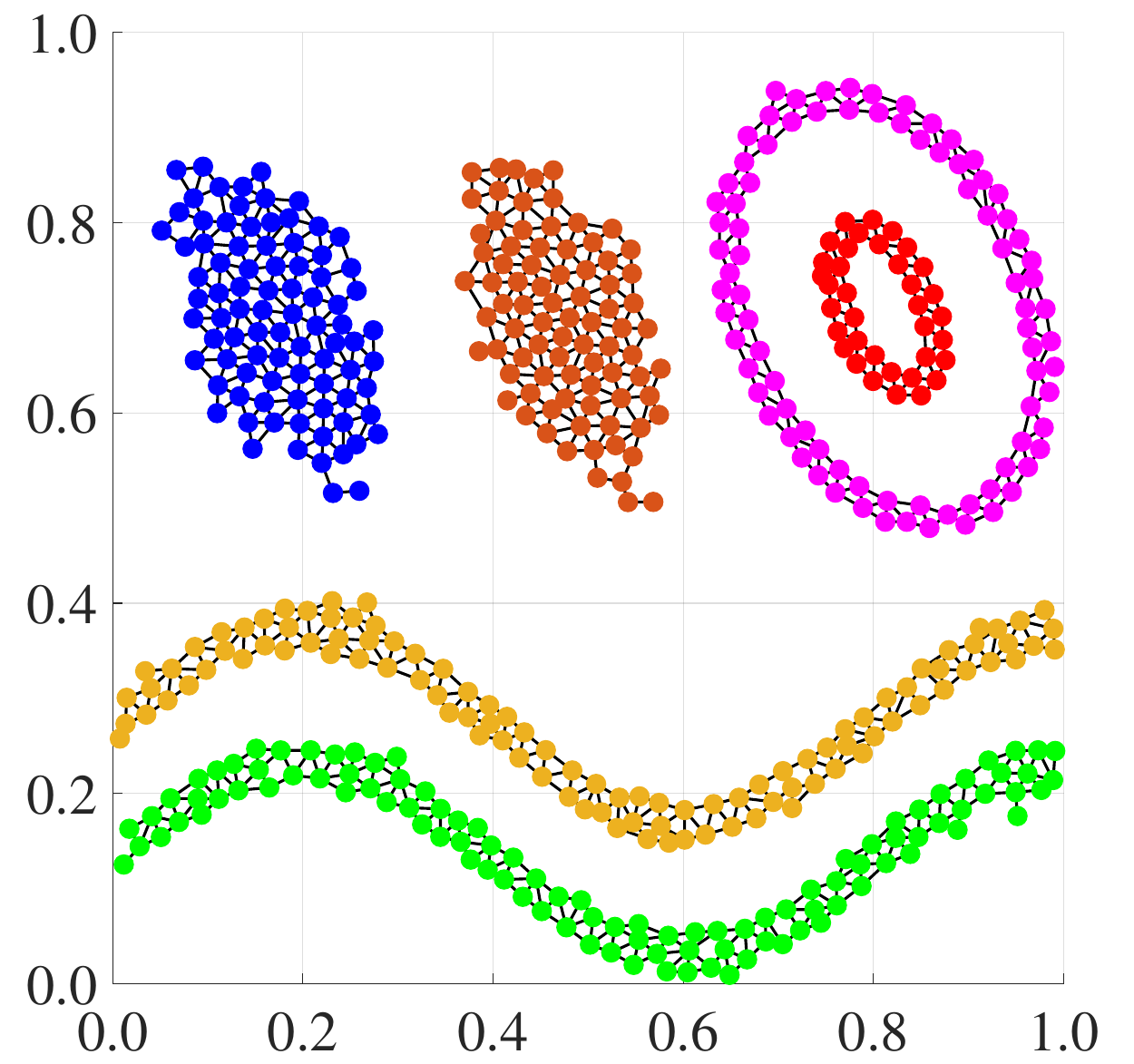}
		\label{fig:Stationary_TCA}
	}\hfil
	\subfloat[CAEA]{
		\includegraphics[width=1.4in]{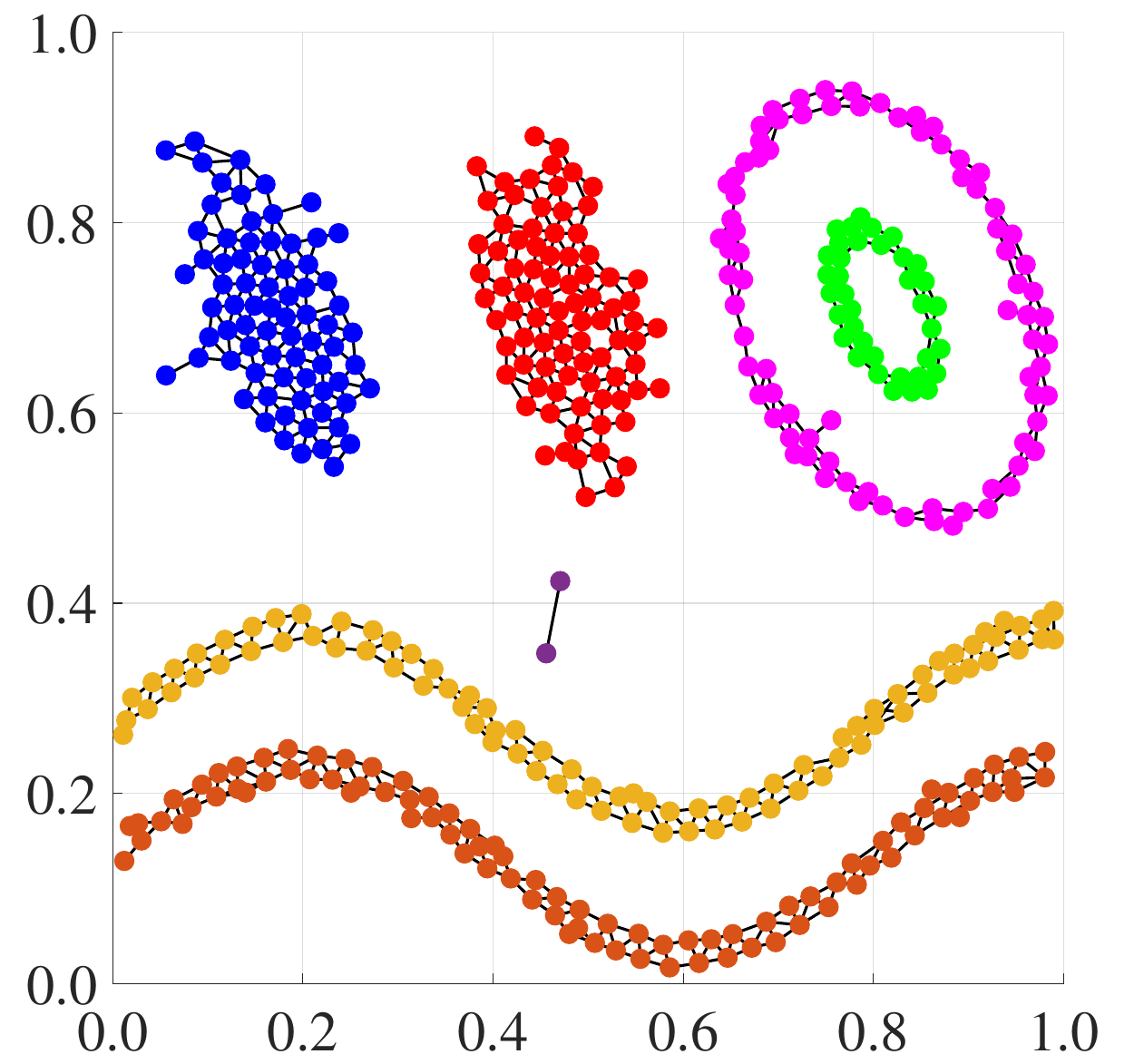}
		\label{fig:Stationary_CAEA}
	}\hfil
	\subfloat[CAE]{
		\includegraphics[width=1.4in]{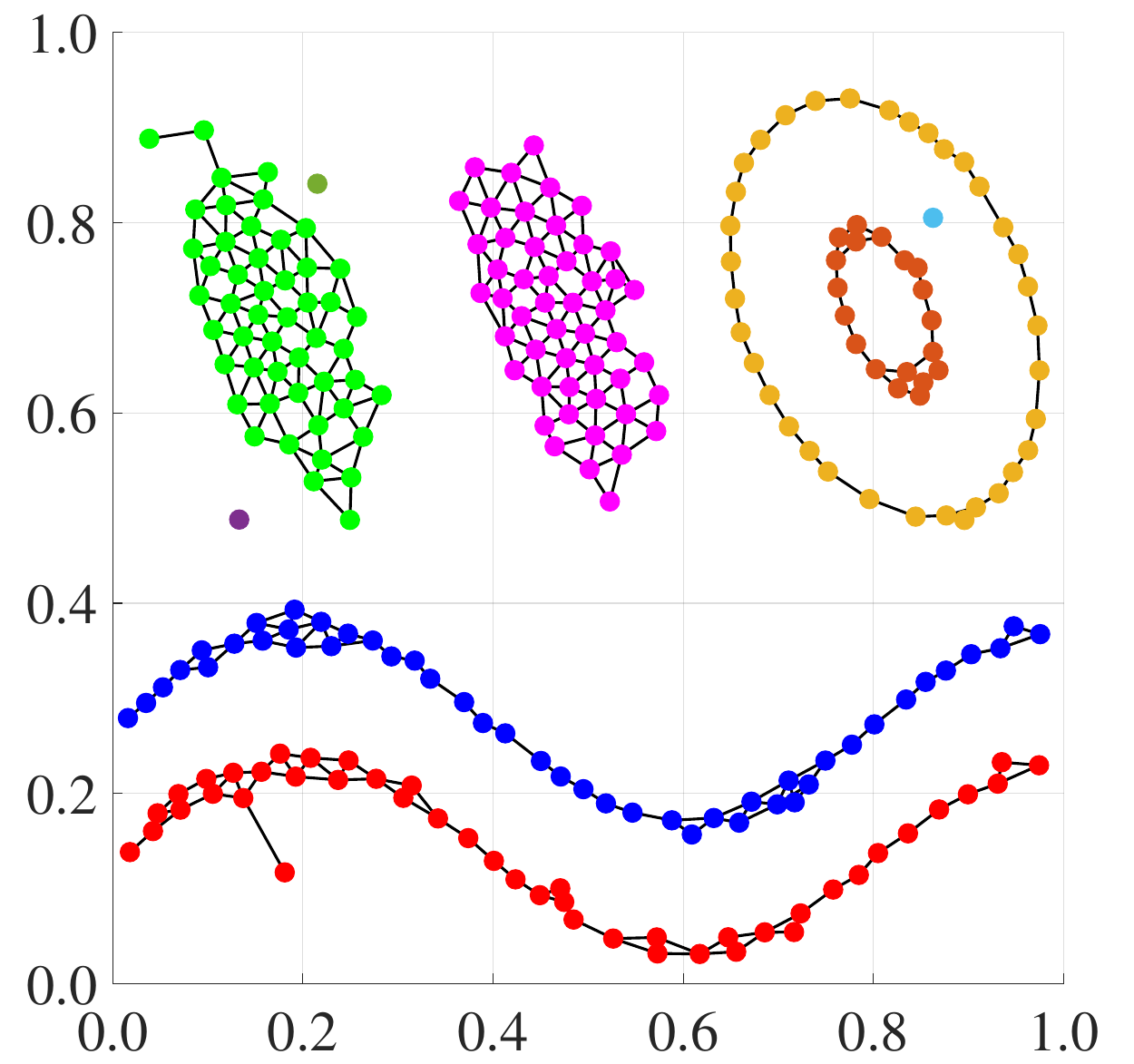}
		\label{fig:Stationary_CAE}
	}
	\vspace{2mm}
	\caption{Visualization of self-organizing results in the stationary environment.}
	\label{fig:Stationary_2d}
\end{figure*}

\begin{figure*}[htbp]
	\vspace{-9mm}
	\centering
	\subfloat[Stream \#1]{
		\includegraphics[width=0.75in]{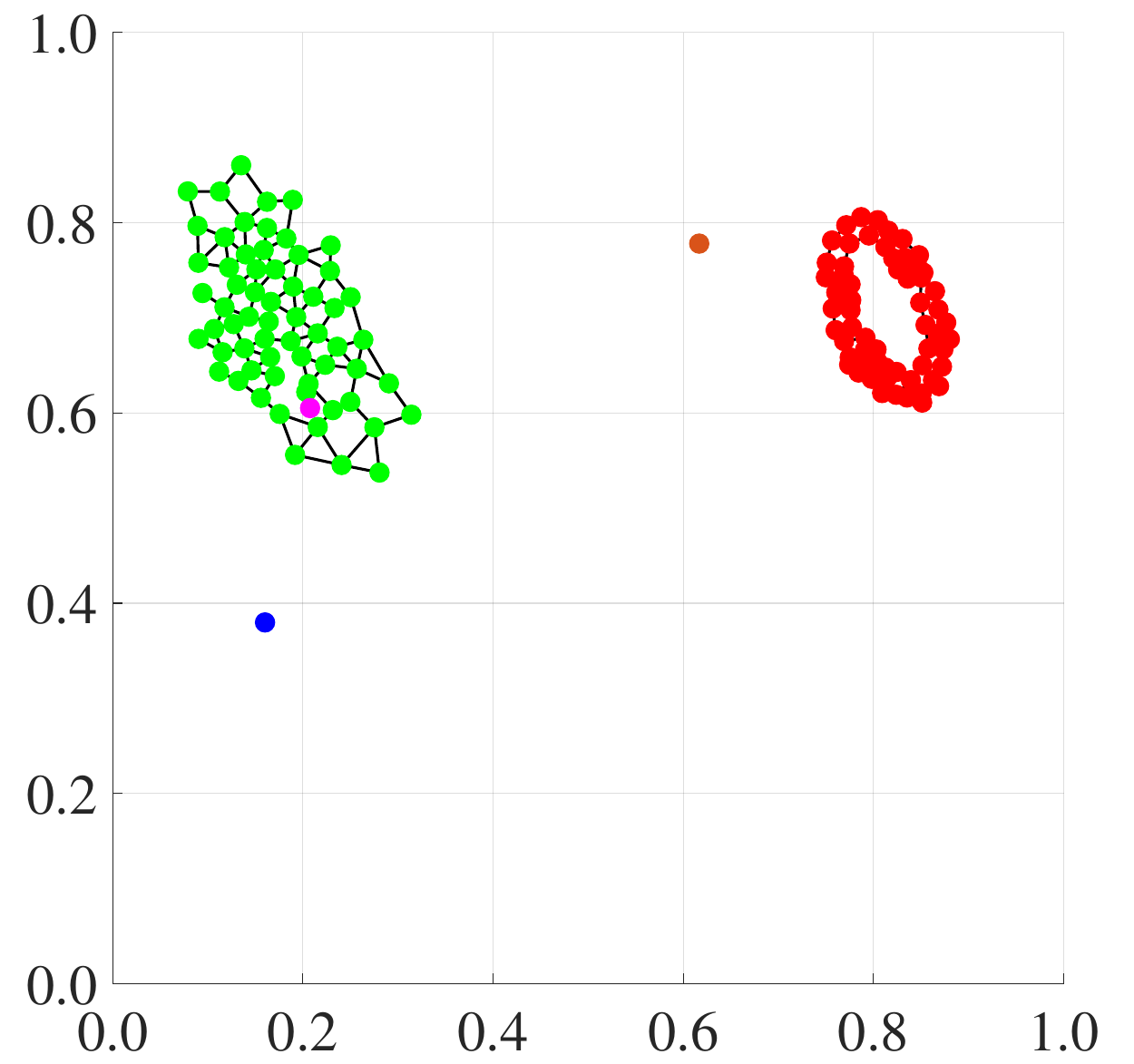}
		\label{fig:SOINNplus_S1}
	}\hfil
	\subfloat[Stream \#2]{
		\includegraphics[width=0.75in]{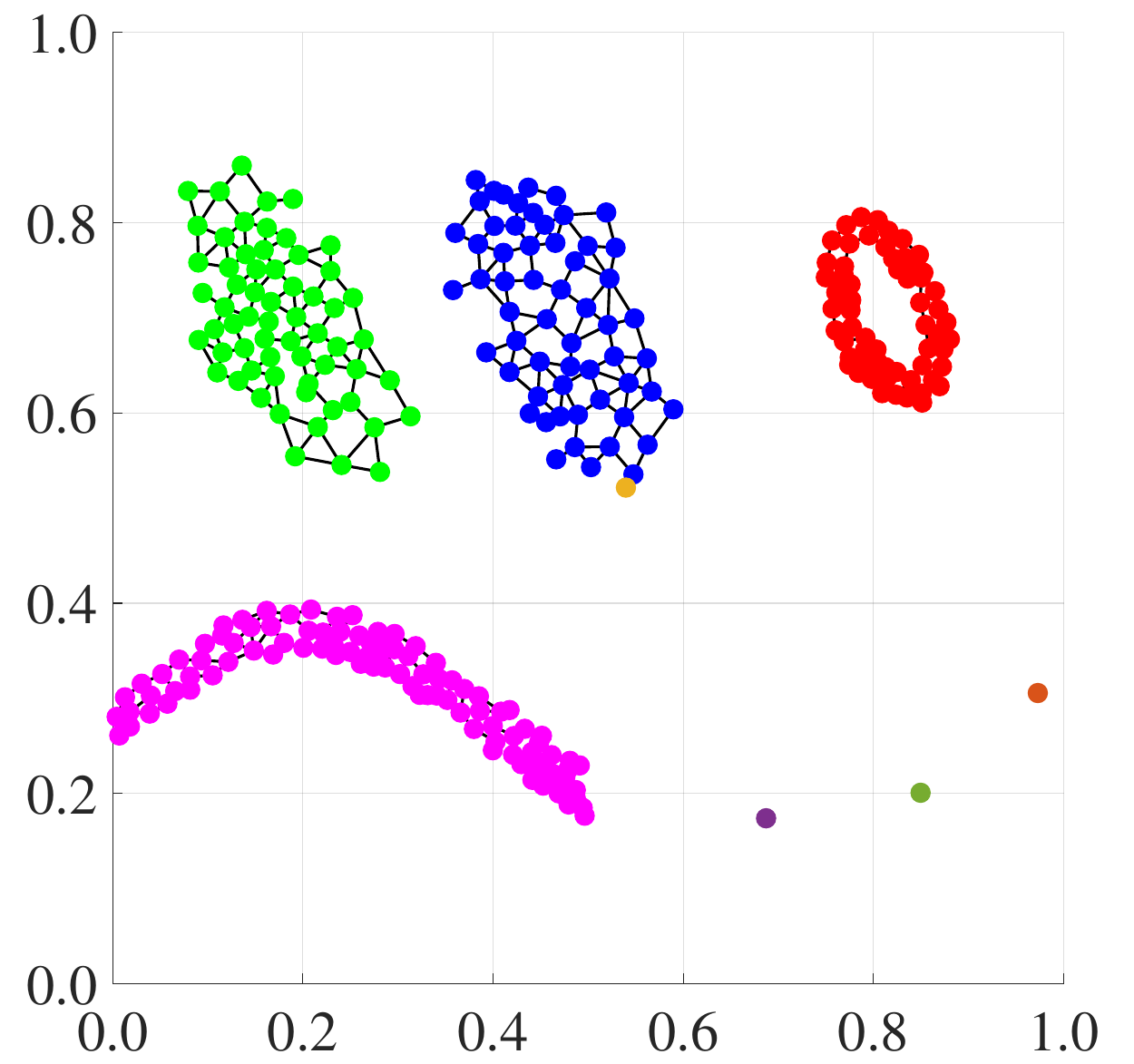}
		\label{fig:SOINNplus_S2}
	}\hfil
	\subfloat[Stream \#3]{
		\includegraphics[width=0.75in]{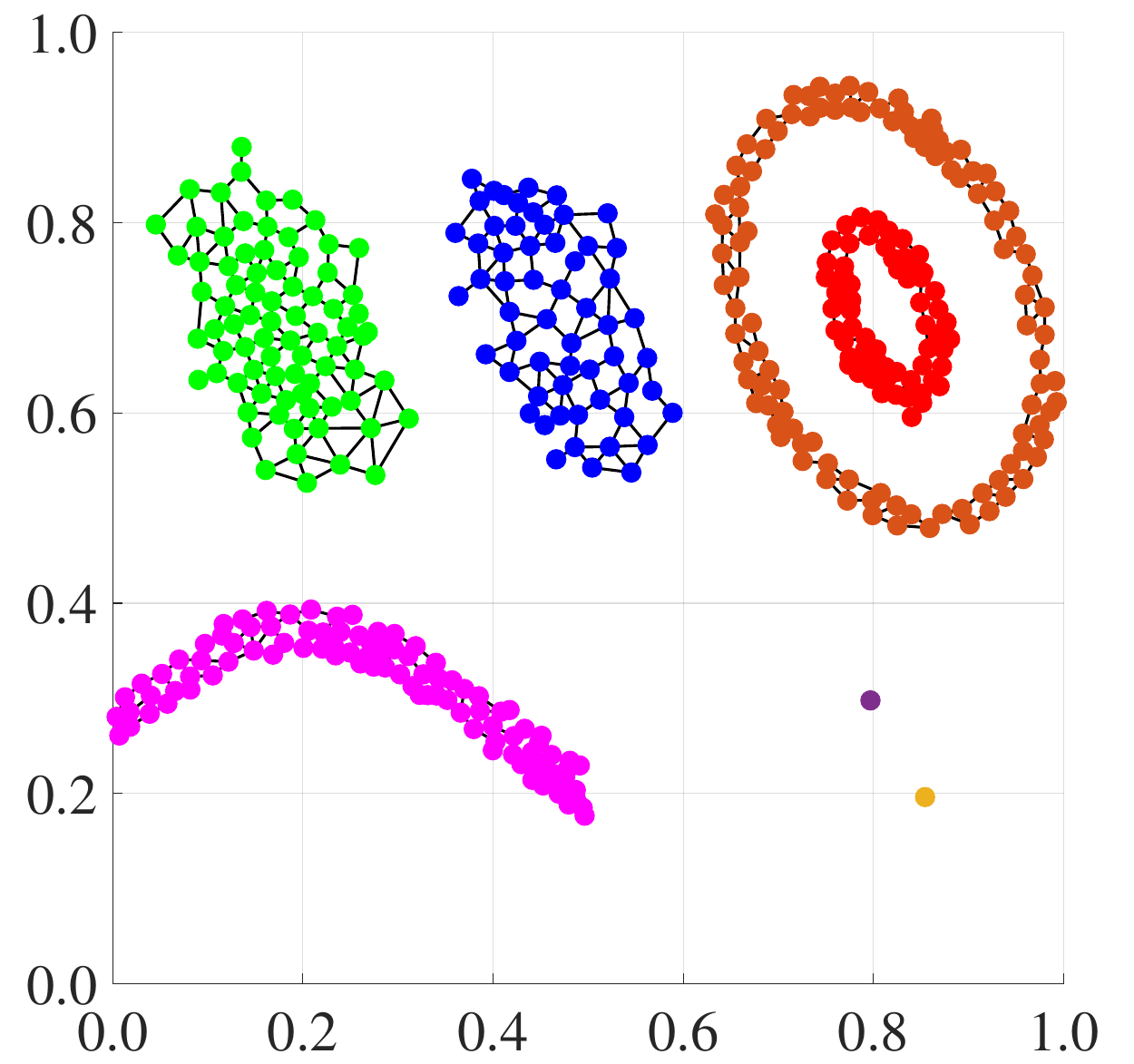}
		\label{fig:SOINNplus_S3}
	}\hfil
	\subfloat[Stream \#4]{
		\includegraphics[width=0.75in]{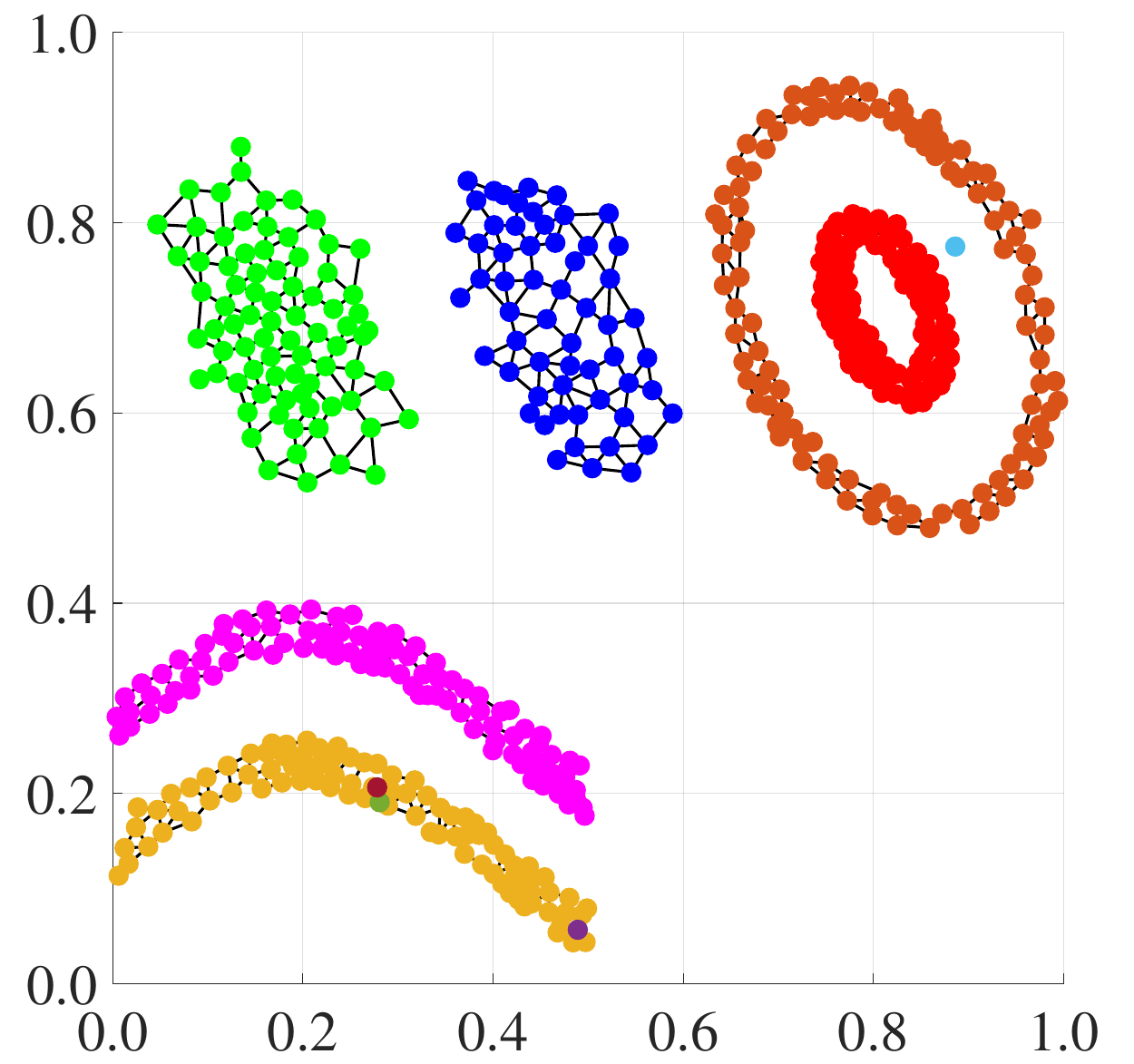}
		\label{fig:SOINNplus_S4}
	}\hfil
	\subfloat[Stream \#5]{
		\includegraphics[width=0.75in]{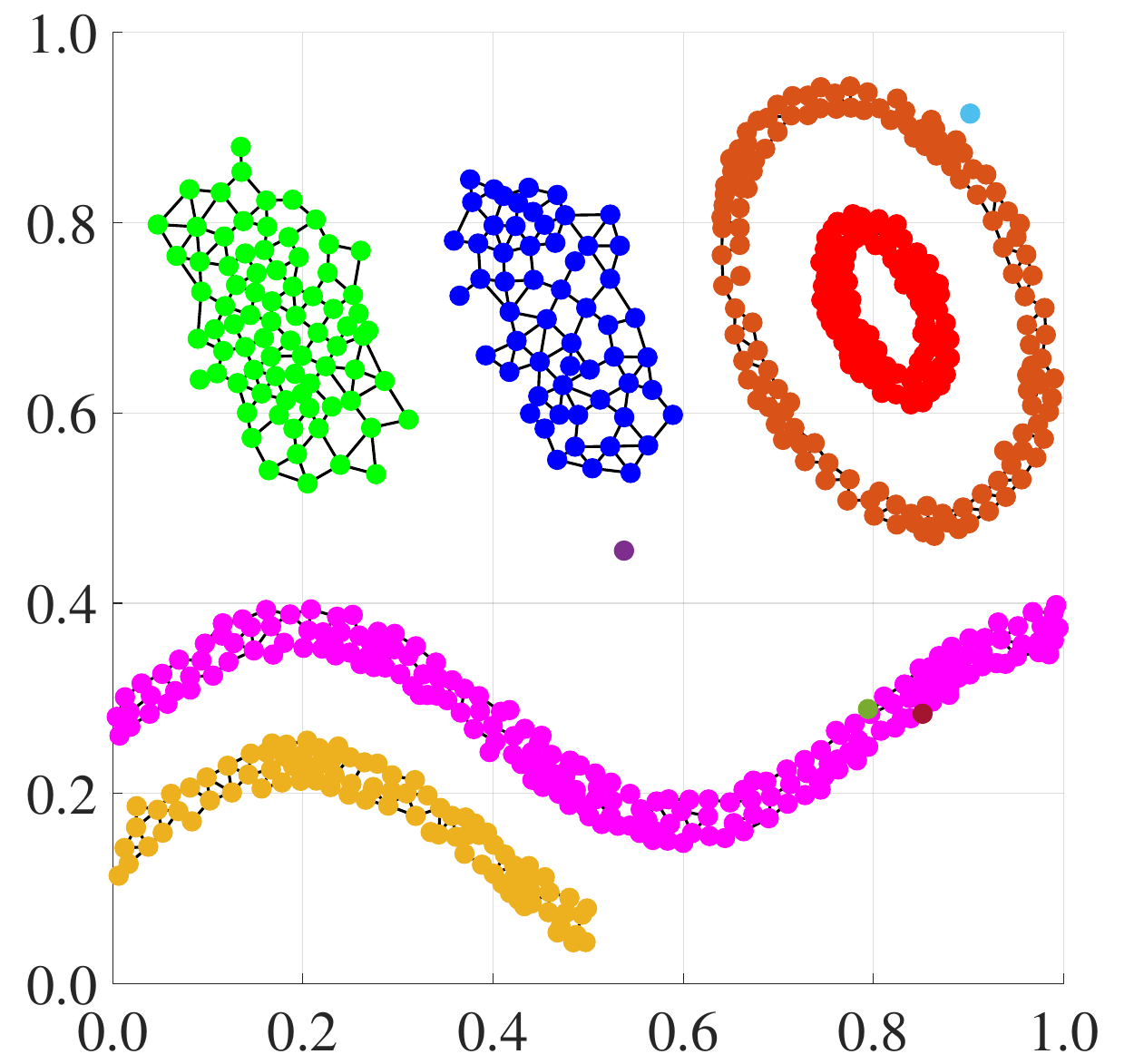}
		\label{fig:SOINNplus_S5}
	}\hfil
	\subfloat[Stream \#6]{
		\includegraphics[width=0.75in]{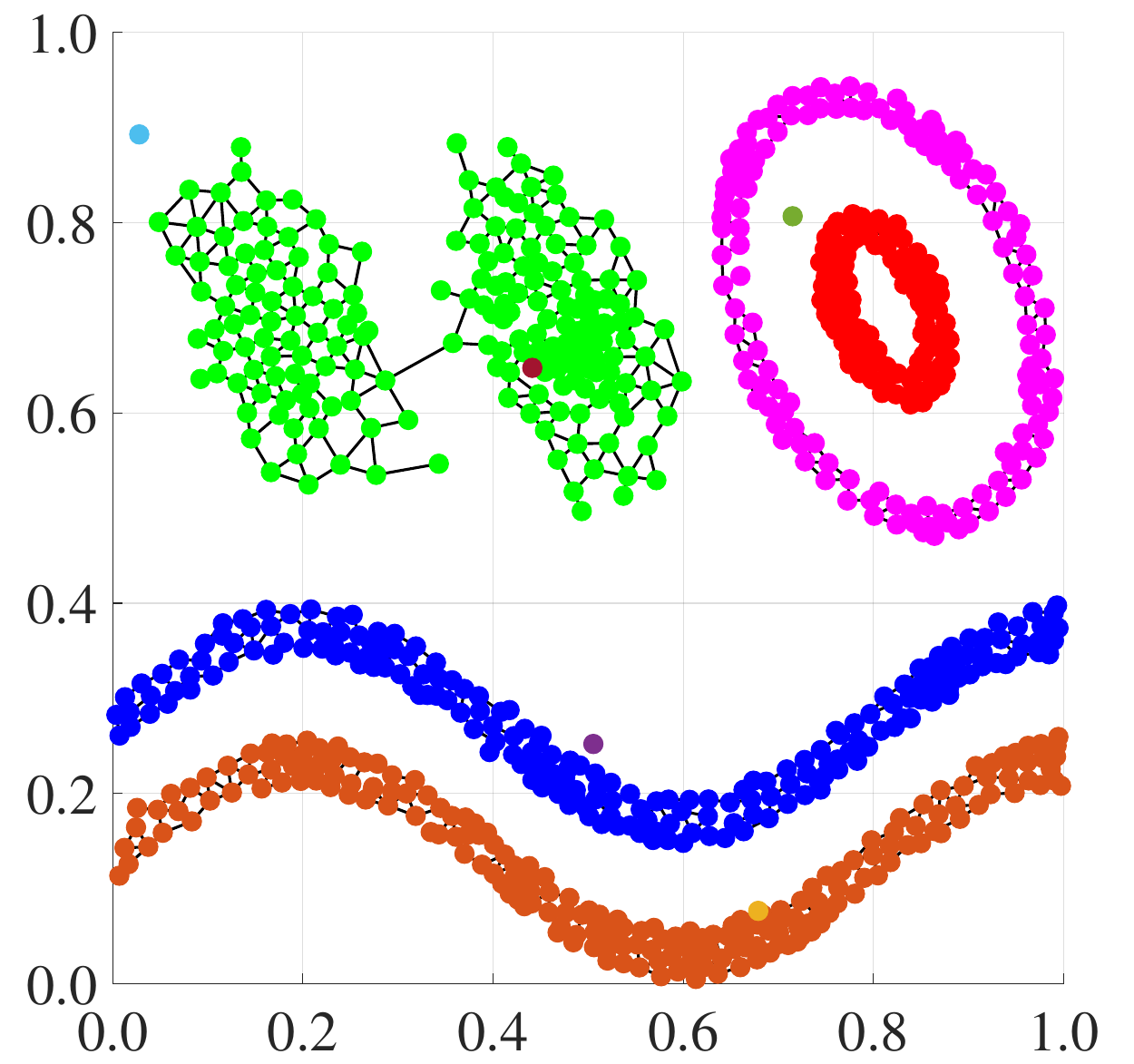}
		\label{fig:SOINNplus_S6}
	}
	\caption{Visualization of self-organizing results of SOINN+ in the non-stationary environment.}
	\label{fig:Nonstationary_2d_SOINNplus}
\end{figure*}

Fig. \ref{fig:Stationary_2d} shows the visualization of self-organizing results in the stationary environment. The figures depict the trial in which the median NMI was obtained. AutoCloud creates and merges data clouds (i.e., cluster-like granular structures) without using edge information. As a result, its clustering results differ from those of all other algorithms. Additionally, since AutoCloud was originally proposed as an online learning algorithm, its clustering result in the stationary environment is less well-organized. ASOINN and SOINN+ tend to create disconnected topological networks. In contrast, thanks to the properties of ART-based algorithms (such as the use of a fixed similarity threshold), TCA, CAEA, and CAE successfully form well-organized topological networks. Comparing CAE to TCA and CAEA, TCA and CAEA tend to generate a larger number of nodes than CAE. In Fig. \ref{fig:Stationary_2d}, some isolated nodes can also be observed in CAE. This occurs because the number of presented data points is generally not a multiple of the deletion cycle (i.e., the deletion process in line 14 of Algorithm \ref{alg:pseudocodeCAE} is typically not executed just before the algorithm terminates for performance evaluation). This phenomenon also occurs in SOINN+ (see Fig. \ref{fig:Nonstationary_2d_SOINNplus}). Moreover, the same issue may arise in TCA and CAEA, depending on the deletion cycle for isolated nodes. A straightforward solution is to delete isolated nodes after the learning procedure. However, since these algorithms are designed for continual learning, they retain isolated nodes to prepare for future learning, rather than removing them after the current cycle. Therefore, we do not consider this a drawback of these algorithms.

\begin{figure*}[htbp]
	\centering
	\subfloat[Stream \#1]{
		\includegraphics[width=0.75in]{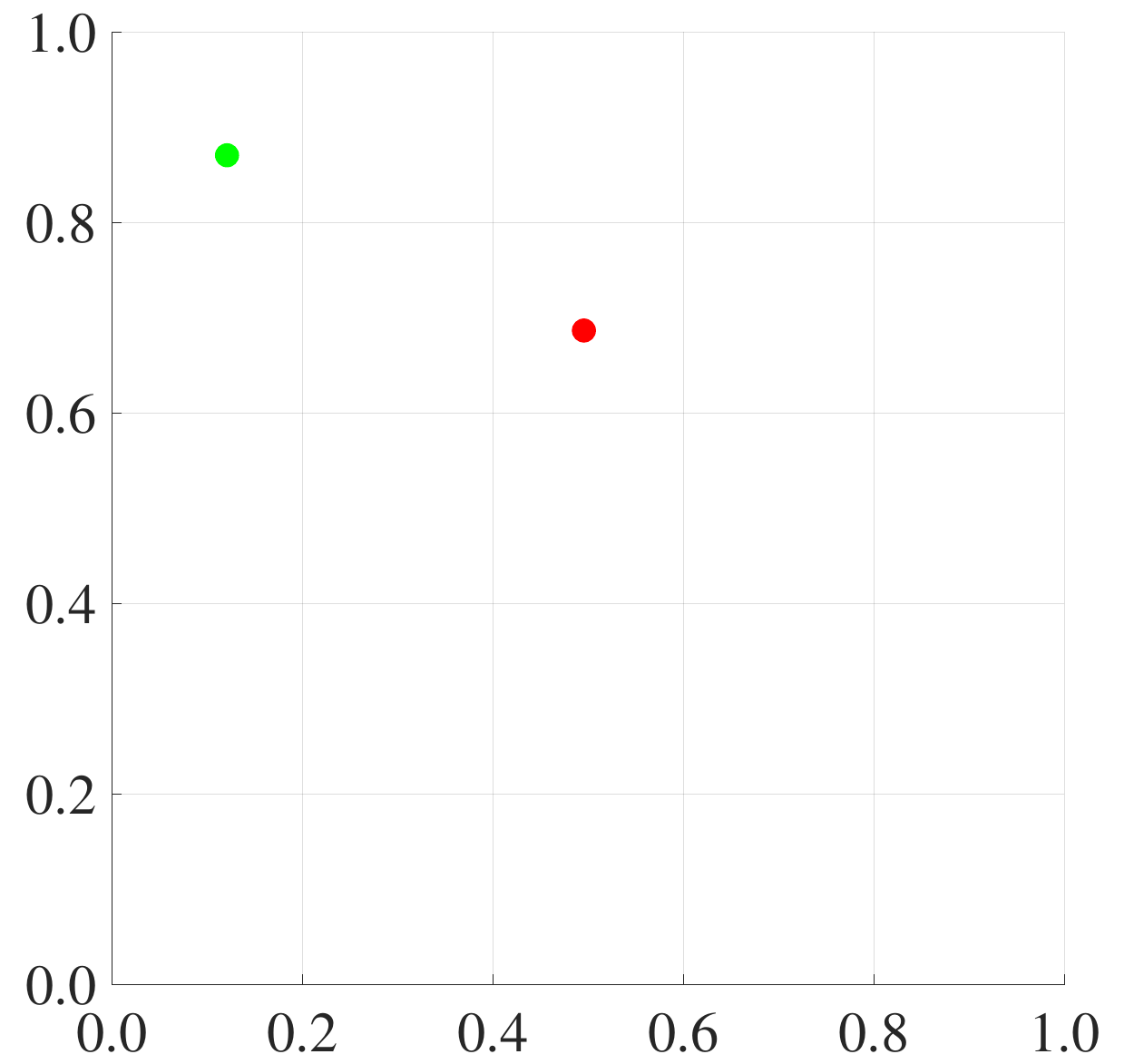}
		\label{fig:AutoCloud_S1}
	}\hfil
	\subfloat[Stream \#2]{
		\includegraphics[width=0.75in]{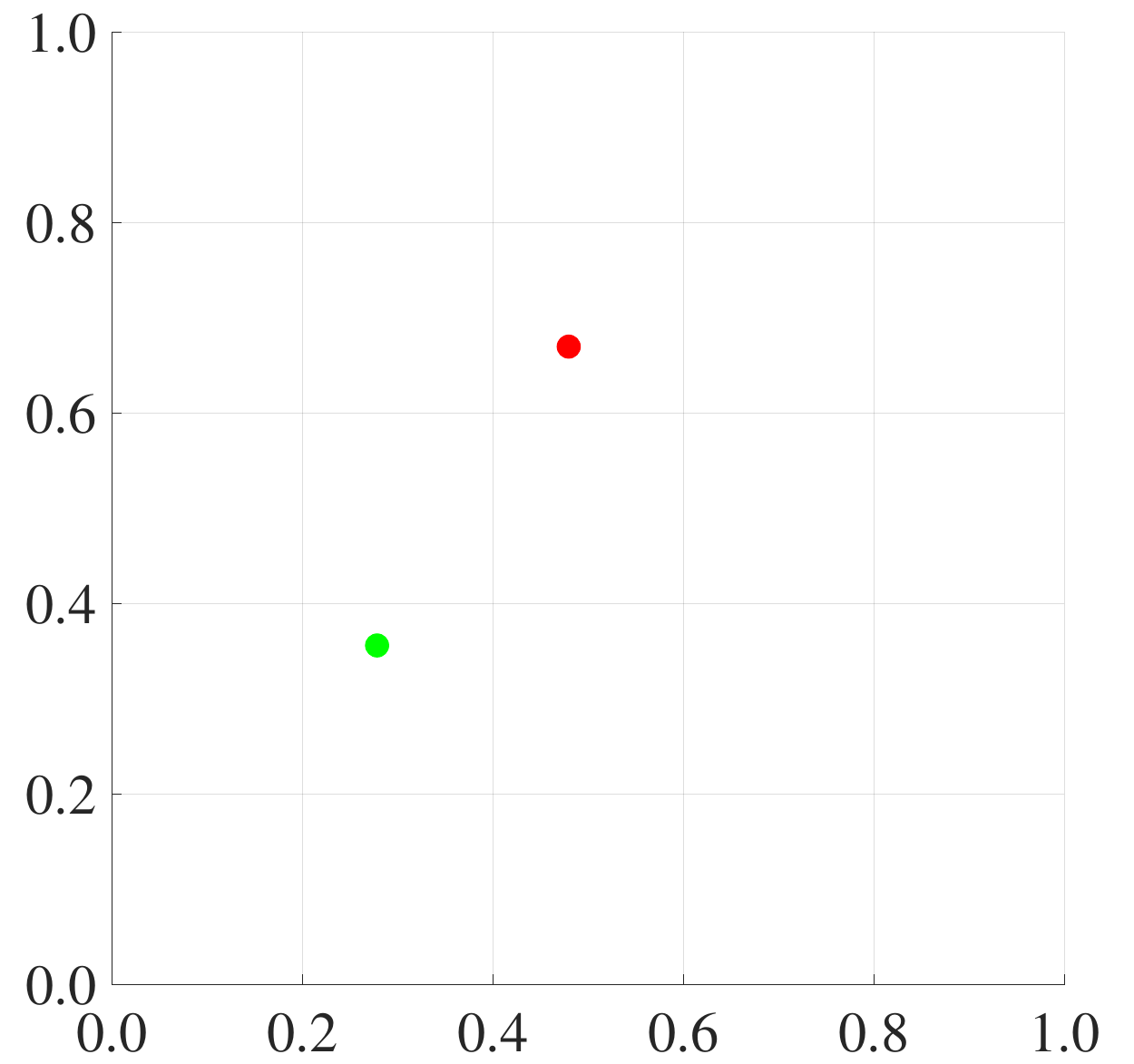}
		\label{fig:AutoCloud_S2}
	}\hfil
	\subfloat[Stream \#3]{
		\includegraphics[width=0.75in]{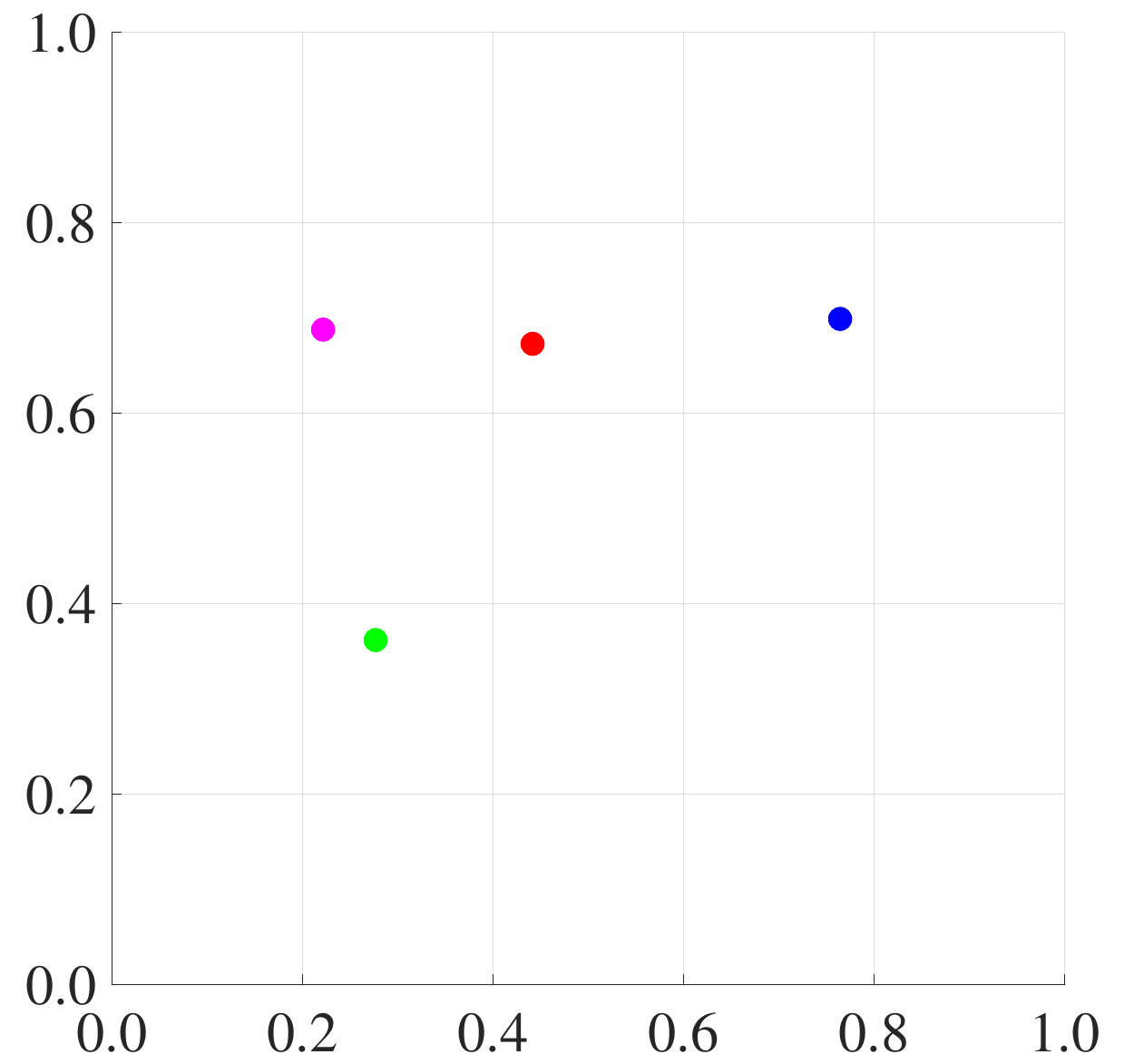}
		\label{fig:AutoCloud_S3}
	}\hfil
	\subfloat[Stream \#4]{
		\includegraphics[width=0.75in]{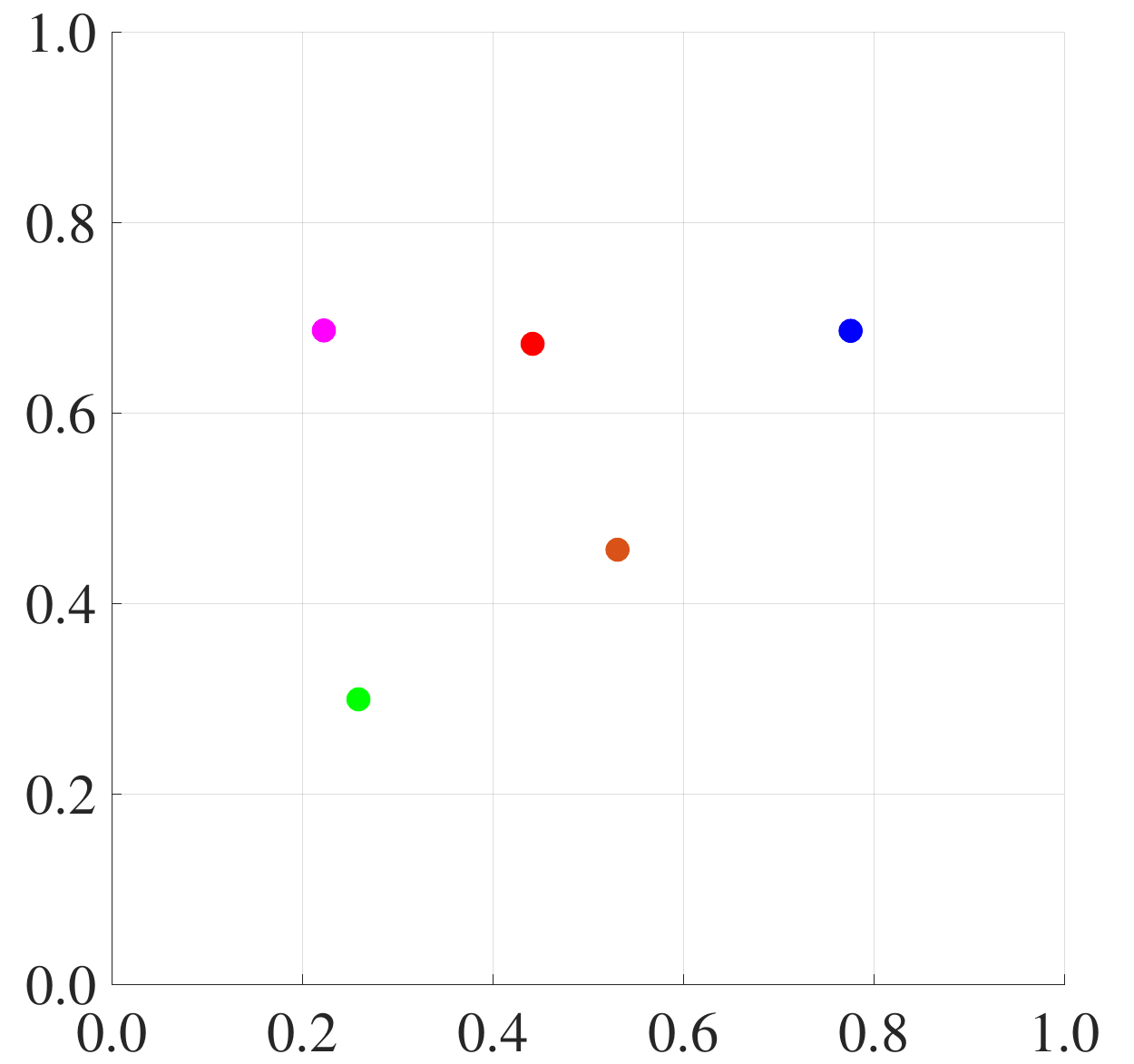}
		\label{fig:AutoCloud_S4}
	}\hfil
	\subfloat[Stream \#5]{
		\includegraphics[width=0.75in]{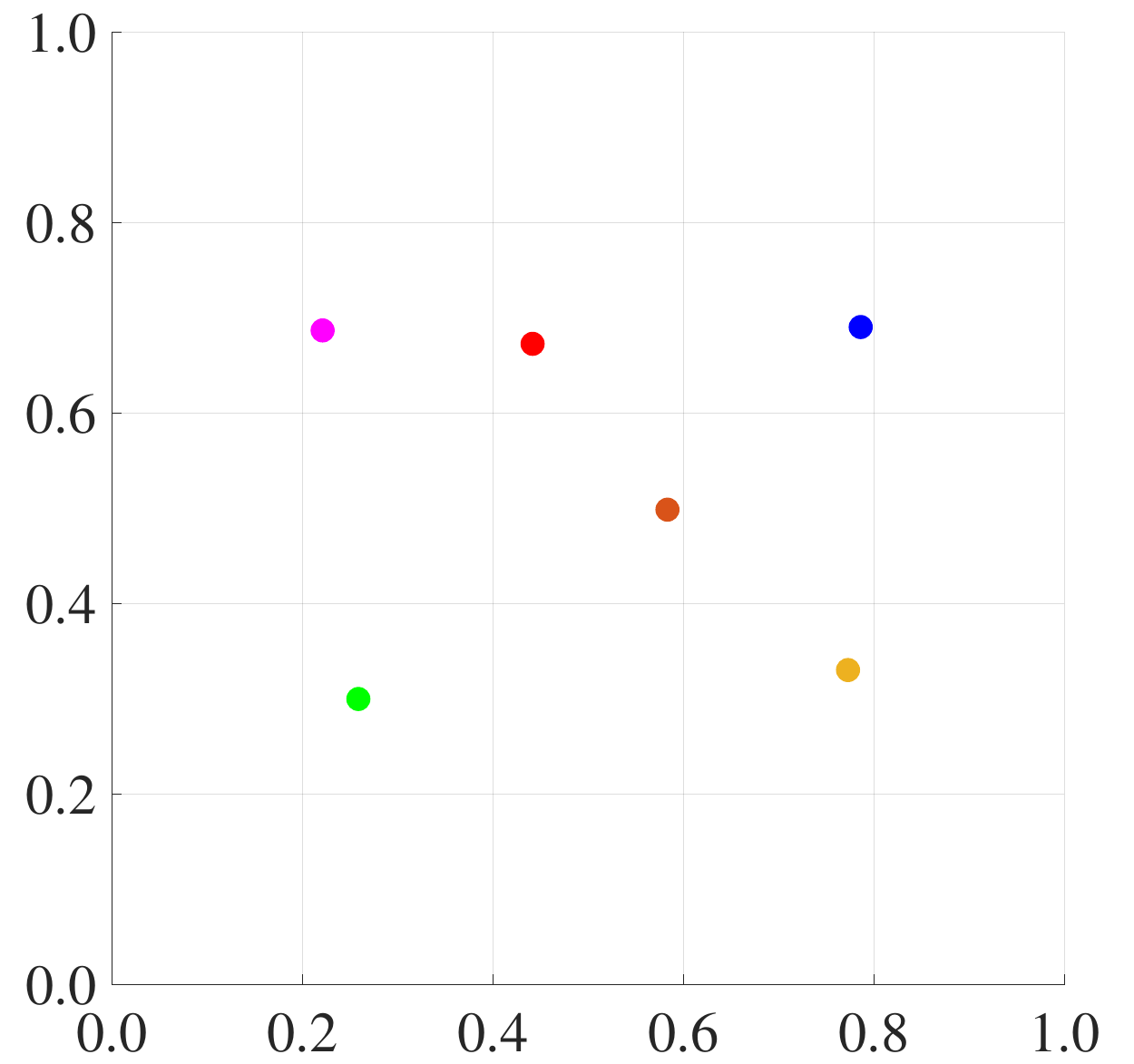}
		\label{fig:AutoCloud_S5}
	}\hfil
	\subfloat[Stream \#6]{
		\includegraphics[width=0.75in]{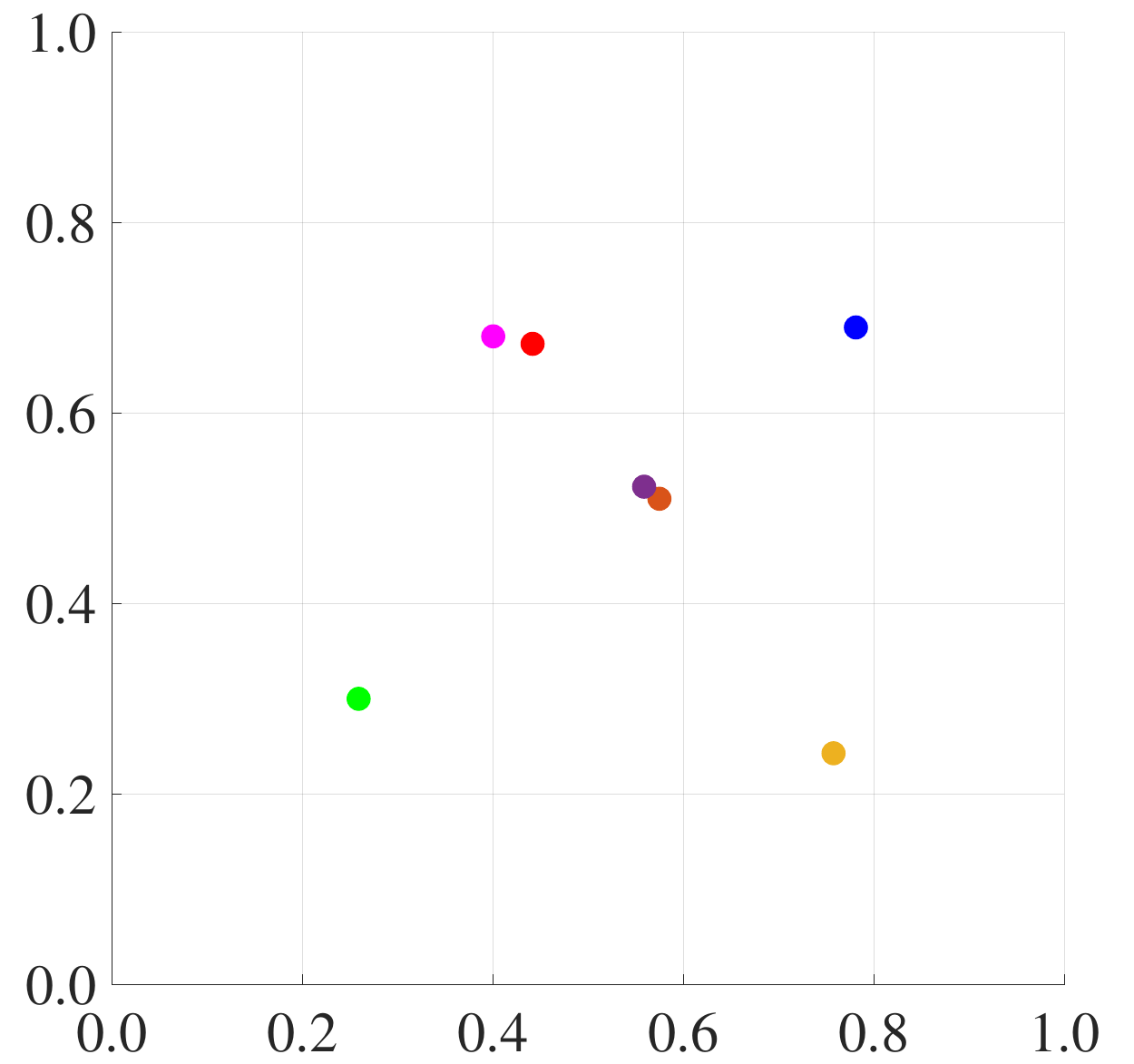}
		\label{fig:AutoCloud_S6}
	}
	\caption{Visualization of self-organizing results of AutoCloud in the non-stationary environment.}
	\label{fig:Nonstationary_2d_AutoCloud}
\end{figure*}

\begin{figure*}[htbp]
	\vspace{-9mm}
	\centering
	\subfloat[Stream \#1]{
		\includegraphics[width=0.75in]{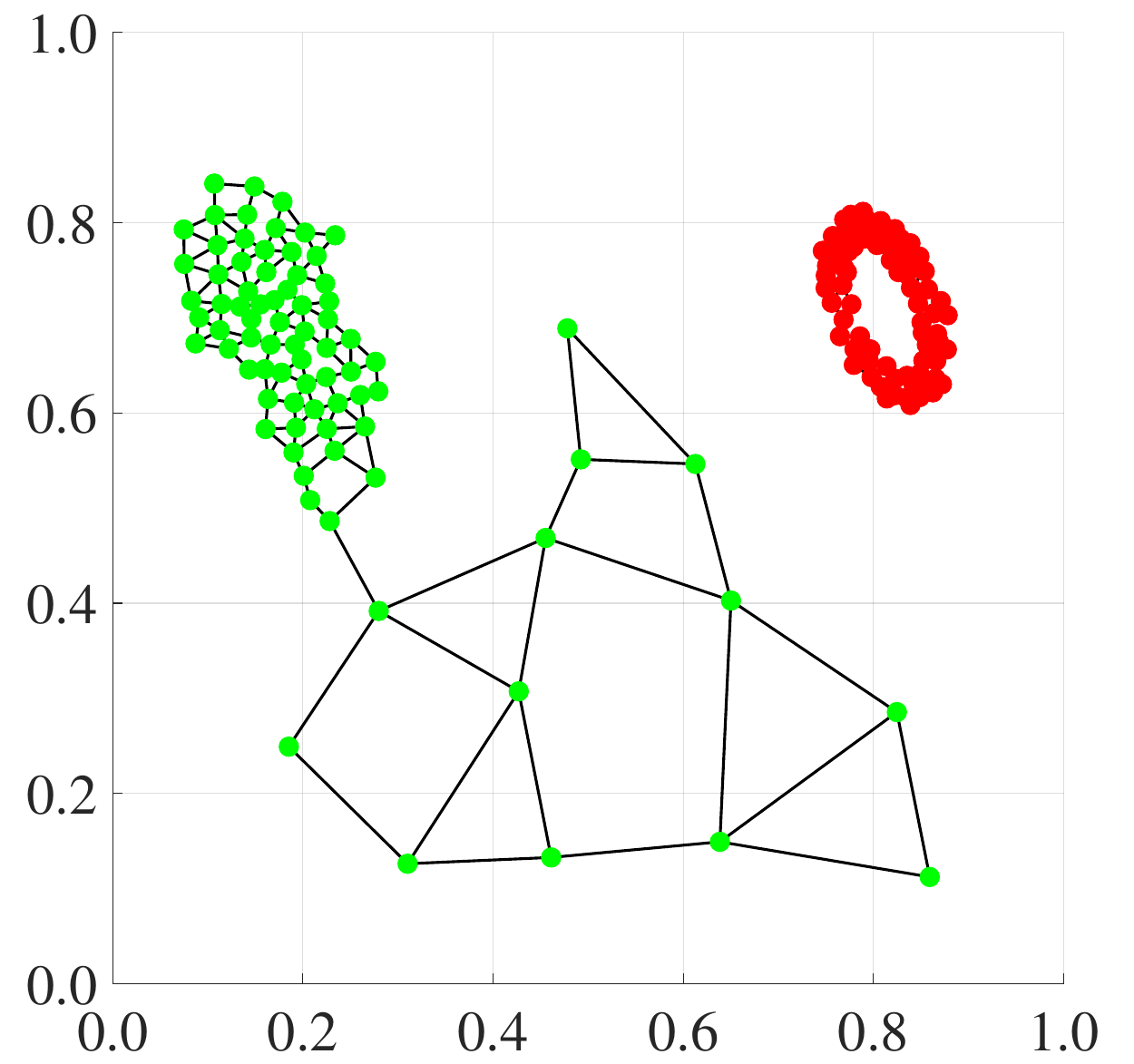}
		\label{fig:ASOINN_S1}
	}\hfil
	\subfloat[Stream \#2]{
		\includegraphics[width=0.75in]{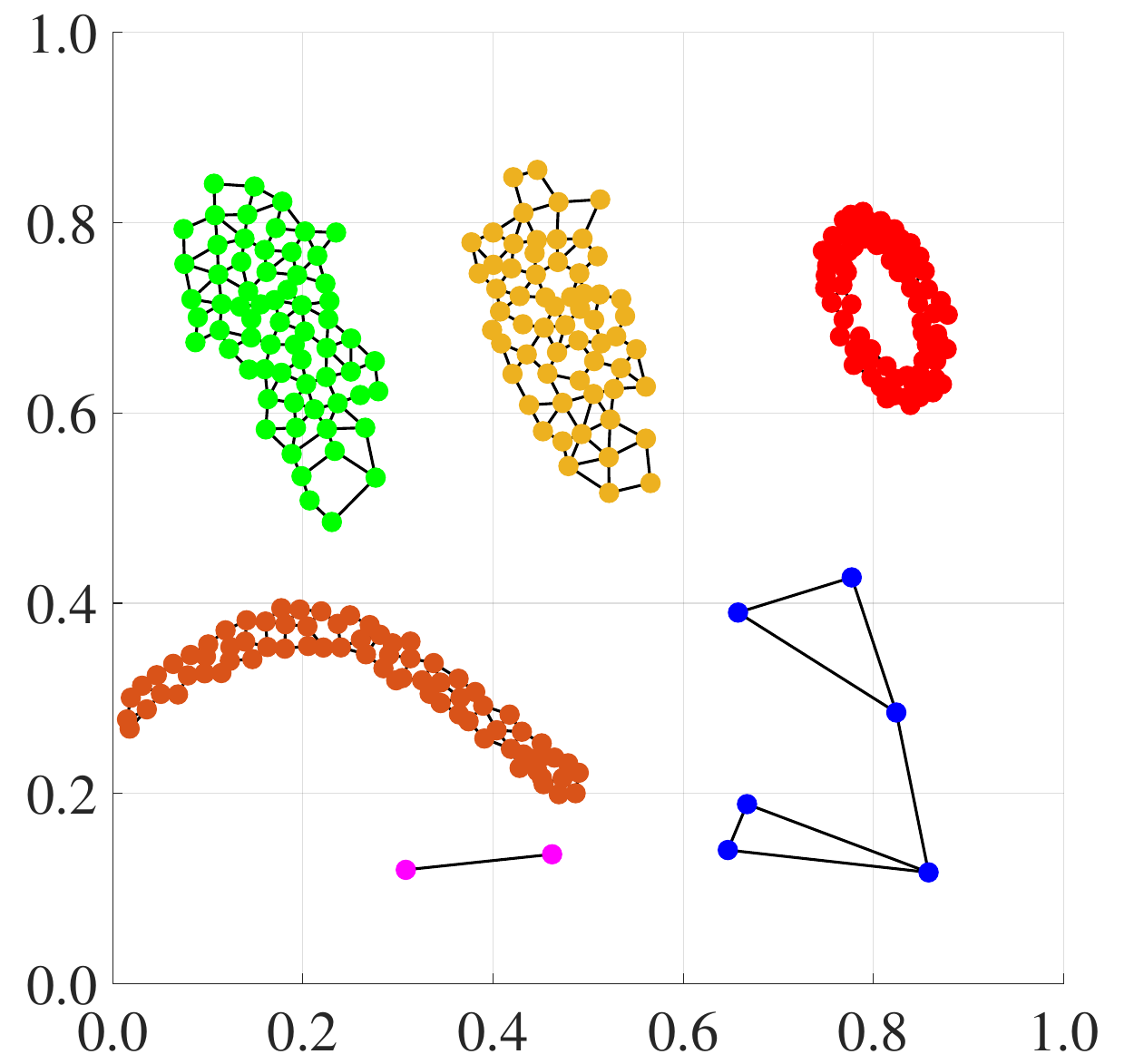}
		\label{fig:ASOINN_S2}
	}\hfil
	\subfloat[Stream \#3]{
		\includegraphics[width=0.75in]{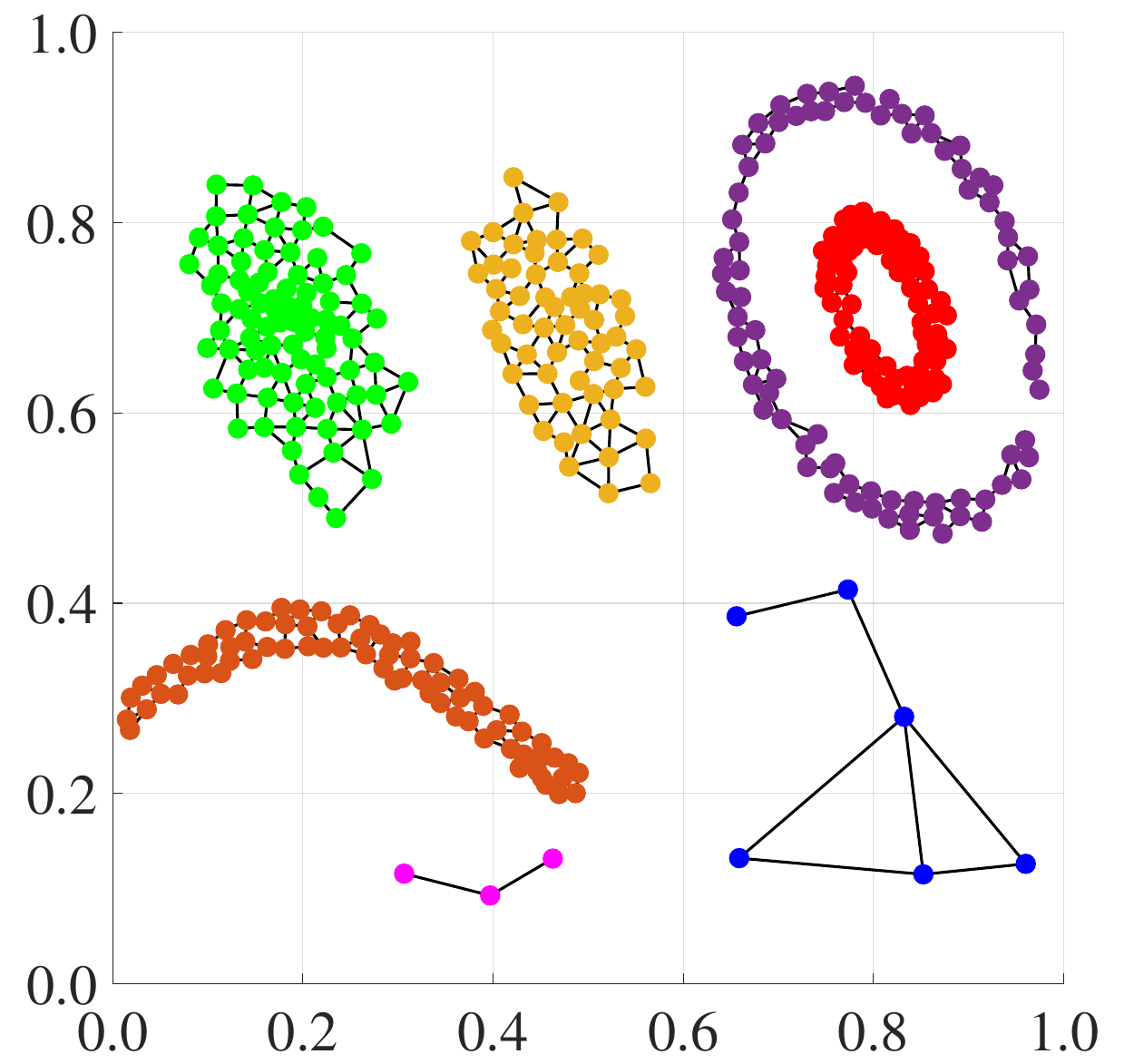}
		\label{fig:ASOINN_S3}
	}\hfil
	\subfloat[Stream \#4]{
		\includegraphics[width=0.75in]{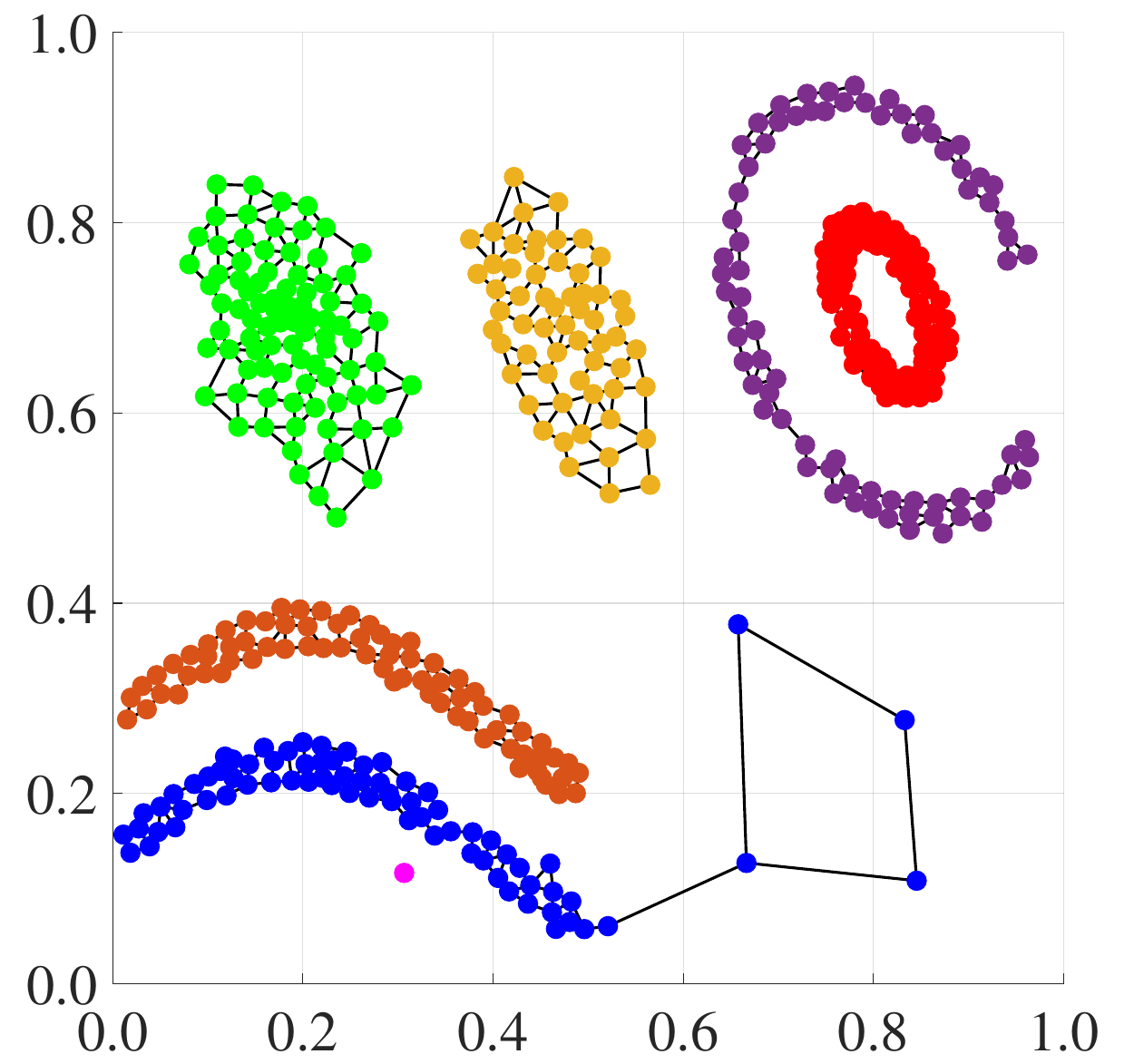}
		\label{fig:ASOINN_S4}
	}\hfil
	\subfloat[Stream \#5]{
		\includegraphics[width=0.75in]{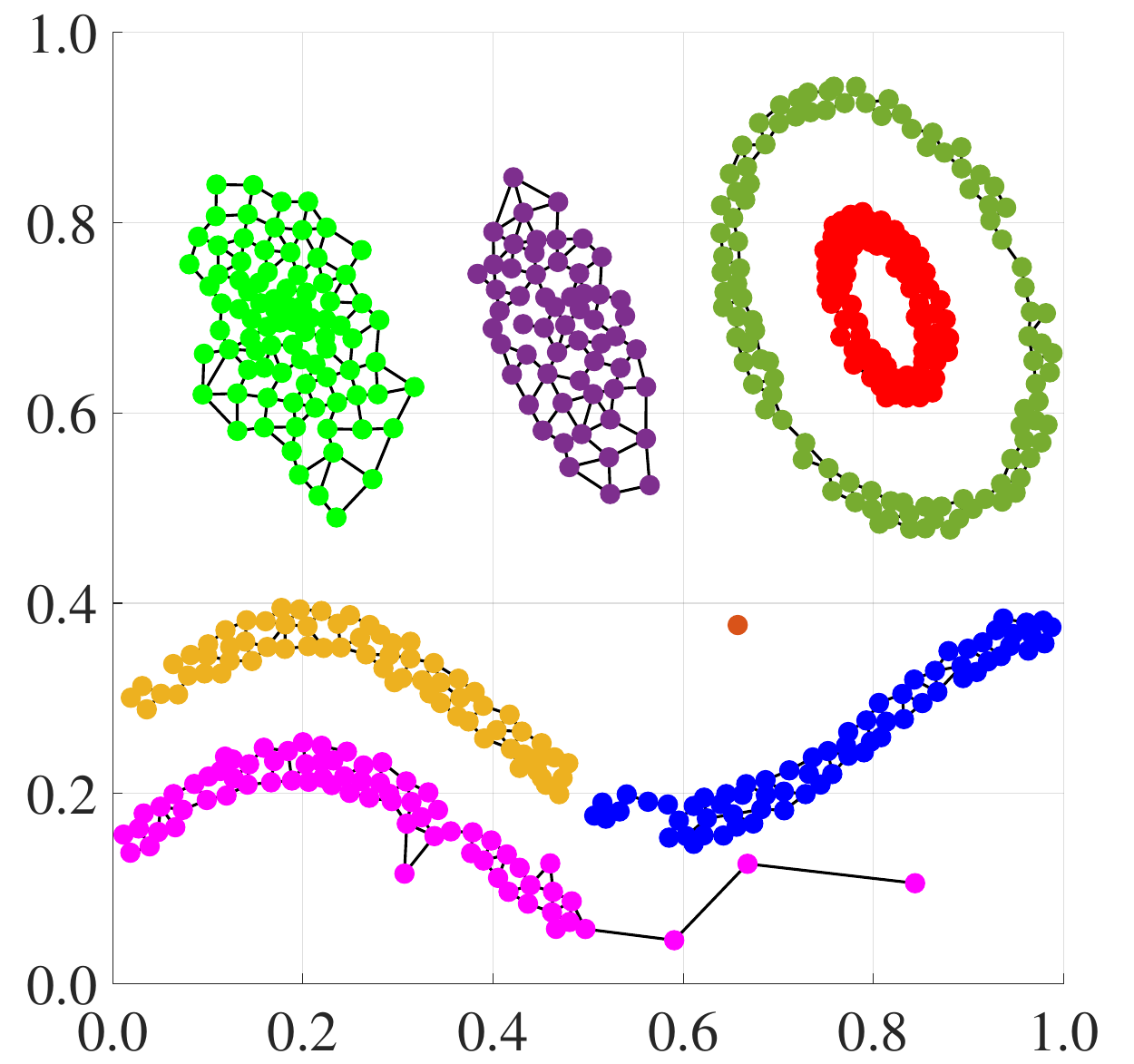}
		\label{fig:ASOINN_S5}
	}\hfil
	\subfloat[Stream \#6]{
		\includegraphics[width=0.75in]{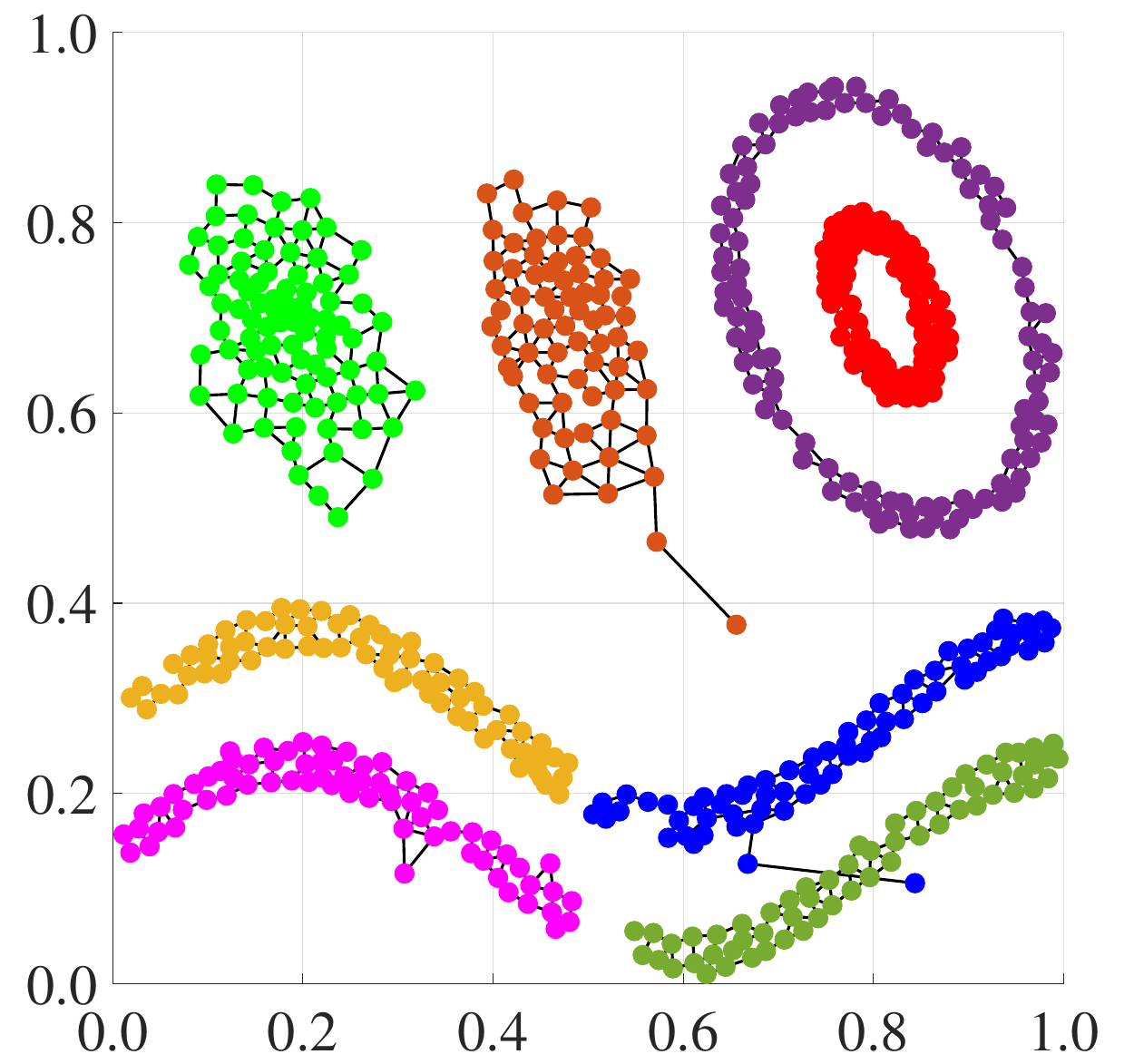}
		\label{fig:ASOINN_S6}
	}
	\caption{Visualization of self-organizing results of ASOINN in the non-stationary environment.}
	\label{fig:Nonstationary_2d_ASOINN}
\end{figure*}

\begin{figure*}[htbp]
	\centering
	\subfloat[Stream \#1]{
		\includegraphics[width=0.75in]{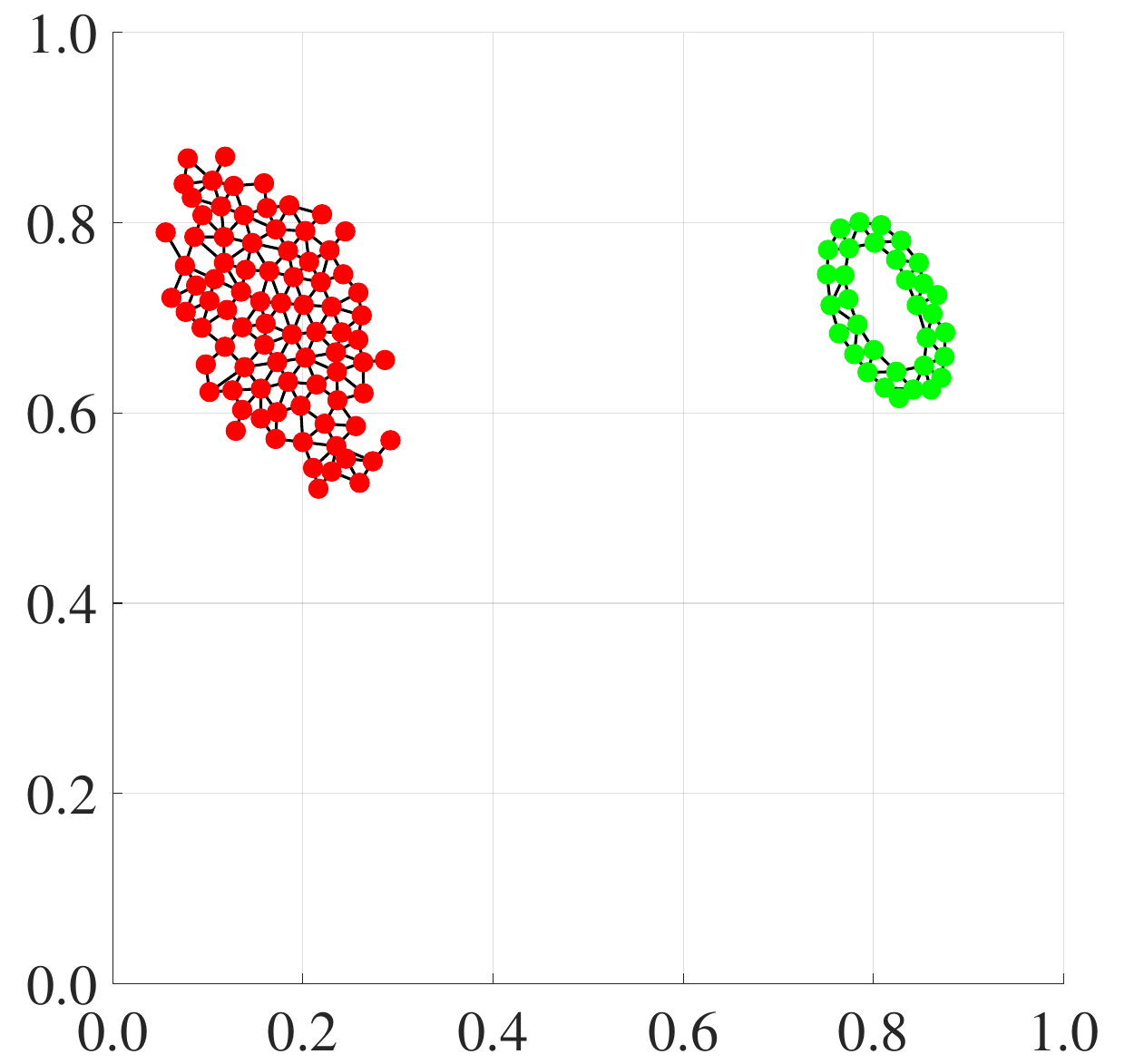}
		\label{fig:TCA_S1}
	}\hfil
	\subfloat[Stream \#2]{
		\includegraphics[width=0.75in]{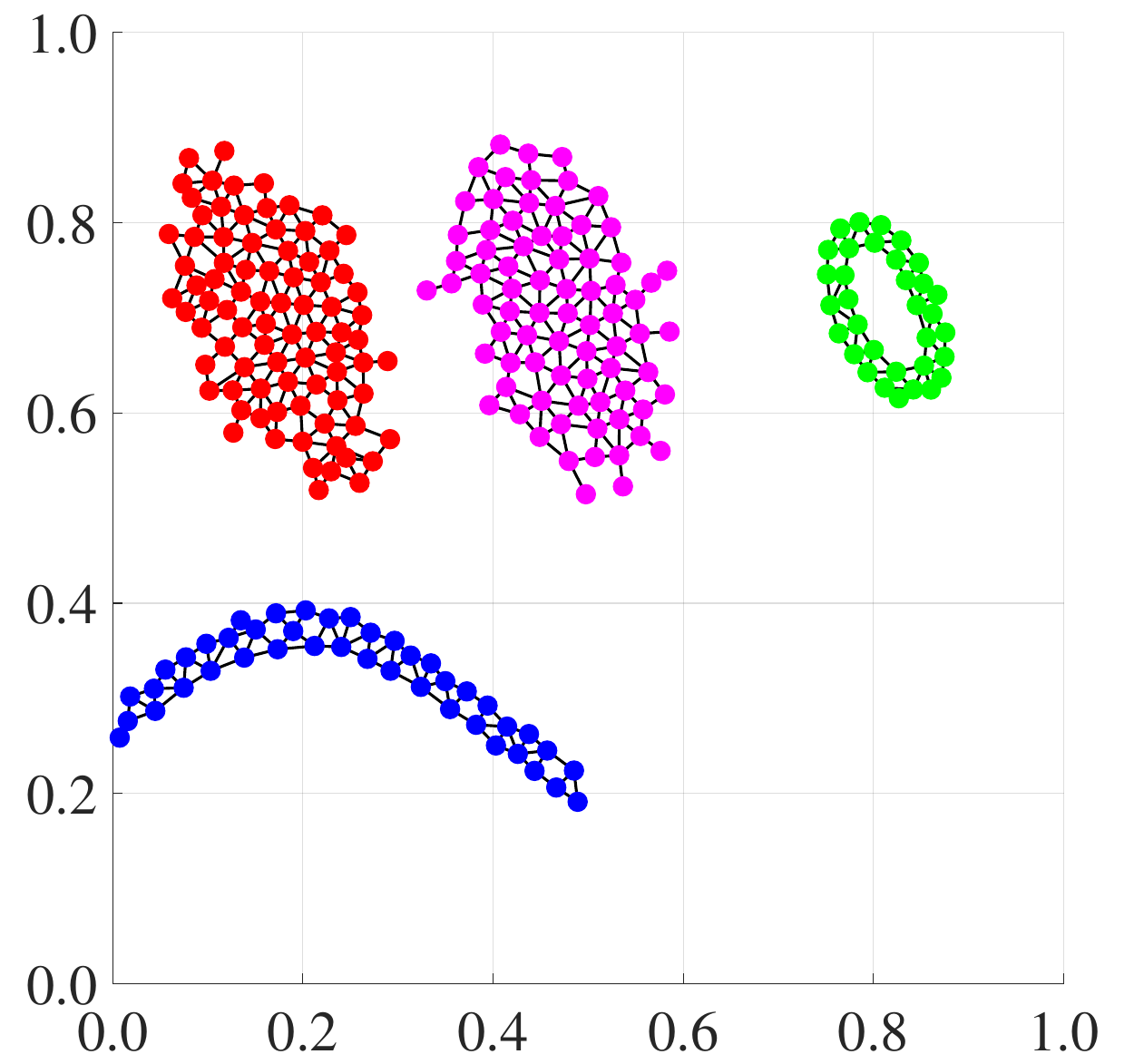}
		\label{fig:TCA_S2}
	}\hfil
	\subfloat[Stream \#3]{
		\includegraphics[width=0.75in]{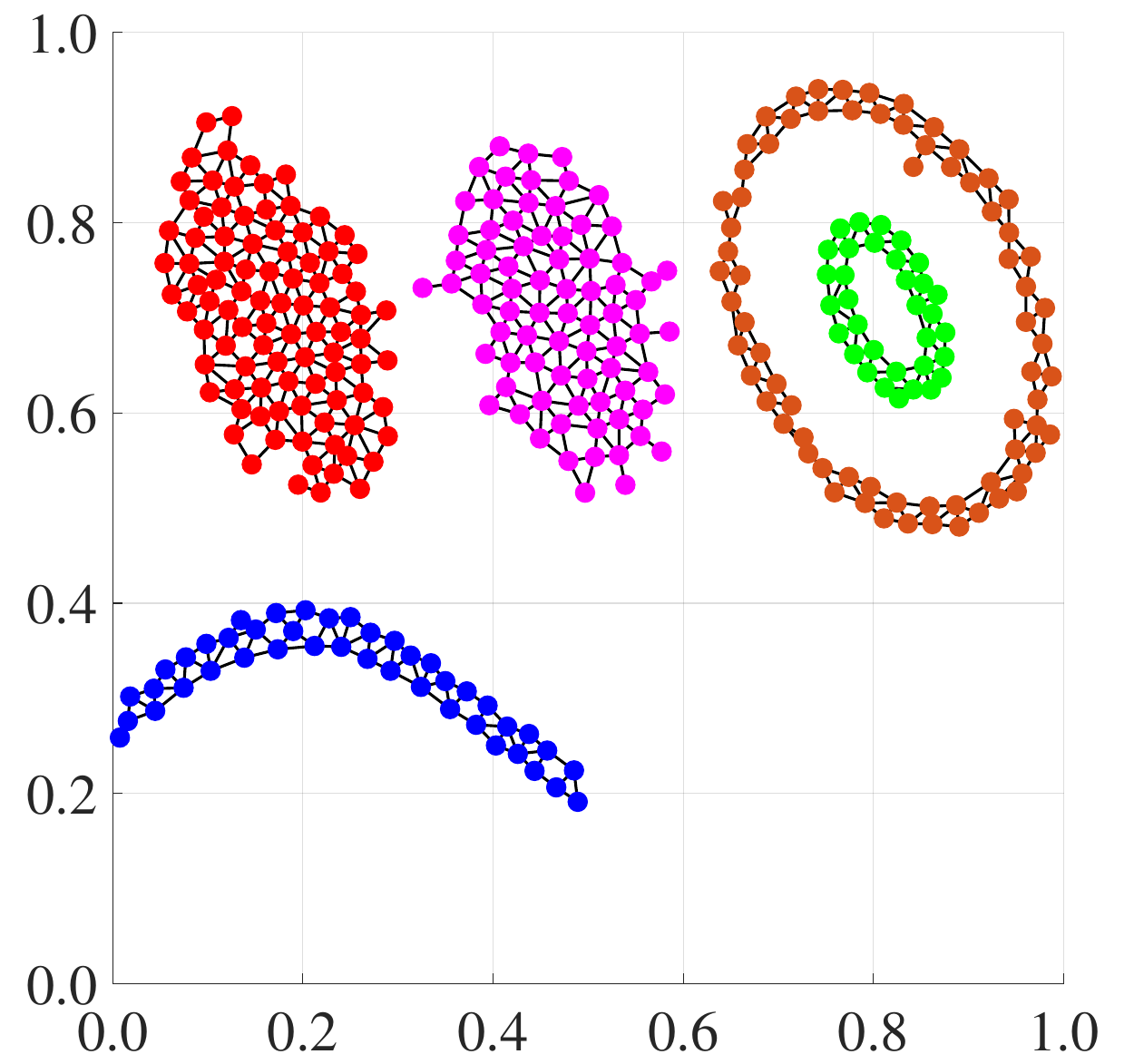}
		\label{fig:TCA_S3}
	}\hfil
	\subfloat[Stream \#4]{
		\includegraphics[width=0.75in]{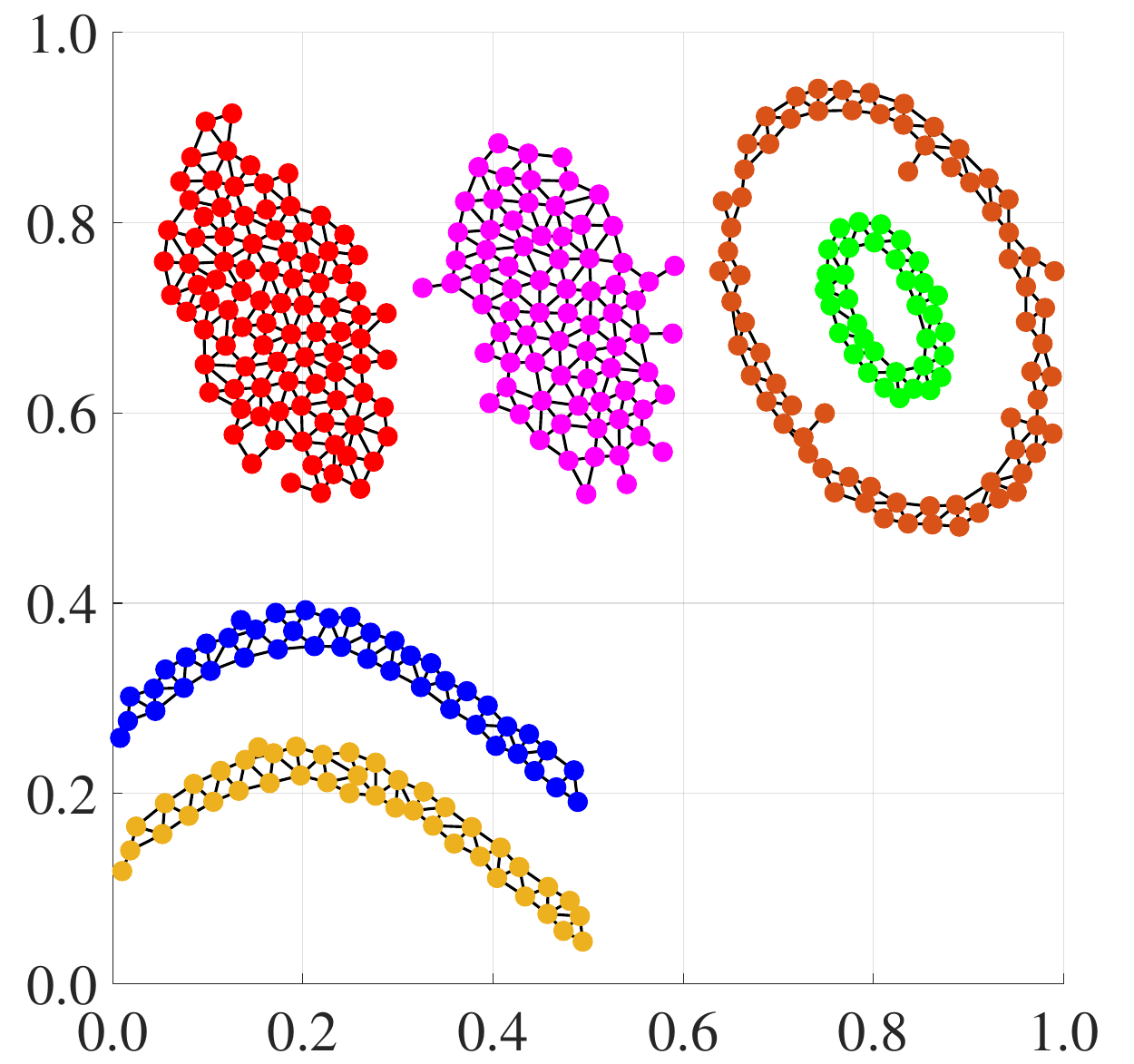}
		\label{fig:TCA_S4}
	}\hfil
	\subfloat[Stream \#5]{
		\includegraphics[width=0.75in]{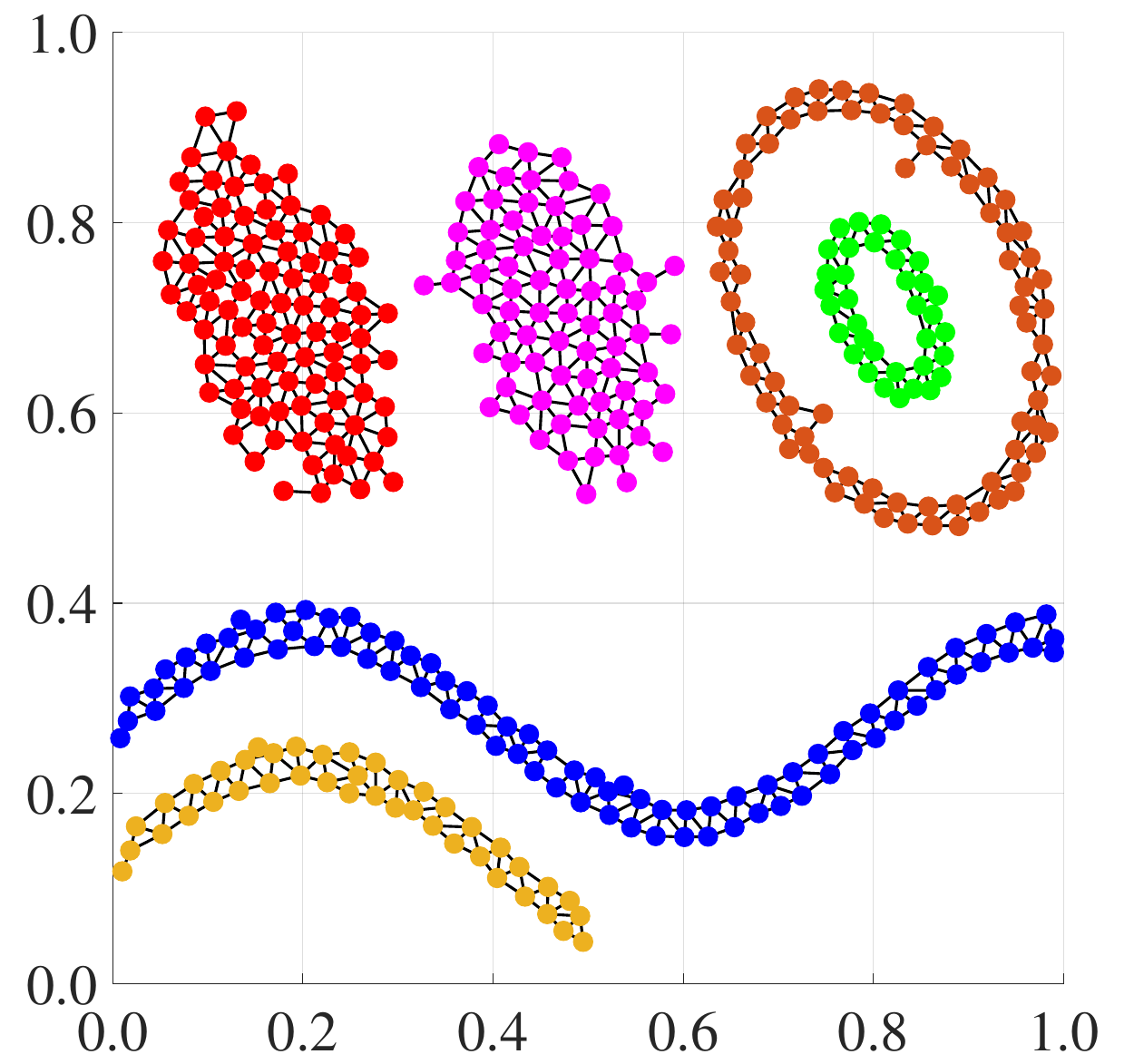}
		\label{fig:TCA_S5}
	}\hfil
	\subfloat[Stream \#6]{
		\includegraphics[width=0.75in]{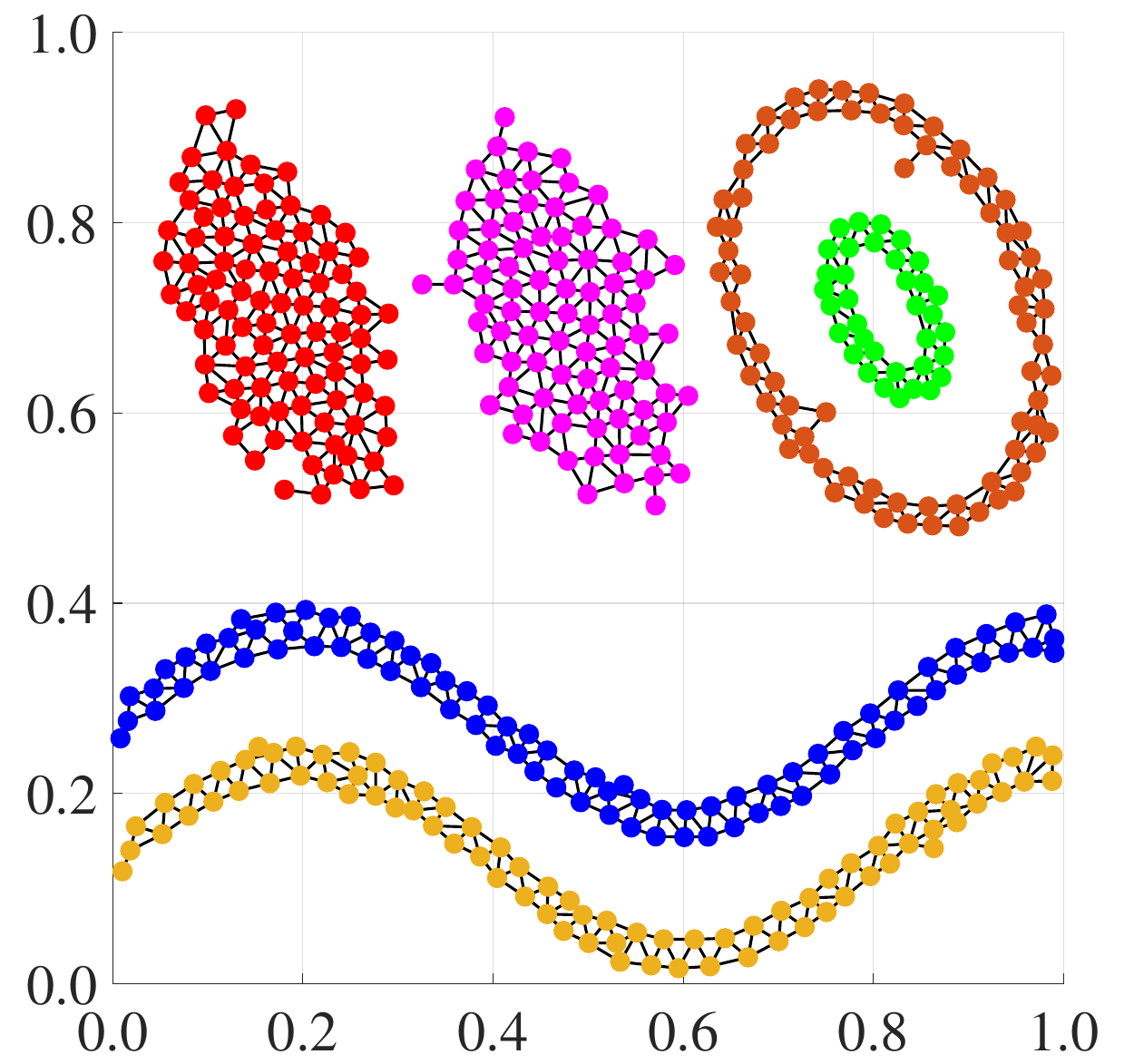}
		\label{fig:TCA_S6}
	}
	\caption{Visualization of self-organizing results of TCA in the non-stationary environment.}
	\label{fig:Nonstationary_2d_TCA}
\end{figure*}

\begin{figure*}[htbp]
	\vspace{-9mm}
	\centering
	\subfloat[Stream \#1]{
		\includegraphics[width=0.75in]{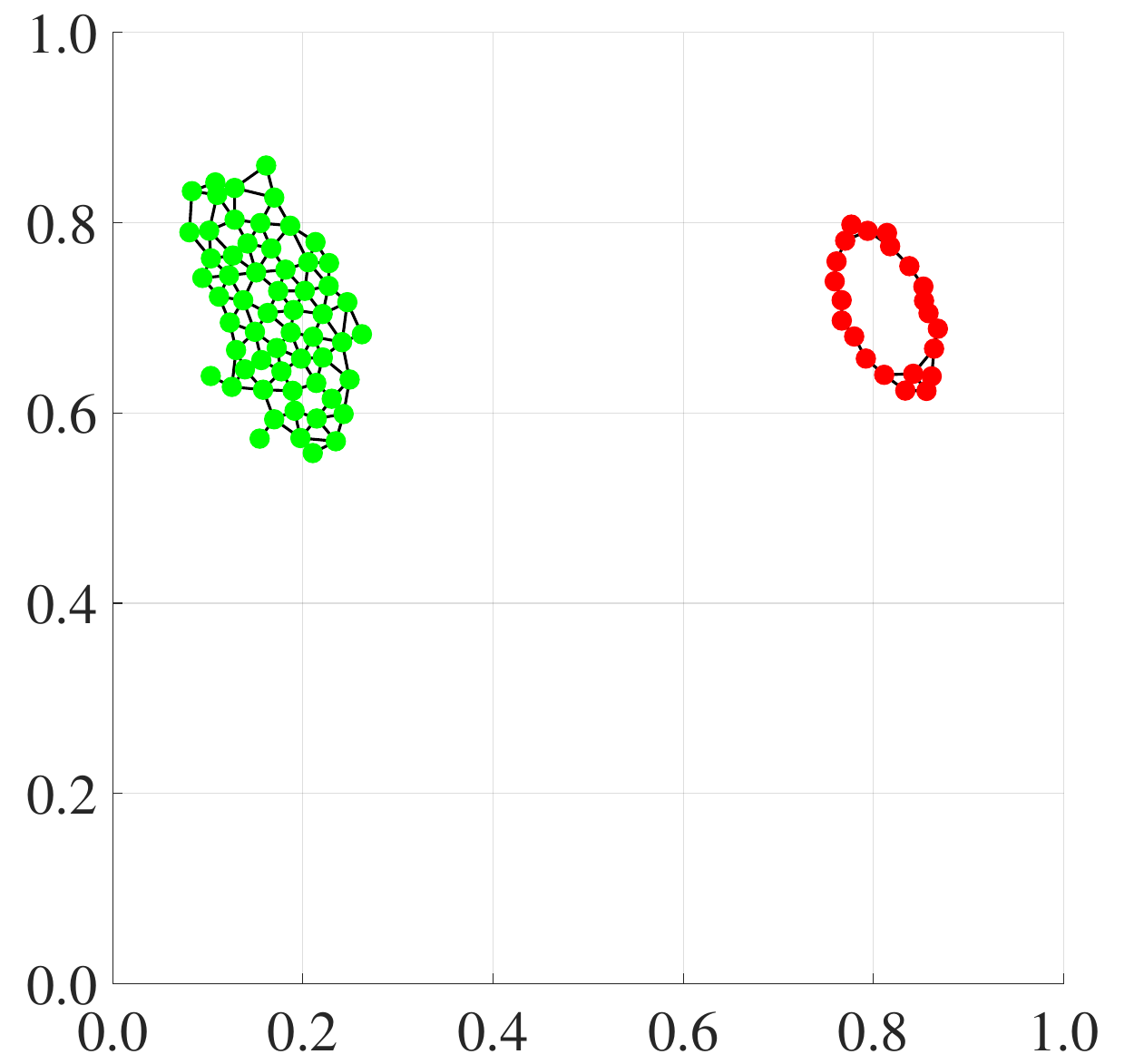}
		\label{fig:CAEA_S1}
	}\hfil
	\subfloat[Stream \#2]{
		\includegraphics[width=0.75in]{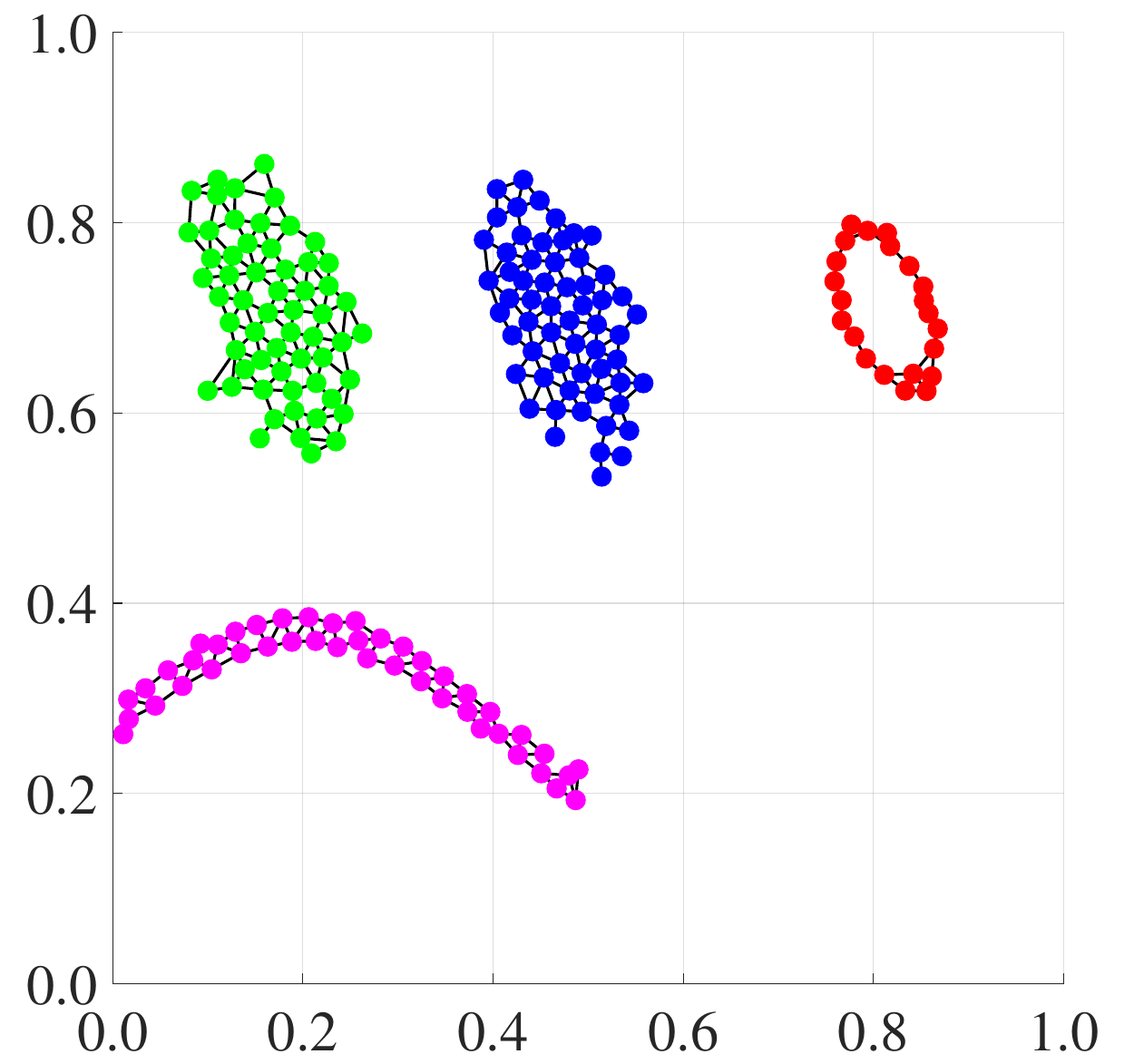}
		\label{fig:CAEA_S2}
	}\hfil
	\subfloat[Stream \#3]{
		\includegraphics[width=0.75in]{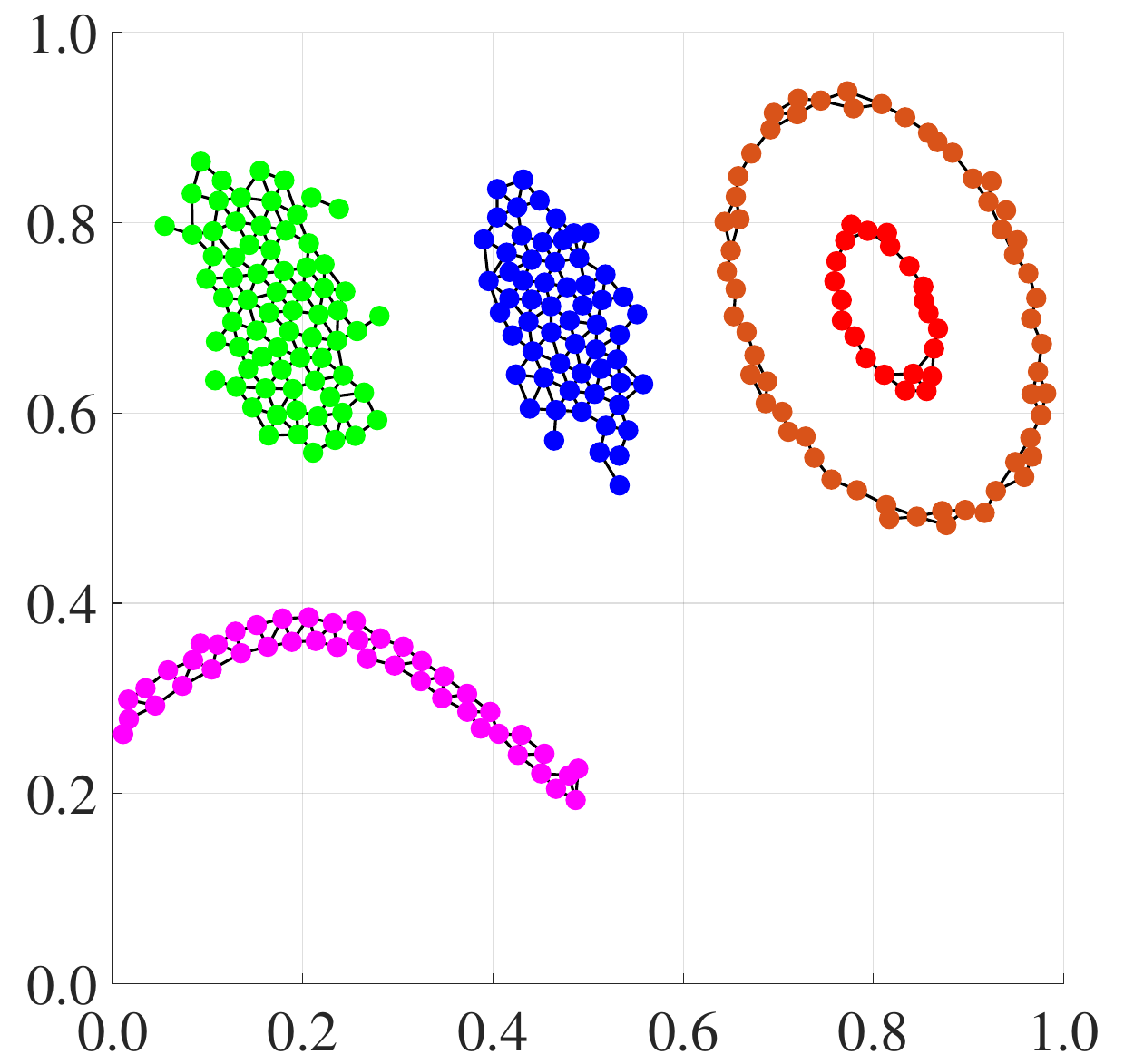}
		\label{fig:CAEA_S3}
	}\hfil
	\subfloat[Stream \#4]{
		\includegraphics[width=0.75in]{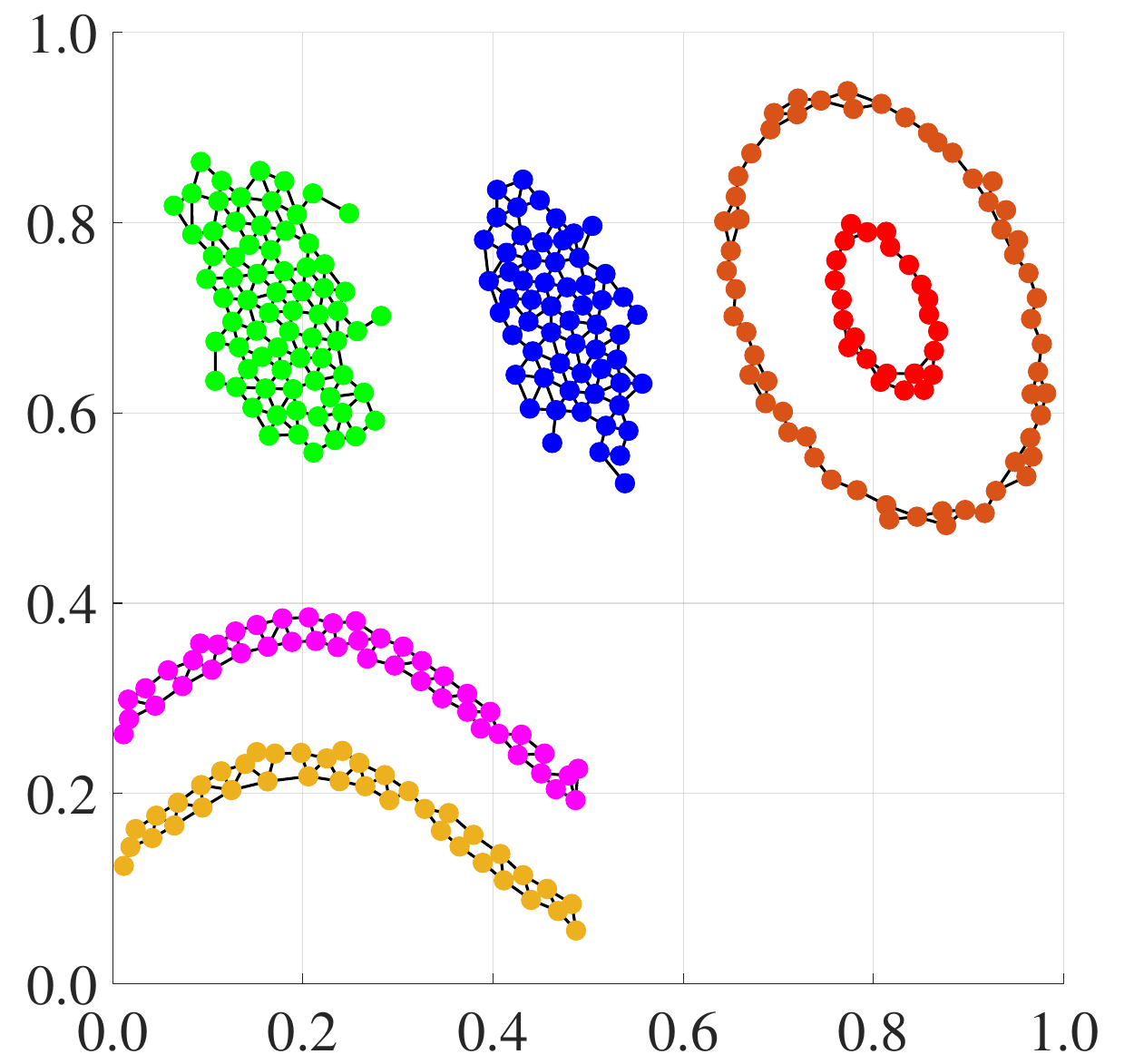}
		\label{fig:CAEA_S4}
	}\hfil
	\subfloat[Stream \#5]{
		\includegraphics[width=0.75in]{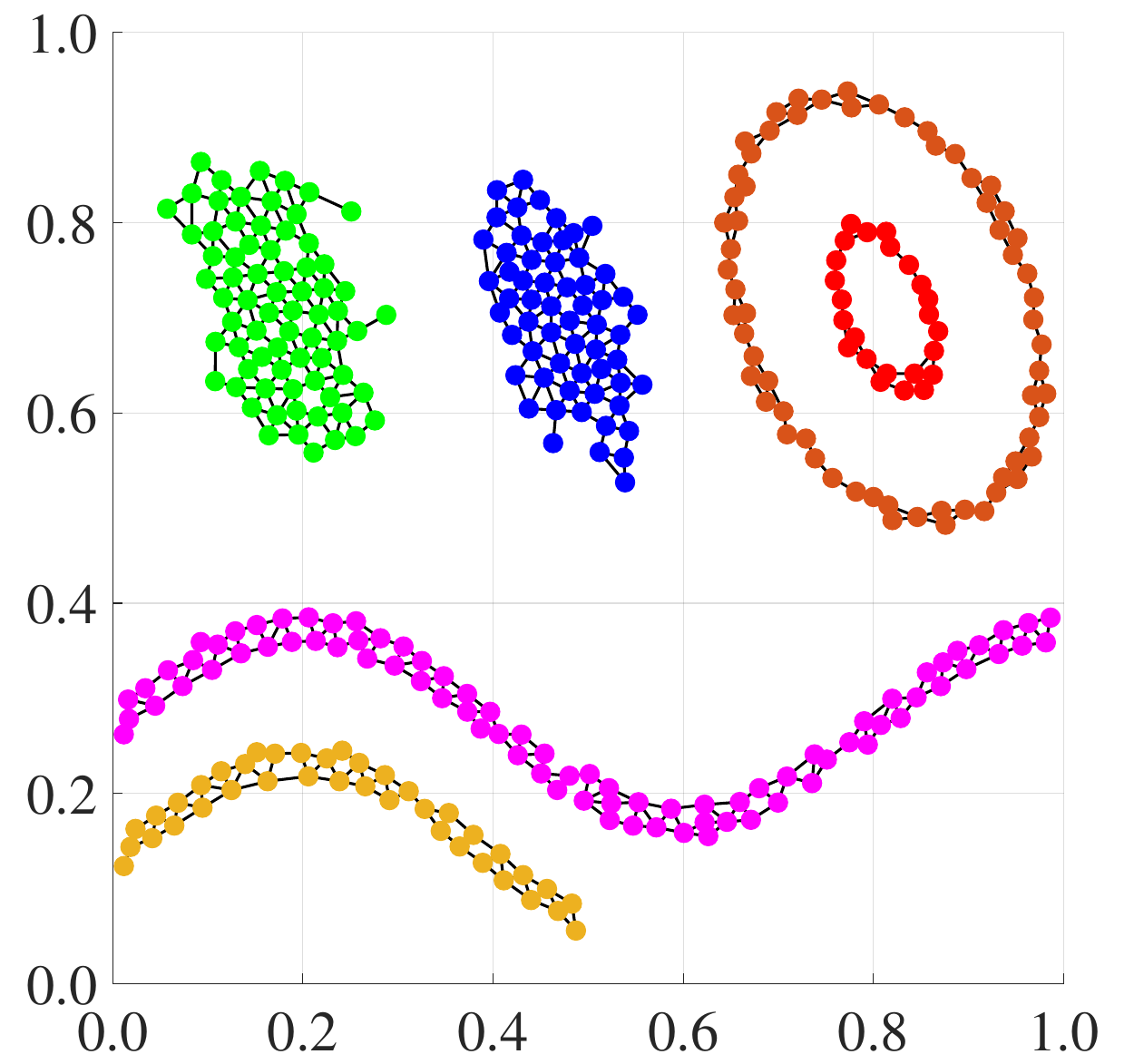}
		\label{fig:CAEA_S5}
	}\hfil
	\subfloat[Stream \#6]{
		\includegraphics[width=0.75in]{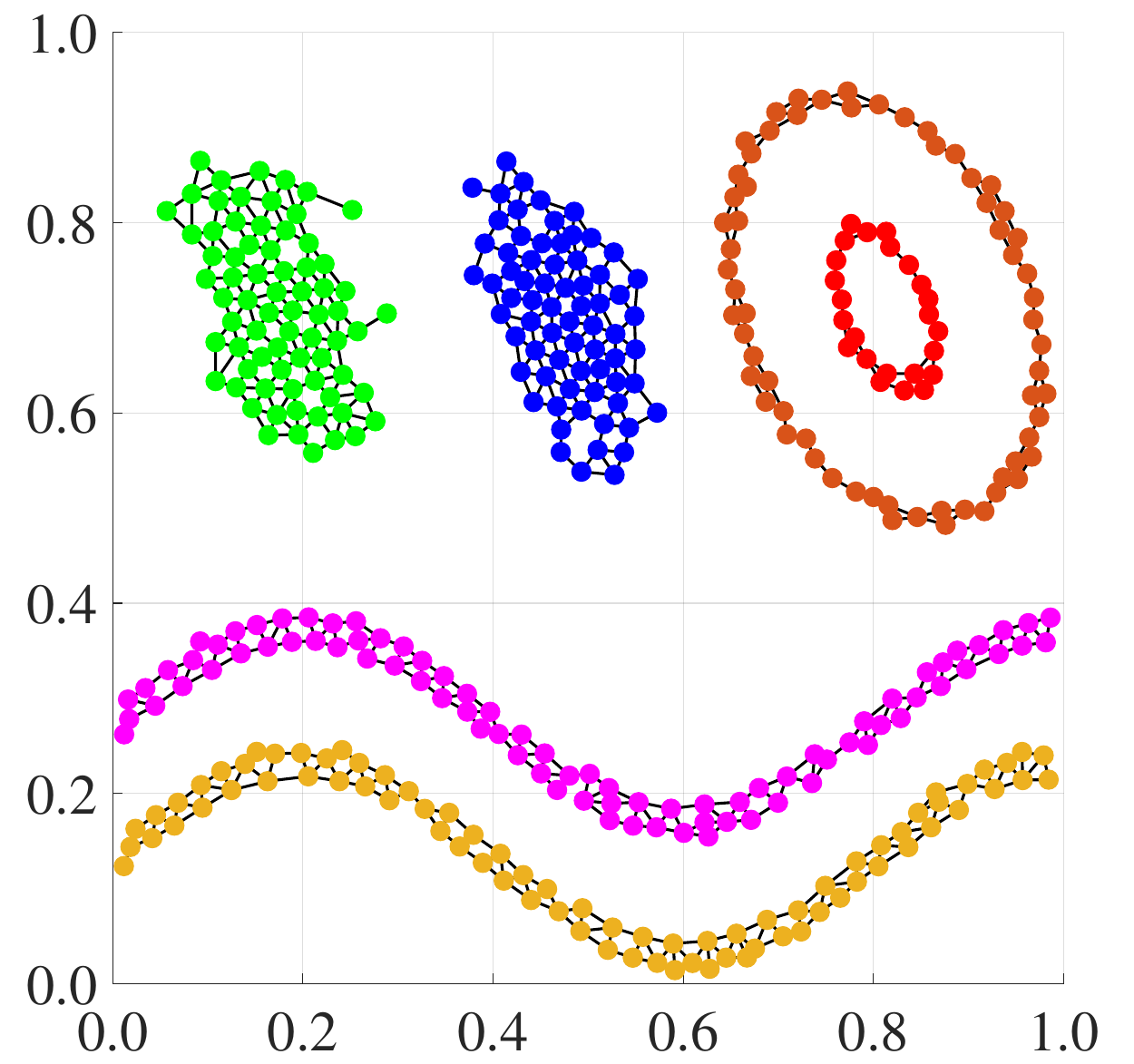}
		\label{fig:CAEA_S6}
	}
	\caption{Visualization of self-organizing results of CAEA in the non-stationary environment.}
	\label{fig:Nonstationary_2d_CAEA}
\end{figure*}

\begin{figure*}[htbp]
	\vspace{-9mm}
	\centering
	\subfloat[Stream \#1]{
		\includegraphics[width=0.75in]{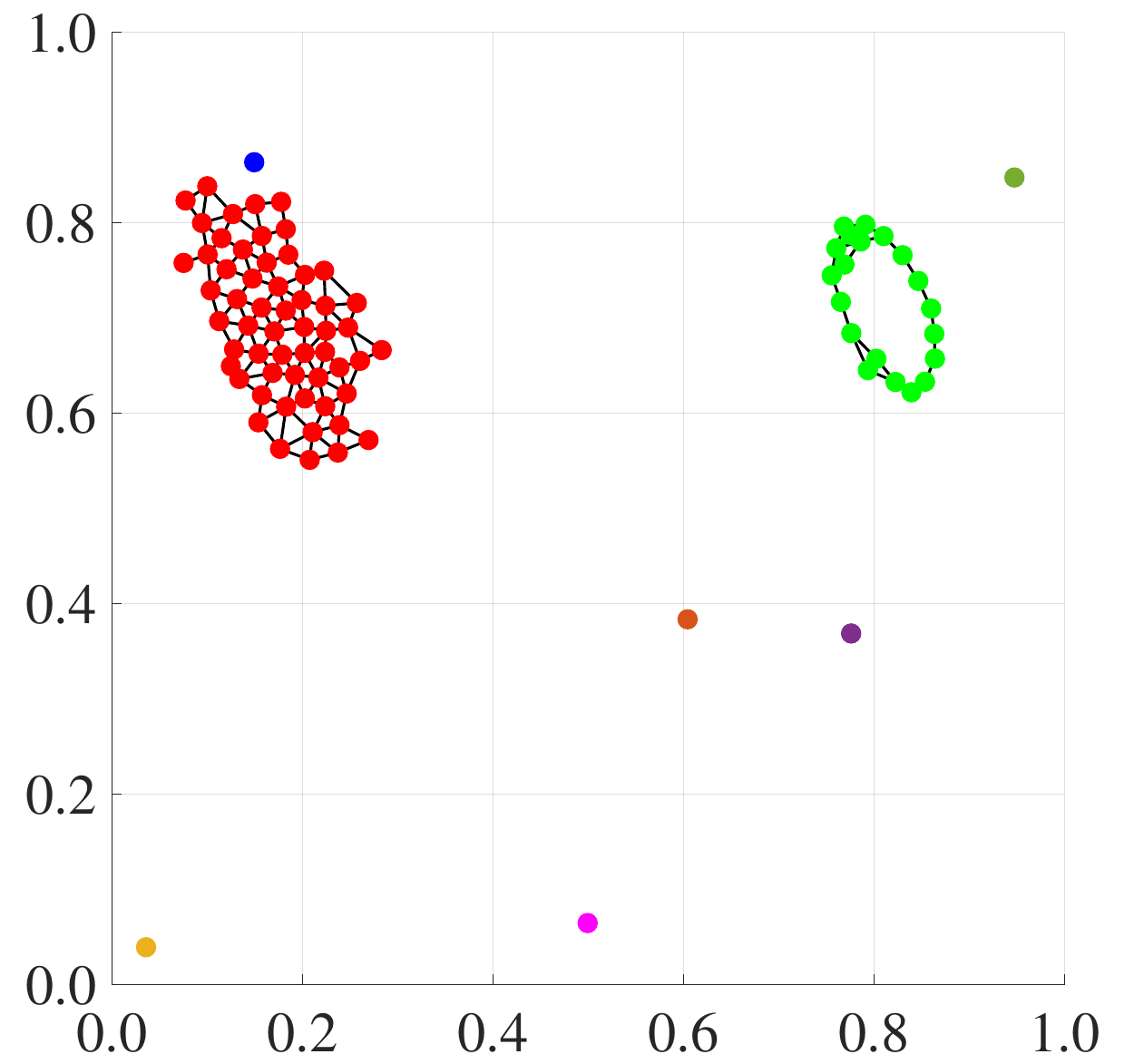}
		\label{fig:CAE_S1}
	}\hfil
	\subfloat[Stream \#2]{
		\includegraphics[width=0.75in]{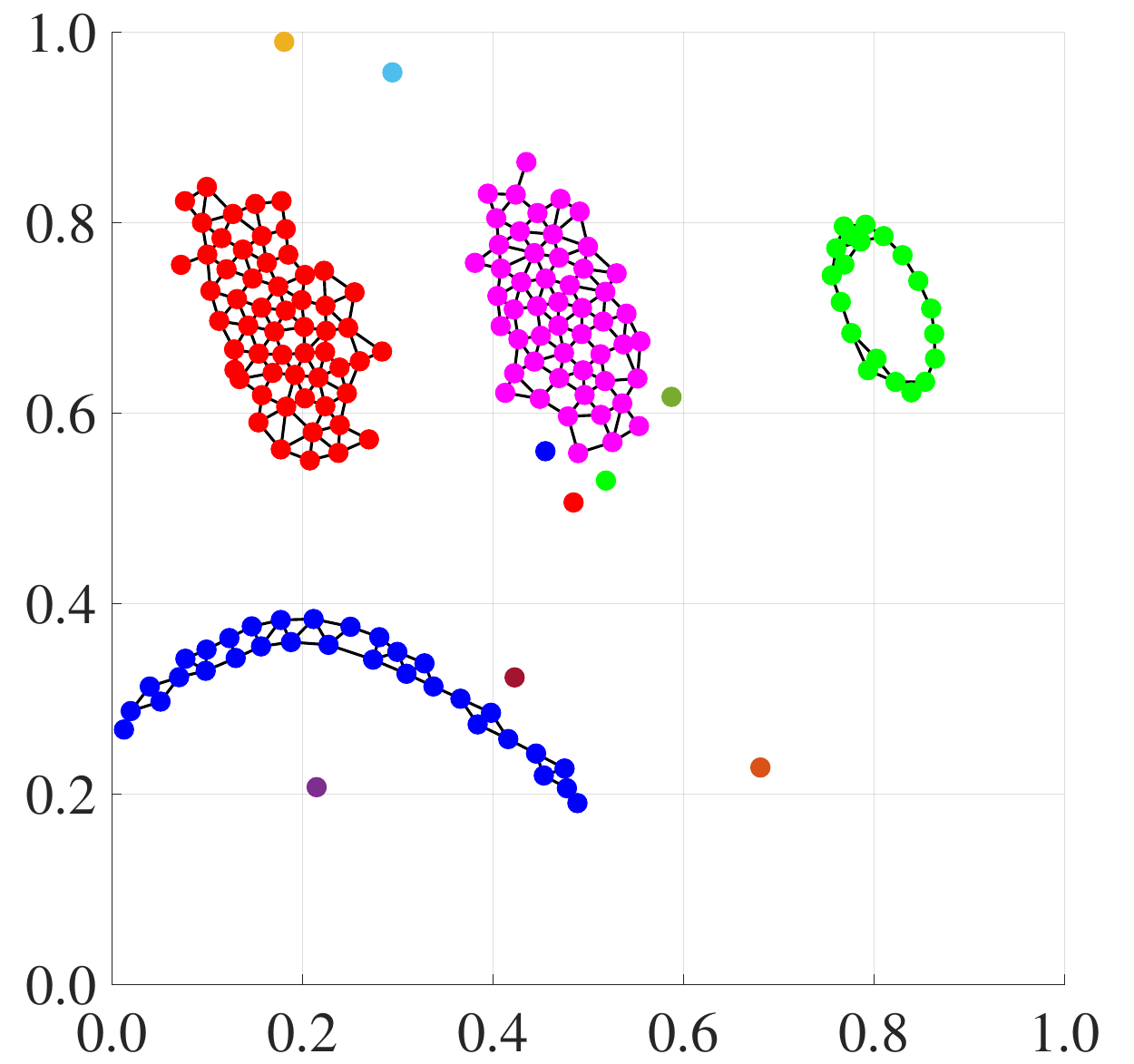}
		\label{fig:CAE_S2}
	}\hfil
	\subfloat[Stream \#3]{
		\includegraphics[width=0.75in]{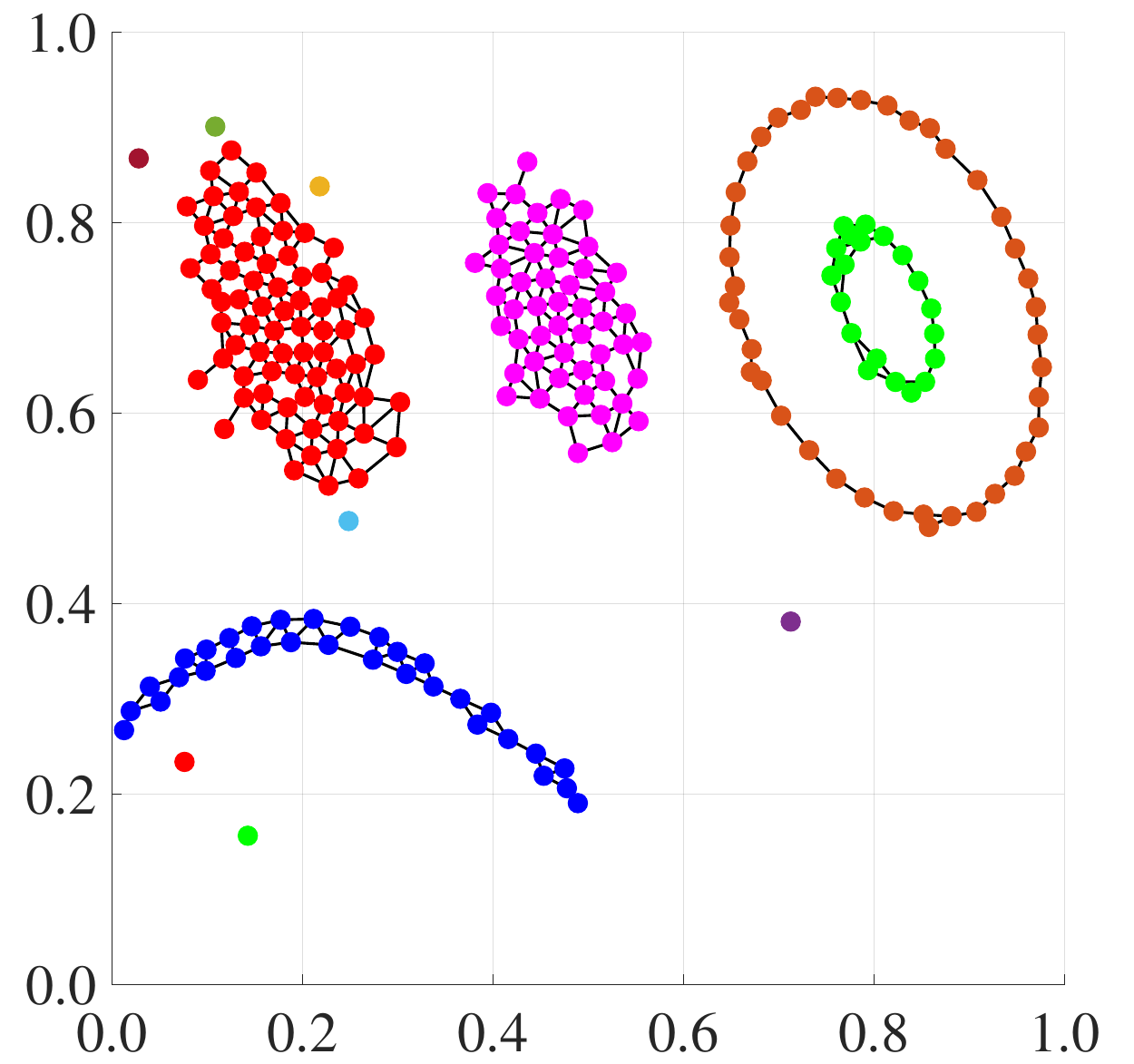}
		\label{fig:CAE_S3}
	}\hfil
	\subfloat[Stream \#4]{
		\includegraphics[width=0.75in]{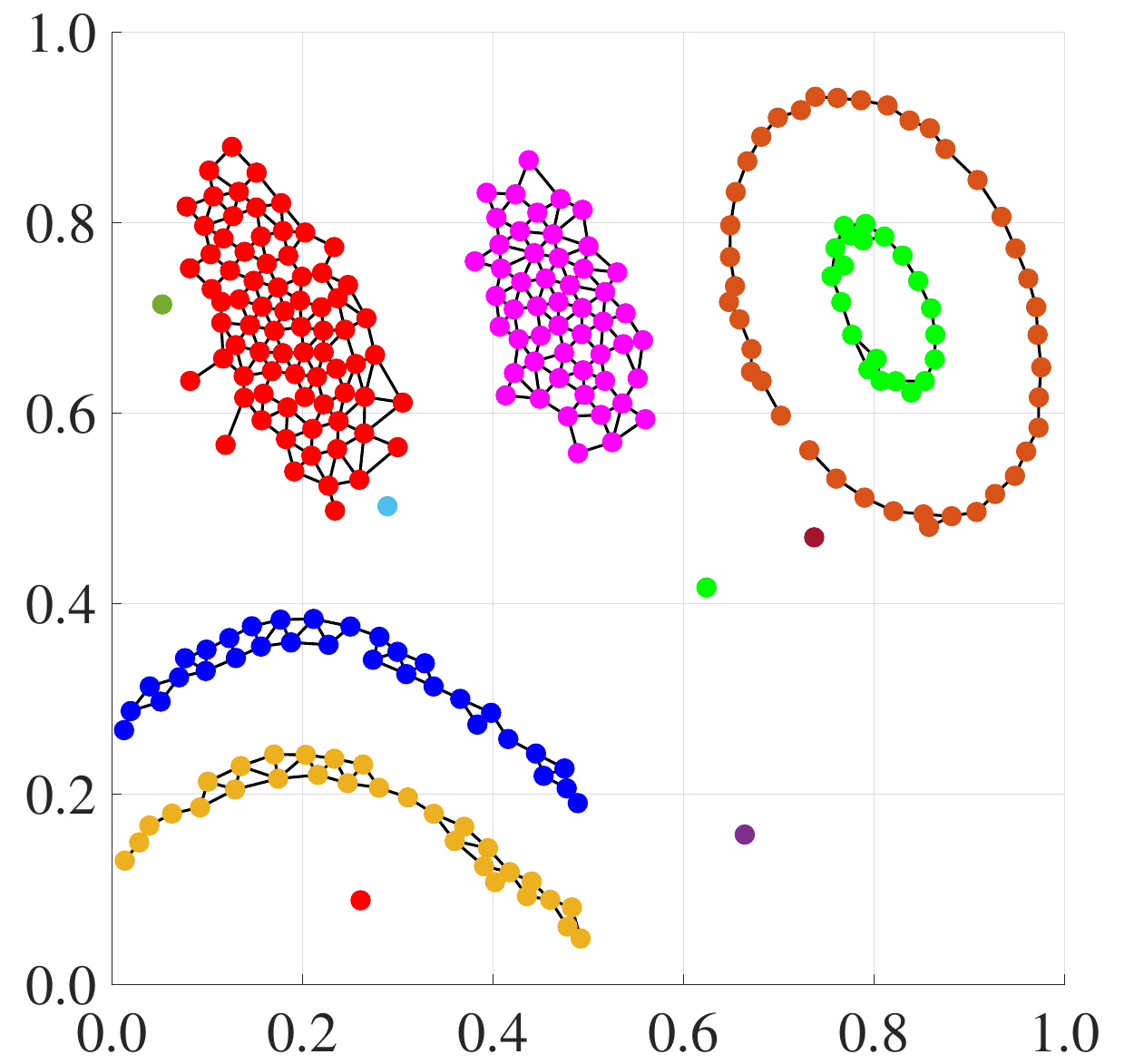}
		\label{fig:CAE_S4}
	}\hfil
	\subfloat[Stream \#5]{
		\includegraphics[width=0.75in]{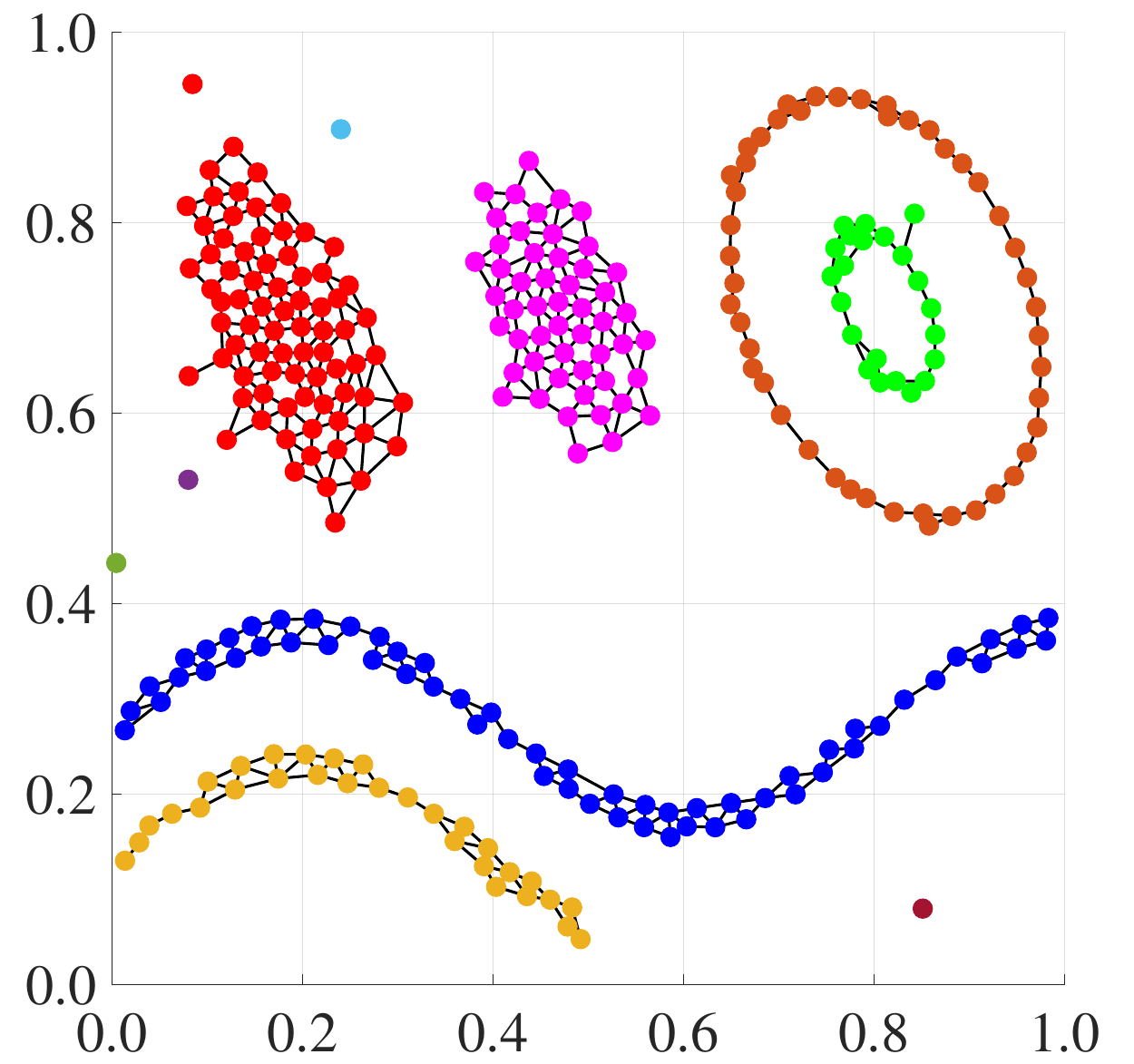}
		\label{fig:CAE_S5}
	}\hfil
	\subfloat[Stream \#6]{
		\includegraphics[width=0.75in]{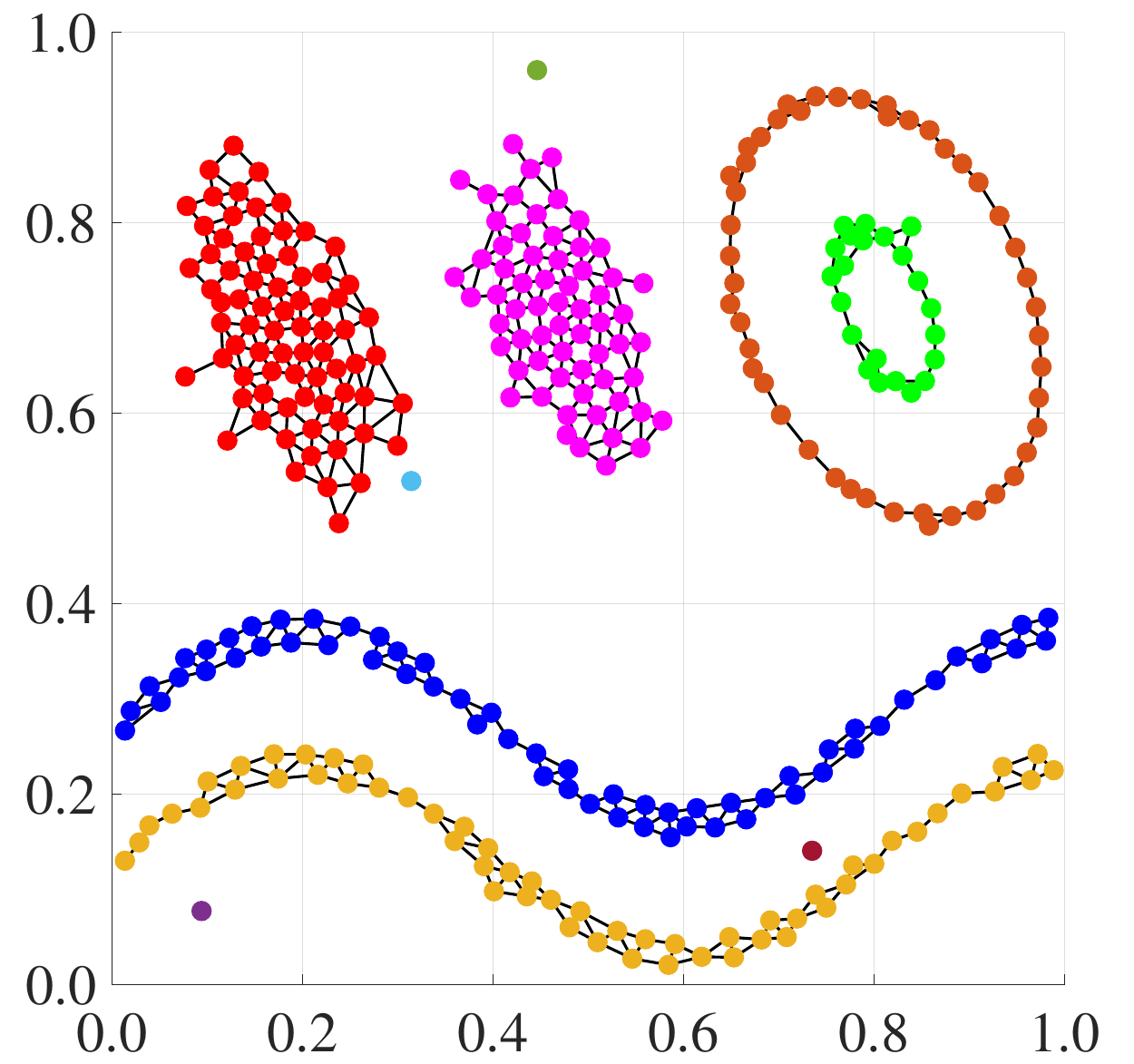}
		\label{fig:CAE_S6}
	}
	\caption{Visualization of self-organizing results of CAE in the non-stationary environment.}
	\label{fig:Nonstationary_2d_CAE}
\end{figure*}

Figs. \ref{fig:Nonstationary_2d_AutoCloud}-\ref{fig:Nonstationary_2d_CAE} show the visualization of self-organizing results in the non-stationary environment. These figures correspond to the trials where the median NMI was achieved. For AutoCloud, clustering performance improves compared to the stationary environment. However, AutoCloud has difficulty handling non-Gaussian distributions. ASOINN tends to generate topological networks from noisy data points and typically produces more nodes than it does in the stationary environment. For SOINN+, two clusters eventually connect, and similar to ASOINN, SOINN+ also tends to generate a greater number of nodes than in the stationary setting.

In contrast, TCA, CAEA, and CAE produce topological networks that are similar to those in the stationary environment. In Fig. \ref{fig:Nonstationary_2d_CAE}, although CAE creates several isolated nodes, these isolated nodes do not overlap with the main topological networks. On the other hand, as seen in Fig. \ref{fig:Nonstationary_2d_SOINNplus}, SOINN+ forms isolated nodes that overlap with its topological networks, indicating that the networks are not properly generated.

Tables \ref{tab:synthetic_quantitative_stationary} and \ref{tab:synthetic_quantitative_nonstationary} present the quantitative results for the synthetic dataset in stationary and non-stationary environments, respectively. We compare the mean values of NMI, Adjusted Rand Index (ARI) \cite{hubert85}, and the number of nodes and clusters produced by each algorithm over 15 runs. The results based on NMI and ARI (in Tables \ref{tab:synthetic_quantitative_stationary} and \ref{tab:synthetic_quantitative_nonstationary}) indicate that clustering performance is comparable among the algorithms in both environments. However, the number of nodes generated by ASOINN and SOINN+ varies significantly depending on the environment, while TCA, CAEA, and CAE produce a similar number of nodes in each case.

The above-mentioned observations of the results in this section indicate that the stability of the self-organizing performance of TCA, CAEA, and CAE is superior to that of AutoCloud, ASOINN, and SOINN+.

\begin{table*}[htbp]
	\vspace{10mm}
	\caption{Results of quantitative comparisons on the synthetic dataset in the stationary environment}
	\label{tab:synthetic_quantitative_stationary}
	\footnotesize
	\centering
	\renewcommand{\arraystretch}{1.2}
  \resizebox{\textwidth}{!}{
		\begin{tabular}{l|r|r|r|r|r|r}
			\hline\hline
			Metric & \multicolumn{1}{c|}{AutoCloud} & \multicolumn{1}{c|}{ASOINN} & \multicolumn{1}{c|}{SOINN+} & \multicolumn{1}{c|}{TCA} & \multicolumn{1}{c|}{CAEA} & \multicolumn{1}{c}{CAE} \\
			\hline
			NMI & 0.000 (0.000) & 0.977 (0.022) & 0.905 (0.077) & 0.998 (0.000) & 0.998 (0.000) & 0.907 (0.254) \\
						ARI & 0.000 (0.000) & 0.959 (0.058) & 0.811 (0.178) & 0.999 (0.000) & 0.999 (0.000) & 0.868 (0.270) \\
						\# of  Nodes & \multicolumn{1}{c|}{---} & 373.5 (14.5) & 451.7 (88.5)& 488.3 (6.8) & 432.3 (82.3) & 243.1 (34.4)\\
						\# of  Clusters & 2.0 (0.0) & 6.4 (0.8) & 10.8 (3.4) & 6.1 (0.3) & 6.0 (0.0)& 7.0 (2.4) \\
			\hline\hline
		\end{tabular}
	}
	\\
	\vspace{1mm}
	\footnotesize \raggedright
	\hspace{-55.5mm} The values in parentheses indicate the standard deviation.
\end{table*}

\begin{table*}[htbp]
	\vspace{8mm}
	\caption{Results of quantitative comparisons on the synthetic dataset in the non-stationary environment}
	\label{tab:synthetic_quantitative_nonstationary}
	\footnotesize
	\centering
	\renewcommand{\arraystretch}{1.2}
	\resizebox{\textwidth}{!}{
		\begin{tabular}{ll|r|r|r|r|r|r}
  \hline\hline
Dataset    & Metric          & \multicolumn{1}{c|}{AutoCloud} & \multicolumn{1}{c|}{ASOINN} & \multicolumn{1}{c|}{SOINN+} & \multicolumn{1}{c|}{TCA} & \multicolumn{1}{c|}{CAEA} & \multicolumn{1}{c}{CAE} \\ \hline
Stream \#1 & NMI             & 0.067 (0.258)                 & 0.800 (0.414)              & 0.980 (0.018)              & 0.998 (0.006)           & 1.000 (0.000)            & 0.987 (0.015)           \\
           & ARI             & 0.067 (0.258)                 & 0.800 (0.414)              & 0.990 (0.010)              & 0.999 (0.003)           & 1.000 (0.000)            & 0.993 (0.008)           \\
           & \# of Nodes    & \multicolumn{1}{c|}{---}                             & 243.4 (16.1)               & 150.9 (48.6)               & 125.1 (4.1)             & 85.2 (11.9)              & 86.2 (12.7)             \\
           & \# of Clusters & 1.2 (0.8)                     & 1.8 (0.4)                  & 5.8 (3.1)                  & 2.1 (0.4)               & 2.0 (0.0)                & 5.5 (2.7)               \\ \hline
Stream \#2 & NMI             & 0.540 (0.183)                 & 0.806 (0.132)              & 0.967 (0.050)              & 0.997 (0.001)           & 0.995 (0.004)            & 0.992 (0.005)           \\
           & ARI             & 0.279 (0.109)                 & 0.625 (0.235)              & 0.953 (0.100)              & 0.999 (0.001)           & 0.997 (0.003)            & 0.996 (0.003)           \\
           & \# of Nodes    & \multicolumn{1}{c|}{---}                             & 450.2 (27.5)               & 339.1 (111.2)              & 255.5 (10.2)            & 190.2 (24.0)             & 180.1 (28.2)            \\
           & \# of Clusters & 4.0 (1.5)                     & 2.9 (0.7)                  & 9.4 (4.8)                  & 4.1 (0.4)               & 4.3 (0.5)                & 7.1 (2.6)               \\ \hline
Stream \#3 & NMI             & 0.467 (0.119)                 & 0.856 (0.093)              & 0.936 (0.080)              & 0.998 (0.000)           & 0.992 (0.010)            & 0.991 (0.013)           \\
           & ARI             & 0.176 (0.075)                 & 0.687 (0.204)              & 0.876 (0.174)              & 0.999(0.000)            & 0.994 (0.011)            & 0.993 (0.013)           \\
           & \# of Nodes    & \multicolumn{1}{c|}{---}                             & 598.7 (30.6)               & 504.4 (151.4)              & 344.5 (16.6)            & 263.1 (37.2)             & 242.7 (44.3)            \\
           & \# of Clusters & 4.7 (0.5)                     & 3.8 (0.9)                  & 12.0 (4.5)                 & 5.1 (0.3)               & 5.5 (0.6)                & 8.0 (2.6)               \\ \hline
Stream \#4 & NMI             & 0.465 (0.110)                 & 0.915 (0.076)              & 0.951 (0.053)              & 0.998 (0.000)           & 0.995(0.007)             & 0.994 (0.009)           \\
           & ARI             & 0.219 (0.079)                 & 0.809 (0.172)              & 0.903 (0.122)              & 0.999 (0.000)           & 0.996 (0.007)            & 0.996 (0.008)           \\
           & \# of Nodes    & \multicolumn{1}{c|}{---}                             & 727.7 (32.4)               & 688.2 (207.6)              & 390.9 (19.8)            & 306.2 (45.7)             & 279.0 (54.0)            \\
           & \# of Clusters & 5.3 (0.9)                     & 5.4 (0.8)                  & 13.1 (5.8)                 & 6.1 (0.3)               & 6.4 (0.5)                & 9.2 (2.6)               \\ \hline
Stream \#5 & NMI             & 0.476 (0.590)                 & 0.881 (0.133)              & 0.951 (0.026)              & 0.999 (0.000)           & 0.994 (0.014)            & 0.994 (0.014)           \\
           & ARI             & 0.267 (0.066)                 & 0.748 (0.261)              & 0.909 (0.056)              & 0.999 (0.000)           & 0.991 (0.032)            & 0.990 (0.032)           \\
           & \# of Nodes    & \multicolumn{1}{c|}{---}                             & 864.7 (39.4)               & 855.3 (259.7)              & 438.3 (22.1)            & 352.1 (55.6)             & 315.9 (65.6)            \\
           & \# of Clusters & 5.2 (0.9)                     & 5.9 (1.9)                  & 15.0 (6)                   & 6.0 (0.0)               & 6.1 (0.3)                & 8.0 (1.7)               \\ \hline
Stream \#6 & NMI             & 0.470 (0.078)                 & 0.891 (0.100)              & 0.941 (0.032)              & 0.999 (0.000)           & 0.998 (0.001)            & 0.988 (0.020)           \\
           & ARI             & 0.234 (0.093)                 & 0.739 (0.221)              & 0.870 (0.083)              & 0.999 (0.000)           & 0.999 (0.000)            & 0.979 (0.047)           \\
           & \# of Nodes    & \multicolumn{1}{c|}{---}                             & 993.4 (41.8)               & 1063.3 (323.4)             & 492.1 (24.3)            & 406.0 (65.2)             & 359.9 (75.9)            \\
           & \# of Clusters & 5.4 (0.5)                     & 6.7 (1.1)                  & 17.7 (9.4)                 & 6.0 (0.0)               & 6.1 (0.4)                & 8.5 (1.9)     \\ \hline \hline      
\end{tabular}
	}
	\\
	\vspace{1mm}
	\footnotesize \raggedright
	\hspace{-55mm} The values in parentheses indicate the standard deviation.
	\vspace{8mm}
\end{table*}

\subsection{Evaluation on Real-world Datasets}
\label{sec:realworld}
Second, we evaluate the clustering performance of CAE on real-world datasets in the stationary and non-stationary environments.

\subsubsection{Dataset}
\label{sec:dataset_real}
We use 12 real-world datasets from public repositories \cite{derrac15, dua19}. Table \ref{tab:datasets} summarizes their statistics. In all experiments, the data points from each dataset are presented to each algorithm only once, without any pre-processing. As in Section \ref{sec:results_2d}, in the stationary environment, all data points are presented to each algorithm in random order. In the non-stationary environment, data points are randomly sampled from a specific class in the dataset, and the class is shifted sequentially. Furthermore, we use the same data points for both training and testing; that is, algorithms are trained on all data points from each dataset and evaluated on the same set. Note that class labels are not used during the training phase.

\begin{table}[htbp]
	\centering
	\renewcommand{\arraystretch}{1.2}
	\caption{Statistics of real-world datasets}
	\label{tab:datasets}
		\begin{tabular}{l|c|c|c}
			\hline\hline
			\multirow{2}{*}{Dataset} & \# of  & \# of   & \# of  \\
			& Data Points & Attributes & Classes \\
			\hline
			Iris              & 150     & 4       & 3        \\
			Ionosphere        & 351     & 34      & 2        \\
			Pima              & 768     & 8       & 2        \\
            Binalpha          & 1,404   & 320     & 36       \\
			Yeast             & 1,484   & 8       & 10       \\
            Semeion           & 1,593   & 256     & 10       \\
			Image Segmentation & 2,310  & 19      & 7        \\
			Phoneme           & 5,404   & 5       & 2        \\
			Texture           & 5,500   & 40      & 11       \\
			PenBased          & 10,992  & 16      & 10       \\
			Letter            & 20,000  & 16      & 26       \\
			Skin              & 245,057 & 3       & 2        \\
			\hline\hline
		\end{tabular}
\end{table}

\subsubsection{Parameter Settings}
\label{sec:results_params}
The parameter ranges used for grid search in AutoCloud, ASOINN, TCA, and CAEA are summarized in Table \ref{tab:paramAlgorithms}. Based on these ranges, we performed grid search to identify the optimal parameter settings for both stationary and non-stationary environments. During grid search, each algorithm is trained on each dataset using all data points for both training and NMI score evaluation. For each parameter setting, the evaluation is repeated 20 times (i.e., 2$\times$10-fold cross-validation) with different random seeds. The parameter configuration yielding the highest NMI score is selected for comparison. Tables \ref{tab:paramGrid_real_stationary} and \ref{tab:paramGrid_real_nonstationary} summarize the grid search results for the real-world datasets in the stationary and non-stationary environments, respectively. These results show that optimal parameter values for each algorithm depend on the dataset and environment. Notably, the proposed CAE does not require any parameter pre-specification.

\begin{table*}[htbp]
	\caption{Parameter specifications by grid search for real-world datasets in the stationary environment}
	\label{tab:paramGrid_real_stationary}
	\footnotesize
	\centering
	\renewcommand{\arraystretch}{1.2}
	\scalebox{1.0}{
		\begin{tabular}{lc|cc|cc|cc}
			\hline\hline
			\multirow{2}{*}{Dataset} & AutoCloud & \multicolumn{2}{c|}{ASOINN} & \multicolumn{2}{c|}{TCA} & \multicolumn{2}{c}{CAEA} \\
			&  $m$ & $\lambda$ & $ a_{\mathrm{max}} $ & $\lambda$ & $ \sigma_{\mathrm{init}} $ & $\lambda$ & $ a_{\mathrm{max}} $ \\
			\hline
			Iris & 1 & 50 & 10 & 100 & 0.3 & 50 & 90  \\
			Ionosphere & 1 & 500  & 10 & 400 & 0.2 & 400 & 10  \\
			Pima & 1 & 450 & 50 & 500 & 0.2 & 400 & 70  \\
			Binalpha & 1 & 500 & 10 & 450 & 0.4 & 500 & 10 \\
			Yeast & 1 & 500 & 90 & 500 & 0.3 & 500 & 40  \\
			Semeion & 1 & 400 & 10 & 500 & 0.4 & 400 & 90 \\
			Image Segmentation & 1 & 100 & 20 & 50 & 0.2 & 100 & 20  \\
			Phoneme & 1 &  50 & 10 & 150 & 0.2 & 500 & 90  \\
			Texture & 1 & 200 & 10 & 250 & 0.2 & 500 & 100  \\
			PenBased & 1 & 150 & 10 & 150 & 0.3 & 100 & 50  \\
			Letter & N/A & 400 & 10 & 500 & 0.2 & 450 & 10  \\
			Skin & N/A &  50 & 20 & 200 & 0.1 & 50 & 80  \\
      		\hline\hline
			\multicolumn{2}{c|}{\multirow{2}{*}{Mean Value}} & \multicolumn{2}{c|}{ASOINN(mean)} & \multicolumn{2}{c|}{TCA(mean)} & \multicolumn{2}{c}{CAEA(mean)} \\
            & & 279.2 & 21.7 & 312.5 & 0.2 & 329.2 & 55.0 \\
			\hline\hline
		\end{tabular}
	}
  \\
	\vspace{1mm}
	\footnotesize \raggedright
	\hspace{-29mm} N/A indicates that an algorithm could not build a predictive model \\
	\hspace{-39mm} within 1 hour under the available computational resources. \\
    \hspace{-18mm} Mean Value represents the average of the parameter values over all datasets.
\end{table*}

\begin{table*}[htbp]
	\vspace{-5mm}
	\caption{Parameter specifications by grid search for real-world datasets in the non-stationary environment}
	\label{tab:paramGrid_real_nonstationary}
	\footnotesize
	\centering
	\renewcommand{\arraystretch}{1.2}
	\scalebox{1.0}{
		\begin{tabular}{lc|cc|cc|cc}
			\hline\hline
			\multirow{2}{*}{Dataset} & AutoCloud & \multicolumn{2}{c|}{ASOINN} & \multicolumn{2}{c|}{TCA} & \multicolumn{2}{c}{CAEA} \\
			&  $m$ & $\lambda$ & $ a_{\mathrm{max}} $ & $\lambda$ & $ \sigma_{\mathrm{init}} $ & $\lambda$ & $ a_{\mathrm{max}} $ \\
			\hline
			Iris & 1 & 300 & 10 & 100 & 0.3 & 50 & 90  \\
			Ionosphere & 1 & 500  & 10 & 500 & 0.2 & 350 & 70  \\
			Pima & 1 & 400 & 50 & 350 & 0.2 & 300 & 60  \\
			Binalpha & 1 & 500 & 10 & 500 & 0.4 & 500 & 100 \\
			Yeast & 1 & 500 & 30 & 500 & 0.3 & 500 & 90  \\
			Semeion & 1 & 400 & 10 & 500 & 0.4 & 150 & 70 \\
			Image Segmentation & 1 & 200 & 20 & 50 & 0.2 & 100 & 60  \\
			Phoneme & 1 & 500 & 10 & 150 & 0.2& 400 & 30  \\
			Texture & 1 & 350 & 10 & 150 & 0.2 & 200 & 90  \\
			PenBased & 1 & 150 & 10 & 150 & 0.3 & 50 & 70  \\
			Letter & N/A & 450 & 10 & 500 & 0.2 & 500 & 50  \\
			Skin & N/A & 50 & 20 & 200 & 0.1 & 50 & 60  \\
            \hline\hline
			\multicolumn{2}{c|}{\multirow{2}{*}{Mean Value}} & \multicolumn{2}{c|}{ASOINN(mean)} & \multicolumn{2}{c|}{TCA(mean)} & \multicolumn{2}{c}{CAEA(mean)} \\
            & & 358.3 & 16.7 & 304.2 & 0.3 & 262.5 & 70.0 \\
			\hline\hline
		\end{tabular}
	}
    \\
	\vspace{1mm}
	\footnotesize \raggedright
	\hspace{-29mm} N/A indicates that an algorithm could not build a predictive model \\
	\hspace{-39mm} within 1 hour under the available computational resources. \\
    \hspace{-18mm} Mean Value represents the average of the parameter values over all datasets.
\end{table*}

To emphasize the difficulty of using a single parameter setting across all datasets, we also report the mean parameter values for ASOINN, TCA, and CAEA in Tables \ref{tab:paramGrid_real_stationary} and \ref{tab:paramGrid_real_nonstationary}. Using these mean values, we define additional compared algorithms, denoted as ASOINN(mean), TCA(mean), and CAEA(mean), whose parameters are fixed across datasets in each environment.

\subsubsection{Results}
\label{sec:results_real}
Tables \ref{tab:ResultsClusteringStationary} and \ref{tab:ResultsClusteringNonStationary} present the results of clustering performance for the 12 real-world datasets in stationary and non-stationary environments, respectively. The values in parentheses indicate the standard deviation. A number to the right of each evaluation metric represents the average rank of each algorithm across 20 evaluations. The smaller the rank, the better the metric score. Additionally, a darker cell tone corresponds to a smaller rank (i.e., better evaluation).

As general trends, CAEA achieves the best clustering performance in both environments when its parameters are carefully tuned. In contrast, CAE does not always yield the highest NMI and ARI values, but it consistently provides stable and competitive clustering performance without the need for parameter tuning. Notably, CAE attains NMI and ARI values close to the best algorithms on large-scale datasets such as Letter and Skin datasets, which is advantageous for practical applications. Moreover, for high-dimensional datasets such as Binalpha and Semeion, CAE also demonstrates competitive performance in both stationary and non-stationary settings. These results suggest that CAE maintains its robustness not only for large-scale datasets but also for high-dimensional data, where parameter sensitivity often leads to unstable clustering performance in other algorithms.

Algorithms using average parameter settings (i.e., CAEA(mean), ASOINN(mean), and TCA(mean)) generally perform worse than their fully tuned counterparts.  CAEA(mean) deteriorates compared with CAEA, confirming the sensitivity of CAEA to parameter specifications, while TCA(mean) and ASOINN(mean) also show consistent degradation relative to TCA and SOINN, respectively. These observations emphasize the importance of careful parameter specification for these algorithms. SOINN+ produces excessively large numbers of nodes and clusters as dataset size increases (e.g., for the Skin dataset with 245,057 samples), which hinders effective data aggregation and clustering. Since AutoCloud is specialized for online learning, its clustering performance is significantly lower in a stationary environment. Furthermore, AutoCloud was unable to build a predictive model within one hour on large datasets (i.e., Letter and Skin) under the available computational resources.

\begin{landscape}

\begin{table*}[htbp]
	\centering
	\caption{Results of quantitative comparisons on 12 real-world datasets in the stationary environment}
	\label{tab:ResultsClusteringStationary}
	\footnotesize
	\renewcommand{\arraystretch}{1.2}
	\scalebox{0.6}{
	\begin{tabular}{ll|*{8}{r C{3.7mm}|} r C{3.7mm}} \hline\hline
Dataset & Metric
& \multicolumn{2}{c|}{AutoCloud}
& \multicolumn{2}{c|}{ASOINN}
& \multicolumn{2}{c|}{ASOINN(mean)}
& \multicolumn{2}{c|}{SOINN+}
& \multicolumn{2}{c|}{TCA}
& \multicolumn{2}{c|}{TCA(mean)}
& \multicolumn{2}{c|}{CAEA}
& \multicolumn{2}{c|}{CAEA(mean)}
& \multicolumn{2}{c}{CAE} \\ \hline

Iris & NMI
& 0.700 (0.094) & \cellcolor{gray!62}{5}
& 0.725 (0.171) & \cellcolor{gray!87}{3}
& 0.596 (0.244) & \cellcolor{gray!37}{7}
& 0.606 (0.048) & \cellcolor{gray!50}{6}
& 0.760 (0.004) & \cellcolor{gray!100}{1}
& 0.750 (0.040) & \cellcolor{gray!0}{2}
& 0.723 (0.045) & \cellcolor{gray!75}{4}
& 0.469 (0.000) & \cellcolor{gray!12}{9}
& 0.561 (0.039) & \cellcolor{gray!25}{8} \\
& ARI
& 0.529 (0.089) & \cellcolor{gray!62}{5}
& 0.557 (0.144) & \cellcolor{gray!75}{4}
& 0.438 (0.195) & \cellcolor{gray!37}{7}
& 0.442 (0.091) & \cellcolor{gray!50}{6}
& 0.572 (0.017) & \cellcolor{gray!87}{2}
& 0.567 (0.031) & \cellcolor{gray!25}{3}
& 0.661 (0.074) & \cellcolor{gray!100}{1}
& 0.000 (0.000) & \cellcolor{gray!0}{9}
& 0.221 (0.136) & \cellcolor{gray!12}{8} \\
& \# of Nodes
& --- &
& 11.3 (2.6) &
& 24.6 (4.8) &
& 26.4 (6.8) &
& 10.1 (1.2) &
& 10.1 (0.9) &
& 20.6 (2.2) &
& 150.0 (0.0) &
& 52.5 (19.3) & \\
& \# of Clusters
& 3.0 (0.2) &
& 2.1 (0.4) &
& 3.1 (1.6) &
& 10.0 (4.5) &
& 2.1 (0.2) &
& 2.1 (0.2) &
& 5.2 (1.0) &
& 150.0 (0.0) &
& 36.5 (18.7) & \\
\hline

Ionosphere & NMI
& 0.003 (0.008) & \cellcolor{gray!0}{9}
& 0.196 (0.072) & \cellcolor{gray!62}{4}
& 0.036 (0.068) & \cellcolor{gray!25}{7}
& 0.143 (0.055) & \cellcolor{gray!50}{5}
& 0.092 (0.054) & \cellcolor{gray!37}{6}
& 0.009 (0.017) & \cellcolor{gray!12}{8}
& 0.362 (0.012) & \cellcolor{gray!100}{1}
& 0.232 (0.039) & \cellcolor{gray!75}{3}
& 0.315 (0.024) & \cellcolor{gray!87}{2} \\
& ARI
& 0.000 (0.001) & \cellcolor{gray!12}{8}
& 0.131 (0.104) & \cellcolor{gray!100}{1}
& 0.014 (0.049) & \cellcolor{gray!37}{7}
& 0.083 (0.059) & \cellcolor{gray!62}{4}
& 0.063 (0.065) & \cellcolor{gray!37}{6}
& 0.000 (0.000) & \cellcolor{gray!12}{8}
& 0.067 (0.032) & \cellcolor{gray!50}{5}
& 0.093 (0.037) & \cellcolor{gray!75}{3}
& 0.118 (0.047) & \cellcolor{gray!87}{2} \\
& \# of Nodes
& --- &
& 43.4 (5.9) &
& 22.5 (6.1) &
& 26.0 (9.5) &
& 12.9 (3.0) &
& 14.5 (2.0) &
& 263.1 (5.2) &
& 86.9 (3.2) &
& 122.5 (25.7) & \\
& \# of Clusters
& 1.3 (0.5) &
& 9.7 (2.3) &
& 2.0 (1.5) &
& 11.8 (4.5) &
& 3.7 (1.2) &
& 1.4 (0.7) &
& 206.2 (7.3) &
& 33.9 (2.4) &
& 79.0 (23.1) & \\
\hline

Pima & NMI
& 0.003 (0.009) & \cellcolor{gray!12}{8.5}
& 0.010 (0.014) & \cellcolor{gray!25}{7}
& 0.003 (0.005) & \cellcolor{gray!12}{8.5}
& 0.039 (0.018) & \cellcolor{gray!50}{5}
& 0.044 (0.017) & \cellcolor{gray!62}{4}
& 0.016 (0.015) & \cellcolor{gray!37}{6}
& 0.186 (0.010) & \cellcolor{gray!100}{1}
& 0.114 (0.009) & \cellcolor{gray!75}{3}
& 0.157 (0.062) & \cellcolor{gray!87}{2} \\
& ARI
& 0.005 (0.019) & \cellcolor{gray!0}{9}
& 0.011 (0.010) & \cellcolor{gray!25}{7}
& 0.006 (0.006) & \cellcolor{gray!12}{8}
& 0.042 (0.022) & \cellcolor{gray!75}{3}
& 0.039 (0.029) & \cellcolor{gray!62}{4.5}
& 0.024 (0.022) & \cellcolor{gray!37}{6}
& 0.056 (0.016) & \cellcolor{gray!87}{2}
& 0.057 (0.018) & \cellcolor{gray!100}{1}
& 0.039 (0.027) & \cellcolor{gray!62}{4.5} \\
& \# of Nodes
& --- &
& 76.7 (12.4) &
& 62.5 (6.4) &
& 58.6 (15.6) &
& 40.9 (5.5) &
& 29.5 (3.6) &
& 337.4 (11.5) &
& 186.8 (5.3) &
& 309.7 (144.4) & \\
& \# of Clusters
& 2.0 (0.0) &
& 2.2 (0.6) &
& 1.6 (0.5) &
& 13.3 (6.3) &
& 7.1 (2.1) &
& 1.9 (0.6) &
& 247.7 (12.2) &
& 105.7 (6.0) &
& 219.1 (136.0) & \\
\hline

Binalpha & NMI
& 0.079 (0.069) & \cellcolor{gray!12}{8}
& 0.286 (0.048) & \cellcolor{gray!37}{6}
& 0.163 (0.055) & \cellcolor{gray!25}{7}
& 0.320 (0.035) & \cellcolor{gray!62}{4}
& 0.317 (0.044) & \cellcolor{gray!50}{5}
& 0.000 (0.000) & \cellcolor{gray!0}{9}
& 0.654 (0.012) & \cellcolor{gray!100}{1}
& 0.553 (0.018) & \cellcolor{gray!75}{3}
& 0.595 (0.057) & \cellcolor{gray!87}{2} \\
& ARI
& 0.003 (0.006) & \cellcolor{gray!12}{8}
& 0.009 (0.005) & \cellcolor{gray!25}{6}
& 0.005 (0.004) & \cellcolor{gray!37}{7}
& 0.020 (0.014) & \cellcolor{gray!50}{5}
& 0.088 (0.023) & \cellcolor{gray!62}{4}
& 0.000 (0.000) & \cellcolor{gray!0}{9}
& 0.181 (0.026) & \cellcolor{gray!100}{1}
& 0.164 (0.034) & \cellcolor{gray!87}{3}
& 0.172 (0.023) & \cellcolor{gray!75}{2} \\
& \# of Nodes
& --- &
& 123.7 (8.5) &
& 48.6 (5.5) &
& 83.5 (24.0) &
& 25.3 (6.0) &
& 0.0 (0.0) &
& 434.1 (8.1) &
& 194.8 (7.8) &
& 284.5 (133.4) & \\
& \# of Clusters
& 36.0 (0.0) &
& 23.9 (4.8) &
& 4.0 (2.1) &
& 42.9 (12.2) &
& 9.2 (2.5) &
& 0.0 (0.0) &
& 315.3 (10.7) &
& 95.1 (5.4) &
& 187.3 (114.7) & \\
\hline

Yeast & NMI
& 0.003 (0.007) & \cellcolor{gray!12}{8}
& 0.037 (0.047) & \cellcolor{gray!36}{6}
& 0.000 (0.000) & \cellcolor{gray!0}{9}
& 0.076 (0.061) & \cellcolor{gray!50}{5}
& 0.080 (0.065) & \cellcolor{gray!62}{4}
& 0.028 (0.041) & \cellcolor{gray!25}{7}
& 0.331 (0.010) & \cellcolor{gray!100}{1}
& 0.264 (0.016) & \cellcolor{gray!87}{2}
& 0.222 (0.059) & \cellcolor{gray!75}{3} \\
& ARI
& 0.000 (0.000) & \cellcolor{gray!12}{8.5}
& 0.003 (0.004) & \cellcolor{gray!37}{6}
& 0.000 (0.000) & \cellcolor{gray!12}{8.5}
& 0.013 (0.032) & \cellcolor{gray!62}{4}
& 0.012 (0.021) & \cellcolor{gray!50}{5}
& 0.002 (0.004) & \cellcolor{gray!25}{7}
& 0.089 (0.014) & \cellcolor{gray!87}{2}
& 0.102 (0.021) & \cellcolor{gray!100}{1}
& 0.076 (0.025) & \cellcolor{gray!75}{3} \\
& \# of Nodes
& --- &
& 101.3 (12.9) &
& 68.4 (8.5) &
& 68.9 (27.0) &
& 12.2 (2.6) &
& 10.5 (2.0) &
& 472.5 (15.8) &
& 254.4 (5.7) &
& 241.4 (103.7) & \\
& \# of Clusters
& 7.2 (1.3) &
& 1.7 (1.0) &
& 1.0 (0.0) &
& 7.7 (5.3) &
& 1.8 (0.6) &
& 1.4 (0.7) &
& 318.9 (16.4) &
& 129.0 (5.5) &
& 111.6 (81.1) & \\
\hline

Semeion & NMI
& 0.188 (0.081) & \cellcolor{gray!12}{8}
& 0.389 (0.068) & \cellcolor{gray!50}{5}
& 0.208 (0.089) & \cellcolor{gray!25}{7}
& 0.356 (0.079) & \cellcolor{gray!37}{6}
& 0.402 (0.048) & \cellcolor{gray!62}{4}
& 0.000 (0.000) & \cellcolor{gray!0}{9}
& 0.603 (0.007) & \cellcolor{gray!100}{1}
& 0.585 (0.008) & \cellcolor{gray!75}{3}
& 0.590 (0.005) & \cellcolor{gray!87}{2} \\
& ARI
& 0.042 (0.041) & \cellcolor{gray!25}{7}
& 0.112 (0.075) & \cellcolor{gray!50}{5}
& 0.023 (0.041) & \cellcolor{gray!12}{8}
& 0.098 (0.083) & \cellcolor{gray!37}{6}
& 0.266 (0.047) & \cellcolor{gray!100}{1.5}
& 0.000 (0.000) & \cellcolor{gray!0}{9}
& 0.266 (0.026) & \cellcolor{gray!100}{1.5}
& 0.263 (0.021) & \cellcolor{gray!75}{3}
& 0.115 (0.024) & \cellcolor{gray!62}{4} \\
& \# of Nodes
& --- &
& 115.9 (8.2) &
& 74.0 (17.9) &
& 83.5 (22.0) &
& 48.8 (6.7) &
& 0.1 (0.5) &
& 417.7 (9.6) &
& 330.6 (10.9) &
& 650.5 (16.4) & \\
& \# of Clusters
& 9.8 (0.5) &
& 26.1 (5.0) &
& 11.3 (8.0) &
& 44.7 (12.0) &
& 12.0 (2.0) &
& 0.1 (0.2) &
& 288.2 (12.3) &
& 218.0 (8.6) &
& 490.8 (15.8) & \\
\hline

Image & NMI
& 0.342 (0.127) & \cellcolor{gray!0}{9}
& 0.632 (0.002) & \cellcolor{gray!62}{4}
& 0.594 (0.077) & \cellcolor{gray!37}{6}
& 0.554 (0.018) & \cellcolor{gray!25}{7}
& 0.641 (0.047) & \cellcolor{gray!87}{2}
& 0.501 (0.172) & \cellcolor{gray!12}{8}
& 0.663 (0.023) & \cellcolor{gray!100}{1}
& 0.622 (0.017) & \cellcolor{gray!50}{5}
& 0.635 (0.029) & \cellcolor{gray!75}{3} \\
Segmentation & ARI
& 0.115 (0.089) & \cellcolor{gray!0}{9}
& 0.236 (0.000) & \cellcolor{gray!37}{6}
& 0.208 (0.057) & \cellcolor{gray!25}{7}
& 0.252 (0.048) & \cellcolor{gray!50}{5}
& 0.437 (0.109) & \cellcolor{gray!87}{2}
& 0.186 (0.115) & \cellcolor{gray!12}{8}
& 0.486 (0.048) & \cellcolor{gray!100}{1}
& 0.352 (0.034) & \cellcolor{gray!62}{4}
& 0.372 (0.100) & \cellcolor{gray!75}{3} \\
& \# of Nodes
& --- &
& 42.2 (5.0) &
& 84.9 (8.7) &
& 128.2 (30.5) &
& 30.7 (4.0) &
& 15.1 (1.7) &
& 77.8 (8.0) &
& 162.1 (5.6) &
& 119.8 (52.3) & \\
& \# of Clusters
& 7.0 (0.0) &
& 3.0 (0.0) &
& 3.0 (0.6) &
& 22.7 (9.2) &
& 5.2 (1.1) &
& 2.7 (0.8) &
& 19.8 (2.2) &
& 29.0 (3.1) &
& 51.2 (33.9) & \\
\hline

Phoneme & NMI
& 0.020 (0.048) & \cellcolor{gray!25}{7}
& 0.048 (0.072) & \cellcolor{gray!50}{5}
& 0.000 (0.000) & \cellcolor{gray!0}{9}
& 0.037 (0.031) & \cellcolor{gray!37}{6}
& 0.117 (0.054) & \cellcolor{gray!62}{4}
& 0.007 (0.019) & \cellcolor{gray!12}{8}
& 0.174 (0.003) & \cellcolor{gray!100}{1}
& 0.155 (0.006) & \cellcolor{gray!87}{2}
& 0.154 (0.012) & \cellcolor{gray!75}{3} \\
& ARI
& 0.024 (0.069) & \cellcolor{gray!37}{6}
& 0.027 (0.053) & \cellcolor{gray!50}{5}
& 0.000 (0.000) & \cellcolor{gray!0}{9}
& 0.015 (0.023) & \cellcolor{gray!25}{7}
& 0.146 (0.054) & \cellcolor{gray!100}{1}
& 0.002 (0.005) & \cellcolor{gray!12}{8}
& 0.037 (0.015) & \cellcolor{gray!62}{4}
& 0.056 (0.02) & \cellcolor{gray!75}{3}
& 0.083 (0.036) & \cellcolor{gray!87}{2} \\
& \# of Nodes
& --- &
& 32.6 (3.3) &
& 116.1 (7.8) &
& 166.6 (47.7) &
& 83.6 (4.8) &
& 21.5 (2.5) &
& 526.4 (11.7) &
& 285.8 (11.1) &
& 245.6 (50.9) & \\
& \# of Clusters
& 2.0 (0.0) &
& 1.7 (0.9) &
& 1.0 (0.0) &
& 11.3 (5.6) &
& 4.5 (1.1) &
& 1.1 (0.2) &
& 280.3 (12.5) &
& 98.9 (7.4) &
& 88.1 (43.1) & \\
\hline

Texture & NMI
& 0.290 (0.099) & \cellcolor{gray!12}{8}
& 0.609 (0.040) & \cellcolor{gray!62}{4}
& 0.545 (0.068) & \cellcolor{gray!25}{7}
& 0.572 (0.049) & \cellcolor{gray!37}{6}
& 0.598 (0.003) & \cellcolor{gray!50}{5}
& 0.004 (0.018) & \cellcolor{gray!0}{9}
& 0.713 (0.016) & \cellcolor{gray!100}{1}
& 0.680 (0.014) & \cellcolor{gray!75}{3}
& 0.688 (0.026) & \cellcolor{gray!87}{2} \\
& ARI
& 0.063 (0.050) & \cellcolor{gray!12}{8}
& 0.164 (0.036) & \cellcolor{gray!62}{4}
& 0.117 (0.035) & \cellcolor{gray!25}{7}
& 0.158 (0.038) & \cellcolor{gray!50}{5}
& 0.148 (0.002) & \cellcolor{gray!37}{6}
& 0.000 (0.001) & \cellcolor{gray!0}{9}
& 0.568 (0.048) & \cellcolor{gray!100}{1}
& 0.486 (0.034) & \cellcolor{gray!75}{3}
& 0.501 (0.083) & \cellcolor{gray!87}{2} \\
& \# of Nodes
& --- &
& 91.4 (6.2) &
& 116.2 (6.8) &
& 163.7 (40.7) &
& 59.4 (3.7) &
& 8.8 (1.0) &
& 262.8 (7.9) &
& 338.4 (8.9) &
& 171.9 (70.4) & \\
& \# of Clusters
& 10.1 (0.3) &
& 4.4 (0.9) &
& 3.7 (0.8) &
& 18.7 (6.6) &
& 4.1 (0.2) &
& 1.0 (0.0) &
& 25.8 (4.3) &
& 163.3 (15.5) &
& 67.0 (48.4) & \\
\hline

PenBased & NMI
& 0.385 (0.018) & \cellcolor{gray!0}{9}
& 0.699 (0.062) & \cellcolor{gray!50}{5}
& 0.541 (0.095) & \cellcolor{gray!12}{8}
& 0.709 (0.068) & \cellcolor{gray!75}{3.5}
& 0.669 (0.061) & \cellcolor{gray!37}{6}
& 0.624 (0.068) & \cellcolor{gray!25}{7}
& 0.739 (0.015) & \cellcolor{gray!100}{1}
& 0.719 (0.014) & \cellcolor{gray!87}{2}
& 0.709 (0.020) & \cellcolor{gray!75}{3.5} \\
& ARI
& 0.151 (0.013) & \cellcolor{gray!0}{9}
& 0.476 (0.154) & \cellcolor{gray!50}{5}
& 0.190 (0.101) & \cellcolor{gray!12}{8}
& 0.492 (0.166) & \cellcolor{gray!62}{4}
& 0.397 (0.115) & \cellcolor{gray!37}{6}
& 0.305 (0.095) & \cellcolor{gray!25}{7}
& 0.647 (0.036) & \cellcolor{gray!100}{1}
& 0.606 (0.031) & \cellcolor{gray!87}{2}
& 0.570 (0.057) & \cellcolor{gray!75}{3} \\
& \# of Nodes
& --- &
& 88.4 (7.2) &
& 157.0 (9.1) &
& 273.4 (67.1) &
& 87.2 (3.0) &
& 120.0 (6.9) &
& 178.9 (6.0) &
& 332.8 (11.5) &
& 378.6 (153.4) & \\
& \# of Clusters
& 10.0 (0.0) &
& 7.9 (1.5) &
& 4.4 (1.4) &
& 32.2 (10.6) &
& 7.7 (1.4) &
& 7.6 (1.0) &
& 50.2 (8.4) &
& 77.2 (7.1) &
& 124.6 (103.2) & \\
\hline

Letter & NMI
& N/A & \cellcolor{gray!0}{9}
& 0.150 (0.061) & \cellcolor{gray!37}{6}
& 0.057 (0.048) & \cellcolor{gray!25}{7}
& 0.153 (0.035) & \cellcolor{gray!50}{5}
& 0.430 (0.021) & \cellcolor{gray!62}{4}
& 0.054 (0.052) & \cellcolor{gray!12}{8}
& 0.499 (0.006) & \cellcolor{gray!100}{1}
& 0.468 (0.011) & \cellcolor{gray!75}{3}
& 0.498 (0.043) & \cellcolor{gray!87}{2} \\
& ARI
& N/A & \cellcolor{gray!0}{9}
& 0.004 (0.004) & \cellcolor{gray!50}{5}
& 0.001 (0.000) & \cellcolor{gray!25}{7.5}
& 0.002 (0.001) & \cellcolor{gray!37}{6}
& 0.086 (0.021) & \cellcolor{gray!62}{4}
& 0.001 (0.001) & \cellcolor{gray!25}{7.5}
& 0.134 (0.010) & \cellcolor{gray!100}{1}
& 0.098 (0.013) & \cellcolor{gray!75}{3}
& 0.114 (0.025) & \cellcolor{gray!87}{2} \\
& \# of Nodes
& --- &
& 208.6 (14.3) &
& 160.8 (9.6) &
& 234.7 (53.0) &
& 222.5 (14.6) &
& 41.7 (4.3) &
& 415.2 (25.0) &
& 423.4 (16.0) &
& 621.6 (223.9) & \\
& \# of Clusters
& N/A &
& 3.2 (1.2) &
& 1.7 (0.6) &
& 11.2 (5.3) &
& 31.5 (3.5) &
& 1.5 (0.5) &
& 216.0 (15.4) &
& 162.4 (12.2) &
& 276.0 (182.5) & \\
\hline

Skin & NMI
& N/A & \cellcolor{gray!0}{9}
& 0.560 (0.082) & \cellcolor{gray!75}{3}
& 0.162 (0.253) & \cellcolor{gray!25}{7}
& 0.338 (0.016) & \cellcolor{gray!37}{6}
& 0.635 (0.020) & \cellcolor{gray!100}{1}
& 0.000 (0.000) & \cellcolor{gray!12}{8}
& 0.563 (0.044) & \cellcolor{gray!87}{2}
& 0.424 (0.029) & \cellcolor{gray!50}{5}
& 0.528 (0.046) & \cellcolor{gray!62}{4} \\
& ARI
& N/A & \cellcolor{gray!0}{9}
& 0.553 (0.204) & \cellcolor{gray!62}{4}
& 0.146 (0.297) & \cellcolor{gray!37}{6}
& 0.081 (0.031) & \cellcolor{gray!25}{7}
& 0.713 (0.021) & \cellcolor{gray!100}{1}
& 0.000 (0.000) & \cellcolor{gray!12}{8}
& 0.674 (0.084) & \cellcolor{gray!87}{2}
& 0.316 (0.116) & \cellcolor{gray!50}{5}
& 0.599 (0.126) & \cellcolor{gray!75}{3} \\
& \# of Nodes
& --- &
& 100.6 (8.7) &
& 381.7 (13.7) &
& 1038.3 (263.6) &
& 202.0 (7.1) &
& 27.7 (1.2) &
& 225.9 (32.4) &
& 836.4 (21.4) &
& 274.75 (57.9) & \\
& \# of Clusters
& N/A &
& 2.7 (1.0) &
& 3.5 (1.1) &
& 122.7 (33.8) &
& 5.7 (1.4) &
& 1.0 (0.0) &
& 9.5 (2.2) &
& 83.7 (10.5) &
& 15.8 (6.0) & \\
\hline

& Average Rank
& \multicolumn{2}{c|}{\cellcolor{gray!0}{8.042 (1.198)}}
& \multicolumn{2}{c|}{\cellcolor{gray!50}{4.833 (1.344)}}
& \multicolumn{2}{c|}{\cellcolor{gray!25}{7.479 (0.848)}}
& \multicolumn{2}{c|}{\cellcolor{gray!37}{5.271 (1.090)}}
& \multicolumn{2}{c|}{\cellcolor{gray!62}{3.708 (1.773)}}
& \multicolumn{2}{c|}{\cellcolor{gray!25}{7.479 (1.753)}}
& \multicolumn{2}{c|}{\cellcolor{gray!100}{1.604 (1.108)}}
& \multicolumn{2}{c|}{\cellcolor{gray!75}{3.458 (1.957)}}
& \multicolumn{2}{c}{\cellcolor{gray!87}{3.125 (1.647)}} \\
\hline\hline
\end{tabular}}
\\
\vspace{1mm}
\footnotesize
\hspace*{8.0mm}
\begin{minipage}{\linewidth}
The values in parentheses indicate the standard deviation. \\
A number to the right of a metric value is the average rank of an algorithm over 20 evaluations. \\
The smaller the rank, the better the metric score. A darker tone in a cell corresponds to a smaller rank. \\
N/A indicates that an algorithm could not build a predictive model within 1 hour under the available computational resources. \\
AutoCloud does not have a node thus it is indicated by a symbol ``---''.
\end{minipage}
\end{table*}

\end{landscape}

\begin{landscape}

\begin{table*}[htbp]
	\centering
	\caption{Results of quantitative comparisons on 12 real-world datasets in the non-stationary environment}
	\label{tab:ResultsClusteringNonStationary}
	\footnotesize
	\renewcommand{\arraystretch}{1.2}
	\scalebox{0.6}{
	\begin{tabular}{ll|*{8}{r C{3.7mm}|} r C{3.7mm}} \hline\hline
Dataset & Metric
& \multicolumn{2}{c|}{AutoCloud}
& \multicolumn{2}{c|}{ASOINN}
& \multicolumn{2}{c|}{ASOINN(mean)}
& \multicolumn{2}{c|}{SOINN+}
& \multicolumn{2}{c|}{TCA}
& \multicolumn{2}{c|}{TCA(mean)}
& \multicolumn{2}{c|}{CAEA}
& \multicolumn{2}{c|}{CAEA(mean)}
& \multicolumn{2}{c}{CAE} \\ \hline

Iris & NMI
& 0.775 (0.055) & \cellcolor{gray!100}{1}
& 0.724 (0.046) & \cellcolor{gray!75}{3}
& 0.422 (0.352) & \cellcolor{gray!0}{9}
& 0.609 (0.042) & \cellcolor{gray!50}{5}
& 0.761 (0.003) & \cellcolor{gray!87}{2}
& 0.719 (0.091) & \cellcolor{gray!62}{4}
& 0.590 (0.061) & \cellcolor{gray!37}{6}
& 0.477 (0.004) & \cellcolor{gray!25}{7}
& 0.457 (0.114) & \cellcolor{gray!12}{8} \\
& ARI
& 0.751 (0.095) & \cellcolor{gray!100}{1}
& 0.554 (0.072) & \cellcolor{gray!62}{4}
& 0.307 (0.283) & \cellcolor{gray!25}{7}
& 0.464 (0.112) & \cellcolor{gray!50}{5}
& 0.572 (0.019) & \cellcolor{gray!87}{2}
& 0.556 (0.089) & \cellcolor{gray!75}{3}
& 0.397 (0.073) & \cellcolor{gray!37}{6}
& 0.012 (0.010) & \cellcolor{gray!0}{9}
& 0.118 (0.137) & \cellcolor{gray!12}{8} \\
& \# of Nodes
& --- & 
& 23.2 (3.9) &
& 24.3 (4.2) &
& 39.3 (8.3) &
& 10.2 (0.8) &
& 10.8 (1.1) &
& 23.7 (5.8) &
& 123.9 (9.0) &
& 88.9 (54.6) & \\
& \# of Clusters
& 3.0 (0.0) &
& 2.8 (0.8) &
& 2.4 (1.6) &
& 17.9 (6.2) &
& 2.1 (0.2) &
& 2.3 (0.6) &
& 9.3 (1.8) &
& 136.2 (5.2) &
& 83.3 (57.2) & \\
\hline

Ionosphere & NMI
& 0.045 (0.078) & \cellcolor{gray!12}{8}
& 0.204 (0.085) & \cellcolor{gray!62}{4}
& 0.149 (0.066) & \cellcolor{gray!37}{6}
& 0.169 (0.080) & \cellcolor{gray!50}{5}
& 0.118 (0.047) & \cellcolor{gray!25}{7}
& 0.007 (0.012) & \cellcolor{gray!0}{9}
& 0.397 (0.053) & \cellcolor{gray!100}{1}
& 0.296 (0.097) & \cellcolor{gray!87}{2}
& 0.248 (0.085) & \cellcolor{gray!75}{3} \\
& ARI
& 0.052 (0.092) & \cellcolor{gray!25}{8}
& 0.150 (0.133) & \cellcolor{gray!87}{2}
& 0.088 (0.066) & \cellcolor{gray!50}{5}
& 0.086 (0.076) & \cellcolor{gray!37}{6}
& 0.103 (0.064) & \cellcolor{gray!75}{3}
& 0.001 (0.004) & \cellcolor{gray!0}{9}
& 0.293 (0.257) & \cellcolor{gray!100}{1}
& 0.062 (0.055) & \cellcolor{gray!25}{7}
& 0.095 (0.089) & \cellcolor{gray!62}{4} \\
& \# of Nodes
& --- &
& 44.0 (5.0) &
& 40.4 (8.3) &
& 41.1 (10.7) &
& 13.1 (3.5) &
& 14.4 (1.5) &
& 232.1 (64.5) &
& 123.9 (9.0) &
& 92.6 (31.9) & \\
& \# of Clusters
& 1.7 (0.5) &
& 10.4 (3.3) &
& 8.3 (3.0) &
& 19.2 (8.5) &
& 3.9 (1.5) &
& 1.5 (0.7) &
& 193.3 (76.3) &
& 97.1 (4.9) &
& 64.1 (30.8) & \\
\hline

Pima & NMI
& 0.004 (0.014) & \cellcolor{gray!12}{8}
& 0.008 (0.010) & \cellcolor{gray!25}{7}
& 0.002 (0.003) & \cellcolor{gray!0}{9}
& 0.045 (0.018) & \cellcolor{gray!62}{4}
& 0.039 (0.027) & \cellcolor{gray!50}{5}
& 0.013 (0.011) & \cellcolor{gray!37}{6}
& 0.171 (0.053) & \cellcolor{gray!100}{1}
& 0.140 (0.023) & \cellcolor{gray!75}{3}
& 0.144 (0.062) & \cellcolor{gray!87}{2} \\
& ARI
& 0.003 (0.011) & \cellcolor{gray!0}{9}
& 0.011 (0.008) & \cellcolor{gray!25}{7}
& 0.005 (0.006) & \cellcolor{gray!12}{8}
& 0.050 (0.050) & \cellcolor{gray!62}{4}
& 0.053 (0.027) & \cellcolor{gray!87}{2}
& 0.018 (0.019) & \cellcolor{gray!37}{6}
& 0.051 (0.021) & \cellcolor{gray!75}{3}
& 0.096 (0.042) & \cellcolor{gray!100}{1}
& 0.044 (0.043) & \cellcolor{gray!50}{5} \\
& \# of Nodes
& --- &
& 80.6 (12.3) &
& 69.9 (8.6) &
& 66.9 (17.1) &
& 31.4 (4.2) &
& 28.7 (2.6) &
& 342.6 (76.9) &
& 258.2 (18.3) &
& 288.9 (153.4) & \\
& \# of Clusters
& 2.0 (0.0) &
& 2.3 (0.9) &
& 1.5 (0.5) &
& 15.6 (5.9) &
& 4.4 (1.6) &
& 2.1 (0.8) &
& 246.4 (87.8) &
& 177.6 (24.1) &
& 210.7 (156.6) & \\
\hline

Binalpha & NMI
& 0.280 (0.091) & \cellcolor{gray!25}{7}
& 0.297 (0.051) & \cellcolor{gray!37}{6}
& 0.246 (0.049) & \cellcolor{gray!12}{8}
& 0.614 (0.026) & \cellcolor{gray!100}{1}
& 0.316 (0.030) & \cellcolor{gray!50}{5}
& 0.000 (0.000) & \cellcolor{gray!0}{9}
& 0.601 (0.011) & \cellcolor{gray!87}{2}
& 0.518 (0.022) & \cellcolor{gray!62}{4}
& 0.550 (0.050) & \cellcolor{gray!75}{3} \\
& ARI
& 0.027 (0.025) & \cellcolor{gray!37}{6}
& 0.011 (0.007) & \cellcolor{gray!25}{7}
& 0.006 (0.006) & \cellcolor{gray!12}{8}
& 0.206 (0.042) & \cellcolor{gray!100}{1}
& 0.088 (0.016) & \cellcolor{gray!50}{5}
& 0.000 (0.000) & \cellcolor{gray!0}{9}
& 0.159 (0.020) & \cellcolor{gray!62}{4}
& 0.168 (0.030) & \cellcolor{gray!87}{2}
& 0.165 (0.042) & \cellcolor{gray!75}{3} \\
& \# of Nodes
& --- &
& 122.8 (8.0) &
& 104.1 (6.1) &
& 372.5 (31.3) &
& 26.8 (4.2) &
& 0.1 (0.5) &
& 411.0 (26.8) &
& 171.2 (14.9) &
& 272.8 (162.0) & \\
& \# of Clusters
& 36.0 (0.0) &
& 23.9 (5.5) &
& 18.4 (4.9) &
& 224.4 (23.9) &
& 9.8 (2.1) &
& 0.1 (0.2) &
& 351.3 (32.4) &
& 101.7 (11.8) &
& 214.8 (173.8) & \\
\hline

Yeast & NMI
& 0.045 (0.054) & \cellcolor{gray!37}{6}
& 0.032 (0.042) & \cellcolor{gray!25}{7}
& 0.000 (0.000) & \cellcolor{gray!0}{9}
& 0.172 (0.055) & \cellcolor{gray!62}{4}
& 0.085 (0.042) & \cellcolor{gray!50}{5}
& 0.026 (0.039) & \cellcolor{gray!12}{8}
& 0.397 (0.021) & \cellcolor{gray!100}{1}
& 0.268 (0.026) & \cellcolor{gray!87}{2}
& 0.196 (0.089) & \cellcolor{gray!75}{3} \\
& ARI
& 0.009 (0.016) & \cellcolor{gray!37}{6}
& 0.003 (0.004) & \cellcolor{gray!25}{7}
& 0.000 (0.000) & \cellcolor{gray!0}{9}
& 0.042 (0.033) & \cellcolor{gray!62}{4}
& 0.011 (0.012) & \cellcolor{gray!50}{5}
& 0.002 (0.003) & \cellcolor{gray!12}{8}
& 0.082 (0.032) & \cellcolor{gray!75}{3}
& 0.113 (0.039) & \cellcolor{gray!100}{1}
& 0.083 (0.056) & \cellcolor{gray!87}{2} \\
& \# of Nodes
& --- &
& 101.2 (13.9) &
& 82.0 (8.7) &
& 196.4 (56.8) &
& 12.4 (2.3) &
& 10.6 (2.1) &
& 888.9 (105.9) &
& 248.7 (11.8) &
& 216.8 (117.2) & \\
& \# of Clusters
& 10.0 (0.2) &
& 1.6 (1.7) &
& 1.0 (0.0) &
& 46.9 (23.4) &
& 2.1 (0.6) &
& 1.4 (0.5) &
& 711.5 (135.1) &
& 140.0 (17.4) &
& 97.9 (102.5) & \\
\hline

Semeion & NMI
& 0.368 (0.121) & \cellcolor{gray!37}{6}
& 0.357 (0.060) & \cellcolor{gray!25}{7}
& 0.224 (0.088) & \cellcolor{gray!12}{8}
& 0.575 (0.022) & \cellcolor{gray!100}{1}
& 0.399 (0.037) & \cellcolor{gray!50}{5}
& 0.000 (0.000) & \cellcolor{gray!0}{9}
& 0.561 (0.035) & \cellcolor{gray!87}{2}
& 0.513 (0.033) & \cellcolor{gray!75}{3}
& 0.486 (0.038) & \cellcolor{gray!62}{4} \\
& ARI
& 0.151 (0.113) & \cellcolor{gray!37}{6}
& 0.075 (0.050) & \cellcolor{gray!25}{7}
& 0.033 (0.048) & \cellcolor{gray!12}{8}
& 0.300 (0.057) & \cellcolor{gray!87}{2}
& 0.261 (0.036) & \cellcolor{gray!62}{4}
& 0.000 (0.000) & \cellcolor{gray!0}{9}
& 0.350 (0.061) & \cellcolor{gray!100}{1}
& 0.262 (0.049) & \cellcolor{gray!75}{3}
& 0.164 (0.051) & \cellcolor{gray!50}{5} \\
& \# of Nodes
& --- &
& 112.6 (10.3) &
& 67.7 (9.7) &
& 320.9 (28.2) &
& 47.8 (5.8) &
& 0.2 (0.6) &
& 226.1 (33.9) &
& 163.7 (23.7) &
& 378.5 (185.5) & \\
& \# of Clusters
& 10.0 (0.0) &
& 26.3 (6.2) &
& 8.5 (4.2) &
& 170.6 (21.0) &
& 12.4 (2.2) &
& 0.1 (0.3) &
& 93.3 (23.3) &
& 62.4 (6.5) &
& 298.3 (188.4) & \\
\hline

Image & NMI
& 0.642 (0.037) & \cellcolor{gray!100}{1}
& 0.630 (0.007) & \cellcolor{gray!87}{2}
& 0.623 (0.014) & \cellcolor{gray!62}{4}
& 0.599 (0.013) & \cellcolor{gray!50}{5}
& 0.624 (0.055) & \cellcolor{gray!75}{3}
& 0.512 (0.146) & \cellcolor{gray!0}{9}
& 0.582 (0.060) & \cellcolor{gray!25}{7}
& 0.562 (0.028) & \cellcolor{gray!12}{8}
& 0.597 (0.064) & \cellcolor{gray!37}{6} \\
Segmentation & ARI
& 0.422 (0.086) & \cellcolor{gray!100}{1}
& 0.235 (0.004) & \cellcolor{gray!37}{6}
& 0.233 (0.008) & \cellcolor{gray!25}{7}
& 0.278 (0.065) & \cellcolor{gray!50}{5}
& 0.411 (0.114) & \cellcolor{gray!87}{2}
& 0.166 (0.079) & \cellcolor{gray!0}{9}
& 0.328 (0.083) & \cellcolor{gray!75}{3}
& 0.193 (0.050) & \cellcolor{gray!12}{8}
& 0.318 (0.103) & \cellcolor{gray!62}{4} \\
& \# of Nodes
& --- &
& 69.2 (7.4) &
& 99.5 (9.5) &
& 572.4 (64.5) &
& 31.6 (4.3) &
& 15.0 (1.5) &
& 146.6 (38.6) &
& 305.7 (45.1) &
& 164.3 (48.2) & \\
& \# of Clusters
& 7.0 (0.0) &
& 3.2 (0.4) &
& 3.6 (0.8) &
& 230.5 (48.4) &
& 5.3 (1.6) &
& 2.5 (0.6) &
& 47.3 (11.5) &
& 180.2 (56.5) &
& 57.3 (27.5) & \\
\hline

Phoneme & NMI
& 0.090 (0.052) & \cellcolor{gray!50}{5}
& 0.021 (0.014) & \cellcolor{gray!25}{7}
& 0.007 (0.013) & \cellcolor{gray!12}{8}
& 0.038 (0.010) & \cellcolor{gray!37}{6}
& 0.115 (0.050) & \cellcolor{gray!62}{4}
& 0.003 (0.009) & \cellcolor{gray!0}{9}
& 0.171 (0.015) & \cellcolor{gray!100}{1}
& 0.158 (0.009) & \cellcolor{gray!87}{2}
& 0.157 (0.011) & \cellcolor{gray!75}{3} \\
& ARI
& 0.169 (0.078) & \cellcolor{gray!100}{1}
& 0.004 (0.011) & \cellcolor{gray!12}{8}
& 0.000 (0.000) & \cellcolor{gray!0}{9}
& 0.016 (0.020) & \cellcolor{gray!37}{6}
& 0.149 (0.077) & \cellcolor{gray!87}{2}
& 0.007 (0.021) & \cellcolor{gray!25}{7}
& 0.052 (0.023) & \cellcolor{gray!50}{5}
& 0.057 (0.026) & \cellcolor{gray!62}{4}
& 0.090 (0.033) & \cellcolor{gray!75}{3} \\
& \# of Nodes
& --- &
& 188.7 (9.8) &
& 142.2 (9.1) &
& 248.8 (72.3) &
& 82.3 (5.0) &
& 21.9 (1.9) &
& 496.0 (69.8) &
& 321.6 (10.4) &
& 272.7 (75.6) & \\
& \# of Clusters
& 2.0 (0.0) &
& 2.5 (1.3) &
& 1.3 (0.4) &
& 21.8 (10.2) &
& 4.9 (1.5) &
& 1.1 (0.3) &
& 251.1 (104.2) &
& 121.4 (26.0) &
& 88.7 (47.3) & \\
\hline

Texture & NMI
& 0.602 (0.026) & \cellcolor{gray!37}{6}
& 0.621 (0.051) & \cellcolor{gray!50}{5}
& 0.566 (0.048) & \cellcolor{gray!12}{8}
& 0.665 (0.027) & \cellcolor{gray!75}{3}
& 0.591 (0.025) & \cellcolor{gray!25}{7}
& 0.000 (0.000) & \cellcolor{gray!0}{9}
& 0.667 (0.045) & \cellcolor{gray!87}{2}
& 0.690 (0.037) & \cellcolor{gray!100}{1}
& 0.645 (0.058) & \cellcolor{gray!62}{4} \\
& ARI
& 0.332 (0.074) & \cellcolor{gray!50}{5}
& 0.175 (0.043) & \cellcolor{gray!37}{6}
& 0.129 (0.028) & \cellcolor{gray!12}{8}
& 0.429 (0.069) & \cellcolor{gray!75}{3}
& 0.145 (0.014) & \cellcolor{gray!25}{7}
& 0.000 (0.000) & \cellcolor{gray!0}{9}
& 0.461 (0.093) & \cellcolor{gray!87}{2}
& 0.515 (0.082) & \cellcolor{gray!100}{1}
& 0.391 (0.146) & \cellcolor{gray!62}{4} \\
& \# of Nodes
& --- &
& 139.1 (9.1) &
& 135.8 (6.4) &
& 1400.3 (96.4) &
& 51.3 (2.5) &
& 8.1 (0.6) &
& 462.2 (65.8) &
& 579.2 (66.7) &
& 521.4 (183.2) & \\
& \# of Clusters
& 11.0 (0.0) &
& 4.9 (1.0) &
& 3.7 (0.5) &
& 483.3 (74.5) &
& 4.0 (0.3) &
& 1.0 (0.0) &
& 112.7 (40.4) &
& 142.3 (70.6) &
& 166.7 (150.1) & \\
\hline

PenBased & NMI
& 0.673 (0.025) & \cellcolor{gray!50}{5}
& 0.681 (0.055) & \cellcolor{gray!75}{3}
& 0.513 (0.130) & \cellcolor{gray!0}{9}
& 0.712 (0.025) & \cellcolor{gray!87}{2}
& 0.642 (0.096) & \cellcolor{gray!25}{7}
& 0.550 (0.156) & \cellcolor{gray!12}{8}
& 0.745 (0.050) & \cellcolor{gray!100}{1}
& 0.679 (0.076) & \cellcolor{gray!62}{4}
& 0.654 (0.059) & \cellcolor{gray!37}{6} \\
& ARI
& 0.511 (0.057) & \cellcolor{gray!75}{3}
& 0.402 (0.139) & \cellcolor{gray!37}{6}
& 0.172 (0.104) & \cellcolor{gray!0}{9}
& 0.565 (0.063) & \cellcolor{gray!87}{2}
& 0.346 (0.143) & \cellcolor{gray!25}{7}
& 0.278 (0.134) & \cellcolor{gray!12}{8}
& 0.626 (0.095) & \cellcolor{gray!100}{1}
& 0.494 (0.159) & \cellcolor{gray!62}{4}
& 0.425 (0.142) & \cellcolor{gray!50}{5} \\
& \# of Nodes
& --- &
& 91.1 (8.6) &
& 191.7 (7.3) &
& 2375.1 (158.0) &
& 86.6 (3.8) &
& 116.5 (6.4) &
& 326.6 (44.7) &
& 745.0 (141.5) &
& 977.1 (319.0) & \\
& \# of Clusters
& 10.0 (0.0) &
& 7.6 (1.4) &
& 4.5 (1.4) &
& 713.3 (82.5) &
& 7.0 (1.6) &
& 7.8 (1.6) &
& 23.2 (7.3) &
& 179.7 (88.7) &
& 268.4 (208.9) & \\
\hline

Letter & NMI
& N/A & \cellcolor{gray!0}{9}
& 0.168 (0.065) & \cellcolor{gray!37}{6}
& 0.060 (0.046) & \cellcolor{gray!12}{8}
& 0.590 (0.015) & \cellcolor{gray!100}{1}
& 0.441 (0.019) & \cellcolor{gray!50}{5}
& 0.066 (0.053) & \cellcolor{gray!25}{7}
& 0.502 (0.023) & \cellcolor{gray!75}{3}
& 0.484 (0.033) & \cellcolor{gray!62}{4}
& 0.504 (0.028) & \cellcolor{gray!87}{2} \\
& ARI
& N/A & \cellcolor{gray!0}{9}
& 0.004 (0.003) & \cellcolor{gray!37}{6}
& 0.001 (0.000) & \cellcolor{gray!25}{7.5}
& 0.091 (0.020) & \cellcolor{gray!62}{4}
& 0.102 (0.029) & \cellcolor{gray!75}{3}
& 0.001 (0.001) & \cellcolor{gray!25}{7.5}
& 0.088 (0.007) & \cellcolor{gray!50}{5}
& 0.105 (0.011) & \cellcolor{gray!87}{2}
& 0.123 (0.024) & \cellcolor{gray!100}{1} \\
& \# of Nodes
& --- &
& 237.7 (9.8) &
& 199.0 (8.6) &
& 6059.2 (308.6) &
& 229.2 (12.8) &
& 40.1 (2.3) &
& 1235.6 (171.6) &
& 828.5 (177.2) &
& 908.85 (266.1) & \\
& \# of Clusters
& N/A &
& 4.5 (1.7) &
& 1.8 (0.7) &
& 2530.4 (315.4) &
& 32.7 (3.6) &
& 1.8 (0.8) &
& 496.8 (28.9) &
& 271.9 (26.6) &
& 275.8 (105.7) & \\
\hline

Skin & NMI
& N/A & \cellcolor{gray!0}{9}
& 0.528 (0.087) & \cellcolor{gray!62}{4}
& 0.276 (0.256) & \cellcolor{gray!25}{7}
& 0.334 (0.023) & \cellcolor{gray!37}{6}
& 0.629 (0.016) & \cellcolor{gray!100}{1}
& 0.000 (0.000) & \cellcolor{gray!12}{8}
& 0.545 (0.048) & \cellcolor{gray!87}{2}
& 0.426 (0.052) & \cellcolor{gray!50}{5}
& 0.543 (0.066) & \cellcolor{gray!75}{3} \\
& ARI
& N/A & \cellcolor{gray!0}{9}
& 0.471 (0.206) & \cellcolor{gray!62}{4}
& 0.264 (0.300) & \cellcolor{gray!37}{6}
& 0.081 (0.034) & \cellcolor{gray!25}{7}
& 0.712 (0.028) & \cellcolor{gray!100}{1}
& 0.000 (0.000) & \cellcolor{gray!12}{8}
& 0.645 (0.086) & \cellcolor{gray!87}{2}
& 0.376 (0.157) & \cellcolor{gray!50}{5}
& 0.639 (0.099) & \cellcolor{gray!75}{3} \\
& \# of Nodes
& --- &
& 92.6 (11.6) &
& 431.8 (15.9) &
& 1461.8 (402.0) &
& 208.0 (9.1) &
& 27.9 (2.0) &
& 335.0 (53.1) &
& 1066.9 (35.5) &
& 390.1 (109.9) & \\
& \# of Clusters
& N/A &
& 3.0 (0.9) &
& 5.4 (2.4) &
& 144.6 (45.7) &
& 6.5 (1.4) &
& 1.0 (0.0) &
& 15.1 (3.7) &
& 66.1 (19.9) &
& 17.7 (9.5) & \\
\hline

& Average Rank
& \multicolumn{2}{c|}{\cellcolor{gray!25}{5.625 (2.826)}}
& \multicolumn{2}{c|}{\cellcolor{gray!37}{5.458 (1.732)}}
& \multicolumn{2}{c|}{\cellcolor{gray!12}{7.688 (1.306)}}
& \multicolumn{2}{c|}{\cellcolor{gray!87}{3.833 (1.818)}}
& \multicolumn{2}{c|}{\cellcolor{gray!50}{4.125 (1.965)}}
& \multicolumn{2}{c|}{\cellcolor{gray!0}{7.812 (1.600)}}
& \multicolumn{2}{c|}{\cellcolor{gray!100}{2.708 (1.814)}}
& \multicolumn{2}{c|}{\cellcolor{gray!87}{3.833 (2.375)}}
& \multicolumn{2}{c}{\cellcolor{gray!62}{3.917 (1.730)}} \\
\hline\hline
\end{tabular}}
\\
\vspace{1mm}
\footnotesize
\hspace*{7.0mm}
\begin{minipage}{\linewidth}
The values in parentheses indicate the standard deviation. \\
A number to the right of a metric value is the average rank of an algorithm over 20 evaluations. \\
The smaller the rank, the better the metric score. A darker tone in a cell corresponds to a smaller rank. \\
N/A indicates that an algorithm could not build a predictive model within 1 hour under the available computational resources. \\
AutoCloud does not have a node thus it is indicated by a symbol ``---''.
\end{minipage}
\end{table*}

\end{landscape}

For statistical comparisons, we use the Friedman test and the Nemenyi post-hoc analysis \cite{demvsar06}. The Friedman test assesses the null hypothesis that all algorithms have equal performance. If the null hypothesis is rejected, we then perform the Nemenyi post-hoc analysis. This post-hoc test is applied for all pairwise algorithm comparisons based on the ranks of performance metrics across all datasets. A difference in performance between two algorithms is considered statistically significant if the $ p $-value from the Nemenyi post-hoc analysis is less than the specified significance level. In this study, the null hypothesis is rejected at a significance level of $ 0.05 $ in both the Friedman test and the Nemenyi post-hoc analysis.

\begin{figure}[htbp]
	\centering
	\includegraphics[width=3.4in]{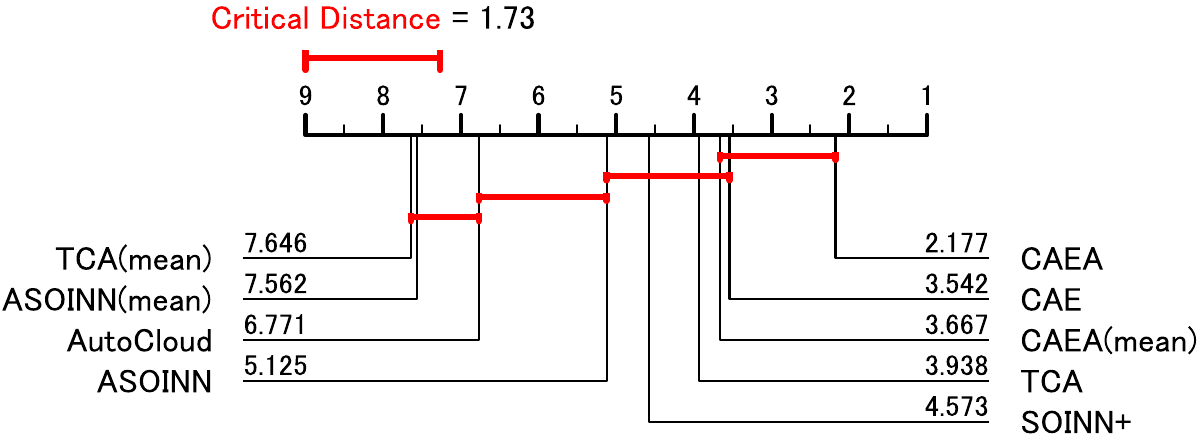}
	\vspace{2mm}
	\caption{Critical difference diagram based on the overall results of NMI and ARI.}
	\label{fig:cd_overall}
\end{figure}

Fig. \ref{fig:cd_overall} presents a critical difference diagram summarizing the overall results (i.e., all NMI and ARI values from both stationary and non-stationary environments). CAEA achieves the lowest average rank (i.e., the best overall performance). However, the rank of CAEA(mean) is lower than that of CAEA, indicating that the performance of CAEA is sensitive to parameter settings. Among the parameter-free or fixed-parameter algorithms (i.e., SOINN+, ASOINN(mean), TCA(mean), and CAEA(mean)), CAE attains the best average rank.

\begin{figure}[htbp]
	\centering
	\includegraphics[width=3.4in]{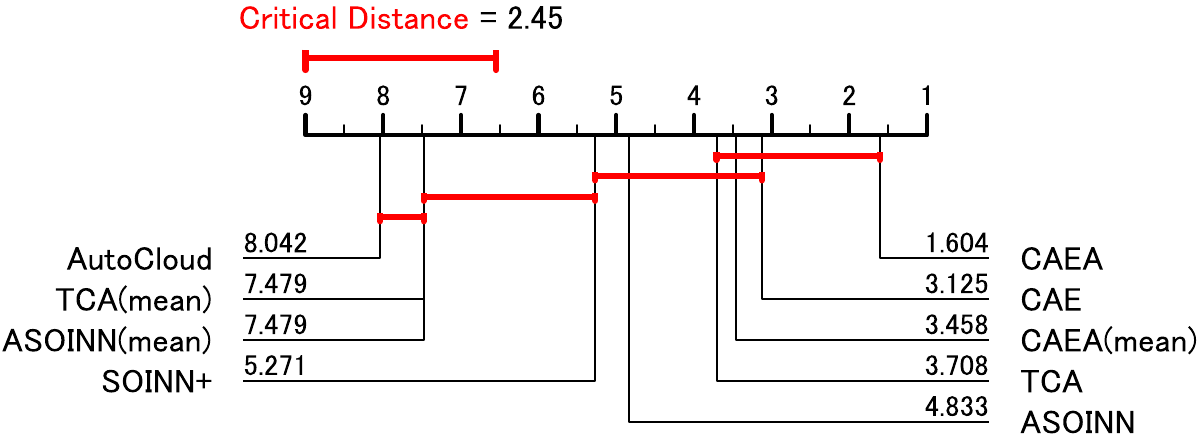}
	\vspace{2mm}
	\caption{Critical difference diagram based on results of NMI and ARI in the stationary environment.}
	\label{fig:cd_stationary}
\end{figure}

\begin{figure}[htbp]
	\vspace{-5mm}
	\centering
	\includegraphics[width=3.4in]{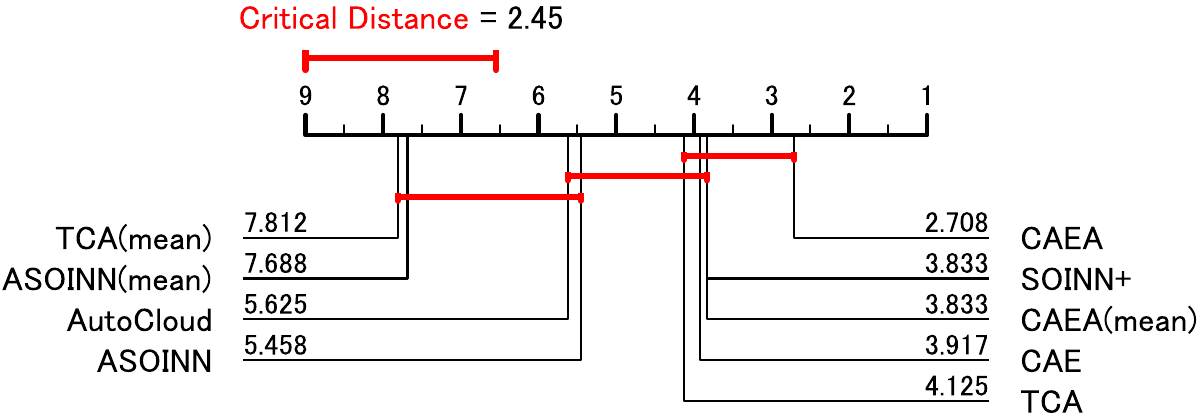}
	\vspace{2mm}
	\caption{Critical difference diagram based on results of NMI and ARI in the non-stationary environment.}
	\label{fig:cd_nonstationary}
\end{figure}

To discuss the features of each algorithm in detail, Figs. \ref{fig:cd_stationary} and \ref{fig:cd_nonstationary} present critical difference diagrams for the stationary and non-stationary environments, respectively. In the stationary environment (Fig. \ref{fig:cd_stationary}), CAEA attains the best (lowest) rank, followed by CAE, with CAEA(mean) and TCA showing comparable rankings. ASOINN ranks in the middle, while SOINN+, ASOINN(mean), and TCA(mean) are positioned in the lower tier alongside AutoCloud. In the non-stationary environment (Fig. \ref{fig:cd_nonstationary}), CAEA again achieves the best rank, with SOINN+, CAE, and CAEA(mean) forming the second tier. TCA follows slightly below, while ASOINN(mean) and TCA(mean) occupy the lowest tier.

These results highlight that ASOINN(mean) and TCA(mean) generally perform worse than their fully tuned counterparts and do not benefit from simple parameter averaging. Overall, CAEA, CAE, CAEA(mean), and TCA consistently achieve relatively higher ranks in both environments, whereas the performance of SOINN+ varies considerably depending on whether the environment is stationary or non-stationary.

In summary, while CAEA achieves the best clustering performance among the seven algorithms, it relies on manually pre-specified parameters, such as the similarity threshold and edge deletion threshold. In contrast, CAE removes the need for manual parameter tuning by estimating these parameters online, which enhances robustness across different datasets and environments and improves the practical usability of ART-based clustering.

\subsection{Continual Learning Performance}
\label{sec:continualperformance}
In addition to the final clustering performance reported in Section \ref{sec:realworld}, we further evaluate the continual learning capabilities of the algorithms. In continual learning scenarios, it is important not only to achieve high final performance but also to preserve previously acquired knowledge without experiencing significant forgetting.

\subsubsection{Evaluation Metrics}
\label{sec:continualMetric}

Average Incremental (AI) performance evaluates the average performance of an algorithm across all incremental stages, thereby measuring both the learning of new classes and the retention of previously learned ones. For a class-incremental learning scenario with $C$ classes, let $M_{c}$ denote the performance after training on the $c$-th class and testing on all classes learned up to that point. The AI is defined as
\begin{equation}
\text{AI} = \frac{1}{C}\sum_{c=1}^{C} M_{c}.
\end{equation}

Backward Transfer (BWT) quantifies how much performance on previously learned classes changes after subsequent classes, thus directly reflecting the degree of forgetting.  
Let $R_{i,j}$ denote the performance on class $i$ after training up to class $j$ $(j \geq i)$. Then, BWT is defined as
\begin{equation}
\text{BWT} = \frac{1}{C-1}\sum_{i=1}^{C-1}\big(R_{i,C} - R_{i,i}\big).
\end{equation}

Here, $R_{i,C}$ measures the final performance on the first $i$ classes after all training, while $R_{i,i}$ measures the performance immediately after learning class $i$. A negative BWT indicates forgetting, while a positive value suggests that later training has improved earlier knowledge.

In this paper, we adopt NMI and ARI as the underlying performance measures for AI and BWT. Accordingly, we report the results as AI-NMI, AI-ARI, BWT-NMI, and BWT-ARI.

\subsubsection{Results}
\label{sec:continualResults}
Table \ref{tab:ResultsAIBWT} summarizes the AI-NMI, AI-ARI, BWT-NMI, and BWT-ARI scores across all real-world datasets. The values in parentheses indicate the standard deviation. A number to the right of each evaluation metric is the average rank of an algorithm over 20 evaluations. The smaller the rank, the better the metric score. Additionally, a darker cell shade indicates a smaller rank (i.e., better performance).

It should be noted that these values are not obtained from separate experiments, but are derived from intermediate results already produced during the real-world dataset evaluations described in Section \ref{sec:realworld}. Specifically, during incremental training on each class, clustering performance (NMI and ARI) is recorded at each step; these values are subsequently used to calculate AI-NMI, AI-ARI, BWT-NMI, and BWT-ARI. Accordingly, the continual learning metrics directly complement the results from Section \ref{sec:realworld} without necessitating additional experiments.

As in Section \ref{sec:realworld}, the Friedman test and Nemenyi post-hoc analysis \cite{demvsar06} are used to facilitate a concise comparison of results in Table \ref{tab:ResultsAIBWT}. The Friedman test examines the null hypothesis that all algorithms perform equally. If this hypothesis is rejected at the $0.05$ significance level, the Nemenyi post-hoc analysis is conducted for all pairwise rank comparisons.

\begin{landscape}

\begin{table*}[htbp]
	\centering
	\caption{Results of quantitative comparisons for continual learning performance}
	\label{tab:ResultsAIBWT}
	\footnotesize
	\renewcommand{\arraystretch}{1.2}
	\scalebox{0.6}{
	\begin{tabular}{ll|*{8}{r C{3.7mm}|} r C{3.7mm}} \hline\hline
Dataset           & Metric
& \multicolumn{2}{c|}{AutoCloud}
& \multicolumn{2}{c|}{ASOINN}
& \multicolumn{2}{c|}{ASOINN(mean)}
& \multicolumn{2}{c|}{SOINN+}
& \multicolumn{2}{c|}{TCA}
& \multicolumn{2}{c|}{TCA(mean)}
& \multicolumn{2}{c|}{CAEA}
& \multicolumn{2}{c|}{CAEA(mean)}
& \multicolumn{2}{c}{CAE} \\ \hline
Iris               & AI-NMI   
&  0.871 (0.006) & \cellcolor{gray!100}{1}
&  0.805 (0.018) & \cellcolor{gray!75}{3}
&  0.732 (0.017) & \cellcolor{gray!50}{5}
&  0.722 (0.005) & \cellcolor{gray!37}{6}
&  0.818 (0.022) & \cellcolor{gray!87}{2}
&  0.800 (0.020) & \cellcolor{gray!62}{4}
&  0.683 (0.004) & \cellcolor{gray!25}{7}
&  0.621 (0.000) & \cellcolor{gray!12}{8}
&  0.566 (0.009) & \cellcolor{gray!0}{9}  \\
                   & AI-ARI   
&  0.877 (0.004) & \cellcolor{gray!100}{1}
&  0.749 (0.022) & \cellcolor{gray!75}{3}
&  0.669 (0.019) & \cellcolor{gray!50}{5}
&  0.634 (0.012) & \cellcolor{gray!37}{6}
&  0.758 (0.023) & \cellcolor{gray!87}{2}
&  0.737 (0.023) & \cellcolor{gray!62}{4}
&  0.554 (0.003) & \cellcolor{gray!25}{7}
&  0.337 (0.000) & \cellcolor{gray!0}{9}
&  0.388 (0.005) & \cellcolor{gray!12}{8}  \\
                   & BWT-NMI           
& -0.134 (0.015) & \cellcolor{gray!50}{5}
& -0.105 (0.038) & \cellcolor{gray!62}{4}
&  0.001 (0.040) & \cellcolor{gray!100}{1}
& -0.169 (0.006) & \cellcolor{gray!25}{7}
& -0.086 (0.050) & \cellcolor{gray!75}{3}
& -0.059 (0.046) & \cellcolor{gray!87}{2}
& -0.208 (0.008) & \cellcolor{gray!12}{8}
& -0.219 (0.000) & \cellcolor{gray!0}{9}
& -0.163 (0.014) & \cellcolor{gray!37}{6}  \\
                   & BWT-ARI           
& -0.172 (0.021) & \cellcolor{gray!87}{2}
& -0.223 (0.042) & \cellcolor{gray!75}{3.5}
& -0.126 (0.041) & \cellcolor{gray!100}{1}
& -0.256 (0.007) & \cellcolor{gray!37}{6}
& -0.254 (0.054) & \cellcolor{gray!50}{5}
& -0.223 (0.062) & \cellcolor{gray!75}{3.5}
& -0.311 (0.005) & \cellcolor{gray!25}{7}
& -0.491 (0.000) & \cellcolor{gray!0}{9}
& -0.404 (0.015) & \cellcolor{gray!12}{8}  \\
\hline
Ionosphere         & AI-NMI   
&  0.508 (0.000) & \cellcolor{gray!0}{9}
&  0.577 (0.002) & \cellcolor{gray!62}{4}
&  0.554 (0.002) & \cellcolor{gray!37}{6}
&  0.584 (0.002) & \cellcolor{gray!75}{3}
&  0.544 (0.001) & \cellcolor{gray!25}{7}
&  0.511 (0.001) & \cellcolor{gray!12}{8}
&  0.559 (0.001) & \cellcolor{gray!50}{5}
&  0.619 (0.002) & \cellcolor{gray!87}{2}
&  0.624 (0.002) & \cellcolor{gray!100}{1}  \\
                   & AI-ARI   
&  0.507 (0.001) & \cellcolor{gray!0}{9}
&  0.547 (0.002) & \cellcolor{gray!100}{1.5}
&  0.527 (0.001) & \cellcolor{gray!37}{6}
&  0.543 (0.001) & \cellcolor{gray!75}{3}
&  0.536 (0.001) & \cellcolor{gray!50}{5}
&  0.509 (0.001) & \cellcolor{gray!12}{8}
&  0.541 (0.001) & \cellcolor{gray!62}{4}
&  0.516 (0.000) & \cellcolor{gray!25}{7}
&  0.547 (0.002) & \cellcolor{gray!100}{1.5}  \\
                   & BWT-NMI           
& -0.985 (0.002) & \cellcolor{gray!0}{9}
& -0.846 (0.006) & \cellcolor{gray!62}{4}
& -0.893 (0.009) & \cellcolor{gray!37}{6}
& -0.831 (0.006) & \cellcolor{gray!75}{3}
& -0.912 (0.003) & \cellcolor{gray!25}{7}
& -0.977 (0.002) & \cellcolor{gray!12}{8}
& -0.882 (0.005) & \cellcolor{gray!50}{5}
& -0.763 (0.008) & \cellcolor{gray!87}{2}
& -0.752 (0.008) & \cellcolor{gray!100}{1} \\
                   & BWT-ARI           
& -0.986 (0.002) & \cellcolor{gray!0}{9}
& -0.906 (0.007) & \cellcolor{gray!87}{2}
& -0.947 (0.006) & \cellcolor{gray!37}{6}
& -0.914 (0.006) & \cellcolor{gray!75}{3}
& -0.928 (0.004) & \cellcolor{gray!50}{5}
& -0.981 (0.004) & \cellcolor{gray!12}{8}
& -0.918 (0.004) & \cellcolor{gray!62}{4}
& -0.968 (0.001) & \cellcolor{gray!25}{7}
& -0.905 (0.008) & \cellcolor{gray!100}{1}  \\
\hline
Pima               & AI-NMI   
&  0.509 (0.000) & \cellcolor{gray!25}{7}
&  0.502 (0.000) & \cellcolor{gray!12}{8.5}
&  0.502 (0.000) & \cellcolor{gray!12}{8.5}
&  0.522 (0.000) & \cellcolor{gray!50}{5}
&  0.524 (0.000) & \cellcolor{gray!62}{4}
&  0.512 (0.000) & \cellcolor{gray!37}{6}
&  0.565 (0.000) & \cellcolor{gray!75}{3}
&  0.571 (0.000) & \cellcolor{gray!87}{2}
&  0.572 (0.001) & \cellcolor{gray!100}{1}  \\
                   & AI-ARI   
&  0.513 (0.001) & \cellcolor{gray!37}{6}
&  0.504 (0.000) & \cellcolor{gray!0}{9}
&  0.505 (0.000) & \cellcolor{gray!12}{8}
&  0.525 (0.001) & \cellcolor{gray!62}{4}
&  0.530 (0.000) & \cellcolor{gray!75}{3}
&  0.507 (0.000) & \cellcolor{gray!25}{7}
&  0.542 (0.000) & \cellcolor{gray!87}{2}
&  0.550 (0.001) & \cellcolor{gray!100}{1}
&  0.522 (0.000) & \cellcolor{gray!50}{5}  \\
                   & BWT-NMI           
& -0.982 (0.001) & \cellcolor{gray!37}{7}
& -0.996 (0.000) & \cellcolor{gray!12}{8.5}
& -0.996 (0.000) & \cellcolor{gray!12}{8.5}
& -0.955 (0.000) & \cellcolor{gray!50}{5}
& -0.953 (0.000) & \cellcolor{gray!62}{4}
& -0.977 (0.001) & \cellcolor{gray!25}{6}
& -0.869 (0.000) & \cellcolor{gray!75}{3}
& -0.857 (0.001) & \cellcolor{gray!87}{2}
& -0.856 (0.004) & \cellcolor{gray!100}{1} \\
                   & BWT-ARI           
& -0.975 (0.003) & \cellcolor{gray!37}{6}
& -0.993 (0.000) & \cellcolor{gray!0}{9}
& -0.989 (0.000) & \cellcolor{gray!12}{8}
& -0.950 (0.003) & \cellcolor{gray!62}{4}
& -0.941 (0.001) & \cellcolor{gray!75}{3}
& -0.986 (0.000) & \cellcolor{gray!25}{7}
& -0.917 (0.002) & \cellcolor{gray!87}{2}
& -0.901 (0.003) & \cellcolor{gray!100}{1}
& -0.956 (0.002) & \cellcolor{gray!50}{5}  \\
\hline
Binalpha           & AI-NMI   
&  0.362 (0.004) & \cellcolor{gray!12}{8}
&  0.517 (0.006) & \cellcolor{gray!50}{5}
&  0.514 (0.007) & \cellcolor{gray!37}{6}
&  0.625 (0.002) & \cellcolor{gray!100}{1}
&  0.589 (0.000) & \cellcolor{gray!87}{2}
&  0.143 (0.050) & \cellcolor{gray!0}{9}
&  0.571 (0.000) & \cellcolor{gray!75}{3}
&  0.533 (0.000) & \cellcolor{gray!62}{4}
&  0.486 (0.002) & \cellcolor{gray!25}{7}  \\
                   & AI-ARI   
&  0.131 (0.002) & \cellcolor{gray!12}{8}
&  0.275 (0.007) & \cellcolor{gray!87}{2}
&  0.272 (0.008) & \cellcolor{gray!75}{3}
&  0.315 (0.005) & \cellcolor{gray!100}{1}
&  0.189 (0.001) & \cellcolor{gray!62}{4}
&  0.032 (0.003) & \cellcolor{gray!0}{9}
&  0.150 (0.001) & \cellcolor{gray!25}{7}
&  0.185 (0.001) & \cellcolor{gray!50}{5}
&  0.158 (0.006) & \cellcolor{gray!37}{6}  \\
                   & BWT-NMI           
& -0.084 (0.010) & \cellcolor{gray!50}{5}
& -0.110 (0.002) & \cellcolor{gray!37}{6}
& -0.138 (0.002) & \cellcolor{gray!25}{7}
& -0.012 (0.001) & \cellcolor{gray!62}{4}
& -0.178 (0.001) & \cellcolor{gray!0}{9}
& -0.147 (0.053) & \cellcolor{gray!12}{8}
&  0.029 (0.000) & \cellcolor{gray!87}{2}
& -0.010 (0.000) & \cellcolor{gray!75}{3}
&  0.066 (0.005) & \cellcolor{gray!100}{1}  \\
                   & BWT-ARI           
& -0.107 (0.002) & \cellcolor{gray!37}{6}
& -0.177 (0.004) & \cellcolor{gray!12}{8}
& -0.204 (0.003) & \cellcolor{gray!0}{9}
& -0.112 (0.003) & \cellcolor{gray!25}{7}
& -0.048 (0.001) & \cellcolor{gray!50}{5}
& -0.033 (0.003) & \cellcolor{gray!62}{4}
&  0.017 (0.000) & \cellcolor{gray!100}{1}
&  0.003 (0.001) & \cellcolor{gray!75}{3}
&  0.007 (0.002) & \cellcolor{gray!87}{2}  \\
\hline
Yeast              & AI-NMI   
&  0.230 (0.009) & \cellcolor{gray!25}{7}
&  0.178 (0.008) & \cellcolor{gray!12}{8}
&  0.176 (0.009) & \cellcolor{gray!0}{9}
&  0.308 (0.002) & \cellcolor{gray!75}{3}
&  0.248 (0.003) & \cellcolor{gray!50}{5}
&  0.242 (0.003) & \cellcolor{gray!37}{6}
&  0.345 (0.001) & \cellcolor{gray!100}{1}
&  0.325 (0.001) & \cellcolor{gray!87}{2}
&  0.303 (0.004) & \cellcolor{gray!62}{4}  \\
                   & AI-ARI   
&  0.212 (0.010) & \cellcolor{gray!87}{2}
&  0.163 (0.005) & \cellcolor{gray!12}{8}
&  0.164 (0.008) & \cellcolor{gray!25}{7}
&  0.248 (0.004) & \cellcolor{gray!100}{1}
&  0.175 (0.002) & \cellcolor{gray!62}{4.5}
&  0.175 (0.002) & \cellcolor{gray!62}{4.5}
&  0.132 (0.000) & \cellcolor{gray!0}{9}
&  0.168 (0.001) & \cellcolor{gray!37}{6}
&  0.178 (0.003) & \cellcolor{gray!75}{3}  \\
                   & BWT-NMI           
& -0.202 (0.010) & \cellcolor{gray!0}{9}
& -0.187 (0.007) & \cellcolor{gray!12}{8}
& -0.186 (0.009) & \cellcolor{gray!25}{7}
& -0.151 (0.005) & \cellcolor{gray!62}{4}
& -0.175 (0.004) & \cellcolor{gray!50}{5}
& -0.179 (0.003) & \cellcolor{gray!37}{6}
& -0.052 (0.003) & \cellcolor{gray!100}{1}
& -0.077 (0.003) & \cellcolor{gray!87}{2}
& -0.115 (0.006) & \cellcolor{gray!75}{3}  \\
                   & BWT-ARI           
& -0.223 (0.011) & \cellcolor{gray!12}{8}
& -0.180 (0.007) & \cellcolor{gray!62}{4.5}
& -0.182 (0.009) & \cellcolor{gray!37}{6}
& -0.228 (0.007) & \cellcolor{gray!0}{9}
& -0.186 (0.002) & \cellcolor{gray!25}{7}
& -0.180 (0.002) & \cellcolor{gray!62}{4.5}
& -0.074 (0.001) & \cellcolor{gray!87}{2}
& -0.059 (0.002) & \cellcolor{gray!100}{1}
& -0.103 (0.008) & \cellcolor{gray!75}{3}  \\
\hline
Semeion            & AI-NMI   
&  0.470 (0.011) & \cellcolor{gray!12}{8}
&  0.624 (0.002) & \cellcolor{gray!100}{1}
&  0.604 (0.006) & \cellcolor{gray!87}{2}
&  0.603 (0.002) & \cellcolor{gray!75}{3}
&  0.559 (0.001) & \cellcolor{gray!50}{5}
&  0.284 (0.037) & \cellcolor{gray!0}{9}
&  0.564 (0.001) & \cellcolor{gray!62}{4}
&  0.537 (0.001) & \cellcolor{gray!37}{6}
&  0.487 (0.001) & \cellcolor{gray!25}{7}  \\
                   & AI-ARI   
&  0.322 (0.015) & \cellcolor{gray!50}{5}
&  0.497 (0.005) & \cellcolor{gray!100}{1}
&  0.475 (0.015) & \cellcolor{gray!87}{2}
&  0.430 (0.007) & \cellcolor{gray!75}{3}
&  0.312 (0.004) & \cellcolor{gray!37}{6}
&  0.142 (0.010) & \cellcolor{gray!0}{9}
&  0.365 (0.004) & \cellcolor{gray!62}{4}
&  0.283 (0.003) & \cellcolor{gray!25}{7}
&  0.213 (0.003) & \cellcolor{gray!12}{8}  \\
                   & BWT-NMI           
&  -0.113 (0.006) & \cellcolor{gray!12}{8}
&  -0.082 (0.006) & \cellcolor{gray!25}{7}
&  -0.065 (0.003) & \cellcolor{gray!37}{6}
&  -0.031 (0.002) & \cellcolor{gray!87}{2}
&  -0.058 (0.001) & \cellcolor{gray!50}{5}
&  -0.237 (0.034) & \cellcolor{gray!0}{9}
&  -0.046 (0.003) & \cellcolor{gray!62}{4}
&  -0.032 (0.002) & \cellcolor{gray!75}{3}
&  -0.002 (0.002) & \cellcolor{gray!100}{1}  \\
                   & BWT-ARI           
&  -0.191 (0.008) & \cellcolor{gray!0}{9}
&  -0.184 (0.016) & \cellcolor{gray!12}{8}
&  -0.159 (0.007) & \cellcolor{gray!25}{7}
&  -0.144 (0.003) & \cellcolor{gray!37}{6}
&  -0.008 (0.005) & \cellcolor{gray!100}{1}
&  -0.122 (0.009) & \cellcolor{gray!50}{5}
&  -0.071 (0.005) & \cellcolor{gray!62}{4}
&  -0.037 (0.003) & \cellcolor{gray!87}{2}
&  -0.055 (0.003) & \cellcolor{gray!75}{3}  \\
\hline
Image  & AI-NMI   
&  0.780 (0.006) & \cellcolor{gray!100}{1}
&  0.707 (0.011) & \cellcolor{gray!87}{2}
&  0.667 (0.016) & \cellcolor{gray!62}{4}
&  0.640 (0.001) & \cellcolor{gray!50}{5}
&  0.686 (0.006) & \cellcolor{gray!75}{3}
&  0.564 (0.031) & \cellcolor{gray!0}{9}
&  0.588 (0.003) & \cellcolor{gray!25}{7}
&  0.584 (0.000) & \cellcolor{gray!12}{8}
&  0.620 (0.002) & \cellcolor{gray!37}{6}  \\
Segmentation                   & AI-ARI   
&  0.718 (0.008) & \cellcolor{gray!100}{1}
&  0.550 (0.016) & \cellcolor{gray!87}{2}
&  0.517 (0.023) & \cellcolor{gray!62}{4}
&  0.429 (0.008) & \cellcolor{gray!50}{5}
&  0.547 (0.009) & \cellcolor{gray!75}{3}
&  0.394 (0.022) & \cellcolor{gray!37}{6}
&  0.347 (0.003) & \cellcolor{gray!12}{8}
&  0.279 (0.002) & \cellcolor{gray!0}{9}
&  0.377 (0.006) & \cellcolor{gray!25}{7}  \\
                   & BWT-NMI           
&  -0.141 (0.008) & \cellcolor{gray!0}{9}
&  -0.112 (0.014) & \cellcolor{gray!12}{8}
&  -0.070 (0.024) & \cellcolor{gray!50}{5}
&  -0.048 (0.001) & \cellcolor{gray!62}{4}
&  -0.109 (0.006) & \cellcolor{gray!25}{7}
&  -0.098 (0.049) & \cellcolor{gray!37}{6}
&  -0.027 (0.001) & \cellcolor{gray!87}{2.5}
&  -0.025 (0.001) & \cellcolor{gray!100}{1}
&  -0.027 (0.002) & \cellcolor{gray!87}{2.5}  \\
                   & BWT-ARI           
&  -0.287 (0.012) & \cellcolor{gray!12}{8}
&  -0.316 (0.023) & \cellcolor{gray!0}{9}
&  -0.267 (0.037) & \cellcolor{gray!37}{6}
&  -0.176 (0.002) & \cellcolor{gray!62}{4}
&  -0.285 (0.009) & \cellcolor{gray!25}{7}
&  -0.264 (0.027) & \cellcolor{gray!50}{5}
&  -0.046 (0.002) & \cellcolor{gray!100}{1}
&  -0.091 (0.002) & \cellcolor{gray!67}{3}
&  -0.069 (0.003) & \cellcolor{gray!87}{2}  \\
\hline
Phoneme            & AI-NMI   
&  0.536 (0.001) & \cellcolor{gray!50}{5}
&  0.513 (0.000) & \cellcolor{gray!25}{7}
&  0.505 (0.000) & \cellcolor{gray!12}{8}
&  0.519 (0.000) & \cellcolor{gray!37}{6}
&  0.539 (0.000) & \cellcolor{gray!62}{4}
&  0.501 (0.000) & \cellcolor{gray!0}{9}
&  0.583 (0.000) & \cellcolor{gray!100}{1}
&  0.581 (0.000) & \cellcolor{gray!87}{2}
&  0.579 (0.000) & \cellcolor{gray!75}{3}  \\
                   & AI-ARI   
&  0.565 (0.003) & \cellcolor{gray!100}{1}
&  0.501 (0.000) & \cellcolor{gray!25}{7.5}
&  0.500 (0.000) & \cellcolor{gray!0}{9}
&  0.508 (0.000) & \cellcolor{gray!37}{6}
&  0.561 (0.001) & \cellcolor{gray!87}{2}
&  0.501 (0.000) & \cellcolor{gray!25}{7.5}
&  0.521 (0.000) & \cellcolor{gray!50}{5}
&  0.533 (0.000) & \cellcolor{gray!62}{4}
&  0.545 (0.000) & \cellcolor{gray!75}{3}  \\
                   & BWT-NMI           
&  -0.928 (0.005) & \cellcolor{gray!50}{5}
&  -0.974 (0.000) & \cellcolor{gray!25}{7}
&  -0.990 (0.000) & \cellcolor{gray!12}{8}
&  -0.962 (0.000) & \cellcolor{gray!37}{6}
&  -0.922 (0.002) & \cellcolor{gray!62}{4}
&  -0.997 (0.000) & \cellcolor{gray!0}{9}
&  -0.834 (0.000) & \cellcolor{gray!100}{1}
&  -0.838 (0.000) & \cellcolor{gray!87}{2}
&  -0.843 (0.000) & \cellcolor{gray!75}{3}  \\
                   & BWT-ARI           
&  -0.870 (0.012) & \cellcolor{gray!100}{1}
&  -0.999 (0.000) & \cellcolor{gray!12}{8.5}
&  -0.999 (0.000) & \cellcolor{gray!12}{8.5}
&  -0.984 (0.000) & \cellcolor{gray!37}{6}
&  -0.877 (0.006) & \cellcolor{gray!87}{2}
&  -0.998 (0.000) & \cellcolor{gray!25}{7}
&  -0.957 (0.000) & \cellcolor{gray!50}{5}
&  -0.934 (0.001) & \cellcolor{gray!62}{4}
&  -0.910 (0.001) & \cellcolor{gray!75}{3}  \\
\hline
Texture            & AI-NMI   
&  0.709 (0.004) & \cellcolor{gray!75}{3}
&  0.726 (0.015) & \cellcolor{gray!87}{2}
&  0.744 (0.010) & \cellcolor{gray!100}{1}
&  0.696 (0.004) & \cellcolor{gray!62}{4}
&  0.669 (0.011) & \cellcolor{gray!50}{5}
&  0.124 (0.004) & \cellcolor{gray!0}{9}
&  0.648 (0.002) & \cellcolor{gray!37}{6}
&  0.635 (0.003) & \cellcolor{gray!12}{8}
&  0.644 (0.002) & \cellcolor{gray!25}{7}  \\
                   & AI-ARI   
&  0.585 (0.009) & \cellcolor{gray!100}{1}
&  0.479 (0.021) & \cellcolor{gray!62}{4}
&  0.502 (0.011) & \cellcolor{gray!75}{3}
&  0.578 (0.010) & \cellcolor{gray!87}{2}
&  0.403 (0.012) & \cellcolor{gray!12}{8}
&  0.119 (0.003) & \cellcolor{gray!0}{9}
&  0.453 (0.010) & \cellcolor{gray!50}{5}
&  0.417 (0.015) & \cellcolor{gray!25}{7}
&  0.445 (0.013) & \cellcolor{gray!37}{6}  \\
                   & BWT-NMI           
&  -0.107 (0.006) & \cellcolor{gray!50}{5}
&  -0.138 (0.015) & \cellcolor{gray!12}{8}
&  -0.143 (0.016) & \cellcolor{gray!0}{9}
&  -0.034 (0.004) & \cellcolor{gray!62}{4}
&  -0.110 (0.020) & \cellcolor{gray!37}{6}
&  -0.137 (0.005) & \cellcolor{gray!25}{7}
&  -0.008 (0.001) & \cellcolor{gray!75}{3}
&  0.003 (0.001) & \cellcolor{gray!100}{1}
&  0.001 (0.003) & \cellcolor{gray!87}{2}  \\
                   & BWT-ARI           
&  -0.262 (0.011) & \cellcolor{gray!37}{6}
&  -0.351 (0.020) & \cellcolor{gray!12}{8}
&  -0.370 (0.017) & \cellcolor{gray!0}{9}
&  -0.164 (0.007) & \cellcolor{gray!50}{5}
&  -0.297 (0.017) & \cellcolor{gray!25}{7}
&  -0.131 (0.004) & \cellcolor{gray!62}{4}
&  -0.064 (0.004) & \cellcolor{gray!75}{3}
&  -0.036 (0.004) & \cellcolor{gray!100}{1}
&  -0.059 (0.013) & \cellcolor{gray!87}{2}  \\
\hline
PenBased           & AI-NMI  
&  0.760 (0.001) & \cellcolor{gray!75}{3}
&  0.815 (0.006) & \cellcolor{gray!100}{1}
&  0.769 (0.008) & \cellcolor{gray!87}{2}
&  0.743 (0.002) & \cellcolor{gray!50}{5}
&  0.710 (0.005) & \cellcolor{gray!37}{6}
&  0.690 (0.005) & \cellcolor{gray!25}{7}
&  0.744 (0.006) & \cellcolor{gray!62}{4}
&  0.674 (0.005) & \cellcolor{gray!12}{8}
&  0.636 (0.003) & \cellcolor{gray!0}{9}  \\
                   & AI-ARI  
&  0.701 (0.002) & \cellcolor{gray!87}{2}
&  0.733 (0.020) & \cellcolor{gray!100}{1}
&  0.636 (0.020) & \cellcolor{gray!50}{5}
&  0.678 (0.007) & \cellcolor{gray!75}{3}
&  0.543 (0.010) & \cellcolor{gray!37}{6}
&  0.505 (0.010) & \cellcolor{gray!12}{8}
&  0.670 (0.012) & \cellcolor{gray!62}{4}
&  0.531 (0.014) & \cellcolor{gray!25}{7}
&  0.445 (0.016) & \cellcolor{gray!0}{9}  \\
                   & BWT-NMI           
&  -0.080 (0.004) & \cellcolor{gray!37}{6}
&  -0.047 (0.002) & \cellcolor{gray!50}{5}
&  -0.096 (0.014) & \cellcolor{gray!25}{7}
&  -0.035 (0.001) & \cellcolor{gray!62}{4}
&  -0.162 (0.015) & \cellcolor{gray!0}{9}
&  -0.127 (0.011) & \cellcolor{gray!12}{8}
&  -0.020 (0.002) & \cellcolor{gray!37}{3}
&  0.016 (0.001) & \cellcolor{gray!87}{2}
&  0.020 (0.001) & \cellcolor{gray!100}{1}  \\
                   & BWT-ARI           
&  -0.184 (0.008) & \cellcolor{gray!37}{6}
&  -0.151 (0.010) & \cellcolor{gray!50}{5}
&  -0.282 (0.033) & \cellcolor{gray!25}{7}
&  -0.125 (0.003) & \cellcolor{gray!62}{4}
&  -0.360 (0.029) & \cellcolor{gray!0}{9}
&  -0.317 (0.013) & \cellcolor{gray!12}{8}
&  -0.094 (0.006) & \cellcolor{gray!75}{3}
&  -0.016 (0.002) & \cellcolor{gray!100}{1}
&  -0.023 (0.006) & \cellcolor{gray!87}{2}  \\
\hline
Letter             & AI-NMI   
&  N/A \hspace{3.6mm} & \cellcolor{gray!0}{9}
&  0.441 (0.002) & \cellcolor{gray!37}{6}
&  0.365 (0.003) & \cellcolor{gray!25}{7}
&  0.571 (0.001) & \cellcolor{gray!100}{1}
&  0.565 (0.001) & \cellcolor{gray!87}{2}
&  0.048 (0.000) & \cellcolor{gray!12}{8}
&  0.482 (0.000) & \cellcolor{gray!62}{4}
&  0.473 (0.001) & \cellcolor{gray!50}{5}
&  0.494 (0.001) & \cellcolor{gray!75}{3}  \\
                   & AI-ARI   
&  N/A \hspace{3.6mm} & \cellcolor{gray!0}{9}
&  0.124 (0.001) & \cellcolor{gray!50}{5}
&  0.104 (0.001) & \cellcolor{gray!25}{7}
&  0.179 (0.001) & \cellcolor{gray!87}{2}
&  0.230 (0.001) & \cellcolor{gray!100}{1}
&  0.040 (0.000) & \cellcolor{gray!12}{8}
&  0.115 (0.000) & \cellcolor{gray!37}{6}
&  0.135 (0.000) & \cellcolor{gray!62}{4}
&  0.157 (0.001) & \cellcolor{gray!75}{3}  \\
                   & BWT-NMI           
&  N/A \hspace{3.6mm} & \cellcolor{gray!0}{9}
&  -0.059 (0.001) & \cellcolor{gray!25}{7}
&  -0.074 (0.004) & \cellcolor{gray!12}{8}
&  0.020 (0.000) & \cellcolor{gray!100}{1}
&  -0.050 (0.001) & \cellcolor{gray!50}{5.5}
&  -0.050 (0.000) & \cellcolor{gray!50}{5.5}
&  -0.006 (0.000) & \cellcolor{gray!62}{4}
&  -0.001 (0.000) & \cellcolor{gray!75}{3}
&  0.010 (0.000) & \cellcolor{gray!87}{2}  \\
                   & BWT-ARI           
&  N/A \hspace{3.6mm} & \cellcolor{gray!0}{9}
&  -0.110 (0.001) & \cellcolor{gray!25}{7}
&  -0.100 (0.001) & \cellcolor{gray!37}{6}
&  -0.092 (0.000) & \cellcolor{gray!50}{5}
&  -0.157 (0.001) & \cellcolor{gray!12}{8}
&  -0.042 (0.000) & \cellcolor{gray!62}{4}
&  -0.029 (0.000) & \cellcolor{gray!100}{1}
&  -0.035 (0.000) & \cellcolor{gray!87}{2.5}
&  -0.035 (0.000) & \cellcolor{gray!87}{2.5}  \\
\hline
Skin               & AI-NMI   
&  N/A \hspace{3.6mm} & \cellcolor{gray!0}{9}
&  0.765 (0.001) & \cellcolor{gray!62}{4}
&  0.715 (0.025) & \cellcolor{gray!37}{6}
&  0.667 (0.000) & \cellcolor{gray!25}{7}
&  0.813 (0.000) & \cellcolor{gray!100}{1}
&  0.501 (0.000) & \cellcolor{gray!12}{8}
&  0.770 (0.001) & \cellcolor{gray!75}{3}
&  0.717 (0.001) & \cellcolor{gray!50}{5}
&  0.772 (0.001) & \cellcolor{gray!87}{2}  \\
                   & AI-ARI   
&  N/A \hspace{3.6mm} & \cellcolor{gray!0}{9}
&  0.758 (0.006) & \cellcolor{gray!62}{4}
&  0.714 (0.034) & \cellcolor{gray!50}{5}
&  0.540 (0.000) & \cellcolor{gray!25}{7}
&  0.846 (0.000) & \cellcolor{gray!100}{1}
&  0.500 (0.000) & \cellcolor{gray!12}{8}
&  0.821 (0.003) & \cellcolor{gray!87}{2}
&  0.698 (0.006) & \cellcolor{gray!37}{6}
&  0.820 (0.003) & \cellcolor{gray!75}{3}  \\
                   & BWT-NMI          
&  N/A \hspace{3.6mm} & \cellcolor{gray!0}{9}
&  -0.471 (0.006) & \cellcolor{gray!62}{4}
&  -0.571 (0.099) & \cellcolor{gray!37}{6}
&  -0.666 (0.001) & \cellcolor{gray!25}{7}
&  -0.374 (0.000) & \cellcolor{gray!100}{1}
&  -0.999 (0.000) & \cellcolor{gray!12}{8}
&  -0.460 (0.003) & \cellcolor{gray!75}{3}
&  -0.566 (0.002) & \cellcolor{gray!50}{5}
&  -0.457 (0.005) & \cellcolor{gray!87}{2}  \\
                   & BWT-ARI           
&  N/A \hspace{3.6mm} & \cellcolor{gray!0}{9}
&  -0.483 (0.026) & \cellcolor{gray!62}{4}
&  -0.573 (0.135) & \cellcolor{gray!50}{5}
&  -0.919 (0.001) & \cellcolor{gray!25}{7}
&  -0.309 (0.000) & \cellcolor{gray!100}{1}
&  -1.000 (0.000) & \cellcolor{gray!12}{8}
&  -0.358 (0.013) & \cellcolor{gray!87}{2}
&  -0.605 (0.024) & \cellcolor{gray!37}{6}
&  -0.361 (0.010) & \cellcolor{gray!75}{3}  \\
\hline
                          & Average Rank 
& \multicolumn{2}{c|}{\cellcolor{gray!12}{6.021 (2.919)}}
& \multicolumn{2}{c|}{\cellcolor{gray!37}{5.260 (2.588)}}
& \multicolumn{2}{c|}{\cellcolor{gray!25}{5.948 (2.267)}}
& \multicolumn{2}{c|}{\cellcolor{gray!62}{4.354 (1.920)}}
& \multicolumn{2}{c|}{\cellcolor{gray!50}{4.521 (2.321)}}
& \multicolumn{2}{c|}{\cellcolor{gray!0}{6.885 (1.852)}}
& \multicolumn{2}{c|}{\cellcolor{gray!87}{3.865 (2.103)}}
& \multicolumn{2}{c|}{\cellcolor{gray!75}{4.323 (2.639)}}
& \multicolumn{2}{c}{\cellcolor{gray!100}{3.823 (2.509)}} \\
            \hline\hline
\end{tabular}}
	\\
	\vspace{1mm}
	\footnotesize
	\hspace*{6mm}
	\begin{minipage}{\linewidth}
	The values in parentheses indicate the standard deviation. \\
	A number to the right of a metric value is the average rank of an algorithm over 20 evaluations. \\
	The smaller the rank, the better the metric score. A darker tone in a cell corresponds to a smaller rank. \\
	N/A indicates that an algorithm could not build a predictive model within 1 hour under the available computational resources. \\
	\end{minipage}
\end{table*}

\end{landscape}

\begin{figure}[htbp]
	\centering
	\includegraphics[width=3.4in]{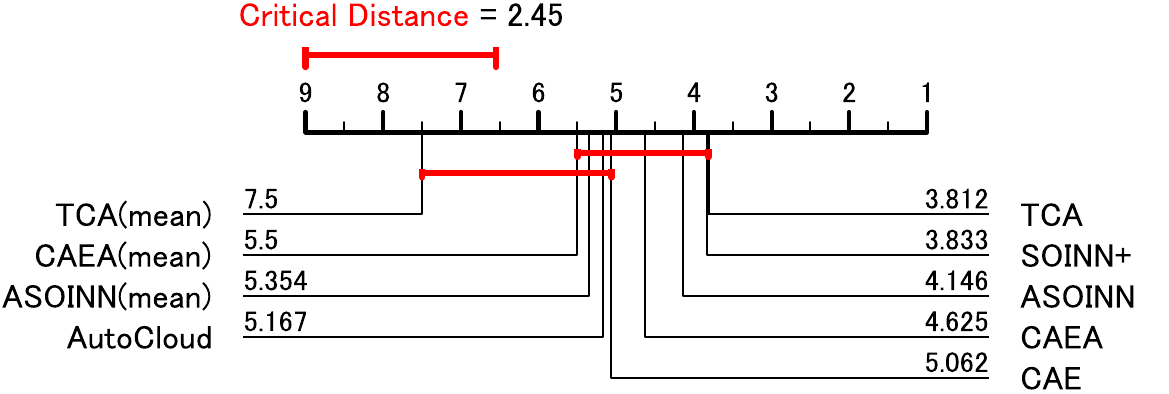}
	\vspace{2mm}
	\caption{Critical difference diagram based on results of AI-NMI and AI-ARI.}
	\label{fig:cd_IncNMIARI}
\end{figure}

As shown in Fig. \ref{fig:cd_IncNMIARI}, algorithms with parameter tuning (TCA, AutoCloud, ASOINN) generally achieve lower (better) ranks, confirming their ability to adapt quickly when parameters are carefully optimized. The parameter-free algorithms (CAE and SOINN+) also attain competitive ranks. In particular, SOINN+ performs comparably to TCA, AutoCloud, and ASOINN, despite not requiring any manual parameter specification.

\begin{figure}[htbp]
	\vspace{-5mm}
	\centering
	\includegraphics[width=3.4in]{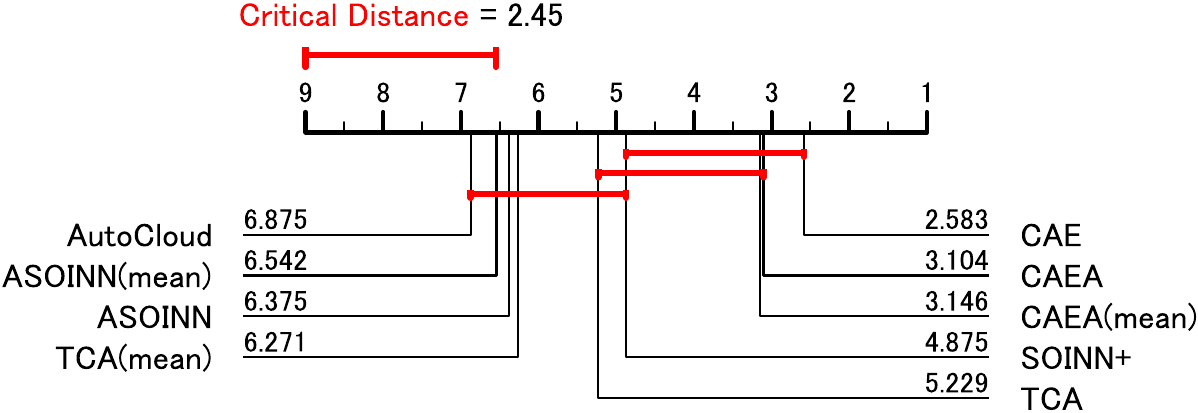}
	\vspace{2mm}
	\caption{Critical difference diagram based on results of BWT-NMI and BWT-ARI.}
	\label{fig:cd_BWT} 
\end{figure}

In contrast, the fixed-parameter variants (ASOINN(mean), TCA(mean), CAEA(mean)) display clearly higher ranks, indicating that simply averaging parameters deteriorates performance. Fig. \ref{fig:cd_BWT} demonstrates that CAE and CAEA constitute the leading group in terms of BWT, exhibiting significantly less forgetting than the other algorithms. Notably, while CAEA requires pre-specified parameters to achieve such robustness, CAE provides comparable or even superior resistance to forgetting in a fully parameter-free manner.


\subsection{Validity of the Number of Active Nodes}
\label{sec:validity}
This section analyzes and discusses the validity of the number of active nodes and its effect on clustering performance (i.e., NMI) and the number of clusters in CAE. First, we manually set the value of $\lambda$ (the number of active nodes) and perform the training process of CAE without the automatic estimation of $\lambda$. We then evaluate the trained network to obtain the NMI value. All other experimental settings are identical to those described in Section \ref{sec:realworld}.

\begin{figure*}[htbp]
	\centering
	\subfloat[Iris]{
		\includegraphics[width=1.1in]{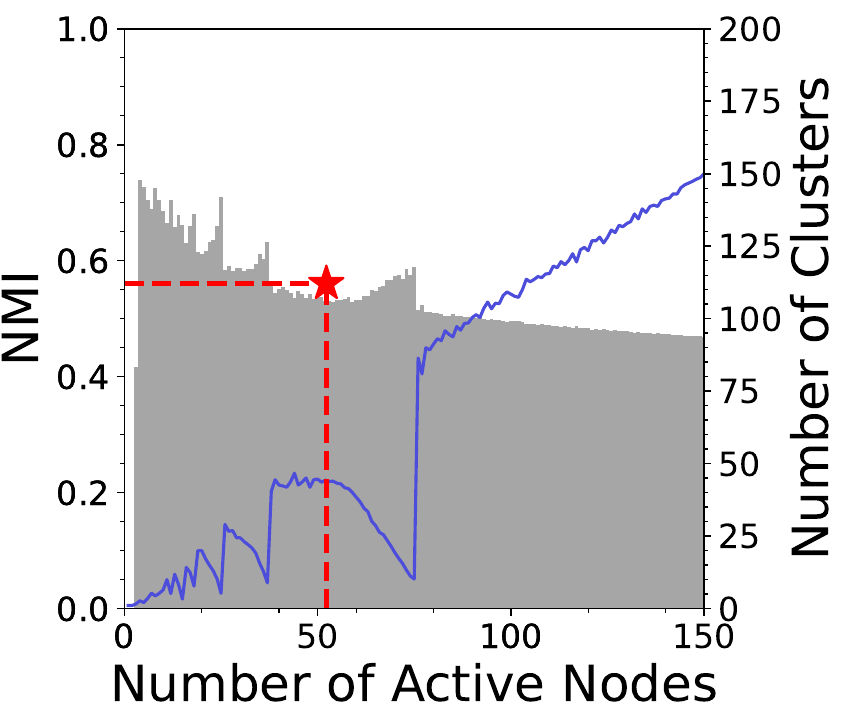}
		\label{fig:active_Iris_stationary}
	}\hfil
	\subfloat[Ionosphere]{
		\includegraphics[width=1.1in]{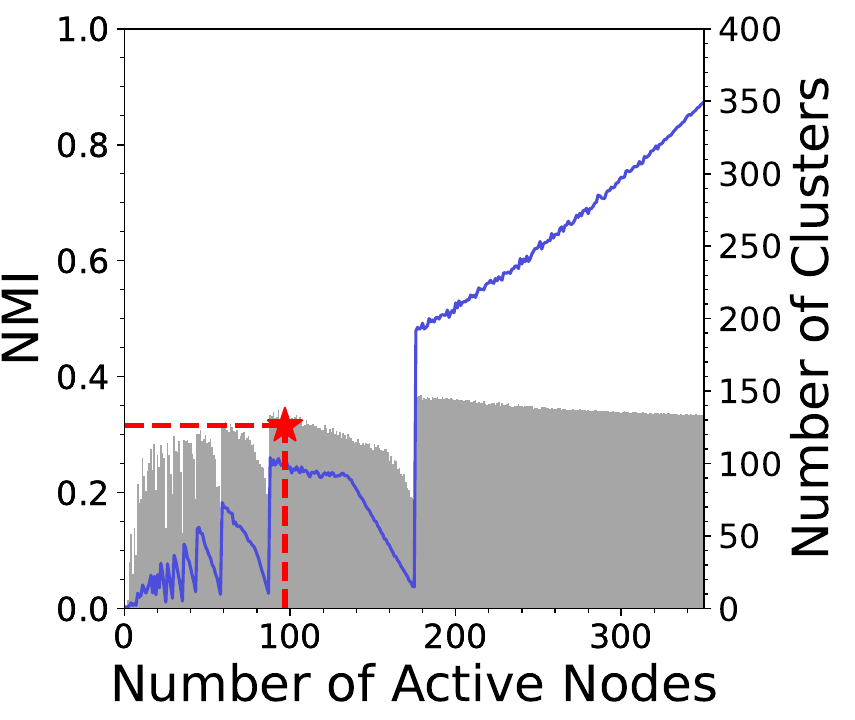}
		\label{fig:active_Ionosphere_stationary}
	}\hfil
	\subfloat[Pima]{
		\includegraphics[width=1.1in]{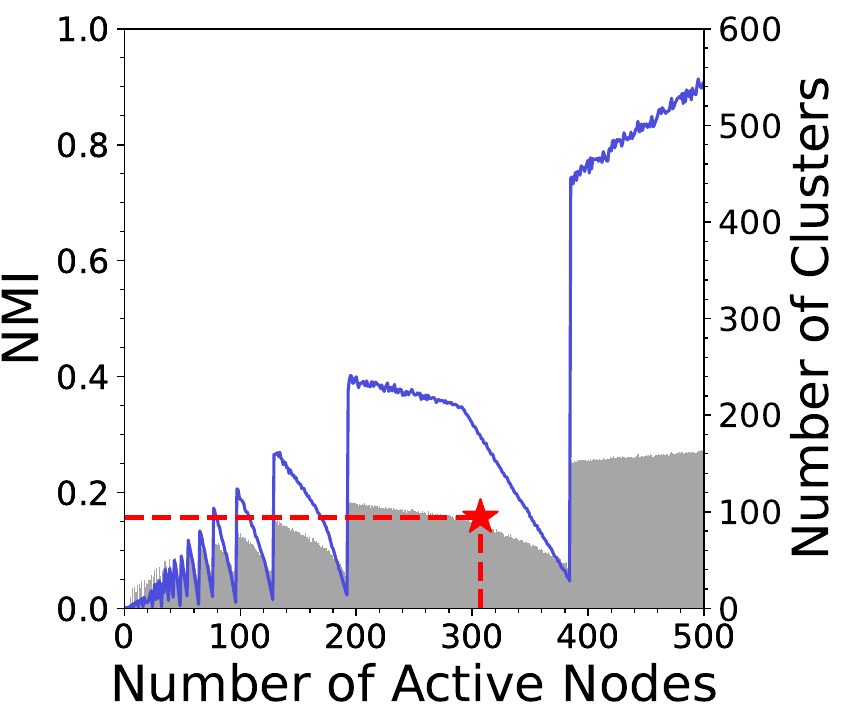}
		\label{fig:active_Pima_stationary}
	}\hfil
	\subfloat[Binalpha]{
		\includegraphics[width=1.1in]{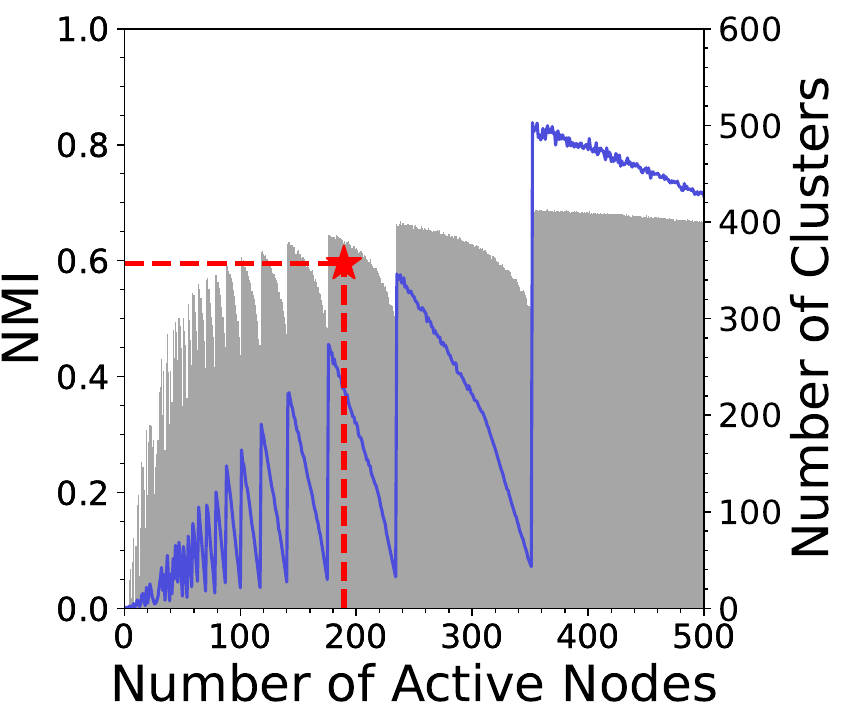}
		\label{fig:active_Binalpha_stationary}
	}\hfil
	\\
	\subfloat[Yeast]{
		\includegraphics[width=1.1in]{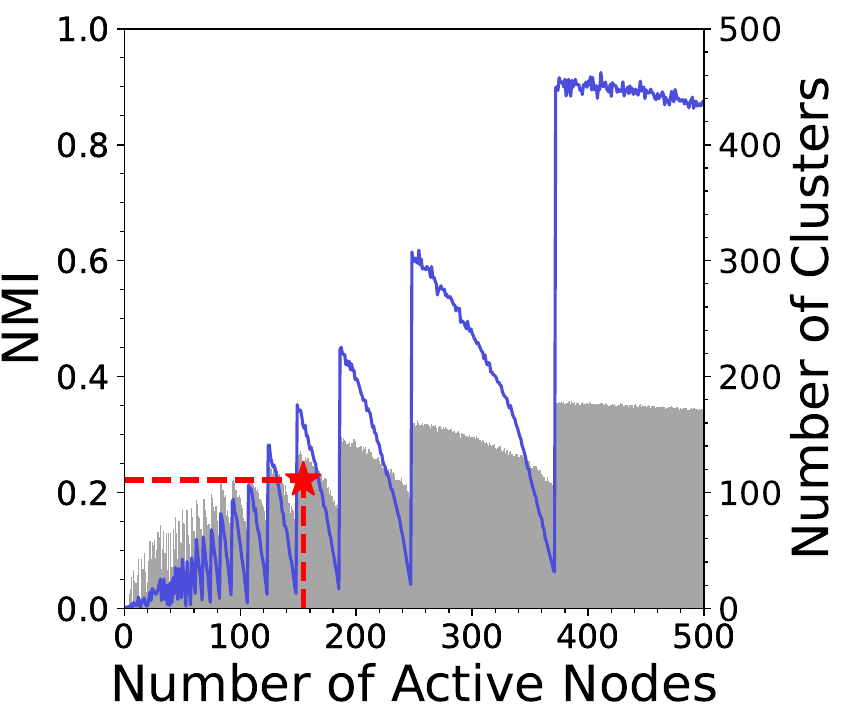}
		\label{fig:active_Yeast_stationary}
	}\hfil
	\subfloat[Semeion]{
		\includegraphics[width=1.1in]{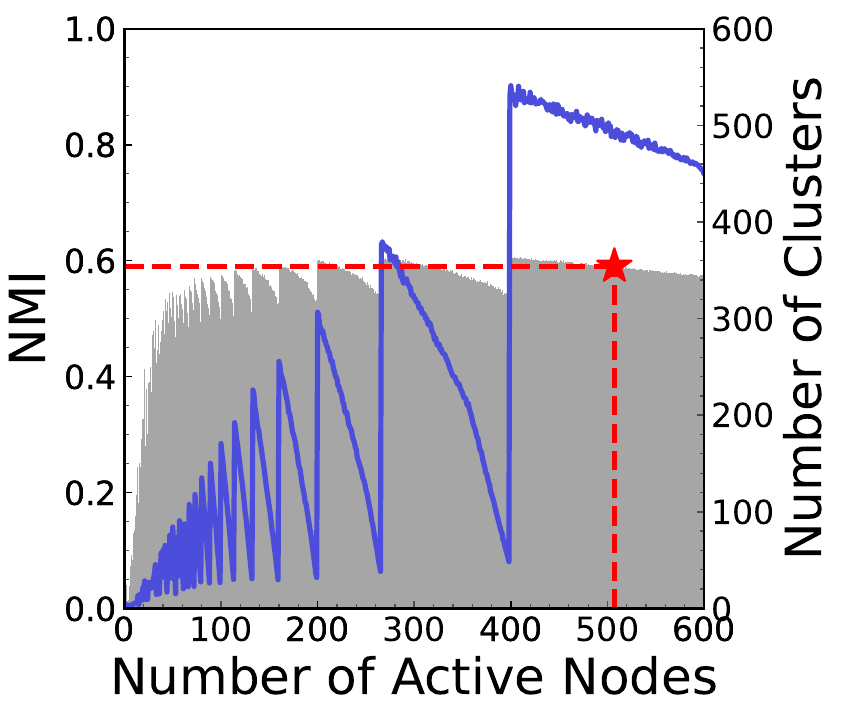}
		\label{fig:active_Semeion_stationary}
	}\hfil
	\subfloat[Image Segmentation]{
		\includegraphics[width=1.1in]{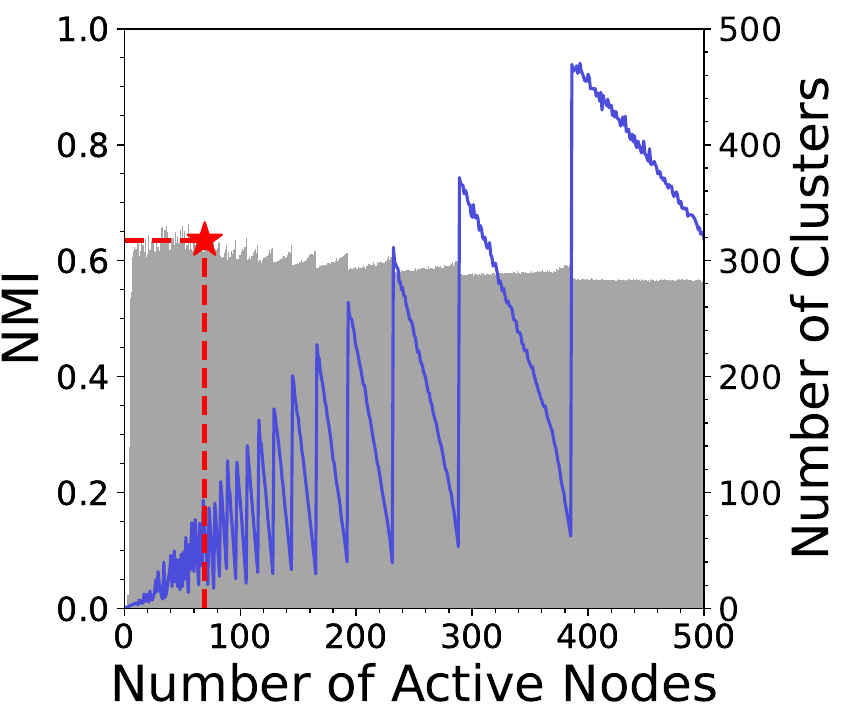}
		\label{fig:active_ImageSegmentation_stationary}
	}\hfil
	\subfloat[Phoneme]{
		\includegraphics[width=1.1in]{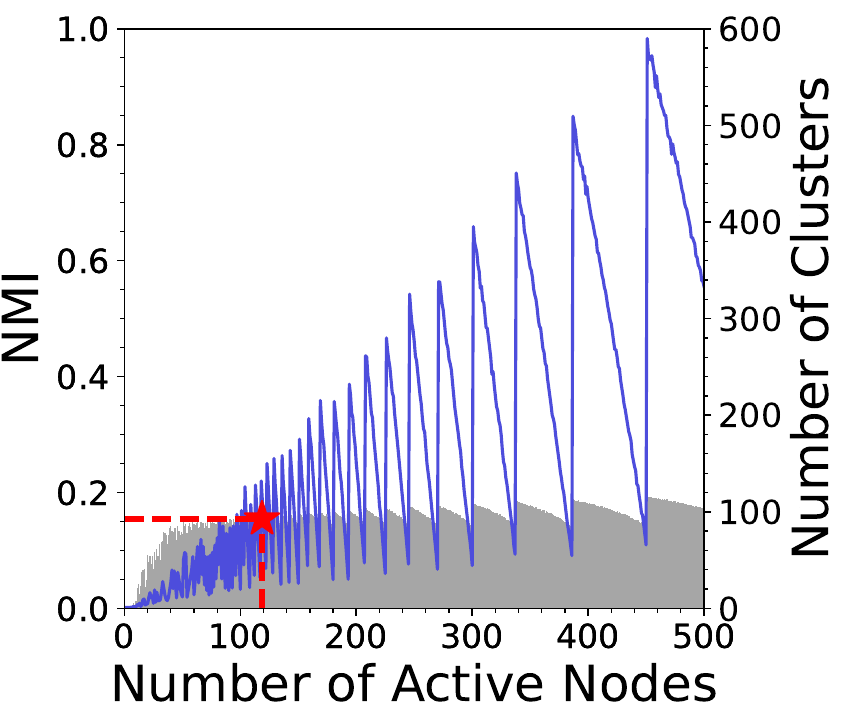}
		\label{fig:active_Phoneme_stationary}
	}\hfil
	\\
	\subfloat[Texture]{
		\includegraphics[width=1.1in]{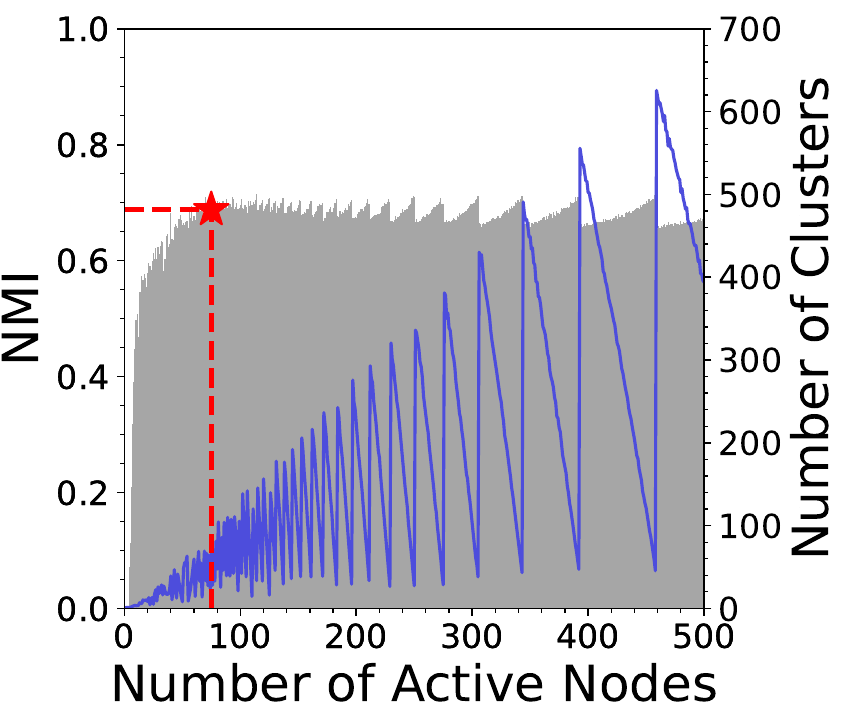}
		\label{fig:active_Texture_stationary}
	}\hfil
	\subfloat[PenBased]{
		\includegraphics[width=1.1in]{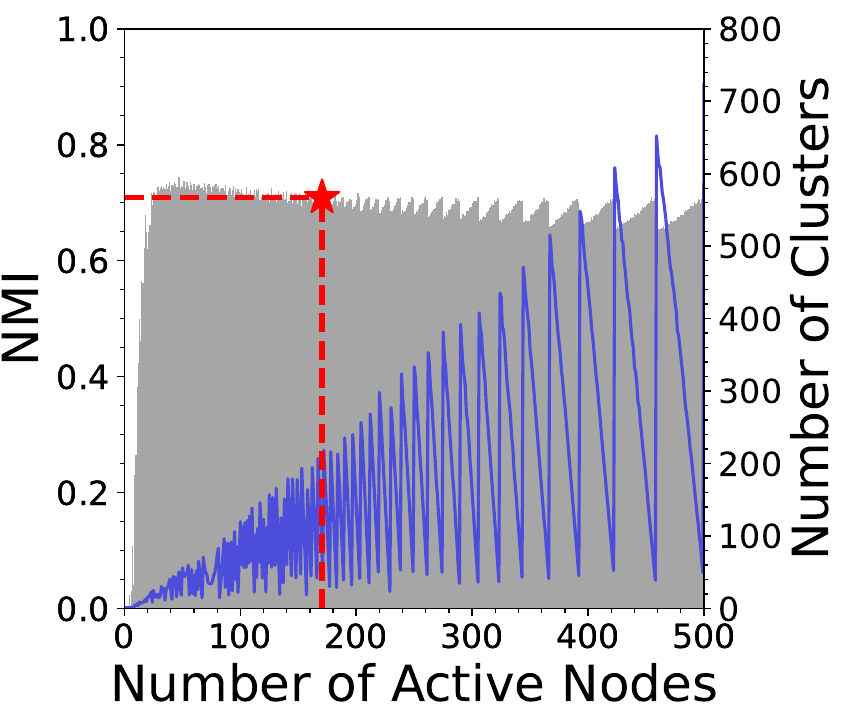}
		\label{fig:active_PenBased_stationary}
	}\hfil
	\subfloat[Letter]{
		\includegraphics[width=1.1in]{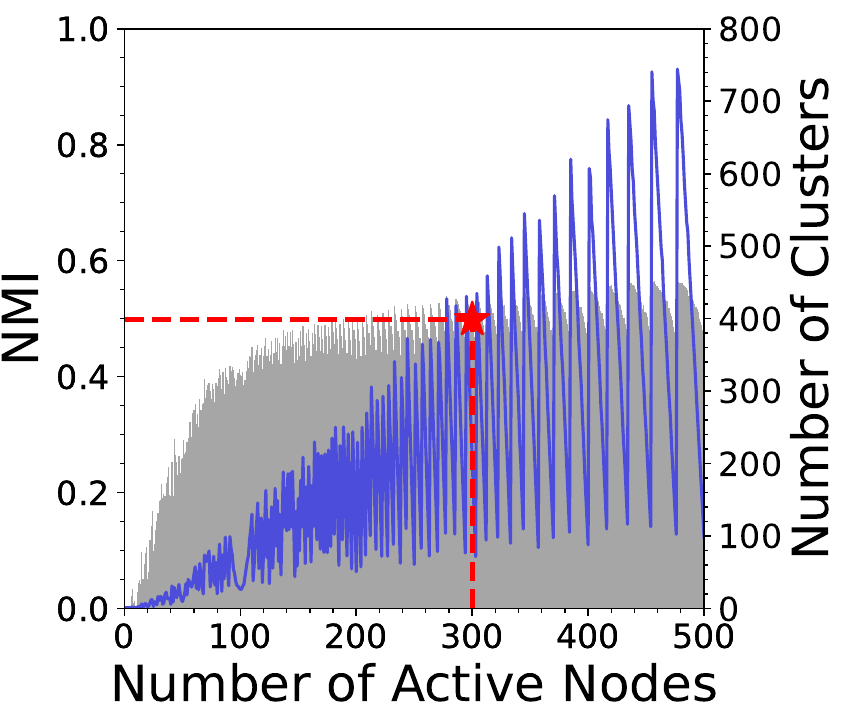}
		\label{fig:active_Letter_stationary}
	}\hfil
	\subfloat[Skin]{
		\includegraphics[width=1.1in]{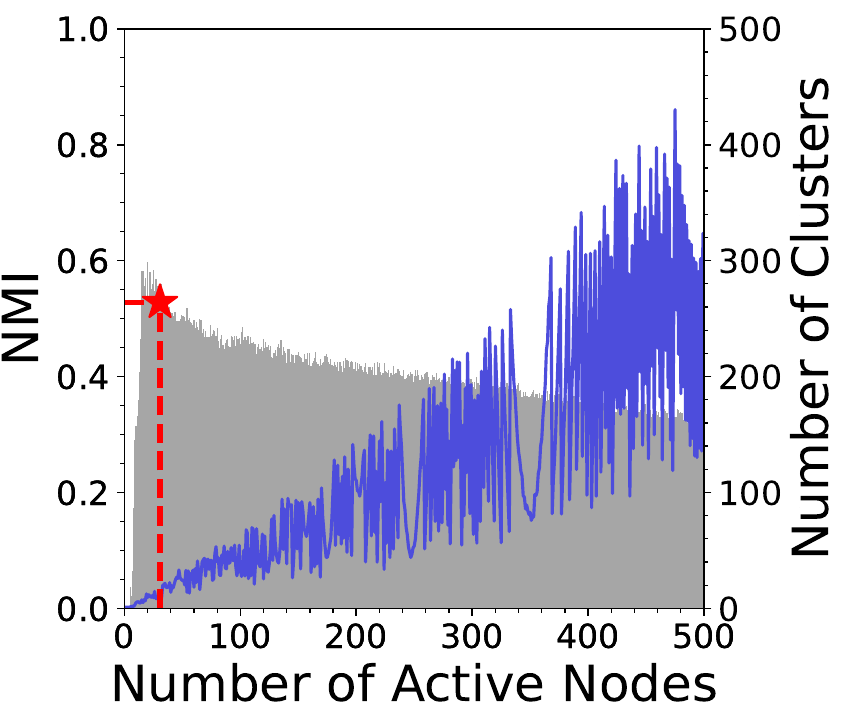}
		\label{fig:active_Skin_stationary}
	}
	\vspace{5pt}
	\includegraphics[width=3.5in]{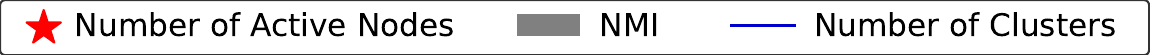}
  \vspace{2mm}
	\caption{Relationships among the number of active nodes, the number of clusters in CAE, and NMI in the stationary environment.}
	\label{fig:active_stationary}
\end{figure*}

\begin{figure*}[htbp]
	\centering
	\subfloat[Iris]{
		\includegraphics[width=1.1in]{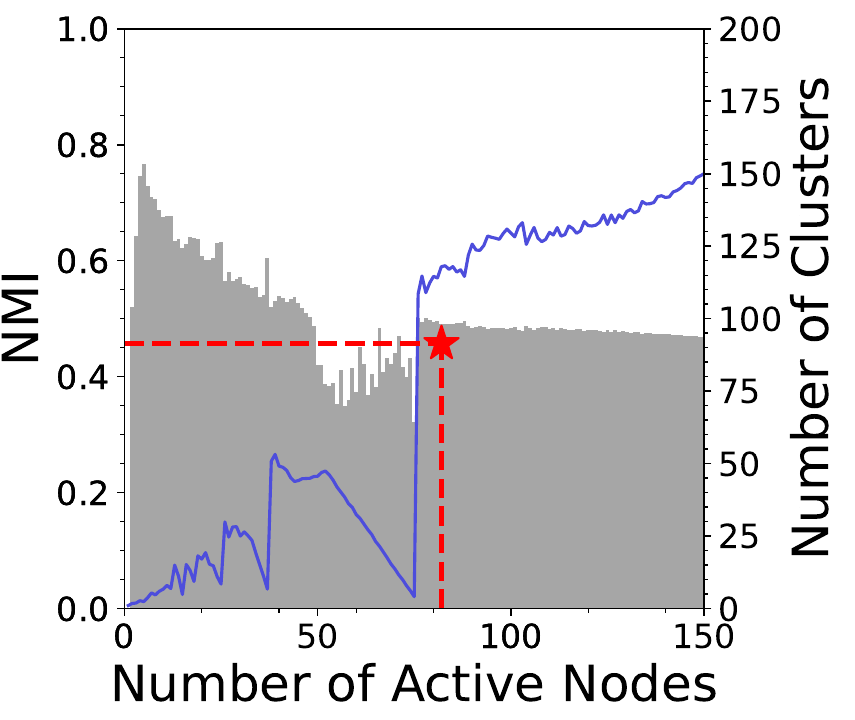}
		\label{fig:active_Iris_nonstationary}
	}\hfil
	\subfloat[Ionosphere]{
		\includegraphics[width=1.1in]{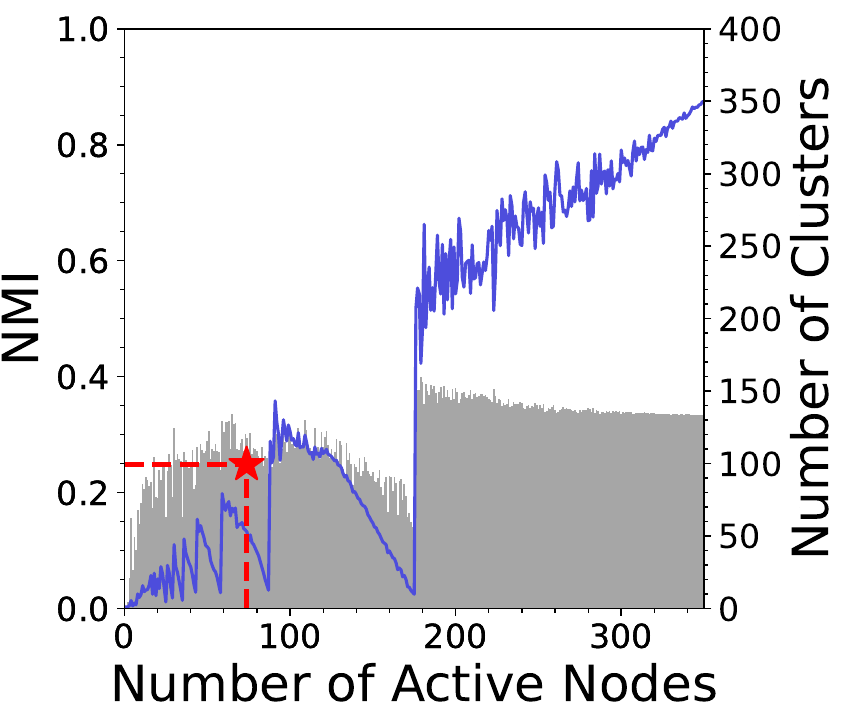}
		\label{fig:active_Ionosphere_nonstationary}
	}\hfil
	\subfloat[Pima]{
		\includegraphics[width=1.1in]{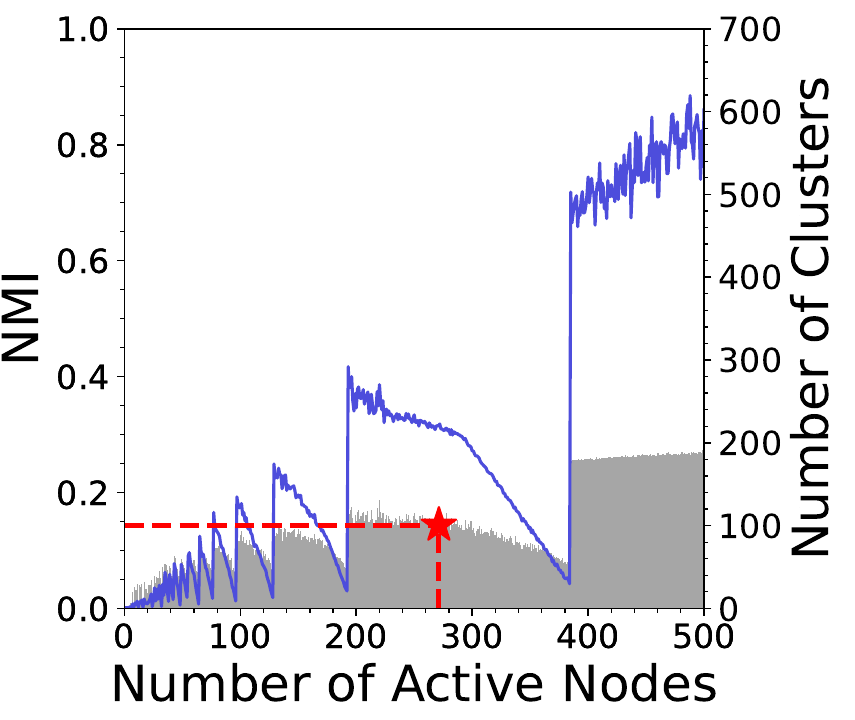}
		\label{fig:active_Pima_nonstationary}
	}\hfil
	\subfloat[Binalpha]{
		\includegraphics[width=1.1in]{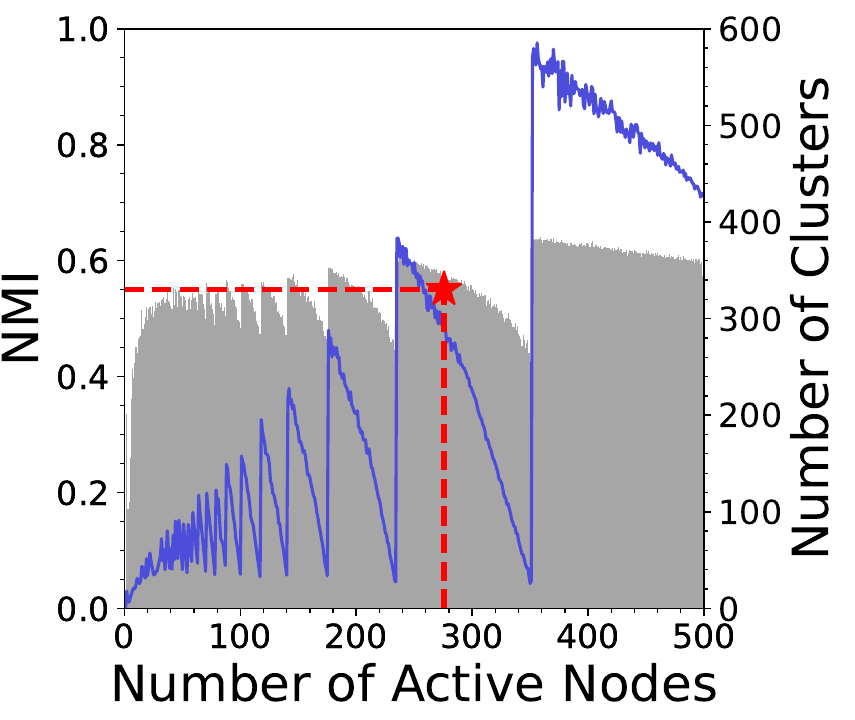}
		\label{fig:active_Binalpha_nonstationary}
	}\hfil
	\\
	\subfloat[Yeast]{
		\includegraphics[width=1.1in]{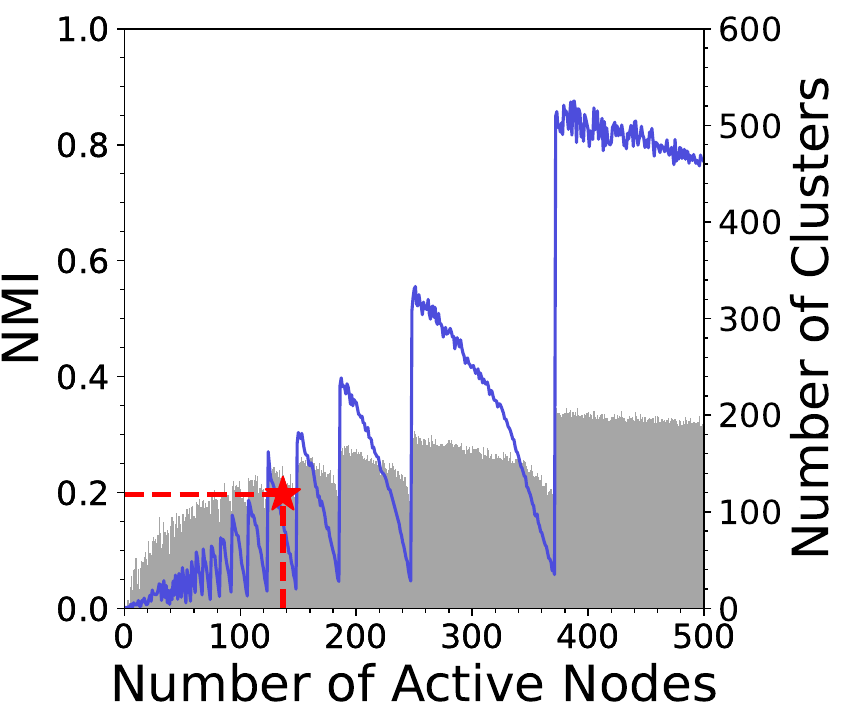}
		\label{fig:active_Yeast_nonstationary}
	}\hfil
	\subfloat[Semeion]{
		\includegraphics[width=1.1in]{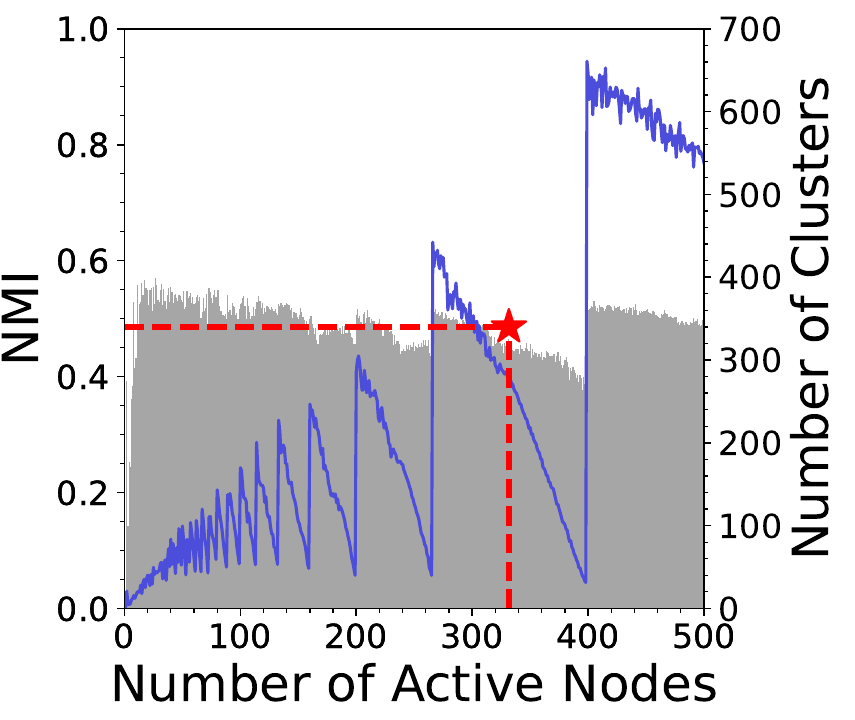}
		\label{fig:active_Semeion_nonstationary}
	}\hfil
	\subfloat[Image Segmentation]{
		\includegraphics[width=1.1in]{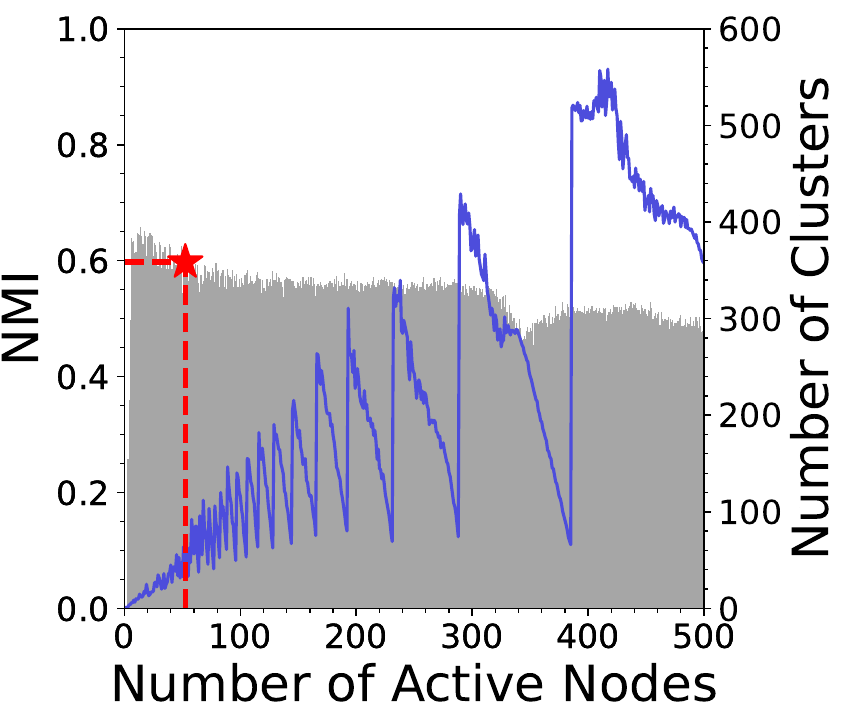}
		\label{fig:active_ImageSegmentation_nonstationary}
	}\hfil
	\subfloat[Phoneme]{
		\includegraphics[width=1.1in]{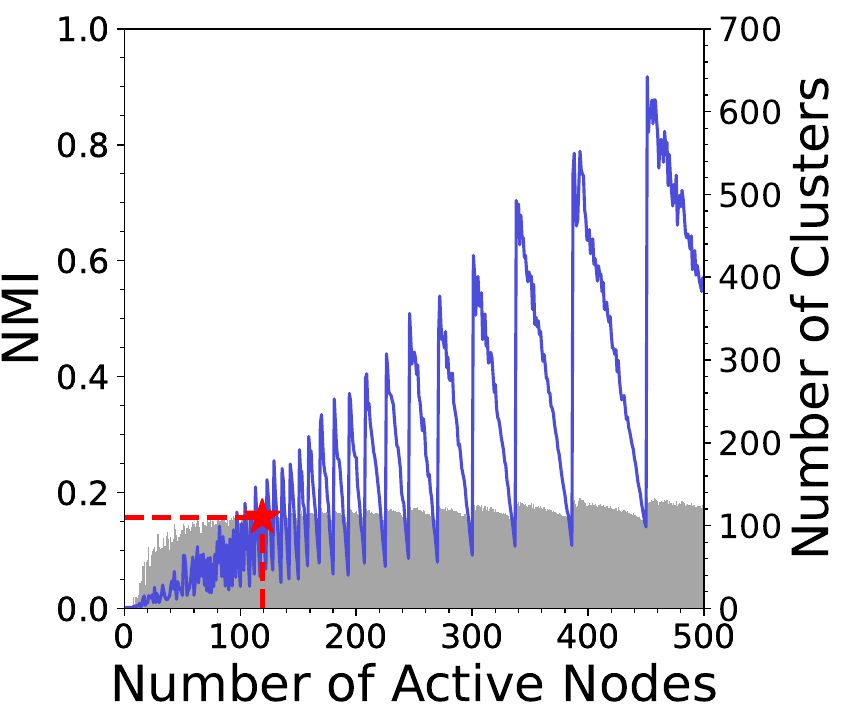}
		\label{fig:active_Phoneme_nonstationary}
	}\hfil
	\\
	\subfloat[Texture]{
		\includegraphics[width=1.1in]{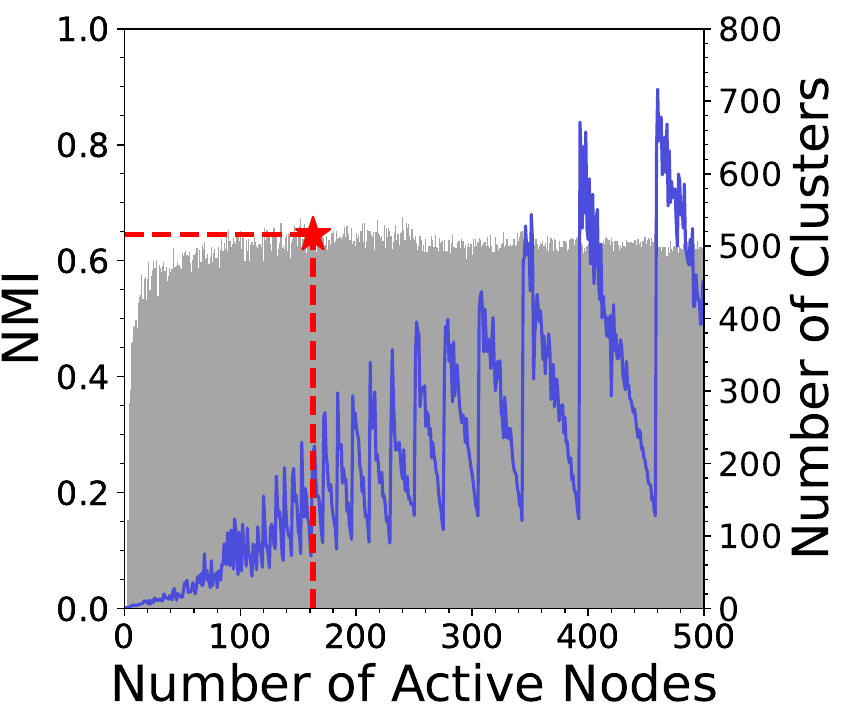}
		\label{fig:active_Texture_nonstationary}
	}\hfil
	\subfloat[PenBased]{
		\includegraphics[width=1.1in]{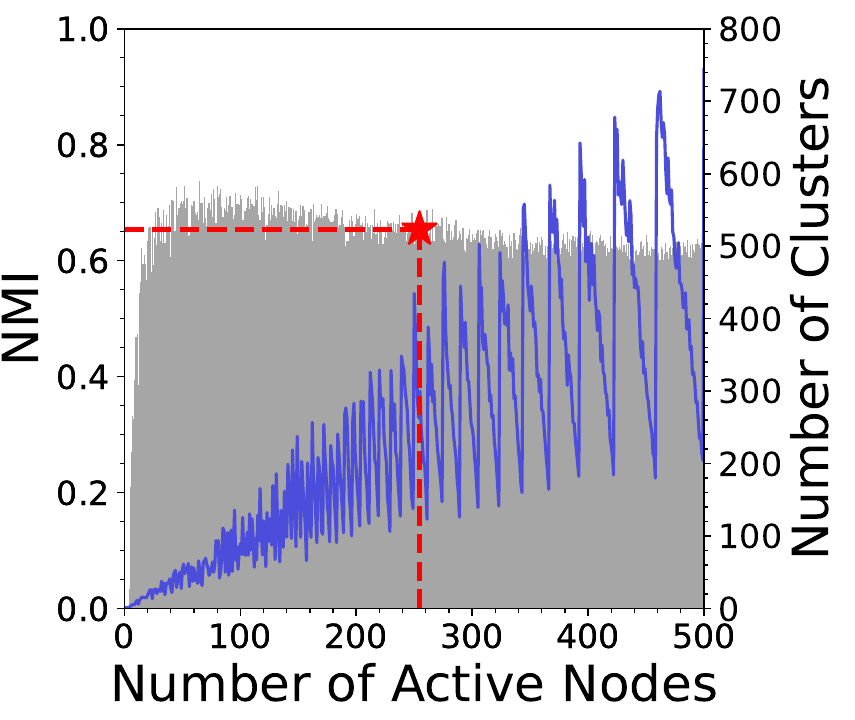}
		\label{fig:active_PenBased_nonstationary}
	}\hfil
	\subfloat[Letter]{
		\includegraphics[width=1.1in]{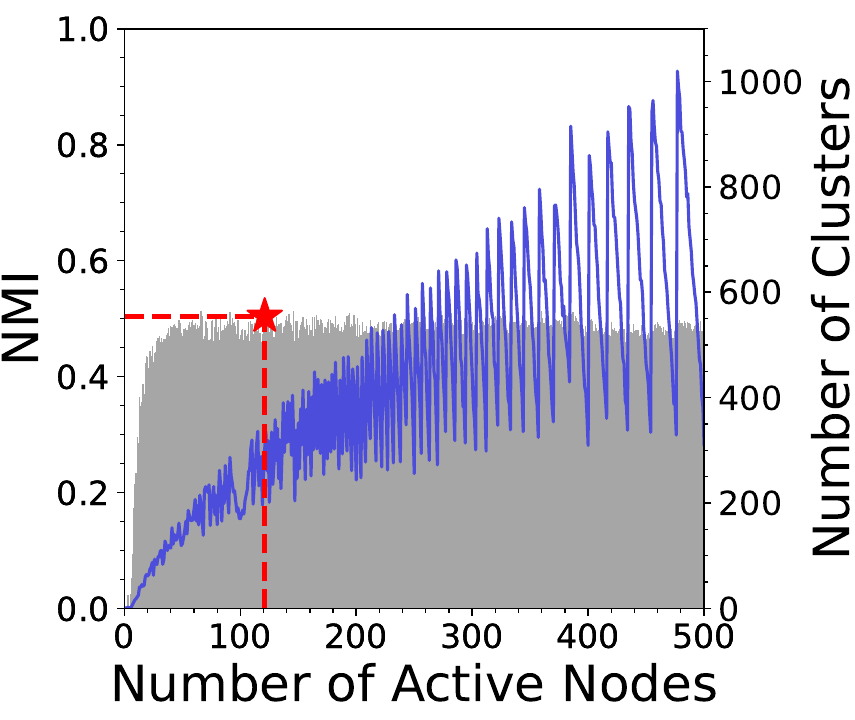}
		\label{fig:active_Letter_nonstationary}
	}\hfil
	\subfloat[Skin]{
		\includegraphics[width=1.1in]{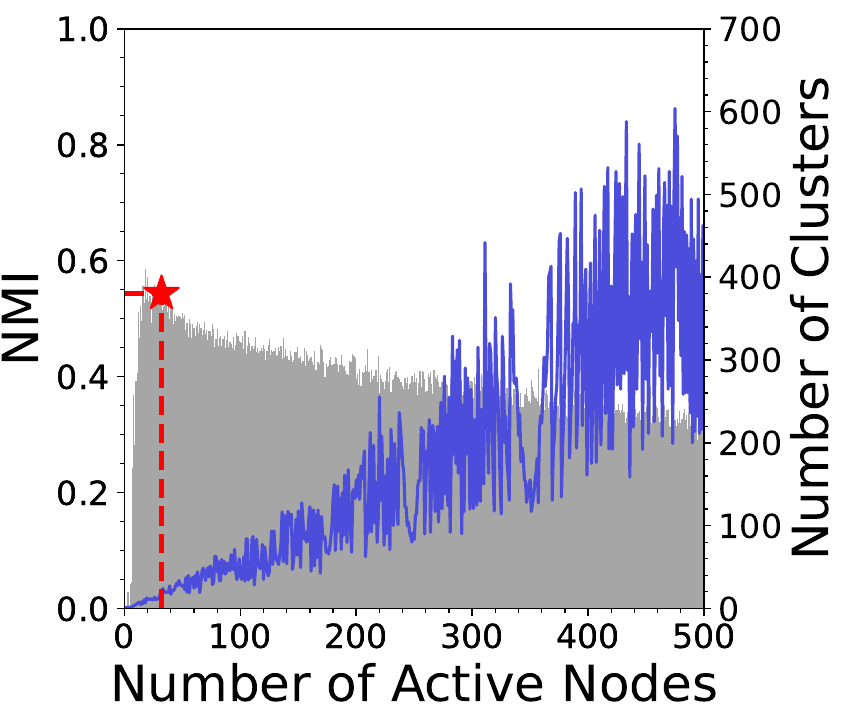}
		\label{fig:active_Skin_nonstationary}
	}
	\vspace{5pt}
	\includegraphics[width=3.5in]{active_legend.pdf}
  \vspace{2mm}
	\caption{Relationships among the number of active nodes, the number of clusters in CAE, and NMI in the non-stationary environment.}
	\label{fig:active_nonstationary}
\end{figure*}

Figs. \ref{fig:active_stationary} and \ref{fig:active_nonstationary} show the relationships among the number of active nodes, NMI, and the number of clusters for each dataset in the stationary and non-stationary environments, respectively. In these figures, the gray bars indicate NMI values, the blue lines represent the number of clusters, and the red stars show the number of active nodes estimated by the DPP-based criterion incorporating CIM.

Note that in Figs. \ref{fig:active_stationary} and \ref{fig:active_nonstationary}, the number of clusters in each dataset oscillates. This phenomenon occurs for the following reason: when the deletion cycle for isolated nodes (i.e., the number of active nodes $\lambda$) is a multiple of the number of presented data points, no isolated nodes remain as clusters after training. In contrast, if the deletion cycle $\lambda$ is not a multiple of the number of presented data points, up to $(\lambda - 1)$ isolated nodes can remain as clusters after training. As mentioned in Section \ref{sec:results_2d}, a simple solution to avoid this phenomenon is to remove isolated nodes after the learning procedure. However, since these algorithms are designed for continual learning, they retain isolated nodes to accommodate future learning, rather than removing them after the current training phase. Therefore, we do not consider this a drawback of these algorithms.

Since the general purpose of clustering is data aggregation or grouping, the ideal outcome is a high NMI value with a small number of clusters. From this perspective, Figs. \ref{fig:active_stationary} and \ref{fig:active_nonstationary} show that the estimated number of active nodes is effective for most datasets, with the exception of Iris. In particular, for the Ionosphere, Image Segmentation, Phoneme, Texture, and Skin datasets, the estimated number of active nodes yields nearly the highest NMI values in both stationary and non-stationary environments.

The above-mentioned observations suggest that the DPP-based criterion incorporating CIM and the calculation of the similarity threshold $V_{\text{threshold}}$ as in (\ref{eq:pairwiseCIM}) are practically effective across a variety of datasets.

\subsection{Computational Complexity}
\label{sec:compComplexity}
For computational complexity analysis, we use the notations in Table \ref{tab:notations}, namely, $ d $ is the dimensionality of a data point, $ n $ is the number of data points, $ K $ is the number of nodes, $ \lambda $ is the number of active nodes, and $ |\mathcal{E}| $ is the number of elements in the ages of edges set $\mathcal{E}$.

The computational complexity of each process in CAE is as follows: for computing a bandwidth of a kernel function in CIM is $ \mathcal{O}(d) $, for computing CIM is $ \mathcal{O}(ndK) $ (line 11 in Alg. \ref{alg:pseudocodeCAE}), for finding nodes which have the 1st and 2nd smallest CIM value is $ \mathcal{O}(K) $ (line 11 in Alg. \ref{alg:pseudocodeCAE}), for calculating a pairwise similarity matrix by using CIM is $ \mathcal{O}((\frac{\lambda}{2})^{2}dK) $ (line 1 in Alg. \ref{alg:ActiveNode}), for calculating the determinant of the pairwise similarity matrix is $ \mathcal{O}((\frac{\lambda}{2})^{3}) $ (line 2 in Alg. \ref{alg:ActiveNode}), and for estimating the edge deletion threshold is $ \mathcal{O}(|\mathcal{E}|\log{|\mathcal{E}|}) $ (Alg. \ref{alg:EstEdgeThreshold}). 


In general, $n < \lambda^{3}$, $K \ll n$, and $\lambda < K$. Consequently, the computational complexity of CAE is $\mathcal{O}(\lambda^{3} d K)$, where $d$ is the data dimensionality, $K$ is the number of nodes, and $\lambda$ is the number of active nodes used in the similarity threshold estimation. The cubic dependence on $\lambda$ implies that the DPP-based diversity estimation constitutes the main computational bottleneck. Nonetheless, in practice, the conditions $\lambda \ll K \ll n$ (where $n$ is the number of data points) typically hold, which prevents this step from dominating the overall runtime. As task complexity and data diversity increase, $K$ will grow accordingly, but continual deletion of edges and isolated nodes mitigates uncontrolled network expansion, supporting stable scalability. For practical deployments in resource-constrained environments, CAE further benefits from its parameter-free nature, which avoids the costly tuning process required in other clustering algorithms. To verify that the theoretical complexity aligns with practical behavior, we measured the empirical computational cost of CAE across multiple benchmark datasets, focusing on total runtime, the DPP share, and peak memory usage.

Table~\ref{tab:runtime} summarizes the results under both stationary and nonstationary environments. The wall-clock time denotes the total learning time per experiment, the DPP share represents the proportion of runtime spent on determinant-based diversity estimation (Algorithm~\ref{alg:ActiveNode}), and the peak memory indicates the maximum footprint observed on a dedicated MATLAB worker. As shown in Table~\ref{tab:runtime}, the DPP step accounted for less than 25\% of the total runtime for most datasets and below 1\% for large-scale cases such as Skin and PenBased. These results confirm that the $\lambda^{3}$ term does not dominate the overall cost. Peak memory scaled roughly with the number of nodes $K$ and stayed within a few megabytes even for high-dimensional data. Overall, the empirical measurements demonstrate that CAE achieves computational efficiency consistent with its theoretical complexity and maintains stable scalability without parameter tuning.

\begin{table}[htbp]
\centering
\caption{Empirical computational cost of CAE under stationary and nonstationary environments}
\label{tab:runtime}
\begin{tabular}{llccc}
\hline\hline
Environment & Dataset & Wall-Clock Time [s] & DPP Share [\%] & Peak Memory [MB] \\
\hline
\multirow[t]{12}{*}{Stationary}
 & Iris              & 0.178 (0.115) & 13.378 (4.142) & 0.287 (1.186) \\
 & Ionosphere        & 0.089 (0.012) & 17.567 (6.564) & 0.249 (1.115) \\
 & Pima              & 0.743 (1.178) & 65.806 (13.935) & 2.740 (2.708) \\
 & Binalpha          & 3.176 (1.440) & 6.374 (4.244) & 2.059 (2.157) \\
 & Yeast             & 0.179 (0.092) & 22.031 (14.775) & 0.502 (0.829) \\
 & Semeion           & 7.440 (0.697) & 24.284 (3.587) & 8.551 (0.936) \\
 & Image Seg.        & 0.174 (0.050) & 2.592 (1.884) & 0.051 (0.224) \\
 & Phoneme           & 0.339 (0.051) & 3.900 (1.319) & 0.351 (0.491) \\
 & Texture           & 0.868 (0.372) & 0.771 (0.591) & 0.153 (0.490) \\
 & PenBased          & 2.124 (0.780) & 1.907 (1.485) & 2.177 (3.058) \\
 & Letter            & 7.869 (3.497) & 3.108 (2.081) & 5.732 (3.754) \\
 & Skin              & 12.324 (1.512) & 0.006 (0.002) & 0.507 (0.508) \\
\hline
\multirow[t]{12}{*}{Nonstationary}
 & Iris              & 0.103 (0.043) & 22.127 (13.647) & 0.326 (1.426) \\
 & Ionosphere        & 0.090 (0.021) & 10.970 (7.211) & 0.094 (0.420) \\
 & Pima              & 0.756 (1.377) & 56.819 (20.346) & 2.347 (3.315) \\
 & Binalpha          & 4.433 (3.366) & 12.587 (11.743) & 3.755 (3.462) \\
 & Yeast             & 0.191 (0.206) & 19.096 (20.720) & 1.074 (1.887) \\
 & Semeion           & 4.142 (1.492) & 14.007 (7.717) & 4.841 (3.835) \\
 & Image Seg.        & 0.188 (0.028) & 1.537 (0.945) & 0.009 (0.029) \\
 & Phoneme           & 0.424 (0.146) & 3.917 (3.034) & 0.706 (0.864) \\
 & Texture           & 2.346 (0.963) & 2.695 (3.316) & 3.439 (2.805) \\
 & PenBased          & 4.963 (2.490) & 3.116 (3.090) & 11.907 (8.092) \\
 & Letter            & 7.344 (2.448) & 0.288 (0.241) & 10.163 (5.752) \\
 & Skin              & 16.753 (5.801) & 0.005 (0.001) & 1.550 (1.381) \\
\hline\hline
\end{tabular}
\vspace{1mm}
\footnotesize
\begin{minipage}{\linewidth}
\setlength{\leftskip}{1mm}
The values are averaged over 20 evaluations. \\
Values in parentheses indicate the standard deviation. \\
Image Seg.\ stands for Image Segmentation. \\
DPP share shows the ratio of wall-clock time used for determinant-based diversity.
\end{minipage}
\end{table}

\section{Concluding Remarks}
\label{sec:conclusion}
This paper proposes a new parameter-free ART-based topological clustering algorithm capable of continual learning. The algorithm introduces two parameter estimation processes: first, estimating the number of active nodes for calculating the similarity threshold using a DPP-based criterion with CIM; second, determining a node deletion threshold based on the age of each edge. Empirical studies using synthetic and real-world datasets demonstrate that CAE achieves superior clustering performance compared to state-of-the-art parameter-free or fixed-parameter algorithms, while maintaining continual learning ability. These capabilities highlight the utility and potential of CAE as a data preprocessing method for various applications.

In real-world applications related to health, finance, and medical fields, a dataset often contain both numerical and categorical attributes and are referred to as mixed datasets \cite{ahmad19}. One potential direction for future research is to develop parameter-free clustering algorithms that can effectively handle mixed datasets while retaining continual learning capabilities.

\backmatter

\bmhead{Acknowledgements}
This work was supported by the Japan Society for the Promotion of Science (JSPS) KAKENHI Grant Number JP19K20358 and 22H03664, National Natural Science Foundation of China (Grant No. 62250710163, 62250710682), Guangdong Provincial Key Laboratory (Grant No. 2020B121201001), the Program for Guangdong Introducing Innovative and Enterpreneurial Teams (Grant No. 2017ZT07X386), The Stable Support Plan Program of Shenzhen Natural Science Fund (Grant No. 20200925174447003), and Shenzhen Science and Technology Program (Grant No. KQTD2016112514355531).

\bmhead{Data Availability}
All the datasets used in this paper are available from public repositories \cite{derrac15, dua19}.

\section*{Declarations}

\bmhead{Conflict of interest}
The authors declare that they have no known competing financial interests or personal relationships that could have appeared to influence the work reported in this paper.

\bmhead{Ethical Approval}
This paper contains no studies with human participants or animals performed by authors.


\bibliography{myref}

\end{document}